\documentclass[10pt,twocolumn,letterpaper]{article}

\usepackage{cvpr}              

\usepackage{url}

\usepackage{amsmath}
\usepackage{amsfonts}
\usepackage{amsthm}
\usepackage{mathtools}
\usepackage{amssymb}
\usepackage{booktabs}
\usepackage{color}
\usepackage{bm}
\usepackage{enumitem}
\usepackage{multirow}
\usepackage{caption}
\usepackage{array}
\usepackage[dvipsnames]{xcolor}

\usepackage[accsupp]{axessibility}

\usepackage{amsmath,amsfonts,bm}

\def\eqref#1{equation~\ref{#1}}

\def\1{\bm{1}}

\DeclareMathAlphabet{\mathsfit}{\encodingdefault}{\sfdefault}{m}{sl}
\SetMathAlphabet{\mathsfit}{bold}{\encodingdefault}{\sfdefault}{bx}{n}

\newcommand{\PreserveBackslash}[1]{\let\temp=\\#1\let\\=\temp}
\newcolumntype{C}[1]{>{\PreserveBackslash\centering}p{#1}}
\newcolumntype{R}[1]{>{\PreserveBackslash\raggedleft}p{#1}}
\newcolumntype{L}[1]{>{\PreserveBackslash\raggedright}p{#1}}

\newcommand{\att}[1]{{A_{\!#1}}}
\newcommand{\prop}[1]{\mathit{#1}}
\newcommand{\zz}{{\textcolor{white}{0}}}
\newcommand{\Godel}{{G\"odel}}
\renewcommand{\eqref}[1]{{(\ref{#1})}}

\definecolor{cvprblue}{rgb}{0.21,0.49,0.74}
\usepackage[pagebackref,breaklinks,colorlinks,citecolor=cvprblue]{hyperref}

\def\paperID{10912} 
\def\confName{CVPR}
\def\confYear{2024}

\title{Predicated Diffusion: Predicate Logic-Based Attention Guidance for Text-to-Image Diffusion Models}

\author{Kota Sueyoshi, Takashi Matsubara\\
Osaka University\\
1-3 Machikaneyama, Toyonaka, Osaka, 560-8531 Japan.\\
{\tt\small sueyoshi@hopf.sys.es.osaka-u.ac.jp, matsubara@sys.es.osaka-u.ac.jp}
}

\begin{document}
\maketitle

\begin{abstract}
    Diffusion models have achieved remarkable success in generating high-quality, diverse, and creative images.
    However, in text-based image generation, they often struggle to accurately capture the intended meaning of the text.
    For instance, a specified object might not be generated, or an adjective might incorrectly alter unintended objects.
    Moreover, we found that relationships indicating possession between objects are frequently overlooked.
    Despite the diversity of users' intentions in text, existing methods often focus on only some aspects of these intentions.
    In this paper, we propose Predicated Diffusion, a unified framework designed to more effectively express users' intentions.
    It represents the intended meaning as propositions using predicate logic and treats the pixels in attention maps as fuzzy predicates.
    This approach provides a differentiable loss function that offers guidance for the image generation process to better fulfill the propositions.
    Comparative evaluations with existing methods demonstrated that Predicated Diffusion excels in generating images faithful to various text prompts, while maintaining high image quality, as validated by human evaluators and pretrained image-text models.
\end{abstract}

\section{Introduction}\label{sec:intro}
Recent advancements in deep learning have paved the way for generating high-quality, diverse, and creative images.
This progress is primarily attributed to diffusion models~\citep{Ho2020,Song2020}, which recursively update images to remove noise and to make them more realistic.
Diffusion models are significantly more stable and scalable compared to previous methods, such as generative adversarial networks~\citep{Goodfellow2014,Radford2015} or autoregressive models~\citep{VanDenOord2016,Kolesnikov2017}.
Moreover, the field of text-based image generation is attracting considerable attention, with the goal being to generate images faithful to a text prompt given as input.
Even in this area, the contributions of diffusion models are notable~\citep{Ramesh2021}.
We can benefit from commercial applications such as DALL-E~\citep{Ramesh2022} and Imagen~\citep{Saharia2022}, as well as the state-of-the-art open-source model, Stable Diffusion~\citep{Rombach2022,Podell2023}.
These models are trained on large-scale and diverse image-text datasets, which allows them to respond to a variety of prompts and to generate images of objects with colors, shapes, and materials not found in the existing datasets.

\begin{figure}[t]
    \centering
    \footnotesize
    \tabcolsep=.1mm
    \begin{tabular}{C{14.5mm}C{17mm}C{17mm}C{17mm}C{17mm}}
                                                                                                               & \textbf{Missing Objects} & \textbf{Object Mixture} & \textbf{Attribute Leakage}                       & \textbf{Possession Failure} \\[1mm]
                                                                                                               & \emph{a~yellow~car}      & \emph{a~bird }          & \scalebox{0.95}[1.0]{\!\emph{a~green balloon}\!} & \emph{a boy}                \\
        \textbf{Prompts}                                                                                       & \emph{and}               & \emph{and}              & \emph{and}                                       & \emph{grasping}             \\
                                                                                                               & \emph{a~blue bird}       & \emph{a~cat}            & \emph{a~purple clock}                            & \emph{a~soccer ball}        \\
        \raisebox{6mm}{\parbox{14.5mm}{\centering\textbf{Stable Diffusion}}}
                                                                                                               &
        \includegraphics[width=16mm]{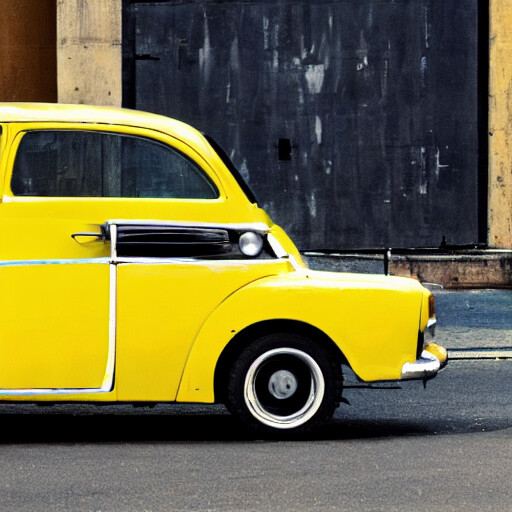}           &
        \includegraphics[width=16mm]{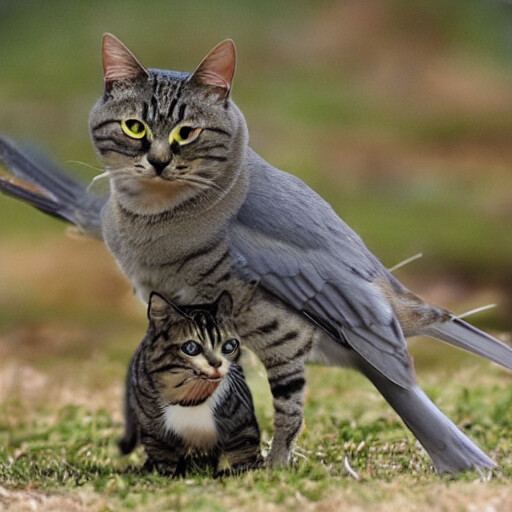}                             &
        \includegraphics[width=16mm]{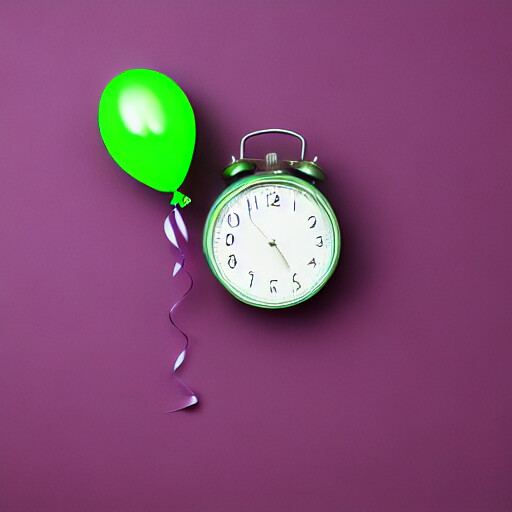}     &
        \includegraphics[width=16mm]{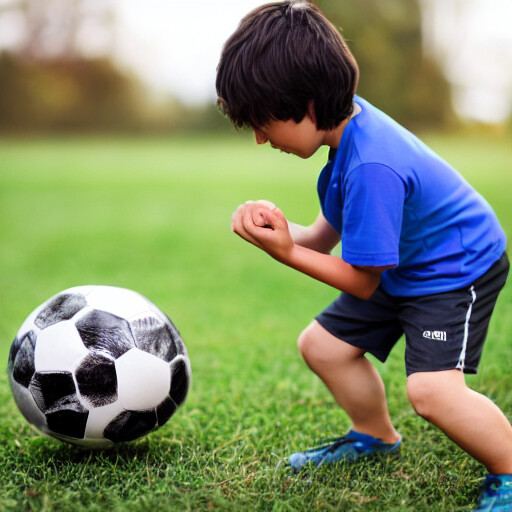}                                                                                                                                                    \\
        \raisebox{6mm}{\parbox{14.5mm}{\centering\textbf{Predicated Diffusion (Ours)}}}
                                                                                                               &
        \includegraphics[width=16mm]{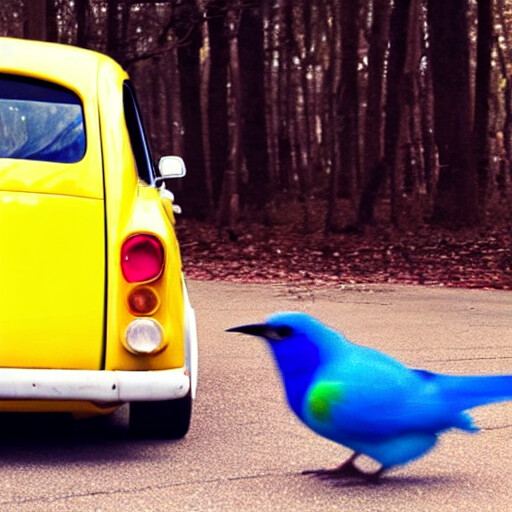}       &
        \includegraphics[width=16mm]{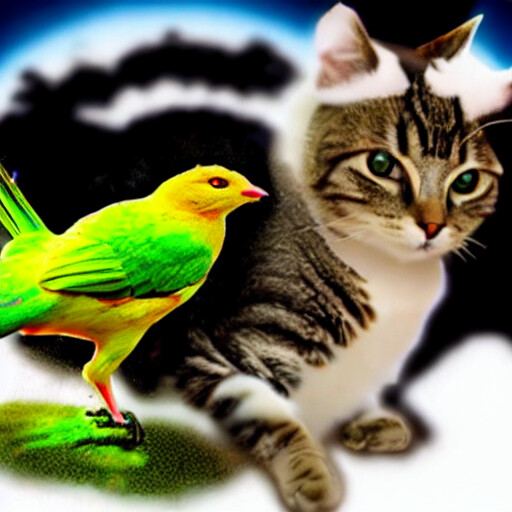}                         &
        \includegraphics[width=16mm]{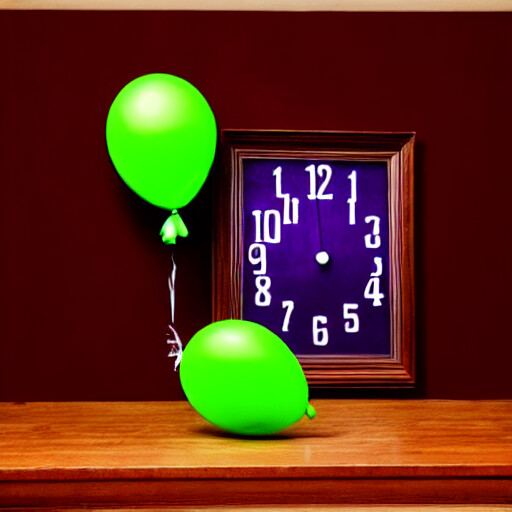} &
        \includegraphics[width=16mm]{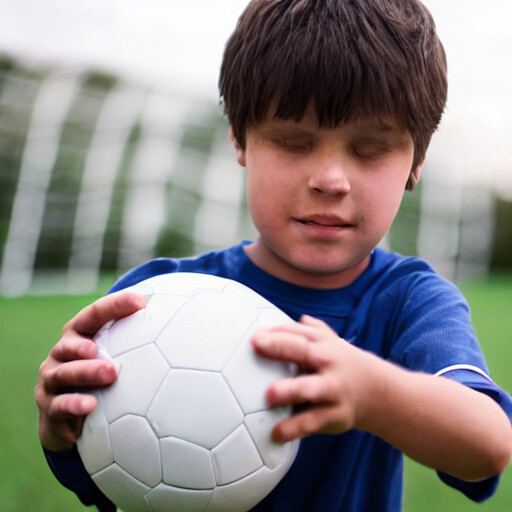}                                                                                                                                                  \\
    \end{tabular}
    \vspace*{-2mm}
    \caption{Visualizations of typical challenges in text-based image generation using diffusion models. The proposed Predicated Diffusion can address all of these challenges.}
    \label{fig:challenges}
\end{figure}

However, many previous studies have pointed out that these models often generate images that ignore the intended meanings of a given prompt, as exemplified in Fig.~\ref{fig:challenges}~\citep{Feng2023,Chefer2023,Rassin2023,Wang2023b}.
When multiple objects are specified in a prompt, only some are generated, with the others disappearing (see the column \emph{missing objects} in Fig.~\ref{fig:challenges}).
Also, two specified objects are sometimes mixed together to form one object in the generated image (\emph{object mixture}).
Given an adjective in a prompt, it alters a different object than the one the adjective was originally intended to modify (\emph{attribute leakage}).
We have found a novel challenge: while a prompt specifies an object being held by someone, the object is depicted as if discarded on the ground (\emph{possession failure}).
Since retraining diffusion models on large-scale datasets is prohibitively expensive, many studies have proposed methods offering guidance for the image generation process of pre-trained diffusion models, ensuring that the images are updated to become more faithful to the prompt.
However, these guidances vary widely, and a unified solution to address the diverse challenges has yet to be established.

\begin{figure}[t]
    \centering
    \includegraphics[scale=0.35,page=1]{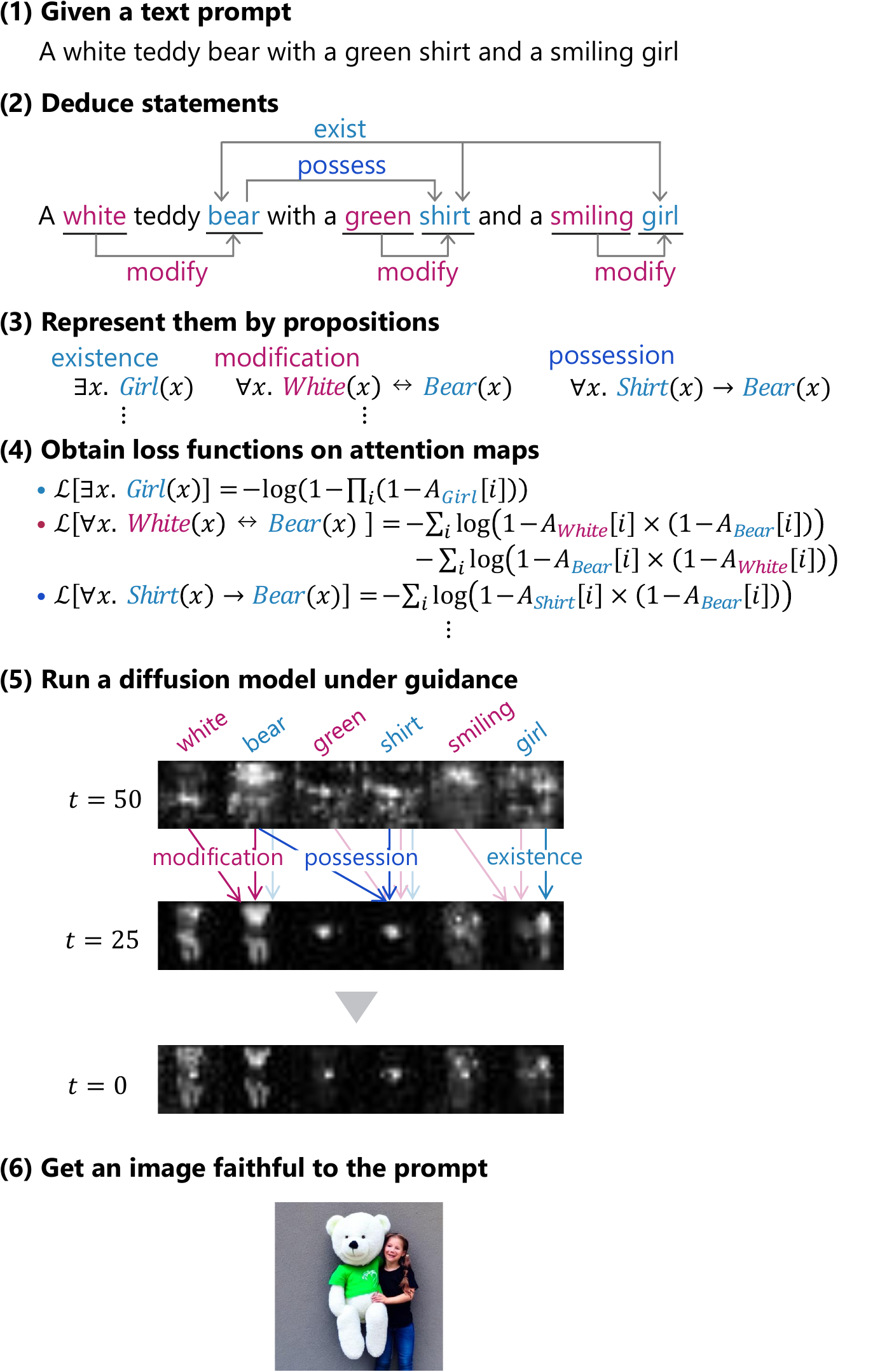}
    \vspace*{-2mm}
    \caption{The conceptual diagram of the proposed Predicated Diffusion, composed of steps (1)--(6).
        One can make propositions manually or using a syntactic dependency parser.
    }
    \label{fig:framework}
    \vspace*{-2mm}
\end{figure}

The root cause of these challenges lies in the diffusion models' inability to accurately capture the logical statements presented in the given prompts, as observed in other image-text models~\cite{Yuksekgonul2023}.
If we could represent such logical statements using predicate logic and integrate it into the diffusion model, the generated images might be more faithful to the statements.
Motivated by this idea, we introduce \emph{Predicated Diffusion} in this paper.
Herein, we represent the relationships between the words in the prompt by propositions using predicate logic.
By employing attention maps and fuzzy logic~\citep{Hajek1998,Prokopowicz2017}, we measure the degree to which the image under generation fulfills the propositions, providing guidance for images to become more faithful to the prompt.
See the conceptual diagram in Fig.~\ref{fig:framework}.
The contribution of this paper is threefold.

\vspace*{1mm}\noindent\textbf{Theoretical Justification and Generality:}
Most existing methods have been formulated based on deep insights, which makes it unclear how to combine them effectively or how to apply them in slightly different situations.
In contrast, Predicated Diffusion can resolve a variety of challenges based on the same foundational theory, allowing us to deductively expand it to address challenges not summarized in Fig.~\ref{fig:challenges}.

\vspace*{1mm}\noindent\textbf{High Fidelity to Prompt:}
The images generated by the proposed Predicated Diffusion and comparison methods were examined by human evaluators and pretrained image-text models~\citep{Radford2021a,Li2022c}.
We observed that Predicated Diffusion generates images more faithful to the prompts and more effectively prevents the issues shown in Fig.~\ref{fig:challenges}, while maintaining or even improving the image quality.

\vspace*{1mm}\noindent\textbf{New Challenge and Solution:}
This paper introduces a new challenge, named \emph{possession failure}, which occurs when the generated image fails to correctly depict a prompt indicating a subject in possession of an object.
Thus, we broaden the horizons of the current research, which has mainly focused on the presence or absence of objects and attributes, to encompass actions.
The fact that Predicated Diffusion can successfully address this new challenge is worthy of attention.

\section{Related Work}
\paragraph{Diffusion Models for Image Generation}
A diffusion model was proposed as a parameterized Markov chain~\citep{Sohl-Dickstein2015,Ho2020}.
Taking a given image $x$ as the initial state $x_0$, the forward process $q(x_{t+1}|x_t)$ adds noise to the state $x_t$ from time $t=0$ to $T$.
The model learns the reverse process $p(x_{t-1}|x_t)$ and thereby the data distribution $p(x)=p(x_0)$.
Intuitively speaking, it repeatedly denoises images to be more realistic.
The reverse process resembles a discretized stochastic differential equation, akin to the Langevin dynamics, which ascends the gradient of the log-probability, $\nabla_x \log p(x)$, where $\nabla_x$ denotes the gradient with respect to image $x$~\citep{Song2020}.

A diffusion model can learn the conditional probability $p(x|c)$.
The condition $c$ might be text, images, or other annotations~\citep{Ramesh2021,Rombach2022}.
One of the leading models, Stable Diffusion, employs the cross-attention mechanism for conditioning~\citep{Vaswani2017}.
A convolutional neural network (CNN), U-Net~\citep{Ronneberger2015}, transforms the image $x$ into an intermediate representation.
For text conditions, a text encoder, based on CLIP~\citep{Radford2021a}, transforms text prompt $c$ into a sequence of intermediate representations, each linked to a word $w$ within the prompt $c$.
Given these representations, the cross-attention mechanism creates an attention map $A_w$ for each word $w$.
U-Net then updates the image $x$ using these maps as weights.
Technically, these processes target not the image $x$ but the latent variable $z$ extracted by a variational autoencoder~\citep{Kingma2014}.

Despite its sophistication, Stable Diffusion sometimes fails to capture the intended meaning of the text prompt, as discussed in the Introduction.
To address this, Structure Diffusion feeds segmented text prompts to the text encoder to emphasize each clause~\citep{Feng2023}.

\paragraph{Training-Free Guidance}
Even when a diffusion model is designed without condition $c$, it can reproduce the conditional probability $p(x|c)$ without retraining.
This is because, from a diffusion model $p(x)$ and a separate classifier $p(c|x)$ for class label $c$, one can obtain the gradient of the conditional log-probability, $\nabla_x \log p(x|c)=\nabla_x \log p(c|x)+\nabla_x \log p(x)$.
Although grounded in probability theory, what it practically offers is additional guidance $\nabla_x \log p(c|x)$ for updating images, which is generalized as \emph{classifier guidance}~\citep{Dhariwal2021}.

Consider the reverse process modeled as a Gaussian distribution $p(x_{t-1}|x_t)=\mathcal N(x_{t-1}|\mu_\theta(x_t,t),\Sigma_\theta(x_t,t))$, where the parameters are determined by neural networks $\mu_\theta$ and $\Sigma_\theta$.
A guidance method can be defined as a method that introduces an adjustment $g(x_t,t)$ to the update in the reverse process as $\mathcal N(x_{t-1}|\mu_\theta(x_t,t)+g(x_t,t),\Sigma_\theta(x_t,t))$.

When the diffusion model is conditioned on $c$, the difference between conditional and unconditional updates serves as \emph{classifier-free guidance}, which can adjust the fidelity of the generated image to condition $c$~\citep{Ho2021}.
\citet{Liu2022b} proposed Composable Diffusion, inspired by energy-based models~\citep{Du2020}.
It generates an image conditioned on two concepts, $c_0$ and $c_1$, by summing their respective conditional updates.
It negates or removes a concept $c_n$ from generated images by subtracting the update conditioned on $c_n$, termed as a \emph{negative prompt}.
Some studies developed guidance using annotations, such as bounding boxes~\citep{Xie2023,Ma2023,Mao2023} and segmentation masks~\citep{Park2023}.
While effective in intentionally controlling image layout, guidance methods based on annotations sometimes limit the diversity of generated images.

\paragraph{Attention Guidance}
Other previous studies developed guidance methods using the attention maps of the cross-attention mechanism, termed as \emph{attention guidance}.
High pixel intensity in an attention map $A_w$ suggests the presence of the corresponding object or concept $w$ at that pixel.
Attend-and-Excite enhances the intensity of at least one pixel in the attention map $A_w$ to ensure the existence of the corresponding object $w$ (that is, address missing objects)~\citep{Chefer2023}.
SynGen equalizes the intensity distributions for related nouns and adjectives, while differentiating others, thus addressing attribute leakage~\citep{Rassin2023}.
While these methods are based on deep insights, they lack comprehensive theoretical justification and generality.

\begin{table}[t]
    \centering
    \tabcolsep=1.5mm
    \footnotesize
    \caption{Propositions and Attention Map.}
    \label{tab:logic}
    \begin{tabular}{ll}
        \toprule
        \textbf{Proposition}    & \textbf{Attention Map}                         \\
        \midrule
        $\mathrm{true}$         & 1                                              \\
        $\mathrm{false}$        & 0                                              \\
        $P(x)$                  & $\att{P}[i]$                                   \\
        $\neg P(x)$             & $1-\att{P}[i]$                                 \\
        $P(x) \land Q(x)$       & $\att{P}[i]\times \att{Q}[i]$                  \\
        $P(x) \rightarrow Q(x)$ & $1-\att{P}[i]\times (1-\att{Q}[i])$            \\
        $P(x) \lor Q(y)$        & $1 - (1 - \att{P}[i]) \times (1 - \att{Q}[j])$ \\
        $\forall x.\,P(x)$      & $\prod_i \att{P}[i]$                           \\
        $\exists x.\,P(x)$      & $1-\prod_i (1-\att{P}[i])$                     \\
        \bottomrule
    \end{tabular}
\end{table}

\section{Method}\label{sec:loss}
\subsection{Predicated Diffusion}
\paragraph{Predicate Logic}
First-order predicate logic is a formal language for expressing knowledge~\citep{Genesereth1987}.
Variables like $x$ and $y$ denote unspecified objects.
Predicates like $P$ and $Q$ indicate properties or relationships between objects.
Using variables and predicates, we can express logical statements that define object properties.
For example, the proposition $P(x)$ represents the statement ``$x$ has property $P$.''
If the predicate $P$ indicates the property ``being a dog,'' the proposition $P(x)$ represents the statement ``$x$ is a dog.''
The existential quantifier, denoted by $\exists$, declares the existence of objects satisfying a given property.
Thus, the proposition $\exists x.\,P(x)$ asserts the existence of at least one object $x$ that satisfies the predicate $P$, representing that ``There is a dog.''

\paragraph{Fuzzy Logic in Attention Map}
Let $\att{P}[i]\in[0,1]$ denote the intensity of the $i$-th pixel in the attention map $\att{P}$ corresponding to a word $P$.
Here, we treat the intensity $\att{P}[i]$ as a continuous version of a proposition $P(x)$.
Specifically, we employ the product fuzzy logic and its operations (the strong conjunction, strong negation, and material implication), summarized in Table~\ref{tab:logic}~\citep{Hajek1998,Prokopowicz2017}.
$\att{P}[i]=1$ indicates that the proposition $P(x)$ holds, whereas $\att{P}[i]=0$ implies that it does not.
$1-\att{P}[i]$ indicates the the negation of the proposition, $\neg P(x)$.
Given another proposition $Q(x)$, we consider the conjunction $Q(x)\land P(x)$ to correspond to the product $\att{Q}[i]\times \att{P}[i]$ in the attention maps.
The implication satisfies $P(x)\rightarrow Q(x)=\neg (P(x)\land \neg Q(x))$, which corresponds to $1-\att{P}[i]\times (1-\att{Q}[i])$.
The disjunction $Q(x)\lor P(x)$ is equivalent to $\neg(\neg Q(x) \land \neg P(x))$ and corresponds to $1-(1-\att{Q}[i])\times(1-\att{P}[i])$.
The universal quantifier $\forall$ asserts that a predicate holds for all objects.
$\forall x.\,P(x)=\land_x P(x)$ corresponds to $\prod_i \att{P}[i]$.
Using this, the existential proposition $\exists x.\,P(x)$ can be re-expressed as $\neg(\forall x.\,\neg P(x))$, corresponding to $1-\prod_i (1-\att{P}[i])$.

\paragraph{Predicated Diffusion}
For simplicity, we will treat italicized words as predicates.
For instance, we will use $\prop{Dog}(x)$ to represent the statement ``$x$ is a dog'' rather than $P(x)$.
We represent the prompt ``There is a dog'' by the proposition $\exists x.\,\prop{Dog}(x)$.
Then, we expect that $1-\prod_i (1-\att{\prop{Dog}}[i])=1$.
To encourage this, we consider its negative logarithm,
\begin{equation}
    \textstyle \mathcal{L}[\exists x.\,\prop{Dog}(x)]= -\log (1-\prod_i(1-\att{\prop{Dog}}[i])),\label{eq:loss_existence}
\end{equation}
and adopt it as the loss function.
Minimizing it makes the intensity of at least one pixel approach 1, ensuring the existence of a dog.
This loss function is inspired by the negative log-likelihood for Bernoulli random variables.

Here, we propose an attention guidance, \emph{Predicated Diffusion}.
Given a proposition $R$ to hold, Predicated Diffusion converts it to an equation of the attention map intensity following Table~\ref{tab:logic}, takes its negative logarithm, uses it as a loss function $\mathcal L[R]$, and integrates it into the reverse process as the guidance term $g(x_t,t)=-\nabla_{x_t}\mathcal L[R]$.
The guidance term decreases the loss function $\mathcal L[R]$ and guides the image toward fulfilling the proposition $R$.
A visual representation is found in Fig.~\ref{fig:framework}.
We provide an overview of prompts and their corresponding loss functions in Table~\ref{tab:logic_loss}.

We develop this idea into the modification by adjectives.
For a prompt such as ``There is a black dog,'' it can be decomposed into: ``There is a dog,'' and ``The dog is black.''
The former statement has been discussed above.
We represent the latter by the proposition $\forall x.\,\prop{Dog}(x)\rightarrow\prop{Black}(x)$.
Then, the loss function is
\begin{equation}\label{eq:loss_adjective}
    \begin{aligned}
        \textstyle \mathcal{L}[\forall x.\,\prop{Dog}(x)\rightarrow\prop{Black}(x)] \hspace*{-16mm}         \\
         & \hspace*{-16mm}\textstyle = -\sum_i\log (1-\att{\prop{Dog}}[i]\times (1-\att{\prop{Black}}[i])).
    \end{aligned}
\end{equation}
Intuitively, this loss function guides all pixels depicting a dog towards a black hue; however, its purpose is not to render the dog entirely in solid black.
We employed product fuzzy logic to imply a tendency rather than enforce a strict property.

\begin{table}[t]
    \centering
    \tabcolsep=1.5mm
    \footnotesize
    \caption{Statements that Predicated Diffusion Can Express.}
    \label{tab:logic_loss}
    \begin{tabular}{lll}
        \toprule
        \textbf{Statements}       & \textbf{Example Prompts}    & \textbf{Loss}                  \\
        \midrule
        Existence                 & There is a dog              & \eqref{eq:loss_existence}      \\
        Modification              & A black dog                 & \eqref{eq:loss_adjective}      \\
        Concurrent existence      & There are a dog and a cat   & \eqref{eq:loss_concurrent}     \\
        One-to-one correspondence & A black dog and a white cat & \eqref{eq:loss_one-to-one}     \\
        Possession                & A man holding a bag         & \eqref{eq:loss_possession}     \\
        Multi-color               & A green and grey bird       & \eqref{eq:loss_multiadjective} \\
        Negation                  & without snow                & \eqref{eq:loss_negation}       \\
        \bottomrule
    \end{tabular}
    \vspace*{-1mm}
\end{table}

\subsection{Addressing Challenges}

\paragraph{Concurrent Existence by Logical Conjunction}
When a text prompt specifies multiple objects, a frequent challenge is the disappearance of one of the objects, known as missing objects.
We address this challenge using Predicated Diffusion.
Take, for example, the prompt ``There are a dog and a cat.''
This prompt can be decomposed into two statements: ``There is a dog,'' and ``There is a cat.''
As noted above, each statement can be represented by a proposition using the existential quantifier.
Since two propositions can be combined using logical conjunction, the original prompt is represented by the proposition $(\exists x.\,\prop{Dog}(x)) \land (\exists x.\,\prop{Cat}(x))$.
The corresponding loss function is
\begin{equation}\label{eq:loss_concurrent}
    \begin{aligned}
        \textstyle \mathcal{L}[(\exists x.\,\prop{Dog}(x)) \land (\exists x.\,\prop{Cat}(x))] \hspace*{-13mm}         \\
         & \hspace*{-13mm}\textstyle = \mathcal{L}[\exists x.\,\prop{Dog}(x)]+\mathcal{L}[\exists x.\,\prop{Cat}(x)].
    \end{aligned}
\end{equation}
Minimizing this loss function encourages the concurrent existence of both a dog and a cat.

\paragraph{One-to-One Correspondence}
When a prompt includes multiple adjectives and nouns, diffusion models often struggle with correct correspondence, leading to the challenge referred to as attribute leakage.
For instance, with the prompt ``a black dog and a white cat,'' leakage could result in the generation of a white dog or a black cat.
To prevent such leakage using Predicated Diffusion, it is essential to deduce statements that are implicitly suggested by the original prompt.
Firstly, we can deduce the statement ``The dog is black'' and its the converse, ``The black object is a dog.''
The latter can be represented by the proposition $\forall x.\,\prop{Black}(x)\rightarrow\prop{Dog}(x)$.
When these two statements are combined, they can be represented using a biimplication: $\forall x.\,\prop{Dog}(x)\leftrightarrow\prop{Black}(x)$.
This leads to the loss function:
\begin{equation}
    \begin{aligned}
         & \hspace*{-10mm} \textstyle \mathcal{L}[\forall x.\,\prop{Dog}(x)\!\leftrightarrow\!\prop{Black}(x)]                                                                                                \\
         & \hspace*{-10mm} \textstyle=\!\mathcal{L}[\forall x.\,\prop{Dog}(x)\!\!\rightarrow\!\!\prop{Black}(x)\!\land\!\forall x.\,\prop{Black}(x)\!\!\rightarrow\!\!\prop{Dog}(x)]          \hspace*{-10mm} \\
         & \hspace*{-10mm} \textstyle=\!\mathcal{L}[\forall x.\,\prop{Dog}(x)\!\!\rightarrow\!\!\prop{Black}(x)]\!+\!\mathcal{L}[\forall x.\,\prop{Black}(x)\!\!\rightarrow\!\!\prop{Dog}(x)] \hspace*{-10mm}
    \end{aligned}
\end{equation}
Next, we can deduce the negative statement ``The dog is not white,'' represented by $\forall x.\,\prop{Dog}(x)\rightarrow\neg \prop{White}(x)$.
A similar deduction applies to the white cat.
Hence, the comprehensive loss function for the original prompt is:
\begin{equation}\label{eq:loss_one-to-one}
    \begin{aligned}
        \textstyle \mathcal{L}_{\mathrm{one-to-one}}
         & \textstyle =\mathcal{L}[\forall x.\,\prop{Dog}(x)\leftrightarrow\prop{Black}(x)]            \\
         & \textstyle\ + \mathcal{L}[\forall x.\,\prop{Cat}(x)\leftrightarrow\prop{White}(x)]          \\
         & \textstyle\ + \alpha\mathcal{L}[\forall x.\,\prop{Dog}(x)\rightarrow\neg \prop{White}(x)]   \\
         & \textstyle\ + \alpha \mathcal{L}[\forall x.\,\prop{Cat}(x)\rightarrow\neg \prop{Black}(x)],
    \end{aligned}
\end{equation}
where the hyperparameter $\alpha\in[0,1]$ adjusts the weight of the negative statements.
To further ensure the existence of objects, the loss function \eqref{eq:loss_concurrent} can also be applied.

\paragraph{Possession by Logical Implication}
Given the prompt ``a man holding a bag,'' diffusion models often depict the bag as not being held by the man.
We refer to this challenge as possession failure.
To address this using Predicated Diffusion, we propose the proposition $\forall x.\,\prop{Bag}(x)\rightarrow\prop{Man}(x)$, leading to the loss function:
\begin{equation}\label{eq:loss_possession}
    \begin{aligned}
        \textstyle \mathcal{L}[\forall x.\,\prop{Bag}(x)\rightarrow\prop{Man}(x)] \hspace*{-15mm}          \\
         & \hspace*{-15mm} \textstyle = -\sum_i\log (1-\att{\prop{Bag}}[i]\times (1-\att{\prop{Man}}[i])).
    \end{aligned}
\end{equation}
The loss function aims to depict the bag as part of the man.
Because attention maps typically have lower resolution than the original image, minor displacements between objects and their possessors are acceptable, eliminating the need for complete overlaps.
This approach applies not just to \textit{holding} but also to any words indicating possession, such as \textit{having}, \textit{grasping}, and \textit{wearing}.

\subsection{Comparisons, Extensions, and Limitations}
Several studies have introduced loss functions or quality measures for machine learning methods by drawing inspiration from fuzzy logic~\citep{Hu2016,Diligenti2017,Mordido2021,Giunchiglia2022,Marra2023}.
In this context, Predicated Diffusion is the first method to establish the correspondence between the attention map and the predicates.

The propositions and corresponding loss functions can be adapted to a variety of scenarios, including, but not limited to, the concurrent existence of more than two objects, a single object modified by multiple adjectives, the combination of one-to-one correspondence and possession, and the negation of existence, modifications, and possessions, as we will show in the following sections.
The definition of conjunction allows us to simply sum up loss functions for all propositions.
These propositions can be formulated manually by users, extracted automatically from prompts using a syntactic dependency parser, or derived from additional data sources such as scene graphs~\citep{Feng2023}.

Predicated Diffusion inherits some of the general limitations of diffusion models, including challenges like the inability to count objects accurately and the tendency for bias toward more typical examples.
Due to this limitation, the loss function \eqref{eq:loss_one-to-one} is inappropriate when multiple instances of the same noun are modified by different adjectives (e.g., ``a black dog and a white dog'').

The (weak) conjunction of \Godel\ fuzzy logic and the product fuzzy logic is achieved by the minimum operation~\citep{Hajek1998,Prokopowicz2017}.
If we employ this operation and define the loss function by taking the negative instead of the negative logarithm, the proposition asserting the concurrent existence, $(\exists x.\,\prop{Dog}(x)) \land (\exists x.\,\prop{Cat}(x))$, leads to the loss function $\max(1-\max_i \att{Dog}[i],1-\max_i \att{Cat}[i])$.
This is equivalent to the one used for Attend-and-Excite~\citep{Chefer2023}.
This comparison suggests that our approach considers Attend-and-Excite as \Godel\ fuzzy logic, replaces the underlying logic with the product fuzzy logic, and broadens the scope of target propositions.
Similar to the loss function \eqref{eq:loss_one-to-one}, SynGen equalizes the attention map intensities for related nouns and adjectives~\citep{Rassin2023}.
SynGen additionally differentiates those for all word pairs except for the adjective-noun pairs.
In contrast, the loss function \eqref{eq:loss_one-to-one} differentiates those for only specific pairs which could trigger attribute leakage based on inferred propositions, thereby preventing the disruption of the harmony, as shown in the following section.

\section{Experiments and Results}
\subsection{Experimental Settings}
We implemented Predicated Diffusion by adapting the official implementation of Attend-and-Excite~\citep{Chefer2023}\footnote{\url{https://github.com/yuval-alaluf/Attend-and-Excite} (MIT license)}.
The reverse process spans 50 steps; we applied the guidance of Predicated Diffusion only to the initial 25 steps, following \citep{Chefer2023,Rassin2023}.
See Appendix~\ref{appendix:implementation} for more details.
For comparison, we also prepared Composable Diffusion~\citep{Liu2022b}, Structure Diffusion~\citep{Feng2023}, and SynGen~\citep{Rassin2023}, in addition to Stable Diffusion and Attend-and-Excite.
All methods used the officially pretrained Stable Diffusion v1.4~\citep{Rombach2022}\footnote{\url{https://github.com/CompVis/stable-diffusion} (CreativeML Open RAIL-M)} as backbones.

We conducted four experiments for assessing each method's performance.
We provided each method with the same prompt and random seed, and then generated a set of images.
Human evaluators were tasked with the visual assessment of these generated images as follows.
\begin{figure*}[p]
    \captionof{table}{Results of Experiments \ref{ex:1} for Concurrent Existence and \ref{ex:2} for One-to-One Correspondence.}
    \label{tab:experiments12}
    \footnotesize
    \centering
    \tabcolsep=1.5mm
    \begin{tabular}{lccccccccc}
        \toprule
                                         & \multicolumn{4}{c}{Experiment \ref{ex:1} for Concurrent Existence} &
        \multicolumn{5}{c}{Experiment \ref{ex:2} for One-to-One Correspondence}                                                                                                                                                                                                                                                                                                                                                               \\
        \cmidrule(lr){2-5}\cmidrule(lr){6-10}
                                         & \multicolumn{2}{c}{\textbf{Human Evaluation}}                      & \multicolumn{2}{c}{\textbf{Automatic Evaluation}} & \multicolumn{3}{c}{\textbf{Human Evaluation}}   & \multicolumn{2}{c}{\textbf{Automatic Evaluation}}                                                                                                                                                                       \\
        \cmidrule(lr){2-3}\cmidrule(lr){4-5}\cmidrule(lr){6-8}\cmidrule(lr){9-10}
        \multirow{2}{*}{\textbf{Methods}} & {\textbf{Missing}$^\dagger$\ }                                     & \multirow{2}{*}{\textbf{Fidelity}}                & \multirow{2}{*}{\textbf{Similarity}$^\ddagger$} & \textbf{CLIP-}                                    & {\textbf{Missing}$^\dagger$\ } & {\textbf{Attribute}\ \ } & \multirow{2}{*}{\textbf{Fidelity}} & \multirow{2}{*}{\textbf{Similarity}$^\ddagger$} & \textbf{CLIP-}   \\
                                         & {\ \textbf{Objects}}                                               &                                                   &                                                 & {\ \textbf{IQA}}                                  & {\ \textbf{Objects}}           & {\ \ \textbf{Leakage}}   &                                    &                                                 & {\ \textbf{IQA}} \\
        \midrule
        Stable Diffusion                 & 54.7         / 66.0                                                & 11.0                                              & 0.326 / 0.767                                   & 0.761                                             & 64.8 / 73.5                    & 88.5                     & \zz6.0                             & 0.345 / 0.744                                   & 0.756            \\
        Composable Diffusion             & 44.5         / 82.3                                                & \zz 2.5                                           & 0.317 / 0.739                                   & 0.764                                             & 49.3 / 83.5                    & 88.5                     & \zz3.8                             & 0.348 / 0.729                                   & 0.757            \\
        Structure Diffusion              & 56.0         / 64.5                                                & 12.0                                              & 0.325 / 0.763                                   & 0.763                                             & 64.3 / 69.5                    & 86.5                     & \zz5.8                             & 0.346 / 0.741                                   & 0.760            \\
        Attend-and-Excite                & 25.3         / 36.3                                                & 29.5                                              & 0.342 / 0.814                                   & 0.766                                             & 28.0 / 35.8                    & 64.5                     & 19.3                               & 0.367 / 0.792                                   & 0.761            \\
        SynGen                           & ---                                                                & ---                                               & ---     / ---                                   & --                                                & 23.3 / 29.3                    & 40.3                     & 36.8                               & 0.367 / 0.801                                   & 0.750            \\
        \midrule
        Predicated Diffusion             & \textbf{18.5}         / \textbf{28.5}                              & \textbf{30.3}                                     & \textbf{0.348} / \textbf{0.825}                 & \textbf{0.775}                                    & \textbf{10.0} / \textbf{16.5}  & \textbf{33.0}            & \textbf{44.8}                      & \textbf{0.379} / \textbf{0.811}                 & \textbf{0.769}   \\
        \bottomrule
        \multicolumn{8}{l}{$^\dagger$Using the lenient and strict criterions. $^\ddagger$Text-image similarity and text-text similarity.}
    \end{tabular}\\[1.5mm]
    \tabcolsep=0.2mm
    \scriptsize
    \centering
    \begin{tabular}{c@{\hspace*{1mm}}ccc@{\hspace*{2mm}}ccc@{\hspace*{2mm}}ccc}
                                                                                           &
        \multicolumn{3}{c}{\emph{a crown and a rabbit}}                                    &

        \multicolumn{3}{c}{\emph{a bird and a cat}}                                        &
        \multicolumn{3}{c}{\emph{an apple and a lion}}                                       \\
        \rotatebox[origin=l]{90}{\parbox{13mm}{\centering Stable Diffusion}}               &
        \includegraphics[width=14mm]{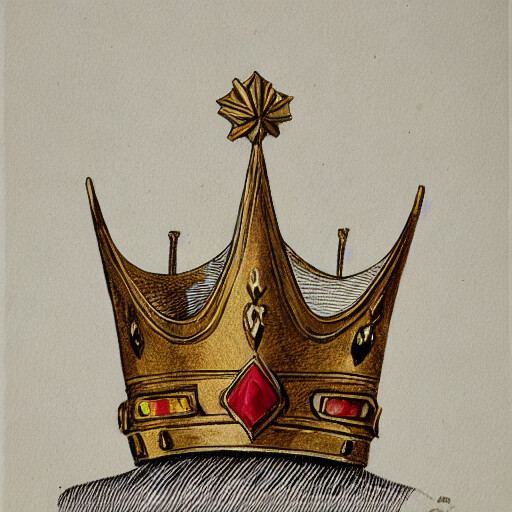}     &
        \includegraphics[width=14mm]{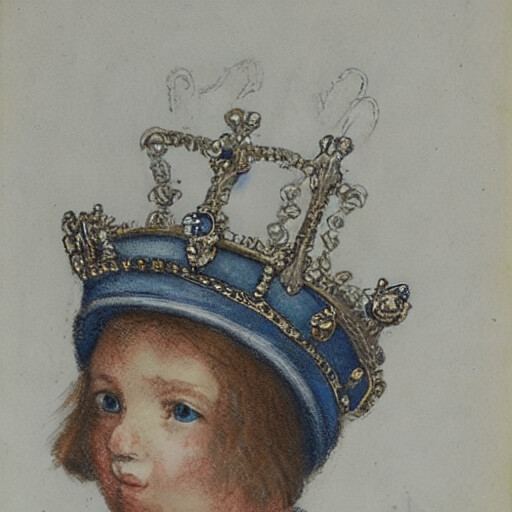}     &
        \includegraphics[width=14mm]{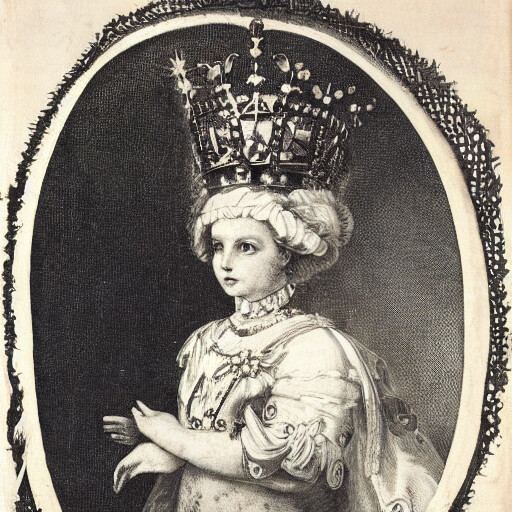}     &
        \includegraphics[width=14mm]{./fig/coexist/a_bird_and_a_cat_61_Stable.jpg}         &
        \includegraphics[width=14mm]{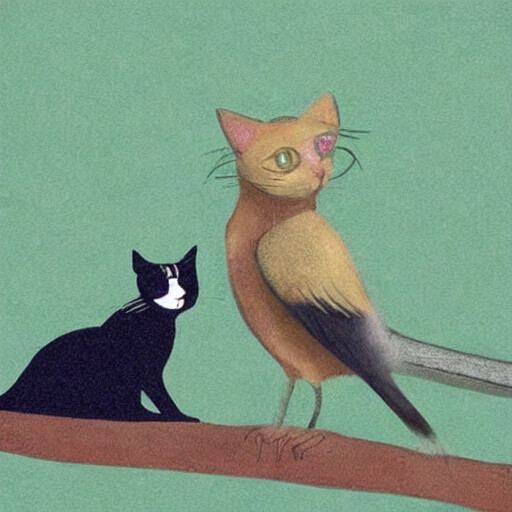}         &
        \includegraphics[width=14mm]{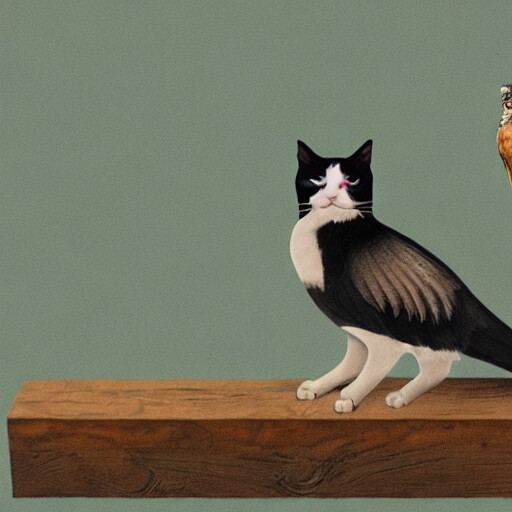}         &
        \includegraphics[width=14mm]{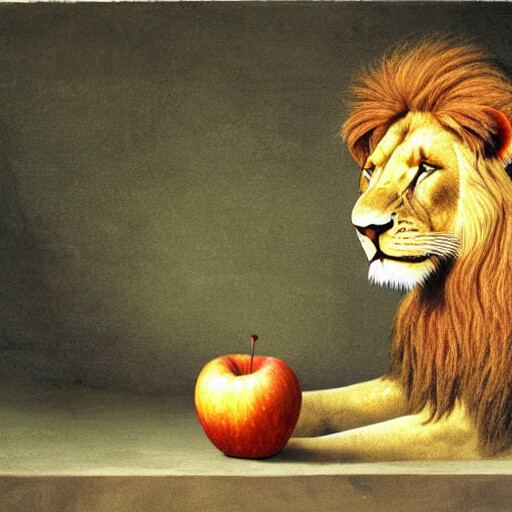}       &
        \includegraphics[width=14mm]{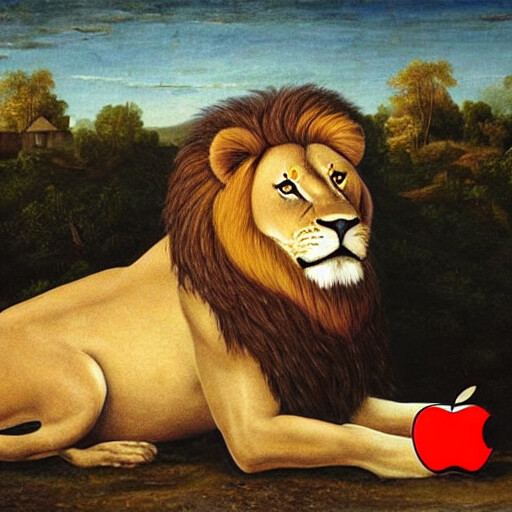}       &
        \includegraphics[width=14mm]{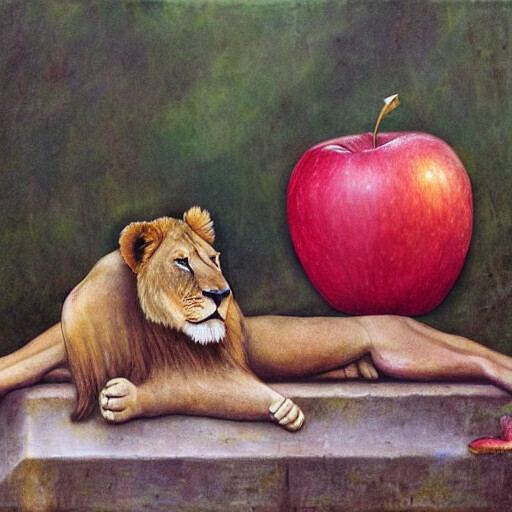}
        \\[-0.3mm]
        \rotatebox[origin=l]{90}{\parbox{13mm}{\centering Composable Diffusion}}           &
        \includegraphics[width=14mm]{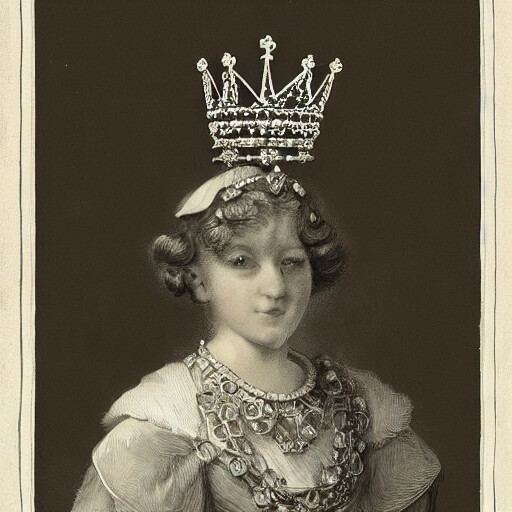} &
        \includegraphics[width=14mm]{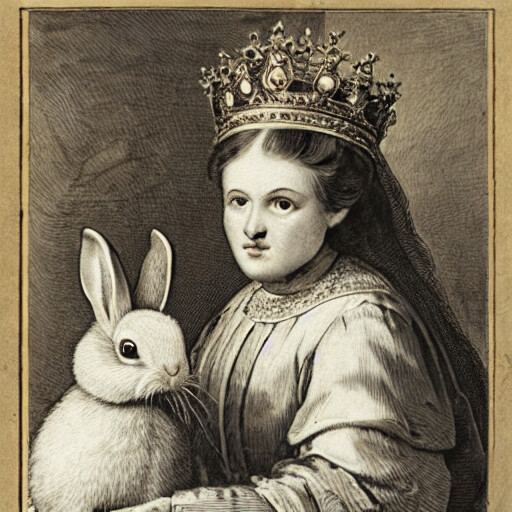} &
        \includegraphics[width=14mm]{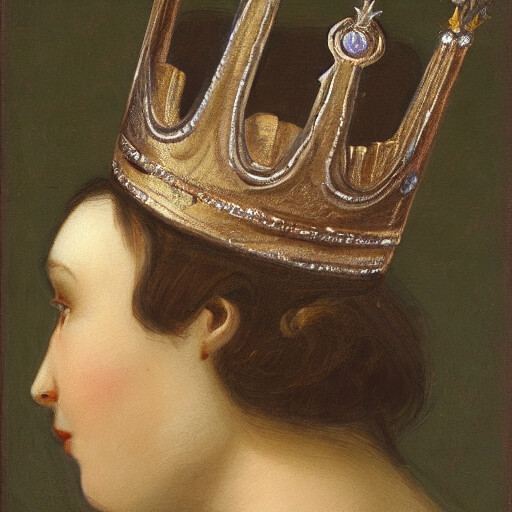} &
        \includegraphics[width=14mm]{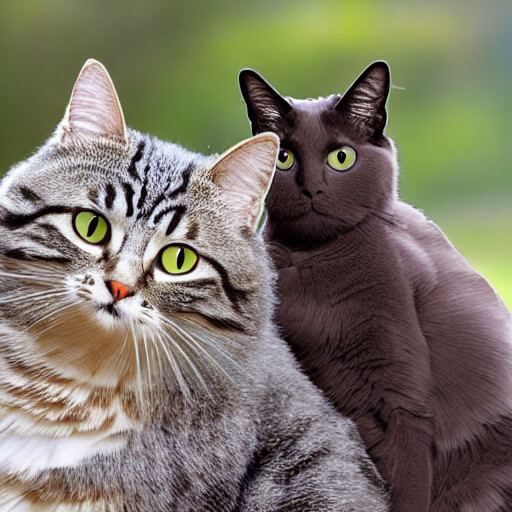}     &
        \includegraphics[width=14mm]{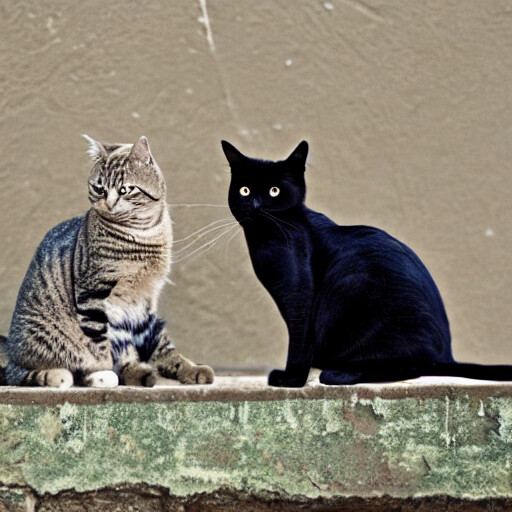}     &
        \includegraphics[width=14mm]{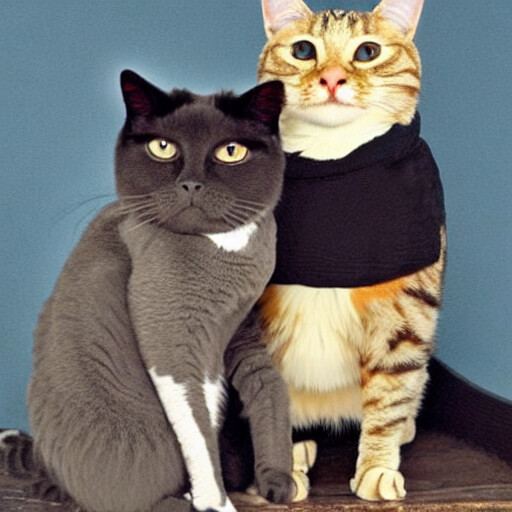}     &
        \includegraphics[width=14mm]{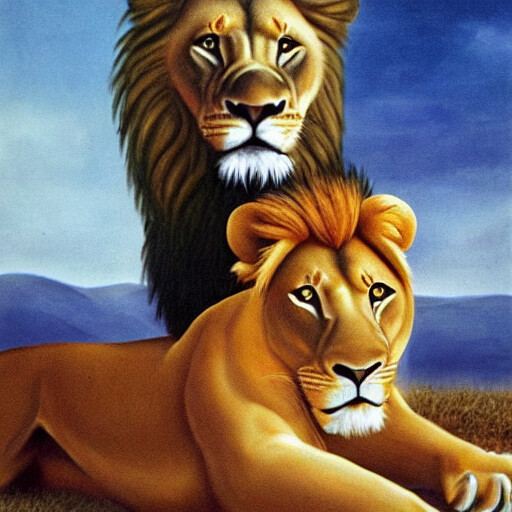}   &
        \includegraphics[width=14mm]{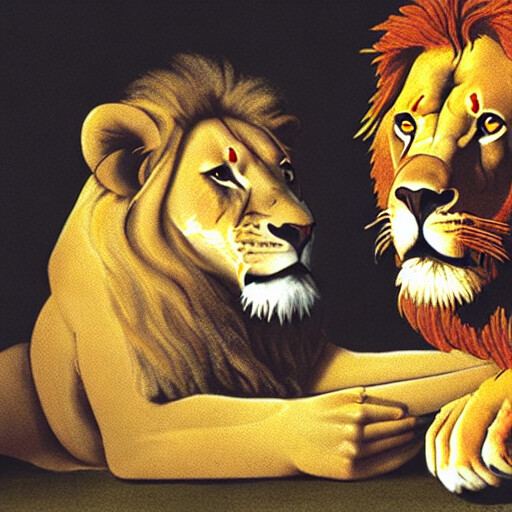}   &
        \includegraphics[width=14mm]{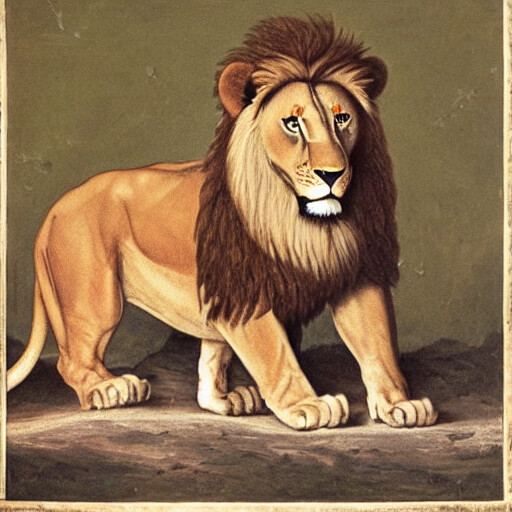}
        \\[-0.3mm]
        \rotatebox[origin=l]{90}{\parbox{13mm}{\centering Structure Diffusion}}            &
        \includegraphics[width=14mm]{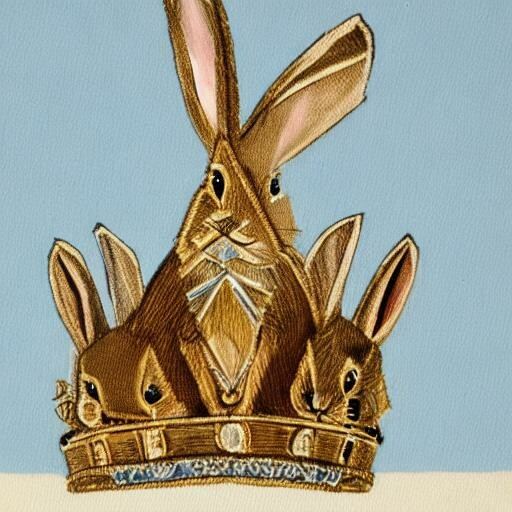} &
        \includegraphics[width=14mm]{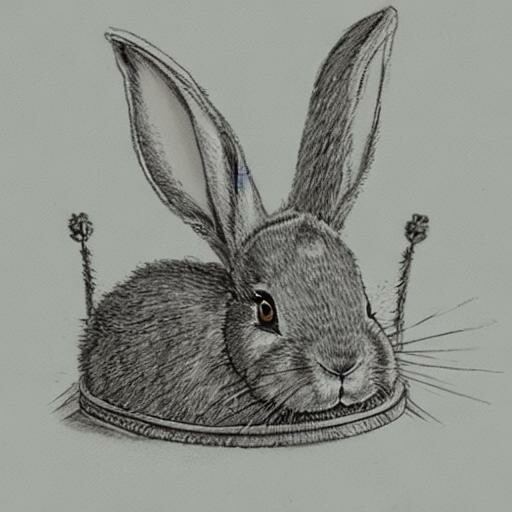} &
        \includegraphics[width=14mm]{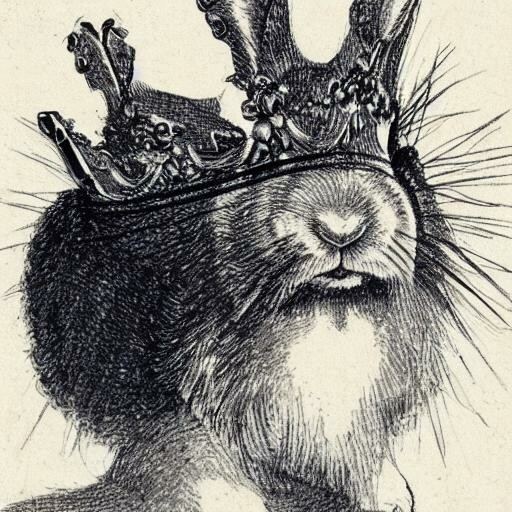} &
        \includegraphics[width=14mm]{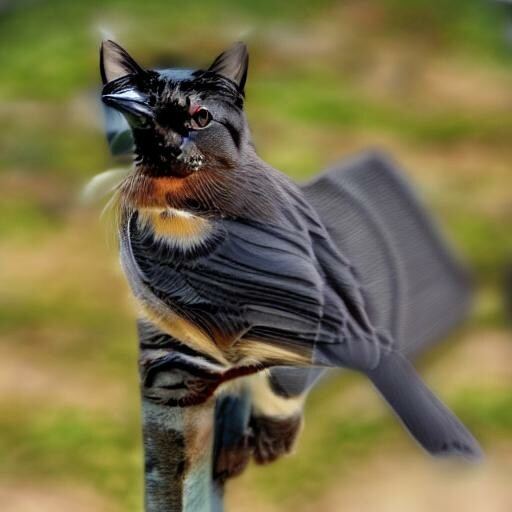}     &
        \includegraphics[width=14mm]{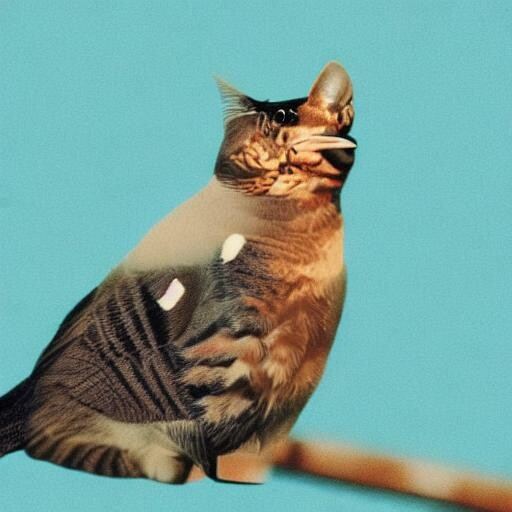}     &
        \includegraphics[width=14mm]{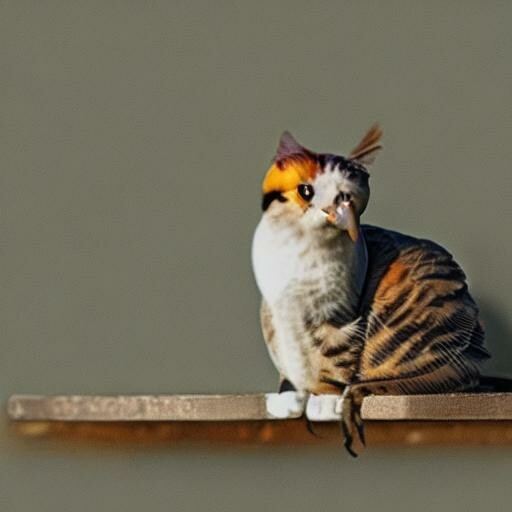}     &
        \includegraphics[width=14mm]{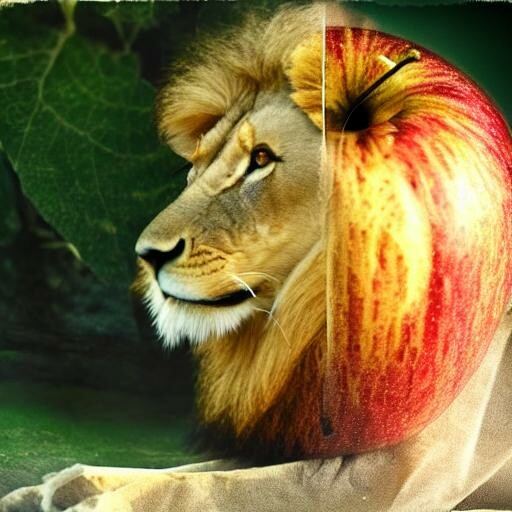}   &
        \includegraphics[width=14mm]{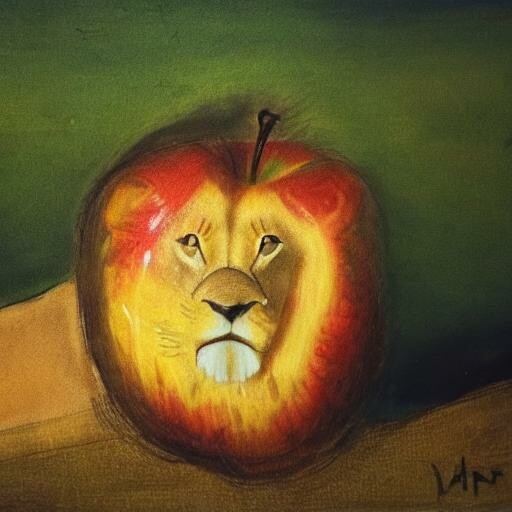}   &
        \includegraphics[width=14mm]{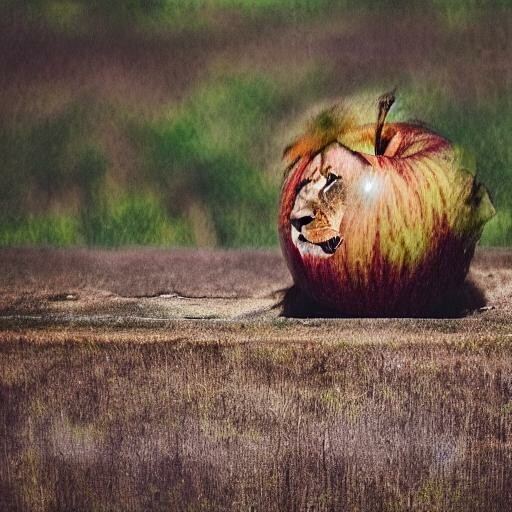}
        \\[-0.3mm]
        \rotatebox[origin=l]{90}{\parbox{13mm}{\centering Attend-and-Excite}}              &
        \includegraphics[width=14mm]{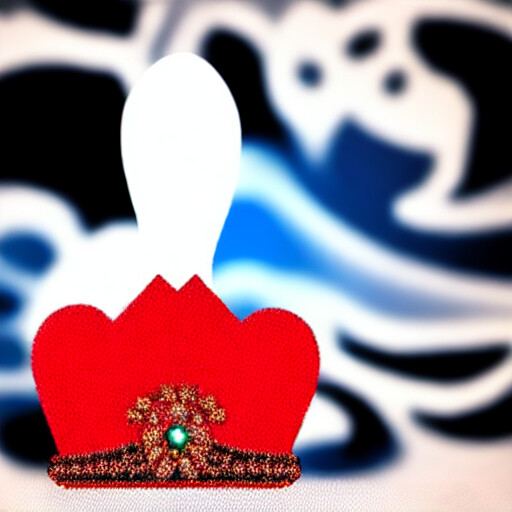}        &
        \includegraphics[width=14mm]{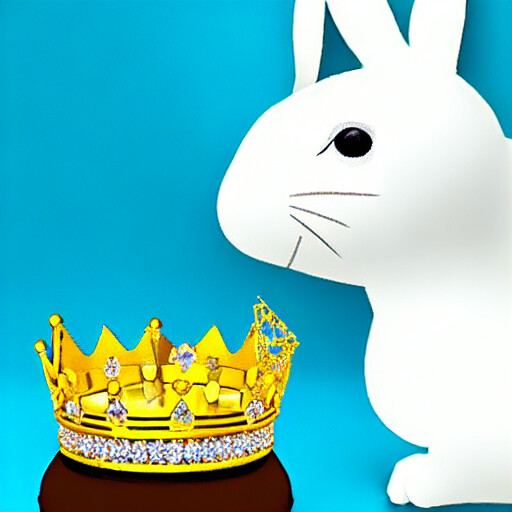}        &
        \includegraphics[width=14mm]{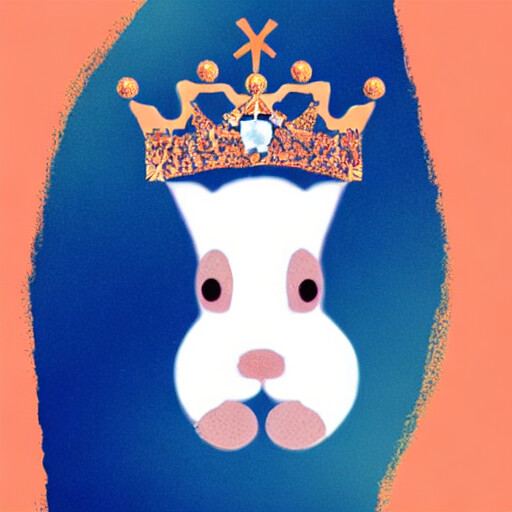}        &
        \includegraphics[width=14mm]{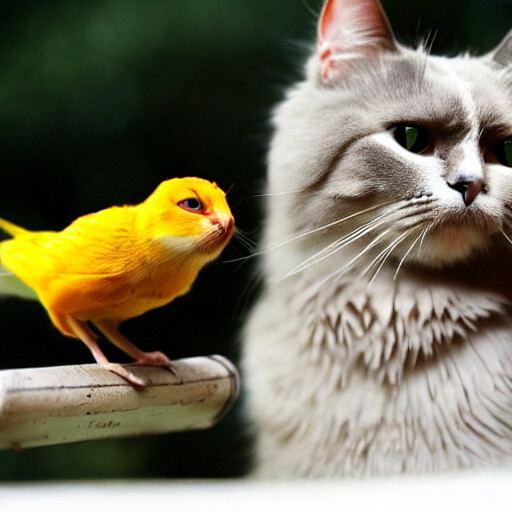}            &
        \includegraphics[width=14mm]{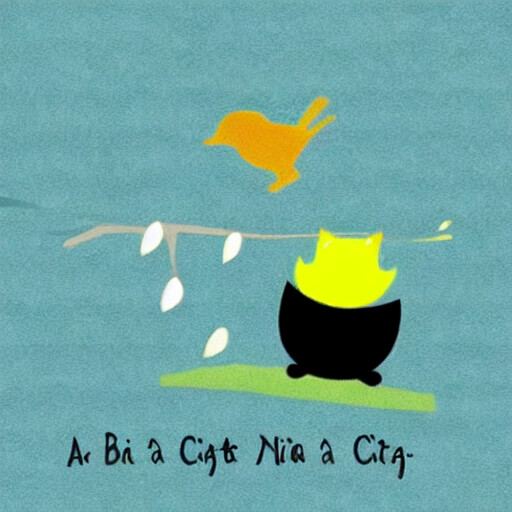}            &
        \includegraphics[width=14mm]{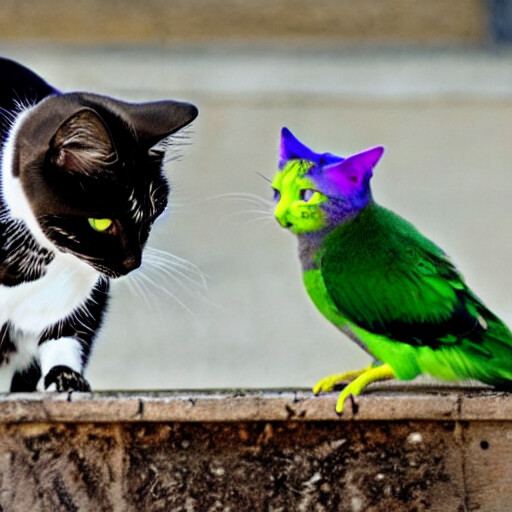}            &
        \includegraphics[width=14mm]{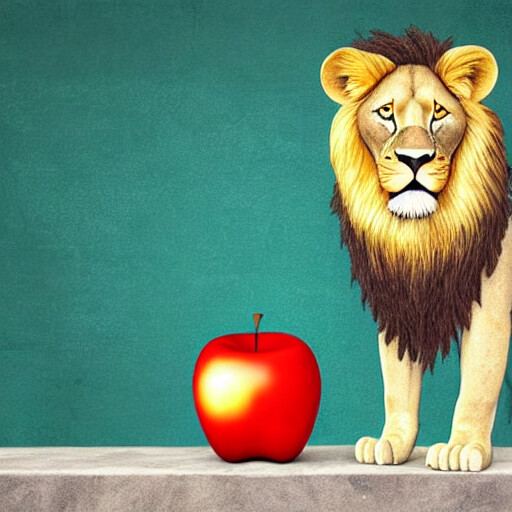}          &
        \includegraphics[width=14mm]{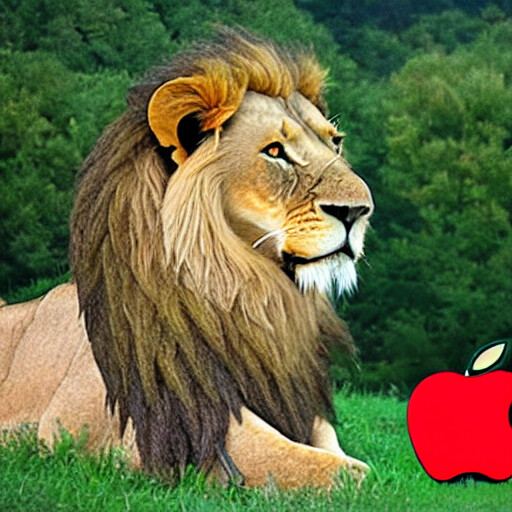}          &
        \includegraphics[width=14mm]{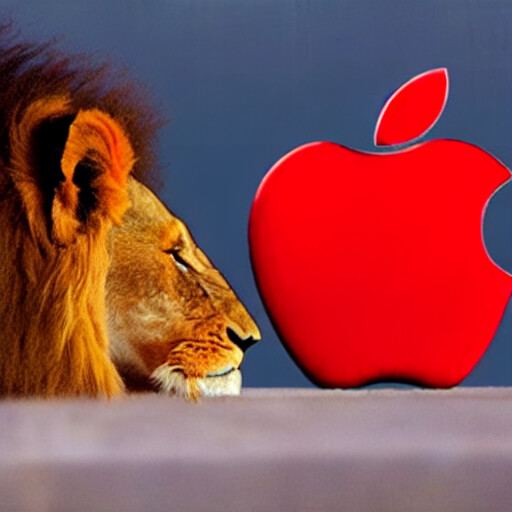}
        \\[-0.3mm]
        \rotatebox[origin=l]{90}{\parbox{13mm}{\centering Predicated Diffusion}}           &
        \includegraphics[width=14mm]{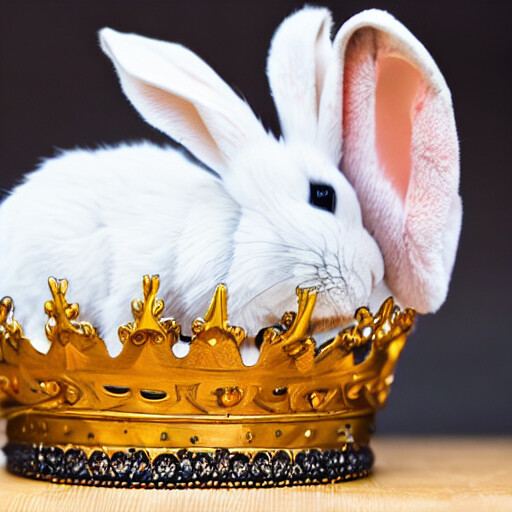} &
        \includegraphics[width=14mm]{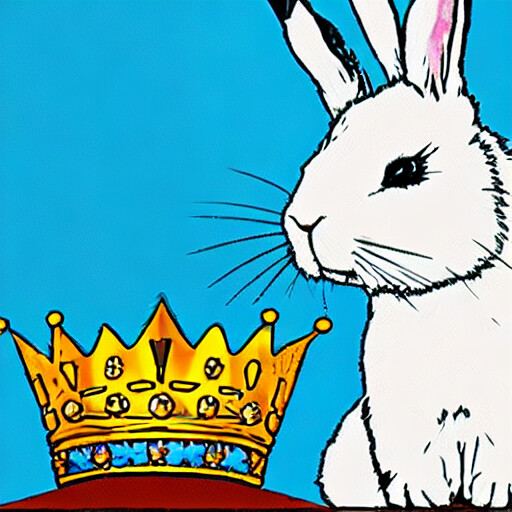} &
        \includegraphics[width=14mm]{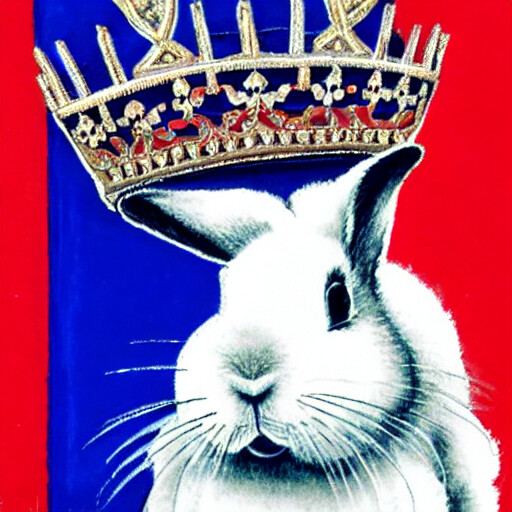} &
        \includegraphics[width=14mm]{./fig/coexist/a_bird_and_a_cat_61_Predicated.jpg}     &
        \includegraphics[width=14mm]{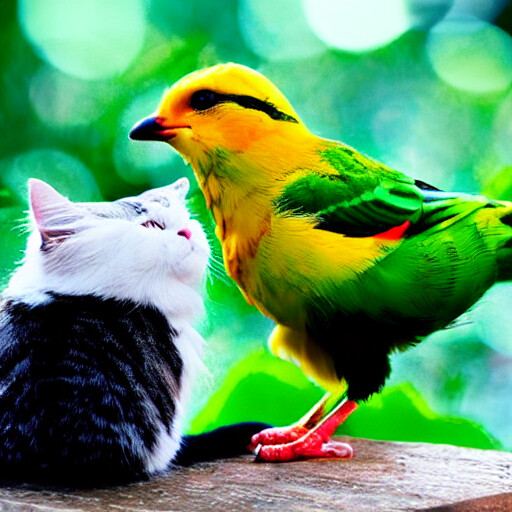}     &
        \includegraphics[width=14mm]{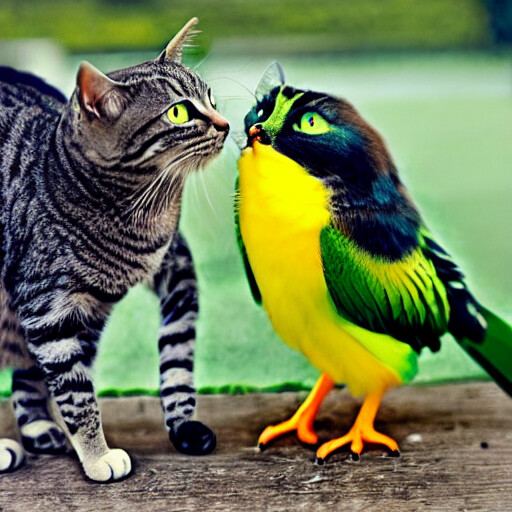}     &
        \includegraphics[width=14mm]{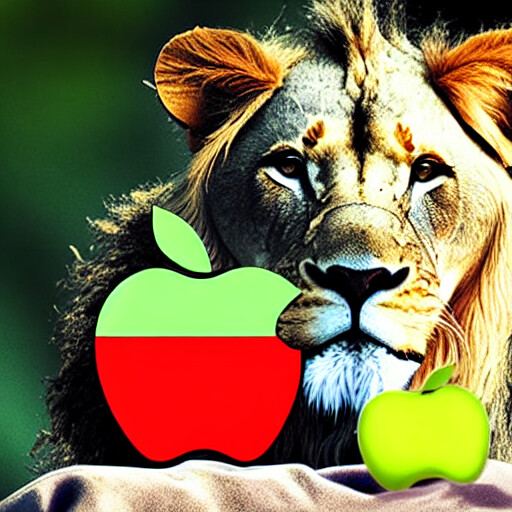}   &
        \includegraphics[width=14mm]{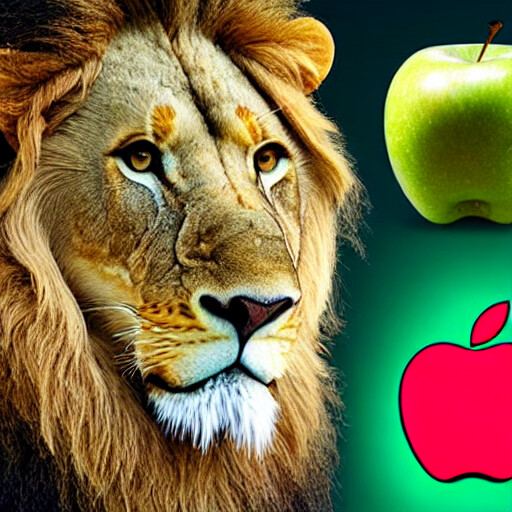}   &
        \includegraphics[width=14mm]{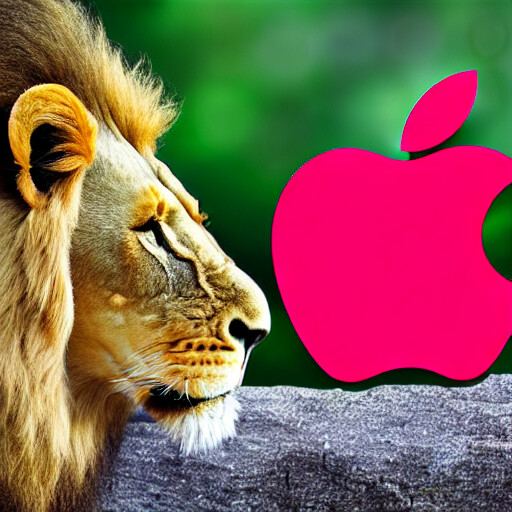}
        \\
    \end{tabular}
    \vspace{-3mm}
    \captionof{figure}{Example results of Experiment \ref{ex:1} for concurrent existence.
        See also Fig.~\ref{fig:experiment1_additional}.
    }\label{fig:experiment1}
    \vspace{1.5mm}
    \tabcolsep=0.2mm
    \scriptsize
    \centering
    \begin{tabular}{c@{\hspace*{1mm}}ccc@{\hspace*{2mm}}ccc@{\hspace*{2mm}}ccc}
                                                                                                               &
        \multicolumn{3}{c}{\emph{a yellow car and a blue bird}}                                                &
        \multicolumn{3}{c}{\emph{a green frog and a gray cat}}                                                 &
        \multicolumn{3}{c}{\emph{a green balloon and a purple clock}}
        \\
        \rotatebox[origin=l]{90}{\parbox{13mm}{\centering Stable Diffusion}}                                   &
        \includegraphics[width=14mm]{./fig/correspondence/a_yellow_car_and_a_blue_bird_3_Stable.jpg}           &
        \includegraphics[width=14mm]{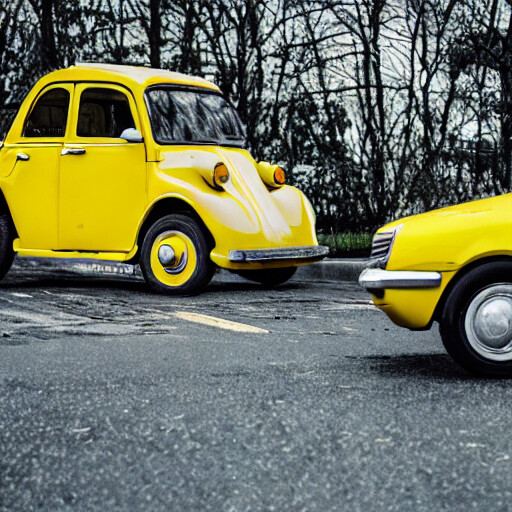}           &
        \includegraphics[width=14mm]{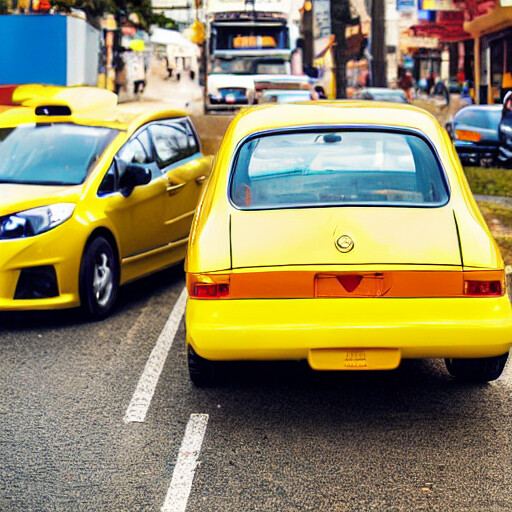}           &
        \includegraphics[width=14mm]{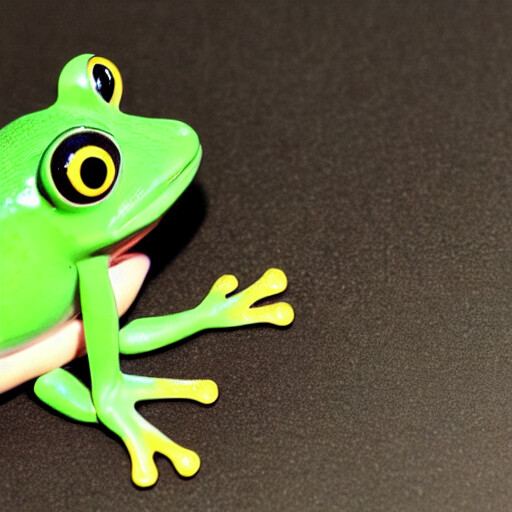}            &
        \includegraphics[width=14mm]{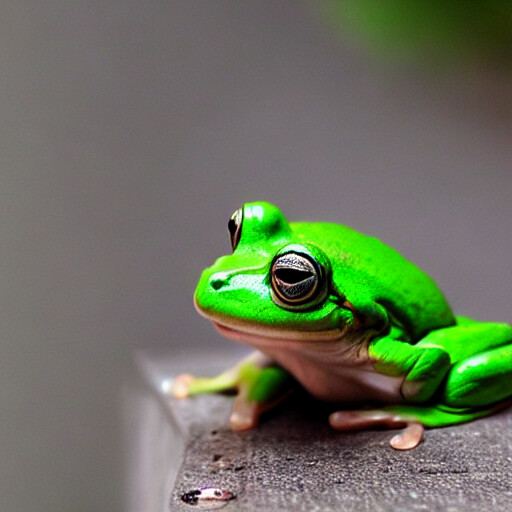}            &
        \includegraphics[width=14mm]{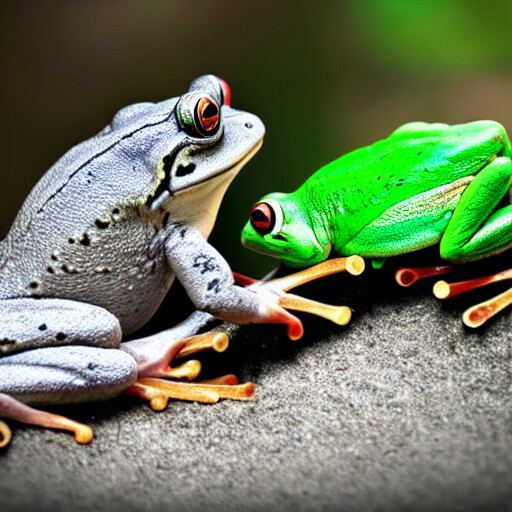}            &
        \includegraphics[width=14mm]{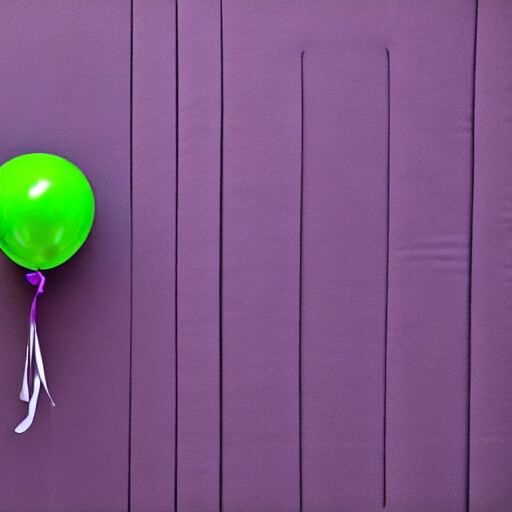}     &
        \includegraphics[width=14mm]{./fig/correspondence/a_green_balloon_and_a_purple_clock_6_Stable.jpg}     &
        \includegraphics[width=14mm]{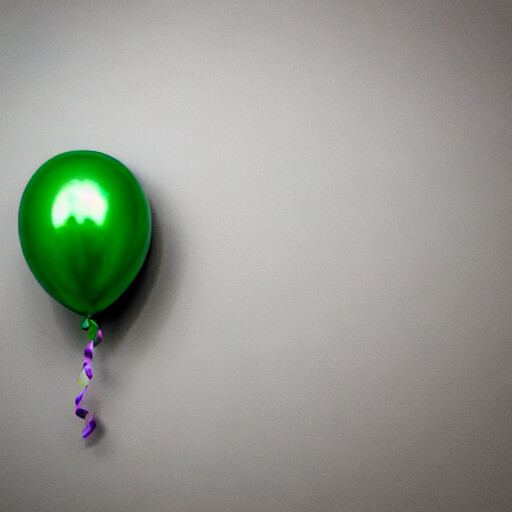}
        \\[-0.3mm]
        \rotatebox[origin=l]{90}{\parbox{13mm}{\centering Composable Diffusion}}                               &
        \includegraphics[width=14mm]{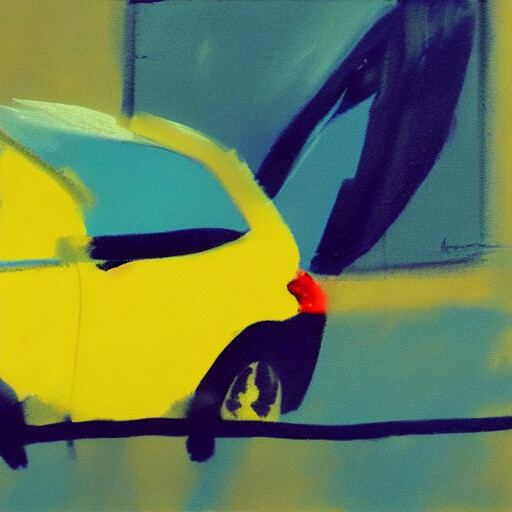}       &
        \includegraphics[width=14mm]{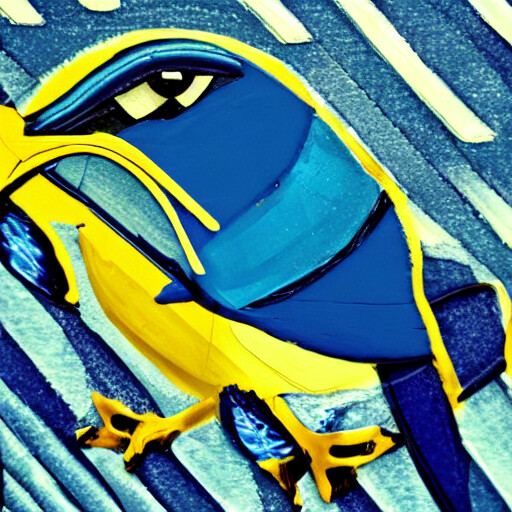}       &
        \includegraphics[width=14mm]{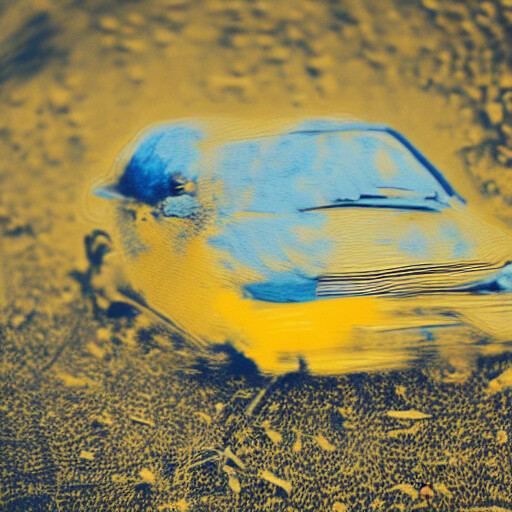}       &
        \includegraphics[width=14mm]{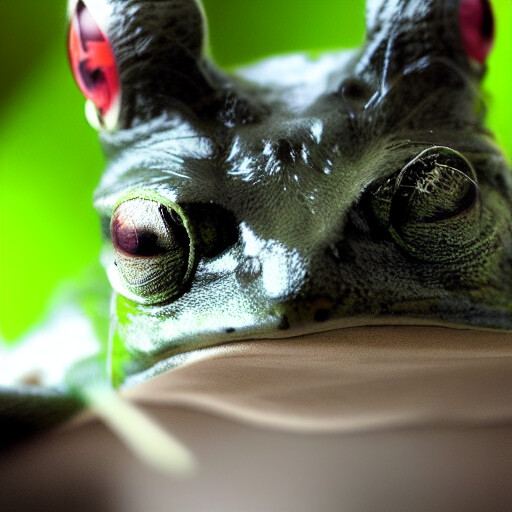}        &
        \includegraphics[width=14mm]{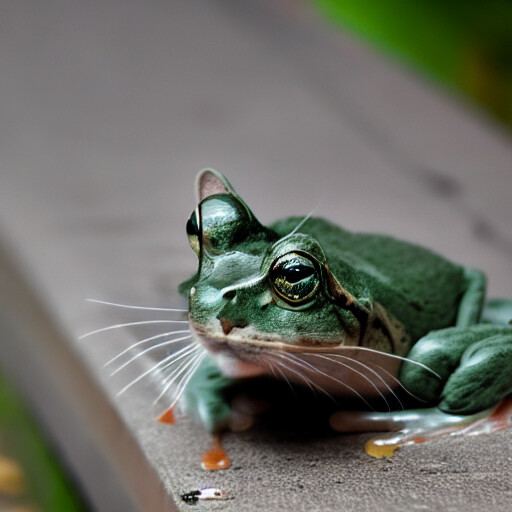}        &
        \includegraphics[width=14mm]{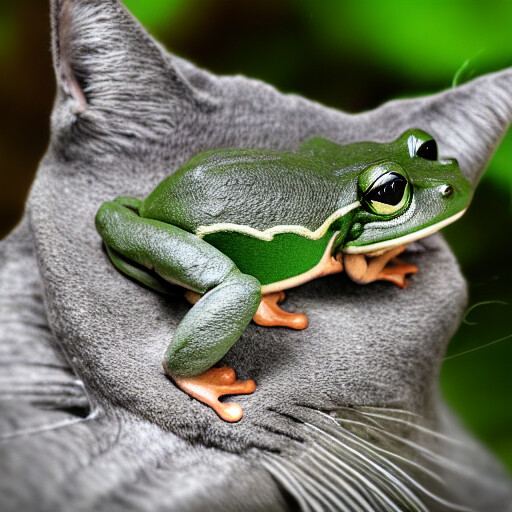}        &
        \includegraphics[width=14mm]{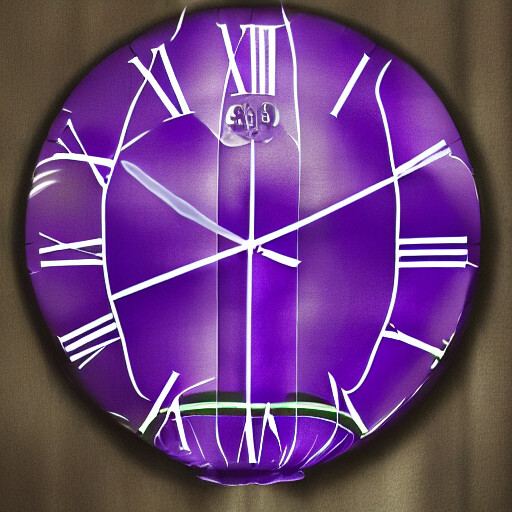} &
        \includegraphics[width=14mm]{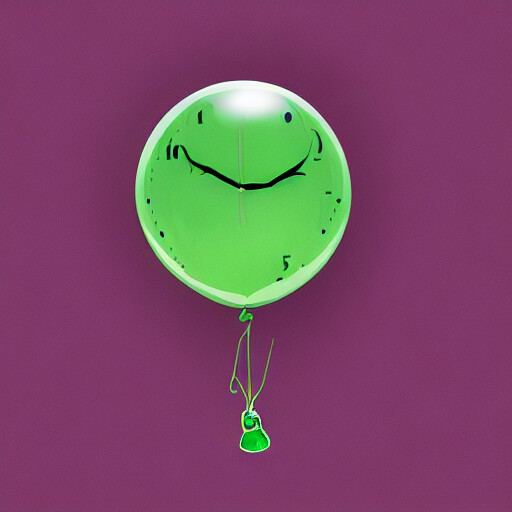} &
        \includegraphics[width=14mm]{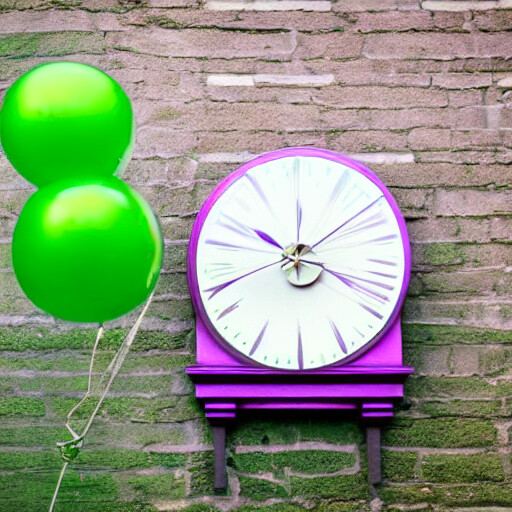}
        \\[-0.3mm]
        \rotatebox[origin=l]{90}{\parbox{13mm}{\centering Structure Diffusion}}                                &
        \includegraphics[width=14mm]{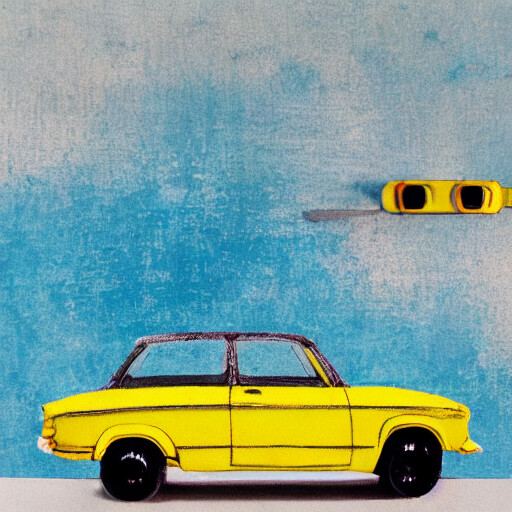}       &
        \includegraphics[width=14mm]{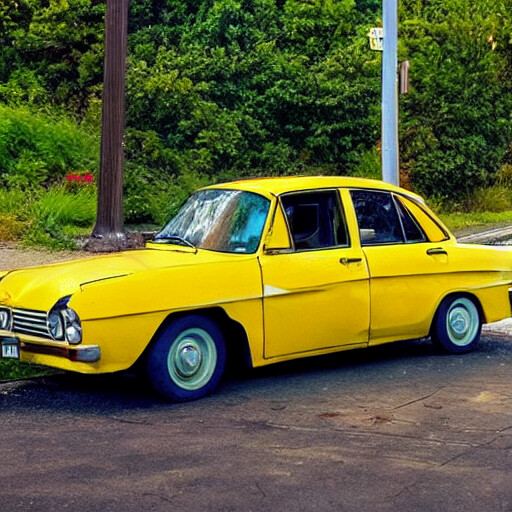}       &
        \includegraphics[width=14mm]{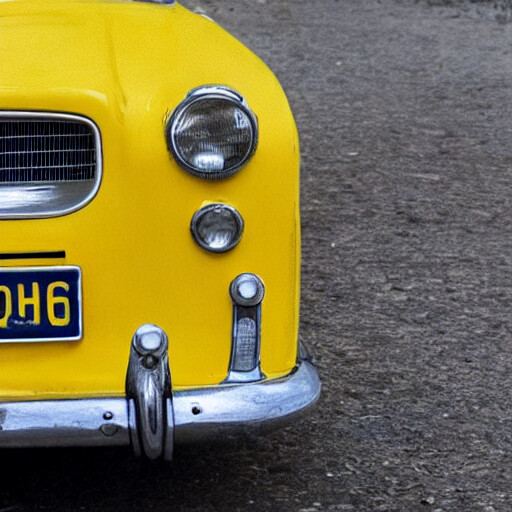}       &
        \includegraphics[width=14mm]{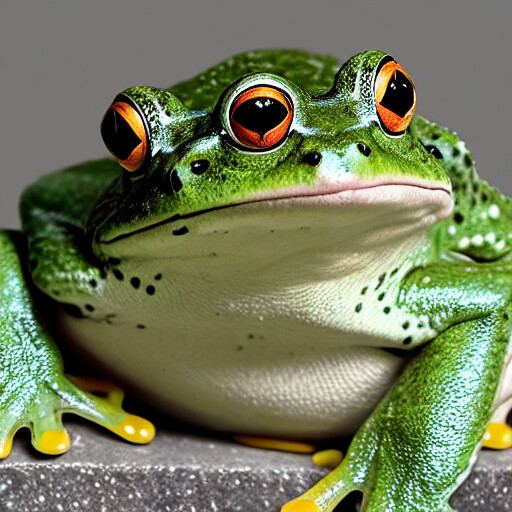}        &
        \includegraphics[width=14mm]{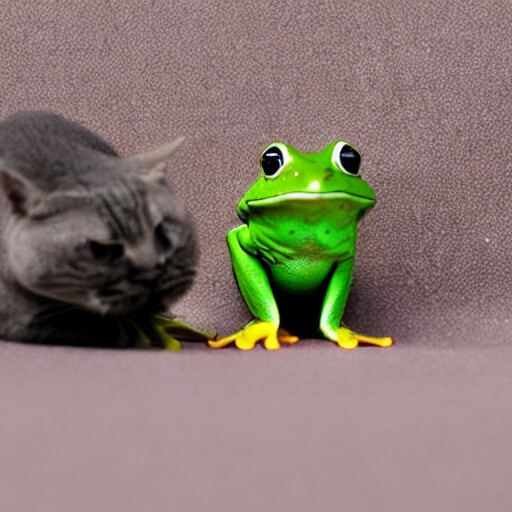}        &
        \includegraphics[width=14mm]{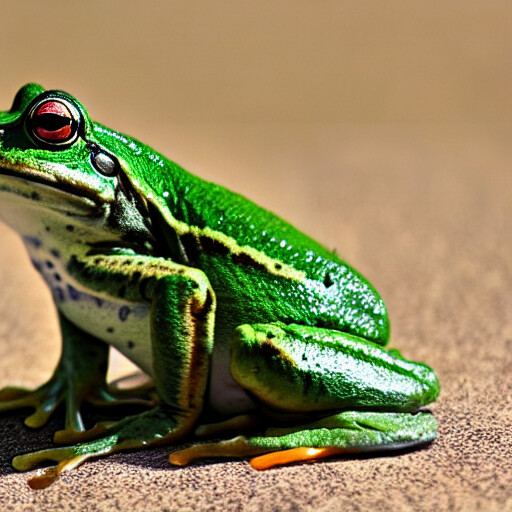}        &
        \includegraphics[width=14mm]{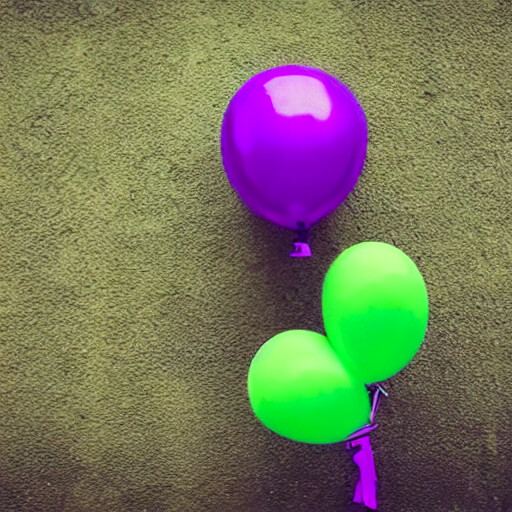} &
        \includegraphics[width=14mm]{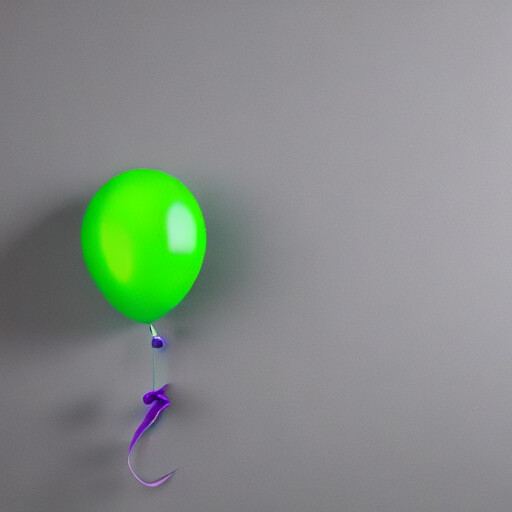} &
        \includegraphics[width=14mm]{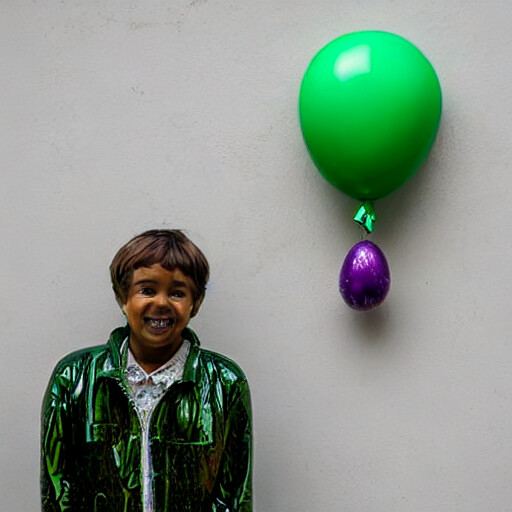}
        \\[-0.3mm]
        \rotatebox[origin=l]{90}{\parbox{13mm}{\centering Attend-and-Excite}}                                  &
        \includegraphics[width=14mm]{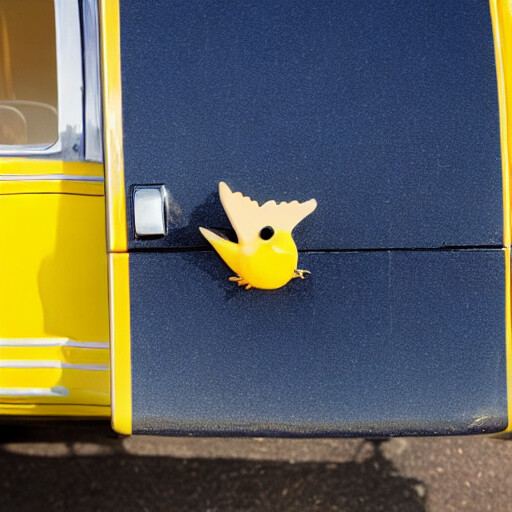}              &
        \includegraphics[width=14mm]{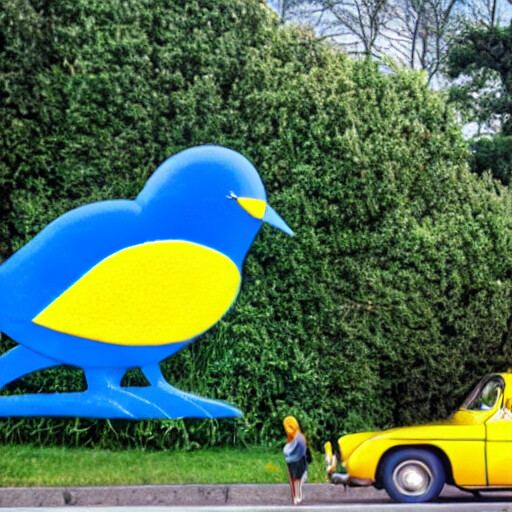}              &
        \includegraphics[width=14mm]{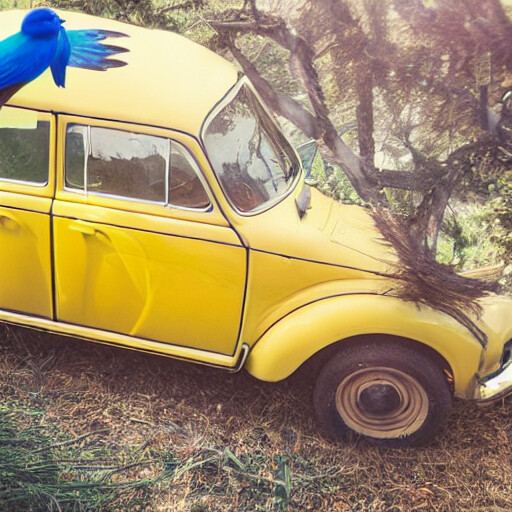}              &
        \includegraphics[width=14mm]{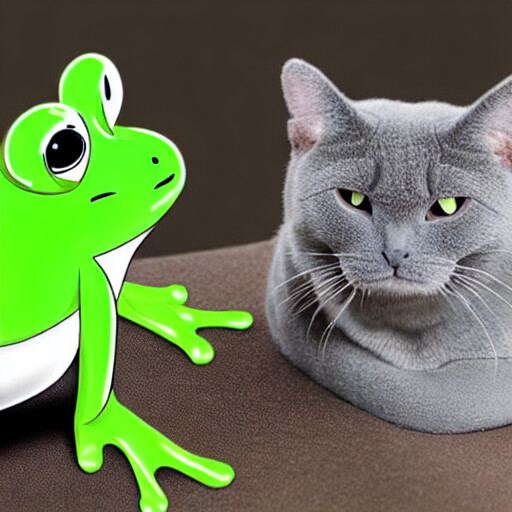}               &
        \includegraphics[width=14mm]{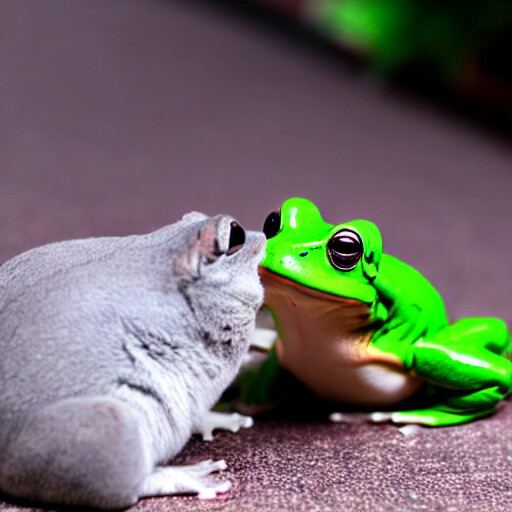}               &
        \includegraphics[width=14mm]{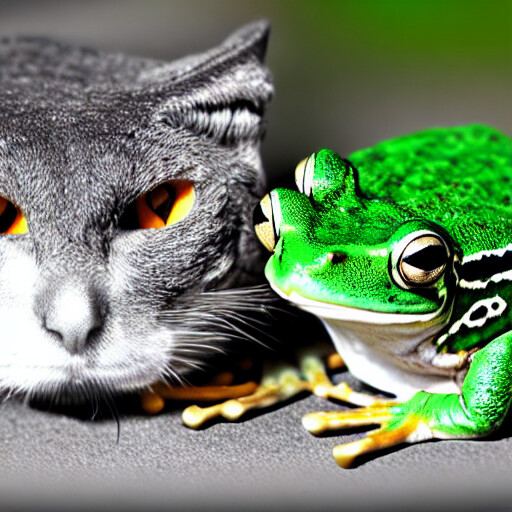}               &
        \includegraphics[width=14mm]{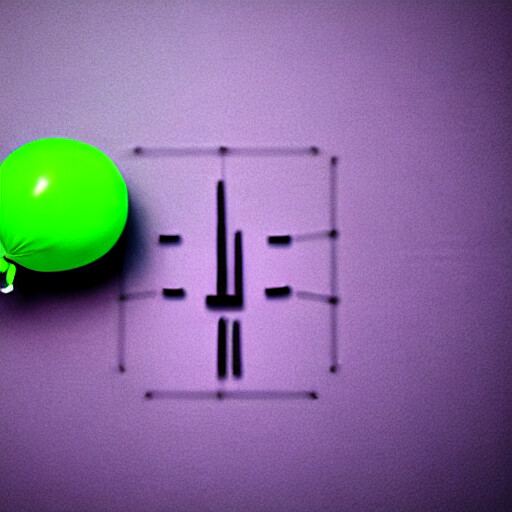}        &
        \includegraphics[width=14mm]{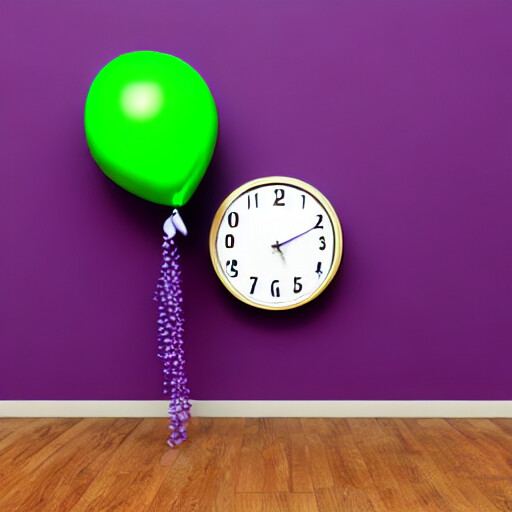}        &
        \includegraphics[width=14mm]{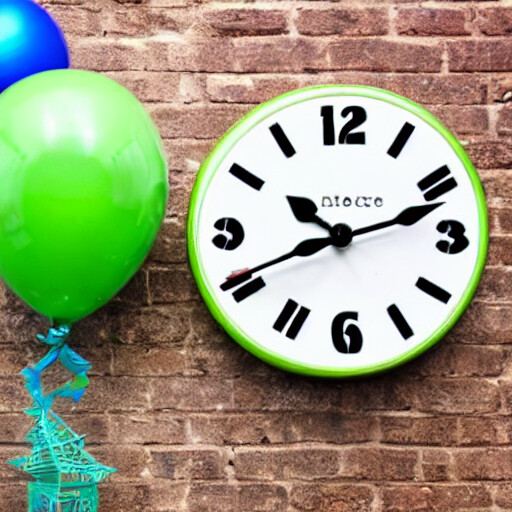}
        \\[-0.3mm]
        \rotatebox[origin=l]{90}{\parbox{13mm}{\centering SynGen}}                                             &
        \includegraphics[width=14mm]{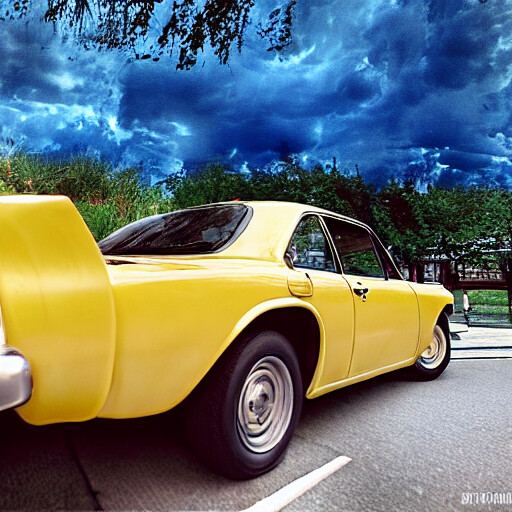}           &
        \includegraphics[width=14mm]{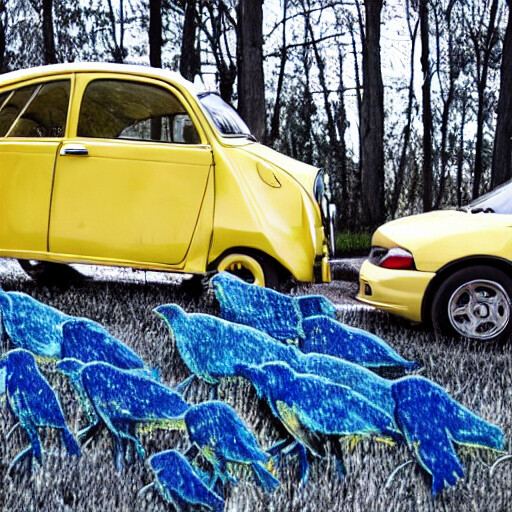}           &
        \includegraphics[width=14mm]{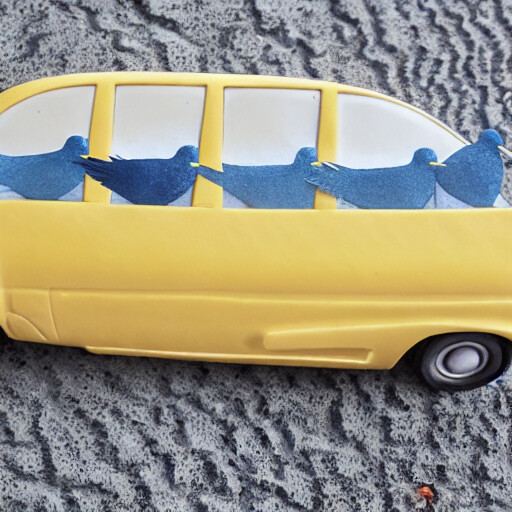}           &
        \includegraphics[width=14mm]{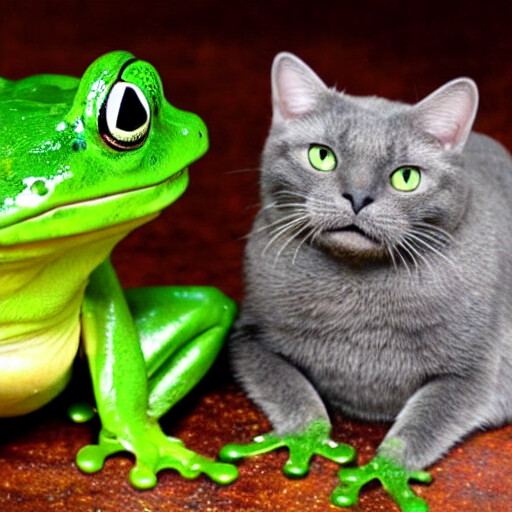}            &
        \includegraphics[width=14mm]{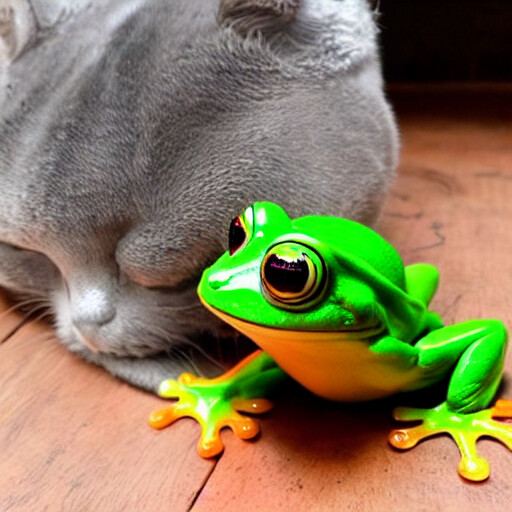}            &
        \includegraphics[width=14mm]{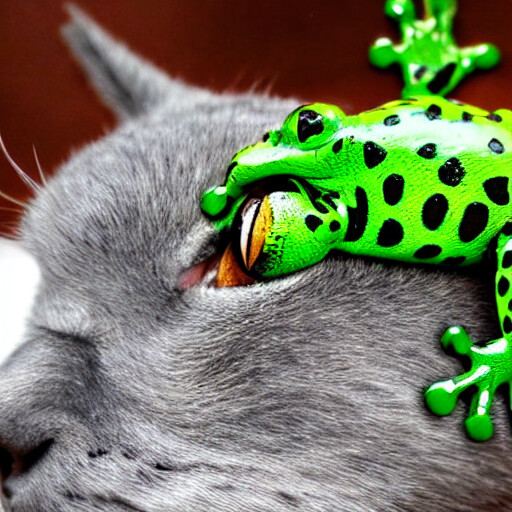}            &
        \includegraphics[width=14mm]{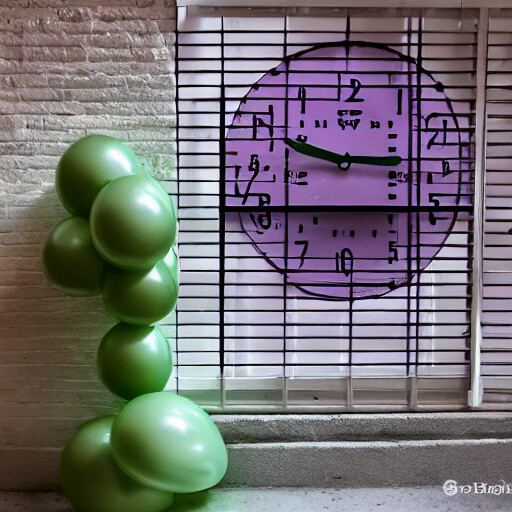}     &
        \includegraphics[width=14mm]{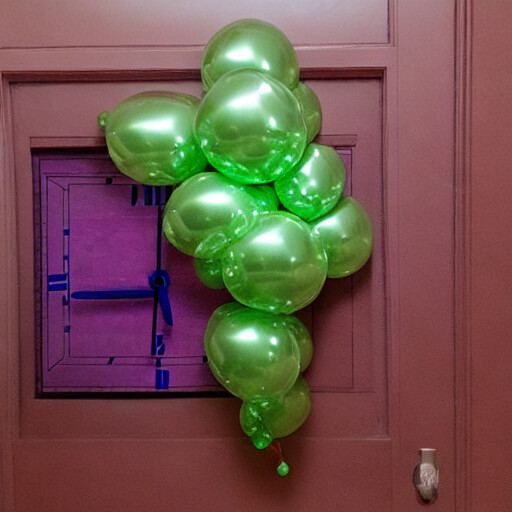}     &
        \includegraphics[width=14mm]{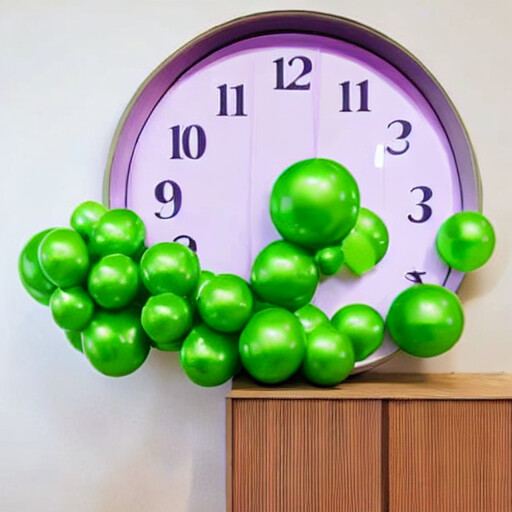}
        \\[-0.3mm]
        \rotatebox[origin=l]{90}{\parbox{13mm}{\centering Predicated Diffusion}}                               &
        \includegraphics[width=14mm]{./fig/correspondence/a_yellow_car_and_a_blue_bird_3_Predicated.jpg}       &
        \includegraphics[width=14mm]{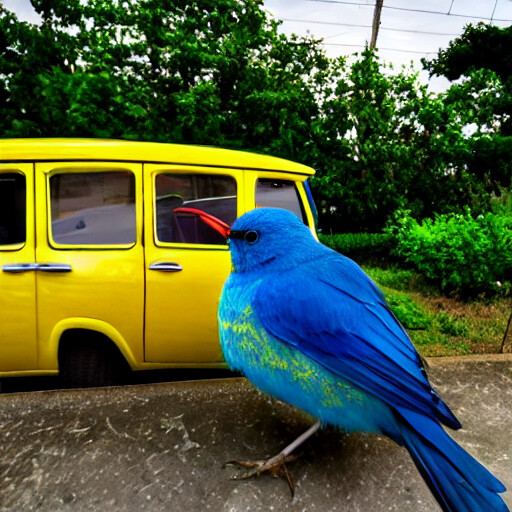}       &
        \includegraphics[width=14mm]{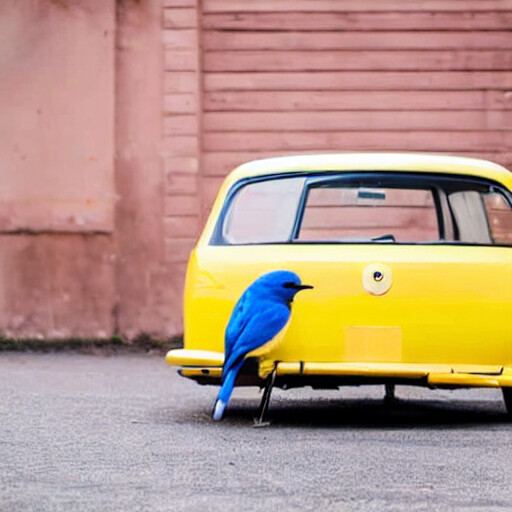}       &
        \includegraphics[width=14mm]{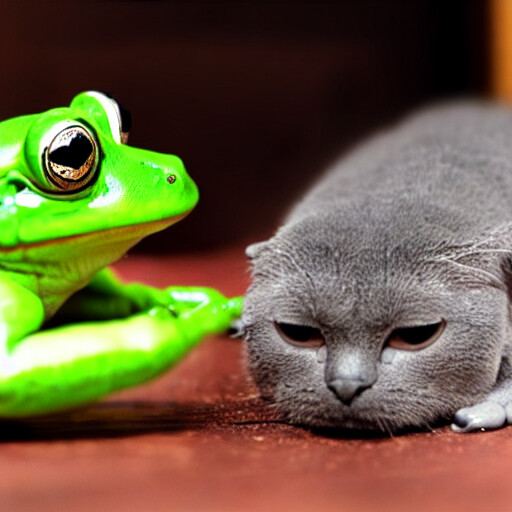}        &
        \includegraphics[width=14mm]{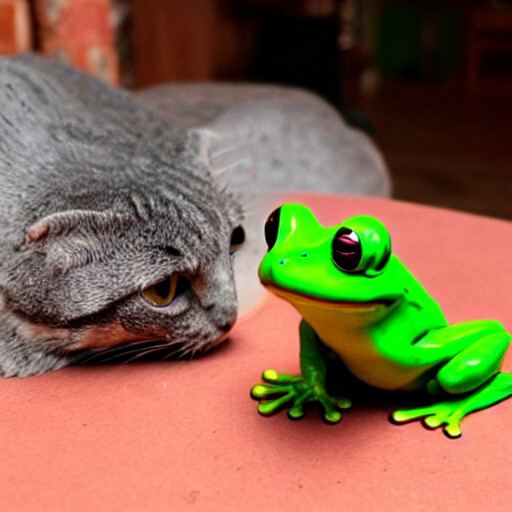}        &
        \includegraphics[width=14mm]{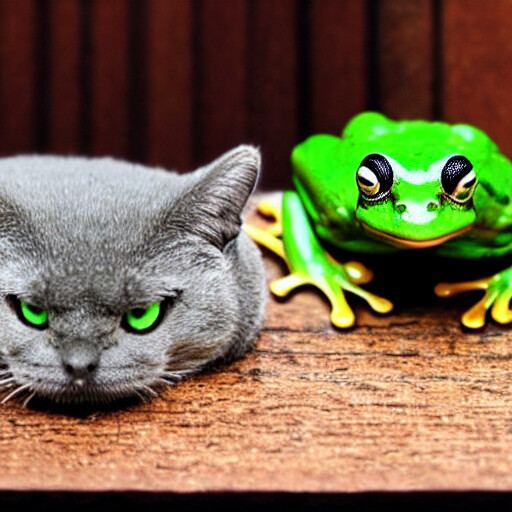}        &
        \includegraphics[width=14mm]{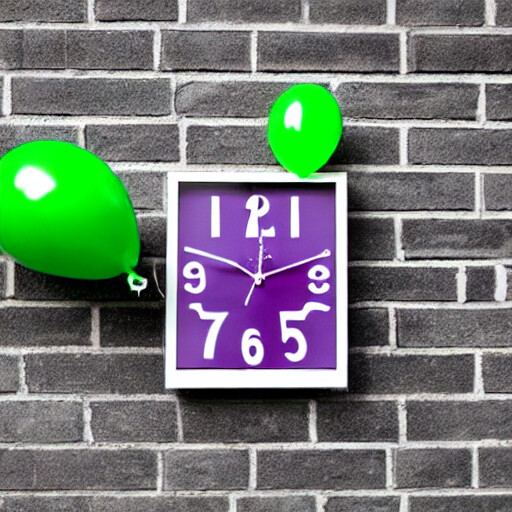} &
        \includegraphics[width=14mm]{./fig/correspondence/a_green_balloon_and_a_purple_clock_6_Predicated.jpg} &
        \includegraphics[width=14mm]{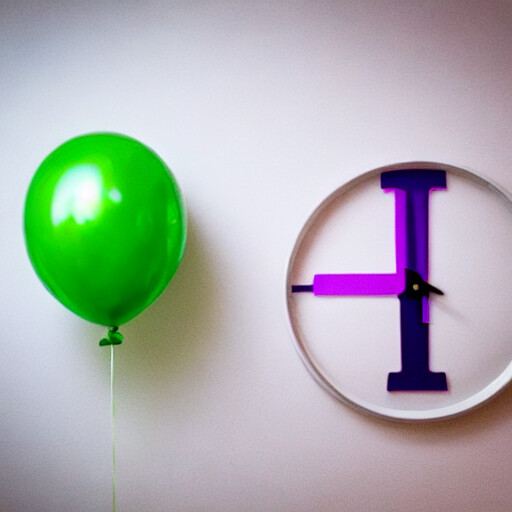}
        \\
    \end{tabular}
    \vspace{-3mm}
    \captionof{figure}{Example results of Experiment \ref{ex:2} for one-to-one correspondence.
        See also Fig.~\ref{fig:experiment2_additional}.
    }  \label{fig:experiment2}
\end{figure*}

\begin{enumerate}[label=(\roman*),topsep=0ex,itemsep=.2ex,partopsep=0ex,parsep=0ex,leftmargin=7mm]
    \item \label{ex:1} \emph{Concurrent Existence}: We prepared 400 random prompts, each mentioning ``[Object A] and [Object B]'', and generated 400 sets of images.
          The evaluators identified cases of ``missing objects'' based on two criteria: a lenient criterion where ``object mixture'' was not counted as ``missing objects'', and a strict criterion where it was.
          For Predicated Diffusion, we used the loss function~\eqref{eq:loss_concurrent}.
    \item \label{ex:2} \emph{One-to-One Correspondence}: Similarly, we prepared 400 random prompts, each mentioning ``[Adjective A] [Object A] and [Adjective B] [Object B]''.
          In addition to identifying missing objects, the evaluators identified the cases of ``attribute leakage''.
          For Predicated Diffusion, we used the loss function~\eqref{eq:loss_concurrent}$+$\eqref{eq:loss_one-to-one} with $\alpha=0.3$.
    \item \label{ex:3} \emph{Possession}: We prepared 10 prompts, each mentioning ``[Subject A] is [Verb C]-ing [Object B]''.
          [Verb C] can be ``have,'' ``hold,'' ``wear,'' or the like.
          We generated 20 images for each of these prompts.
          In addition to identifying missing objects, the evaluators identified the cases of ``possession failure''.
          For Predicated Diffusion, we used the loss function~\eqref{eq:loss_concurrent}$+$\eqref{eq:loss_possession}.
    \item \label{ex:4} \emph{Complicated}: To demonstrate the generality of Predicated Diffusion, we prepared diverse prompts, some of which were taken from the ABC-6K dataset~\citep{Feng2023}.
          Images were generated after manually extracting propositions and their respective loss functions.
          While a summary of generated images is presented, numerical evaluations were not undertaken due to the diversity of the prompts.
\end{enumerate}
Experiments \ref{ex:1} and \ref{ex:2} were inspired by previous works~\citep{Feng2023,Chefer2023,Rassin2023}.
In Experiments \ref{ex:1}--\ref{ex:3}, the evaluators also assessed the fidelity of the generated images to the prompts.
Instructions and evaluation criteria provided to the evaluators are detailed in Appendix~\ref{appendix:instruction}
The candidates of objects, adjectives, and prompts can be found in Appendix~\ref{appendix:prompt}.

We also automatically measured the fidelity and quality of generated images by using the pretrained image-text encoder, CLIP~\citep{Radford2021a} and image-captioning model, BLIP~\citep{Li2022c}.
We prepared 10,000 images for each of Experiments \ref{ex:1}--\ref{ex:3}.
For the fidelity, we evaluated similarity measures proposed by~\citet{Chefer2023}.
The text-image similarity is the cosine similarity between the prompts and the generated images in the embedding space of CLIP.
For the text-text similarity, instead of the generated images, their captions generated by BLIP are used.
For the quality, we performed CLIP image quality assessment (CLIP-IQA)~\citep{Wang2023c}.
This metric evaluates how close an image is to the text ``good photo'' as opposed to ``bad photo'' in the embedding space, and it has been proven to be highly correlated with the human perception of image quality.
Unlike Fr\`echet inception distance (FID)~\citep{Heusel2017a} or the similarity measures introduced above, CLIP-IQA does not require the ground truth dataset or the text prompts, but only measures image quality.

\begin{table*}[t]
    \captionof{table}{Results of Experiment \ref{ex:3} for Possession}
    \label{tab:experiment3}
    \centering
    \footnotesize
\begin{tabular}{lcccccc}
        \toprule
                             & \multicolumn{3}{c}{\textbf{Human Evaluation}} & \multicolumn{2}{c}{\textbf{Automatic Evaluation}}                                                                           \\
        \cmidrule(lr){2-4}\cmidrule(lr){5-6}
        \textbf{Methods}      & \textbf{Missing Objects}$^\dagger$            & \textbf{Possession Failure}                       & \textbf{Fidelity} & \textbf{Similarity}             & \textbf{CLIP-IQA} \\
        \midrule
        Stable Diffusion     & 31.5 / 36.0                                   & 52.5                                              & 33.5              & 0.320 / 0.811                   & 0.762             \\
        Attend-and-Excite    & \zz7.5 / 17.0                                 & 51.5                                              & 27.5              & 0.334 / 0.843                   & 0.760             \\
        \midrule
        Predicated Diffusion & \zz\textbf{4.0} / \zz\textbf{7.0}             & \textbf{29.5}                                     & \textbf{52.0}     & \textbf{0.345} / \textbf{0.855} & \textbf{0.765}    \\
        \bottomrule
        \multicolumn{4}{l}{$^\dagger$Using the lenient and strict criterions. $^\ddagger$Text-image similarity and text-text similarity.}
    \end{tabular}\\[3mm]
    \tabcolsep=0.2mm
    \scriptsize
    \centering
    \begin{tabular}{c@{\hspace*{1mm}}cc@{\hspace*{2mm}}cc@{\hspace*{2mm}}cc@{\hspace*{2mm}}cc}

                                                                                                 &
        \multicolumn{2}{c}{\emph{a bear having an apple}}                                        &
        \multicolumn{2}{c}{\emph{a frog wearing a hat}}                                          &
        \multicolumn{2}{c}{\emph{a man holding a rabbit}}                                        &
        \multicolumn{2}{c}{\emph{a girl holding a suitcase}}                                       \\
        \rotatebox[origin=l]{90}{\parbox{13mm}{\centering Stable Diffusion}}                     &
        \includegraphics[width=14mm]{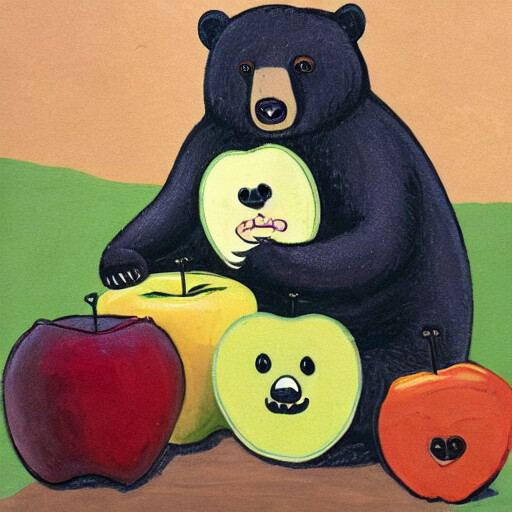}       &
        \includegraphics[width=14mm]{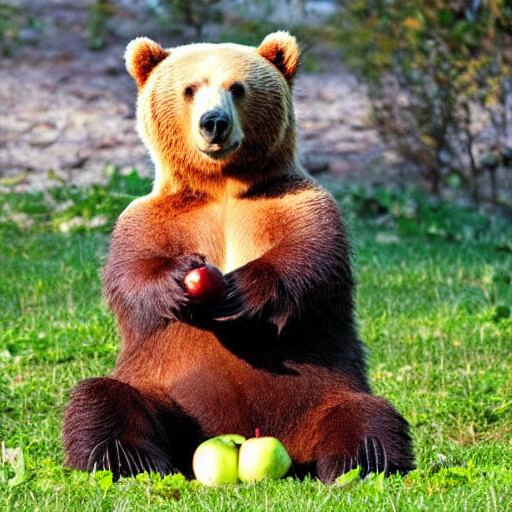}       &
        \includegraphics[width=14mm]{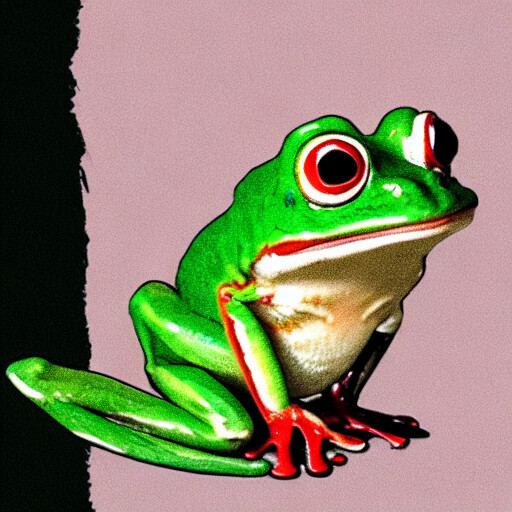}         &
        \includegraphics[width=14mm]{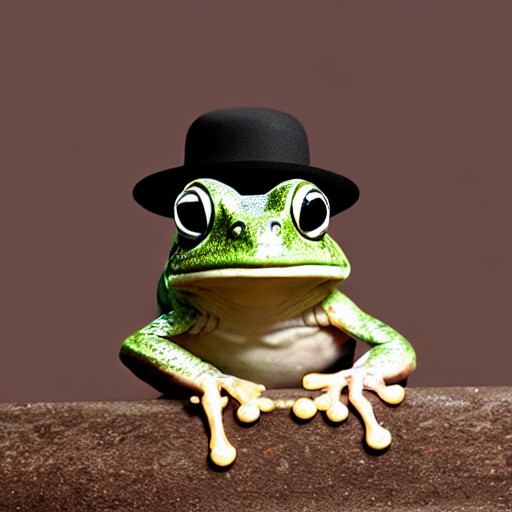}         &
        \includegraphics[width=14mm]{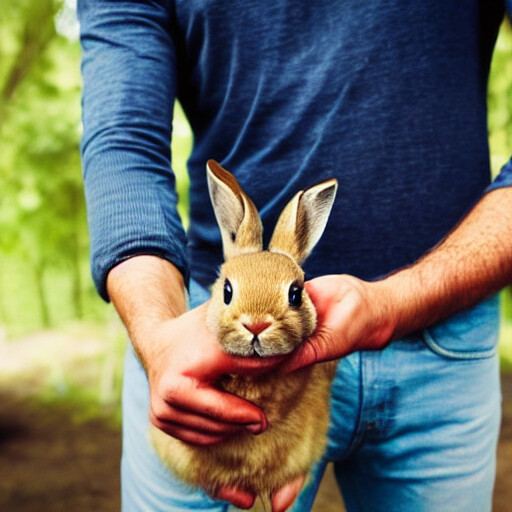}       &
        \includegraphics[width=14mm]{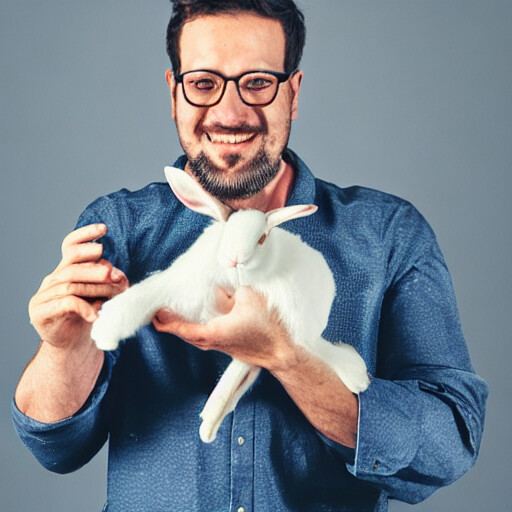}      &
        \includegraphics[width=14mm]{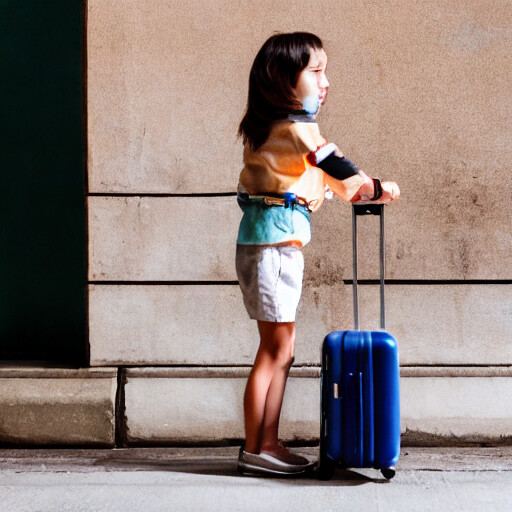}   &
        \includegraphics[width=14mm]{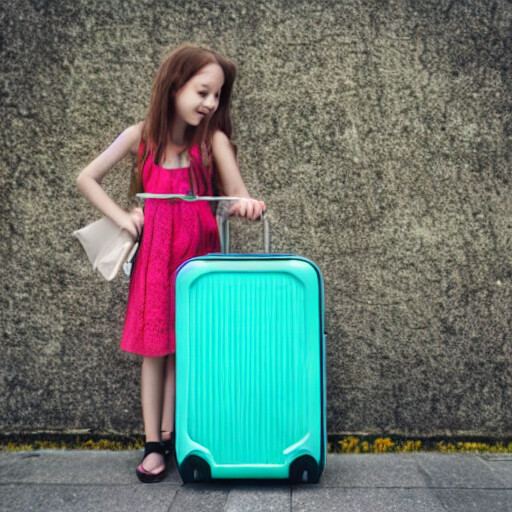}     \\
\end{tabular}
    \vspace*{-2.2mm}
    \captionof{figure}{Example results of Experiment \ref{ex:3} for possession.
        See also Fig.~\ref{fig:experiment3_additional}.
    }
    \label{fig:experiment3}
    \vspace*{-2mm}
\end{table*}

\begin{figure*}[t]
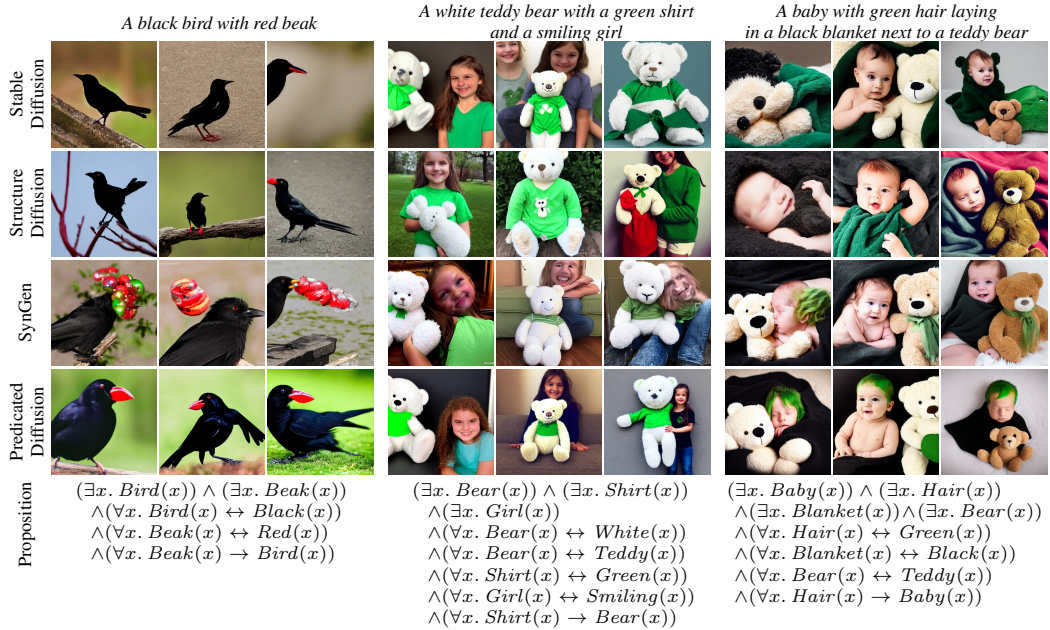

    \tabcolsep=0.2mm
    \scriptsize
    \centering
    \begin{tabular}{c@{\hspace*{1mm}}ccc@{\hspace*{2mm}}ccc@{\hspace*{2mm}}ccc}

                                                                                                                                                                                                                                                                                                                                                                                                                                                                                                                                                                                      &
        \multicolumn{3}{c}{\parbox{42mm}{\centering\emph{A black bird with red beak}}}                                                                                                                                                                                                                                                                                                                                                                                                                                                                                                &
        \multicolumn{3}{c}{\parbox{42mm}{\centering\emph{A white teddy bear with a green shirt and~a~smiling girl}}}                                                                                                                                                                                                                                                                                                                                                                                                                                                                  &
        \multicolumn{3}{c}{\parbox{42mm}{\centering\emph{A baby with green hair laying in~a~black~blanket next to a teddy bear}}}
        \\
\end{tabular}
    \vspace*{-2.2mm}
    \caption{Example results of Experiment \ref{ex:4} using prompts in ABC-6K.
        See also Figs.~\ref{fig:experiment4_additional1} and \ref{fig:experiment4_additional2}.
    }
    \label{fig:experiment4}
\end{figure*}

\subsection{Results}\label{sec:results}
\paragraph{Concurrent Existence and One-to-One Correspondence}
Table~\ref{tab:experiments12} summarizes the results from Experiments \ref{ex:1} and \ref{ex:2}.
Scores of human evaluations are expressed in percentages.
Higher scores are desirable for fidelity, similarities, and CLIP-IQA, while lower values are preferred for the remaining metrics.
Predicated Diffusion notably outperforms other methods, as it achieved the best outcomes across all 13 metrics; it improves both the quality and fidelity of the generated images.
It is worth noting that SynGen degraded the image quality measured by CLIP-IQA compared to the backbone, Stable Diffusion.
Figures~\ref{fig:experiment1} and~\ref{fig:experiment2} show example images for visual evaluation, where images in each column are generated using the same random seed.
See also Figs.~\ref{fig:experiment1_additional} and~\ref{fig:experiment2_additional} in Appendix.
Stable Diffusion, Composable Diffusion, and Structure Diffusion often exhibit missing objects and attribute leakage.
The absence of objects is particularly evident when prompts feature unusual object combinations like ``a crown and a rabbit'' and ``a yellow car and a blue bird.''
When the prompts specify visually similar objects, such as ``a bird and a cat,'' the two objects often get mixed together.
While Attend-and-Excite effectively prevents the issue of missing objects, it struggles with attribute leakage in Experiment \ref{ex:2}, due to its lack of a dedicated mechanism to address this.
While SynGen has achieved relatively good results, Predicated Diffusion outperforms it by further preventing missing objects and attribute leakage and producing images most faithful to the prompt.
Although this aspect was not explicitly part of the evaluation, SynGen often generates multiple instances of small objects, such as birds and balloons.

\paragraph{Possession}
Table~\ref{tab:experiment3} summarizes the results from Experiment \ref{ex:3}.
Predicated Diffusion notably outperforms other methods on all seven metrics.
Compared to Stable Diffusion, Attend-and-Excite succeeds in preventing missing objects but, on the contrary, fails to prevent possession failure, losing human-evaluated fidelity and CLIP-IQA.
Figures~\ref{fig:experiment3} and \ref{fig:experiment3_additional} show visual samples of generated images.
If [Subject A] is an animal, Attend-and-Excite succeeds more frequently than Stable Diffusion in depicting both objects but often depicts [Object B] as discarded on the ground or suspended in the air.
If [Subject A] is a human, the vanilla Stable Diffusion often produces satisfactory results.
Then, Attend-and-Excite, however, tends to deteriorate the overall image quality.
With the possession relationship, [Subject A] and [Object B] often overlap.
Attend-and-Excite makes both stand out competitively and potentially disrupts the overall harmony.
In contrast, the loss function \eqref{eq:loss_possession} is designed to encourage overlap, and hence Predicated Diffusion adeptly depicts subjects in possession of objects.

\paragraph{Qualitative Analysis on Complicated Prompts}
Figures~\ref{fig:experiment4}, \ref{fig:experiment4_additional1}, and \ref{fig:experiment4_additional2} show example results from Experiment \ref{ex:4} along with the propositions used for Predicated Diffusion.
Both the vanilla Stable Diffusion and Structure Diffusion plagued by missing objects and attribute leakage.
When tasked with generating ``A black bird with a red beak,'' SynGen produced multiple red objects, as observed in Experiments \ref{ex:1} and \ref{ex:2}.
When generating ``A white teddy bear with a green shirt and a smiling girl,'' comparison methods other than Predicated Diffusion often mistakenly identified the girl, not the teddy bear, as the owner of the green shirt.
In comparison to the vanilla Stable Diffusion, SynGen reduced the size of the teddy bear's shirt because it differentiates between the intensity distributions on the attention maps of different objects.
A similar tendency is evident in the third case in Fig.~\ref{fig:experiment4}, where the adjective ``green'' often alters wrong objects, and the green hair is not placed on the baby's head.
Predicated Diffusion performed well in these scenarios, which include the concurrent existence of more than two objects with specified colors and possession relationships simultaneously.

See Appendix~\ref{sec:additional_results} for further assessments; visualization of attention maps (\ref{sec:additional_analysis}), assignment of multiple colors to a single object, negation of objects (\ref{appendix:extensions}), automatic extraction of propositions (\ref{appendix:automatic_extraction}), additional analyses (\ref{appendix:analyses}), ablation studies (\ref{appendix:ablation}), and a discussion on more recent models (\ref{appendix:sdxl}).

\section{Conclusion}
This paper proposed Predicated Diffusion, where the intended meanings in a text prompt are represented by propositions using predicate logic, offering guidance for text-based image generation by diffusion models.
Experiments using Stable Diffusion as a backbone have demonstrated that Predicated Diffusion effectively addresses common challenges; missing objects, attribute leakage, and possession failures.
Compared to existing methods, Predicated Diffusion excels in generating images that are more faithful to the prompts and of superior quality.
Moreover, due to the generality of predicate logic, Predicated Diffusion can fulfill complicated prompts that include multiple objects, adjectives, and their relationships.
Although predicates cannot represent all meanings present in natural languages, they can handle most scenarios for adjusting the layout of generated images.
In future work, we plan to combine Predicated Diffusion with other backbones and explore 2-ary predicates asserting relationships, such as $Above(x,y)$, which implies ``$x$ is above $y$.''

\textbf{Acknowledgements:}
This work was partially supported by JST PRESTO (JPMJPR21C7) and Moonshot R\&D (JPMJMS2033-14), Japan.



\clearpage
\setcounter{page}{1}
\maketitlesupplementary

\renewcommand\thesection{\Alph{section}}
\setcounter{section}{0}
\renewcommand\thetable{A\arabic{table}}
\setcounter{table}{0}
\renewcommand\thefigure{A\arabic{figure}}
\setcounter{figure}{0}
\renewcommand\theequation{A\arabic{equation}}
\setcounter{equation}{0}

\section{Detailed Experimental Setting}\label{appendix:experiments}
\subsection{Implementation Details}\label{appendix:implementation}
\paragraph{Guidance Calculation}
We confirmed that Predicated Diffusion that directly employed the loss functions in Section~\ref{sec:loss} worked well.
Nonetheless, to further improve the quality of generated images, we implemented several modifications.

The loss functions are defined using either the summation or product over pixels $i$, which increase or decrease unboundedly as the image is scaled.
To make these loss functions more suitable for neural networks and to prevent excessively large values, we replaced the summation with the arithmetic mean.
Additionally, we replaced the finite product in \eqref{eq:loss_existence} with the geometric mean.
The equation $P(x)\land P(x)=P(x)$ holds for the classical Boolean logic and \Godel\ fuzzy logic but does not for the product fuzzy logic.
Due to our definition of the strong conjunction $\land$ and the assumption of $P(x)\in[0,1]$, the product fuzzy logic leads to the inequality $P(x)\land P(x)\le P(x)$.
This is the reason why the loss functions can increase or decrease unboundedly.

Attention maps are often obtained by performing a softmax operation in the channel direction on the feature maps of a CNN, where each channel is linked to a single word or token.
For calculating the loss function \eqref{eq:loss_existence} for existence, we excluded the start-of-text token.
Because this token is linked to the entire text, its omission ensures capturing the response to each individual word, consistent with the implementation of Attend-and-Excite.
For calculating the loss functions \eqref{eq:loss_adjective}, \eqref{eq:loss_one-to-one}, and \eqref{eq:loss_possession} using (bi)implication, we normalized the intensity of each attention map to a range of 0-1 using the maximum and minimum values.
This allows us to focus on the relative positions rather than the absolute intensity (that is, existence).

Following Attend-and-Excite~\citep{Chefer2023} and SynGen~\citep{Rassin2023}, the gradient of loss function, $\nabla_{x_t}\mathcal L[R]$, was multiplied by 20 for updating images.
Drawing inspiration from Attend-and-Excite~\citep{Chefer2023}, we performed the iterative refinement at $t=T$ (that is, at the beginning of the reverse process).
Specifically, before executing the very first step of the reverse process, we updated the image $x$ under generation as $x\leftarrow x-\nabla_{x}\mathcal L[R]$ five times.
Note that Attend-and-Excite performs the iterative refinement at the 10th and 20th steps until the value of the loss function falls below a certain threshold.
However, we found that the very first step is crucial for modifying the layout of the generated image.
This may be because the reverse process of the diffusion model is similar to gradient descent and may converge to poor local minima given poor initial values; the iterative refinement at $t=T$ helps adjust the initial values for better outcomes.
The results of the ablation study are provided in Table~\ref{tab:experiments12_ablation}, which shows that the iterative refinement slightly improved image fidelity and quality measured as similarities and CLIP-IQA.

In preliminary experiments, we found that the fidelity of the generated images improved more effectively as the number of iterative refinements at $t=T$ was increased.
No degradation in quality or diversity was observed even after several dozen iterations.
However, due to the limitations of available computational resources, we did not adjust for the optimal number of refinements; this is a subject for future research.

\begin{table*}[tb]
    \caption{Ablation Study of Refinement.}
    \label{tab:experiments12_ablation}
    \footnotesize
    \centering
    \tabcolsep=1.5mm
    \begin{tabular}{lcccccc}
        \toprule
                                                          & \multicolumn{2}{c}{Experiment \ref{ex:1}}    & \multicolumn{2}{c}{Experiment \ref{ex:2}} & \multicolumn{2}{c}{Experiment \ref{ex:3}}                                                                           \\
                                                          & \multicolumn{2}{c}{for Concurrent Existence} &
        \multicolumn{2}{c}{for One-to-One Correspondence} &
        \multicolumn{2}{c}{for Possession}                                                                                                                                                                                                                                 \\
        \cmidrule(lr){2-3}\cmidrule(lr){4-5}\cmidrule(lr){6-7}
        \textbf{Methods}                                   & \textbf{Similarity}$^\ddagger$               & \textbf{CLIP-IQA}                         & \textbf{Similarity}$^\ddagger$            & \textbf{CLIP-IQA} & \textbf{Similarity}$^\ddagger$  & \textbf{CLIP-IQA} \\
        \midrule
        Without refinement                                & 0.348 / 0.822                                & 0.771                                     & 0.376 / 0.801                             & 0.764             & 0.339 / 0.854                   & 0.764             \\
        With refinement                                   & 0.348 / \textbf{0.825}                       & \textbf{0.775}                            & \textbf{0.379} / \textbf{0.811}           & \textbf{0.769}    & \textbf{0.345} / \textbf{0.855} & \textbf{0.765}    \\
        \bottomrule
        \multicolumn{7}{l}{$^\ddagger$Text-image similarity and text-text similarity.}
    \end{tabular}\\[1.5mm]
\end{table*}

\paragraph{Computational Cost}
In practice, diffusion models are frequently combined with classifier-free guidance and negative prompts.
Then, the integration of Predicated Diffusion leads to a 33 \% increase in computational cost.

The original diffusion model calculates an image update conditioned on a text prompt.
The classifier-free guidance uses the same diffusion model without any conditions and emphasizes the difference between conditional and unconditional updates.
A negative prompt computes an update using the same diffusion model conditioned on an additional text and subtracts this update from the original one, negating the text.
Consequently, the diffusion model with the classifier-free guidance and a negative prompt requires three times the computational cost of the original model.

Predicated Diffusion reuses attention maps generated during the conditional update, avoiding extra computational cost.
The computational cost of defining the loss function with attention maps is negligible compared to CNNs.
However, computing the gradient of this loss function, $\nabla_{x_t}\mathcal L[R]$, via automatic differentiation incurs a computational cost comparable to the forward computation, equivalent to additional use of the diffusion model.
Therefore, when Predicated Diffusion is employed with classifier-free guidance and negative prompts, the total computational cost is equivalent to four times that of the original diffusion model, with Predicated Diffusion contributing to a 33 \% increase in the overall computation.

\subsection{Instructions to Evaluators}\label{appendix:instruction}
Three, three, and two evaluators joined Experiments (i), (ii), and (iii), respectively.
They are university students aged between 19 and 21, with no background in machine learning or computer vision.
The experiments were conducted in a double-blind manner: the images were presented in a random order, and the evaluators and authors were unaware of which image was generated by which model.
Each image was assessed by a single evaluator to maximize the number of assessed images.

\paragraph{Experiment \ref{ex:1}: Concurrent Existence}
We prepared 400 random prompts, each mentioning ``[Object A] and [Object B]'' with indefinite articles as needed, and generated 400 sets of images.
\begin{enumerate}
    \item[(a)] By showing one image at a time at random, we asked, ``Are both specified objects generated in the image?''
          The evaluators answered this question with one of the following options:
          \begin{enumerate}[label=\arabic*)]
              \item ``No object is generated.''
              \item ``Only one of two objects is generated.''
              \item ``Two objects are generated, but they are mixed together to form one object.''
              \item ``Two objects are generated.''
          \end{enumerate}
          Responses 1) and 2) were categorized as ``missing objects'', and response 3) was categorized as ``object mixture''.
          We tallied the number of responses 1) and 2) under the lenient criterion and that of responses 1)--3) under the strict criterion.
    \item[(b)] By showing a set of images generated with different methods, we asked, ``Which image is the most faithful to the prompt?''
          The evaluators were instructed to select only one image in principle, but were allowed to select more than one image if their fidelities were competitive, or not to select any image if none were faithful.
\end{enumerate}

\paragraph{Experiment \ref{ex:2}: One-to-One Correspondence}
We prepared 400 random prompts, each mentioning ``[Adjective A] [Object A] and [Adjective B] [Object B]'' with indefinite articles as needed.
\begin{enumerate}
    \item[(a)] The same as above.
    \item[(c)] At the same time as question (a), we asked, ``Does each adjective exclusively modify the intended object?''
          The evaluators answered this question with a ``Yes'' or ``No'', and the response ``No'' was categorized as ``attribute leakage''.
          If one or both of the two specified objects were not generated, that is, the response to question (a) was not 4), then question (c) became irrelevant, thereby automatically marking ``No'' as the response.
    \item[(b)] The same as above.
\end{enumerate}

\paragraph{Experiment \ref{ex:3}: Possession}
We prepared 10 prompts, each mentioning ``[Subject A] is [Verb C]-ing [Object B]'' with indefinite articles as needed.
We generated 20 images for each of these prompts.
\begin{enumerate}
    \item[(a)] The same as above.
    \item[(d)] At the same time as question (a), we asked, ``Is the [Subject A] performing [Verb C] with the [Object B]?''
          The evaluators answered this question with a ``Yes'' or ``No'', and the response ``No'' was categorized as ``possession failure.''
          If the response to question (a) was not 4), the response to question (d) was automatically assigned ``No''.
    \item[(b)] The same as above.
\end{enumerate}

\subsection{Prompts}\label{appendix:prompt}
For Experiments \ref{ex:1} and \ref{ex:2}, we randomly selected objects and adjectives from Table~\ref{tab:prompt}, roughly following the experiments conducted by \citet{Chefer2023}.
However, we excluded ``mouse'' and ``backpack'' from the list of objects, and ``orange'' from the list of adjectives.
The term ``mouse'' often led to ambiguity, as it could refer to either the animal or a computer peripheral.
The term ``orange'' also created confusion in all methods, as it could indicate either the color or the fruit.
Furthermore, it is challenging to distinguish ``backpack'' visually from other types of bags.
To ensure the accuracy of the evaluation, we removed these terms from the lists.

We used prompts in Table~\ref{tab:prompt3} for Experiment \ref{ex:3}.

\begin{table}[t]
    \caption{Candidate Words for Generating Prompts in Experiments \ref{ex:1} and \ref{ex:2}}\label{tab:prompt}
    \footnotesize
    \centering
    \begin{tabular}{lc}
        \toprule
        \textbf{Object}    & \parbox{5cm}{
            \hyphenpenalty=10000
            \exhyphenpenalty=10000
            cat, dog, bird, bear, lion, horse, elephant, monkey, frog, turtle, rabbit, glasses, crown, suitcase, chair, balloon, bow, car, bowl, bench, clock, apple} \\
        \midrule
        \textbf{Adjective} & \parbox{5cm}{red, yellow, green, blue, purple, pink, brown, gray, black, white}                                                          \\
        \bottomrule
    \end{tabular}
\end{table}
\begin{table*}[t]
    \caption{Prompts for Experiment \ref{ex:3} and Results for Each Prompt}\label{tab:prompt3}
    \footnotesize
    \centering
    \begin{tabular}{llllrrrr}
        \toprule
        \textbf{[Subject A]}    & \textbf{[Verb C]-ing}     & \textbf{[Object B]}         & \textbf{Methods}                  & $*1$           & $*2$           & $*3$           & $*4$           \\
        \midrule
        \multirow{3}{*}{rabbit} & \multirow{3}{*}{having}   & \multirow{3}{*}{phone}      & \scriptsize Stable Diffusion     & \scriptsize 12 & \scriptsize 12 & \scriptsize 15 & \scriptsize 5  \\[-.5mm]
                                &                           &                             & \scriptsize Attend-and-Excite    & \scriptsize 2  & \scriptsize 2  & \scriptsize 12 & \scriptsize 5  \\[-.5mm]
                                &                           &                             & \scriptsize Predicated Diffusion & \scriptsize 1  & \scriptsize 1  & \scriptsize 10 & \scriptsize 9  \\[-.5mm]
        \midrule
        \multirow{3}{*}{bear}   & \multirow{3}{*}{having}   & \multirow{3}{*}{apple}      & \scriptsize Stable Diffusion     & \scriptsize 2  & \scriptsize 2  & \scriptsize 5  & \scriptsize 10 \\[-.5mm]
                                &                           &                             & \scriptsize Attend-and-Excite    & \scriptsize 0  & \scriptsize 1  & \scriptsize 10 & \scriptsize 1  \\[-.5mm]
                                &                           &                             & \scriptsize Predicated Diffusion & \scriptsize 0  & \scriptsize 1  & \scriptsize 5  & \scriptsize 10 \\[-.5mm]
        \midrule
        \multirow{3}{*}{monkey} & \multirow{3}{*}{having}   & \multirow{3}{*}{bag}        & \scriptsize Stable Diffusion     & \scriptsize 12 & \scriptsize 12 & \scriptsize 12 & \scriptsize 2  \\[-.5mm]
                                &                           &                             & \scriptsize Attend-and-Excite    & \scriptsize 3  & \scriptsize 6  & \scriptsize 6  & \scriptsize 6  \\[-.5mm]
                                &                           &                             & \scriptsize Predicated Diffusion & \scriptsize 0  & \scriptsize 1  & \scriptsize 1  & \scriptsize 12 \\[-.5mm]
        \midrule
        \multirow{3}{*}{panda}  & \multirow{3}{*}{having}   & \multirow{3}{*}{suitcase}   & \scriptsize Stable Diffusion     & \scriptsize 13 & \scriptsize 16 & \scriptsize 19 & \scriptsize 1  \\[-.5mm]
                                &                           &                             & \scriptsize Attend-and-Excite    & \scriptsize 1  & \scriptsize 7  & \scriptsize 11 & \scriptsize 3  \\[-.5mm]
                                &                           &                             & \scriptsize Predicated Diffusion & \scriptsize 1  & \scriptsize 2  & \scriptsize 8  & \scriptsize 7  \\[-.5mm]
        \midrule
        \multirow{3}{*}{lion}   & \multirow{3}{*}{wearing}  & \multirow{3}{*}{crown}      & \scriptsize Stable Diffusion     & \scriptsize 12 & \scriptsize 12 & \scriptsize 12 & \scriptsize 8  \\[-.5mm]
                                &                           &                             & \scriptsize Attend-and-Excite    & \scriptsize 0  & \scriptsize 0  & \scriptsize 0  & \scriptsize 20 \\[-.5mm]
                                &                           &                             & \scriptsize Predicated Diffusion & \scriptsize 0  & \scriptsize 0  & \scriptsize 0  & \scriptsize 20 \\[-.5mm]
        \midrule
        \multirow{3}{*}{frog}   & \multirow{3}{*}{wearing}  & \multirow{3}{*}{hat}        & \scriptsize Stable Diffusion     & \scriptsize 9  & \scriptsize 10 & \scriptsize 10 & \scriptsize 3  \\[-.5mm]
                                &                           &                             & \scriptsize Attend-and-Excite    & \scriptsize 2  & \scriptsize 2  & \scriptsize 5  & \scriptsize 7  \\[-.5mm]
                                &                           &                             & \scriptsize Predicated Diffusion & \scriptsize 1  & \scriptsize 1  & \scriptsize 4  & \scriptsize 11 \\[-.5mm]
        \midrule
        \multirow{3}{*}{man}    & \multirow{3}{*}{holding}  & \multirow{3}{*}{rabbit}     & \scriptsize Stable Diffusion     & \scriptsize 0  & \scriptsize 0  & \scriptsize 2  & \scriptsize 12 \\[-.5mm]
                                &                           &                             & \scriptsize Attend-and-Excite    & \scriptsize 3  & \scriptsize 4  & \scriptsize 12 & \scriptsize 4  \\[-.5mm]
                                &                           &                             & \scriptsize Predicated Diffusion & \scriptsize 4  & \scriptsize 5  & \scriptsize 6  & \scriptsize 7  \\[-.5mm]
        \midrule
        \multirow{3}{*}{woman}  & \multirow{3}{*}{holding}  & \multirow{3}{*}{dog}        & \scriptsize Stable Diffusion     & \scriptsize 1  & \scriptsize 4  & \scriptsize 5  & \scriptsize 12 \\[-.5mm]
                                &                           &                             & \scriptsize Attend-and-Excite    & \scriptsize 3  & \scriptsize 8  & \scriptsize 16 & \scriptsize 2  \\[-.5mm]
                                &                           &                             & \scriptsize Predicated Diffusion & \scriptsize 1  & \scriptsize 3  & \scriptsize 3  & \scriptsize 10 \\[-.5mm]
        \midrule
        \multirow{3}{*}{boy}    & \multirow{3}{*}{grasping} & \multirow{3}{*}{soccerball} & \scriptsize Stable Diffusion     & \scriptsize 1  & \scriptsize 1  & \scriptsize 16 & \scriptsize 4  \\[-.5mm]
                                &                           &                             & \scriptsize Attend-and-Excite    & \scriptsize 1  & \scriptsize 3  & \scriptsize 18 & \scriptsize 2  \\[-.5mm]
                                &                           &                             & \scriptsize Predicated Diffusion & \scriptsize 0  & \scriptsize 0  & \scriptsize 14 & \scriptsize 6  \\[-.5mm]
        \midrule
        \multirow{3}{*}{girl}   & \multirow{3}{*}{holding}  & \multirow{3}{*}{suitcase}   & \scriptsize Stable Diffusion     & \scriptsize 1  & \scriptsize 3  & \scriptsize 9  & \scriptsize 10 \\[-.5mm]
                                &                           &                             & \scriptsize Attend-and-Excite    & \scriptsize 0  & \scriptsize 1  & \scriptsize 13 & \scriptsize 5  \\[-.5mm]
                                &                           &                             & \scriptsize Predicated Diffusion & \scriptsize 0  & \scriptsize 0  & \scriptsize 8  & \scriptsize 12 \\[-.5mm]
        \bottomrule
    \end{tabular}\\
    $*1$ Missing objects in the lenient criterion,
    $*2$ Missing objects in the strict criterion,\\
    $*3$ Possession failure,
    $*4$ Fidelity.
\end{table*}

\section{Additional Experiments and Results}\label{sec:additional_results}

\subsection{Additional Analysis}\label{sec:additional_analysis}

\begin{figure}[t]
    \centering
    \includegraphics[scale=0.4,page=2]{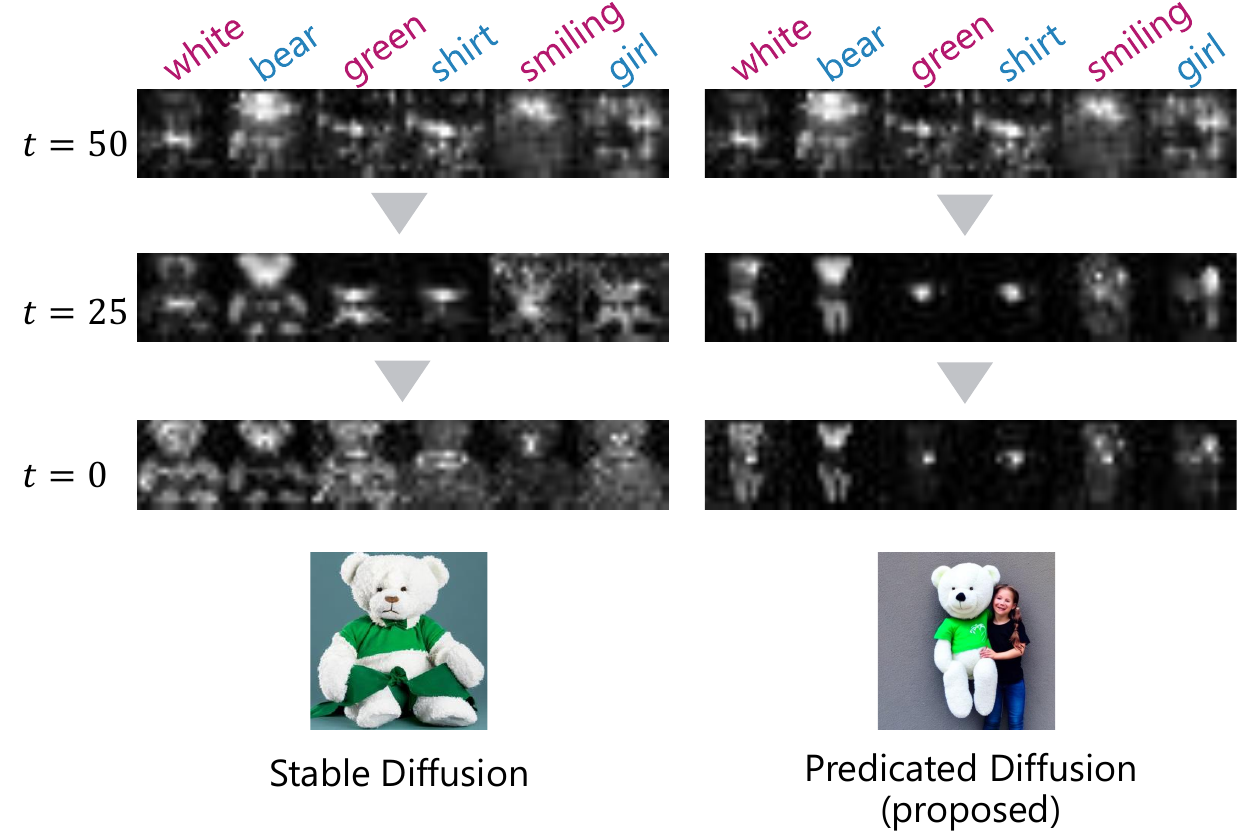}
    \vspace*{-2mm}
    \caption{An example visualization of attention maps during the generation.
        The text prompt was ``A white teddy bear with a green shirt and a smiling girl.''
    }
    \label{fig:attention}
\end{figure}

\paragraph{Visualization of Attention Maps}
We visualized the attention maps linked to the words of interest in a given prompt in Fig.~\ref{fig:attention}.
The attention map is obtained by applying a softmax operation along the token (word) axis of the feature map of the CNN.
Consequently, at most, one attention map can respond strongly at a specific pixel location.
At the start of the reverse process ($t=50$), each attention map responded to the image $x$ to some extent.
As depicted in the left half of Fig.~\ref{fig:attention}, the vanilla Stable Diffusion largely maintained the intensity distributions of the attention maps until the end of the reverse process, $t=0$.
This implies that the layout of the generated image is largely influenced by random initialization, which may not accurately capture the intended meaning of the prompt.
The attention map for ``bear'' dominates the entire space at $t=50$ and leaves no room for the attention map for ``girl'' to respond, resulting in the failure to generate a girl in the image.
In contrast, given the proposition $\exists x.\,\prop{Girl}(x)$, Predicated Diffusion encourages the attention map for ``girl'' to respond, even if it means dampening the response of the attention map for ``bear'', thereby ensuring the presence of a girl.

\begin{figure*}[t]
    \tabcolsep=0.2mm
    \centering
    \scriptsize
\begin{tabular}{c@{\hspace*{1mm}}ccc@{\hspace*{2mm}}ccc@{\hspace*{2mm}}ccc}
                                                                                         &
        \multicolumn{3}{c}{\emph{a frog and a chair}}                                    &
        \multicolumn{3}{c}{\emph{a bear and a dog}}                                      &
        \multicolumn{3}{c}{\emph{a bowl and a bench}}
        \\
        \rotatebox[origin=l]{90}{\parbox{13mm}{\centering Stable Diffusion}}             &
        \includegraphics[width=14mm]{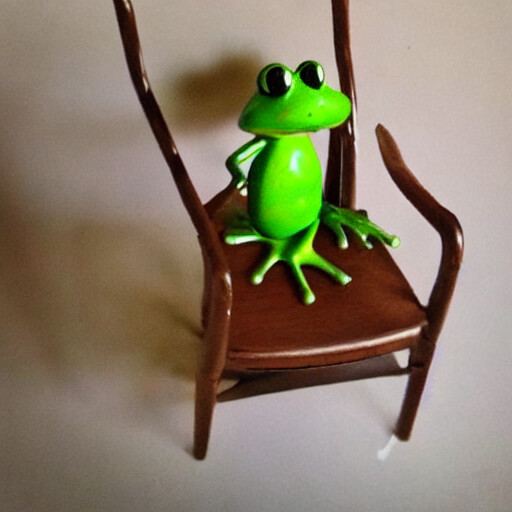}     &
        \includegraphics[width=14mm]{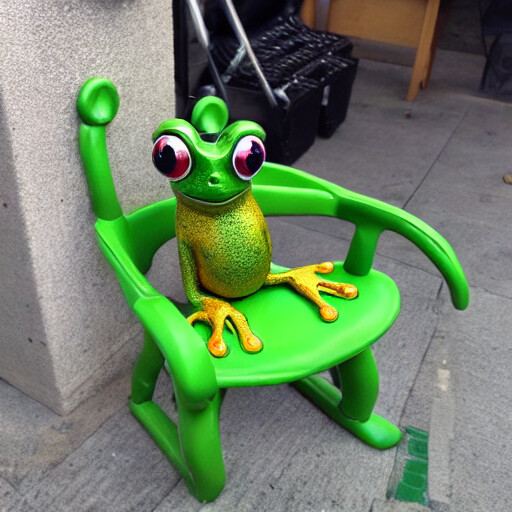}     &
        \includegraphics[width=14mm]{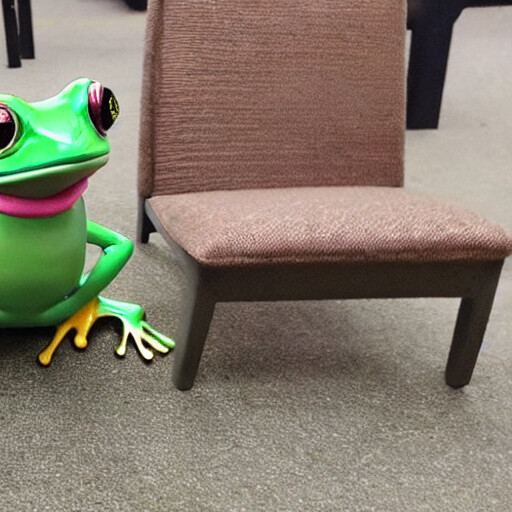}     &
        \includegraphics[width=14mm]{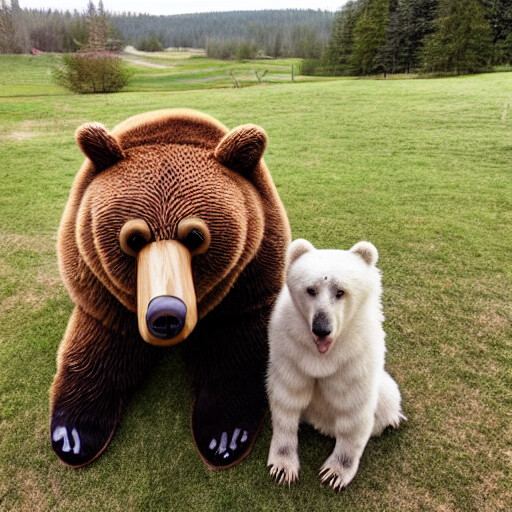}       &
        \includegraphics[width=14mm]{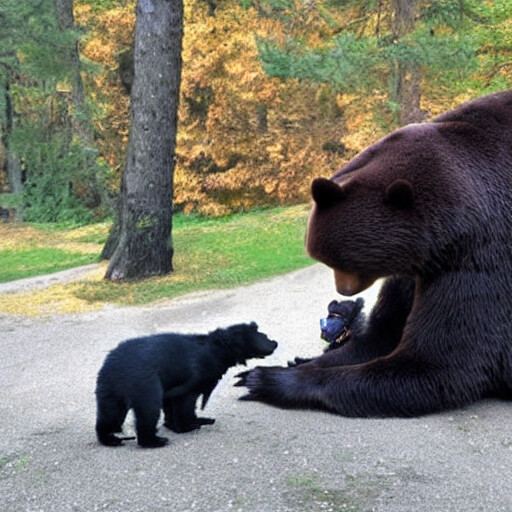}       &
        \includegraphics[width=14mm]{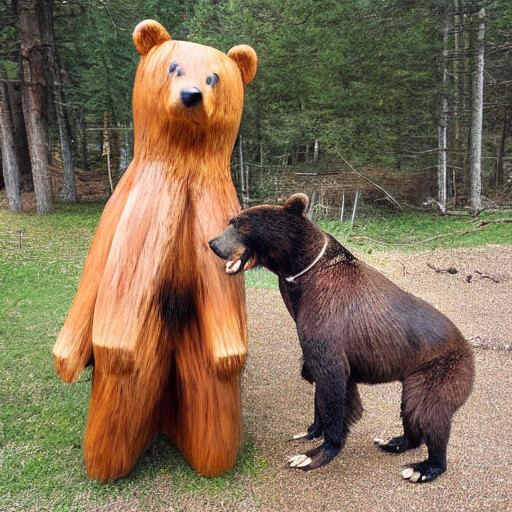}       &
        \includegraphics[width=14mm]{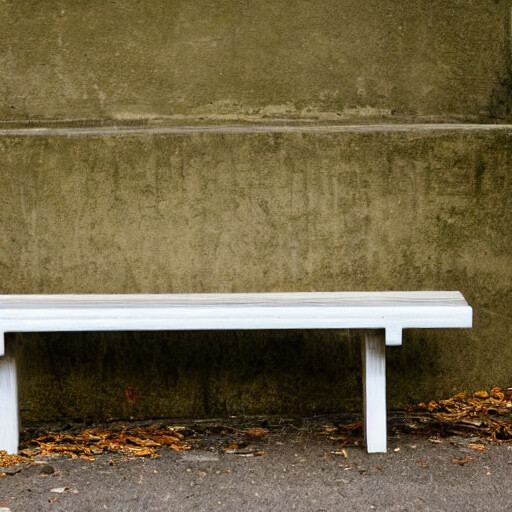}     &
        \includegraphics[width=14mm]{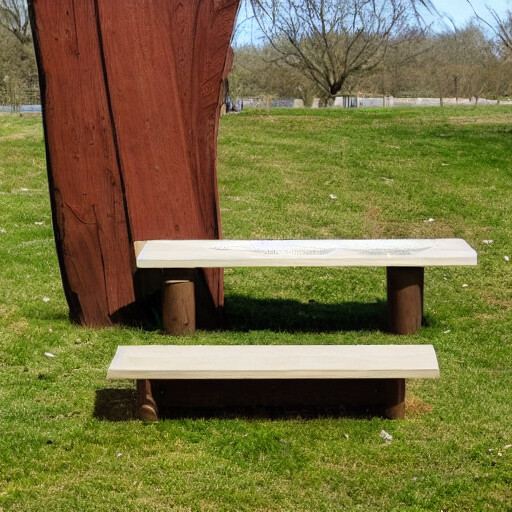}     &
        \includegraphics[width=14mm]{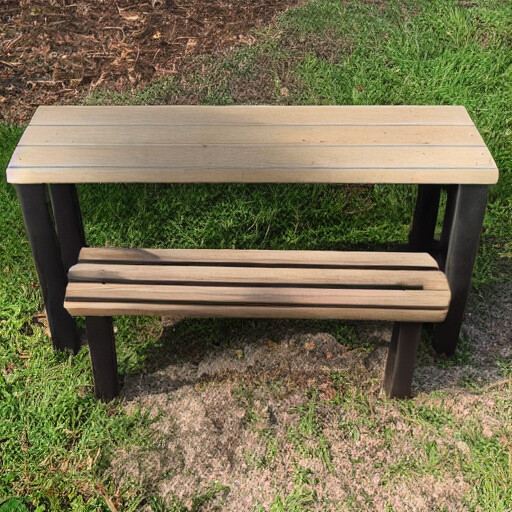}
        \\
\end{tabular}\\[-2mm]
    \caption{Additional results of Experiment \ref{ex:1} for concurrent existence.
        See also Fig.~\ref{fig:experiment1}.}
    \label{fig:experiment1_additional}
    \vspace{5mm}
    \tabcolsep=0.2mm
    \scriptsize
    \begin{tabular}{c@{\hspace*{1mm}}ccc@{\hspace*{2mm}}ccc@{\hspace*{2mm}}ccc}

                                                                                                           &
        \multicolumn{3}{c}{\emph{a black chair and a blue bow}}                                            &
        \multicolumn{3}{c}{\emph{a black cat and a pink crown}}                                            &
        \multicolumn{3}{c}{\emph{a blue turtle and a white bear}}
        \\
        \rotatebox[origin=l]{90}{\parbox{13mm}{\centering Stable Diffusion}}                               &
        \includegraphics[width=14mm]{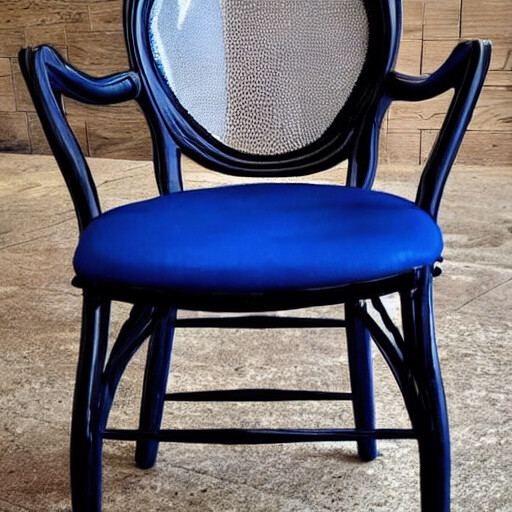}       &
        \includegraphics[width=14mm]{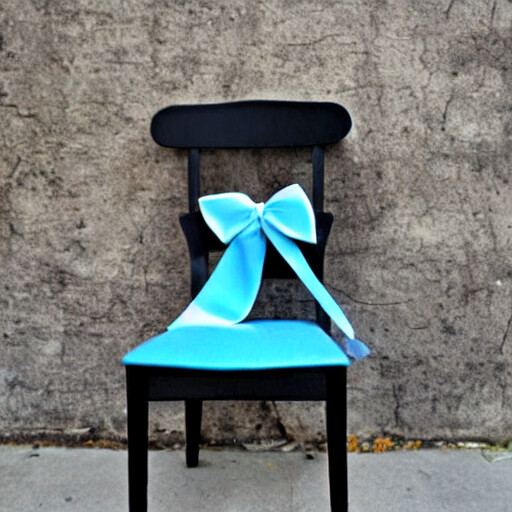}       &
        \includegraphics[width=14mm]{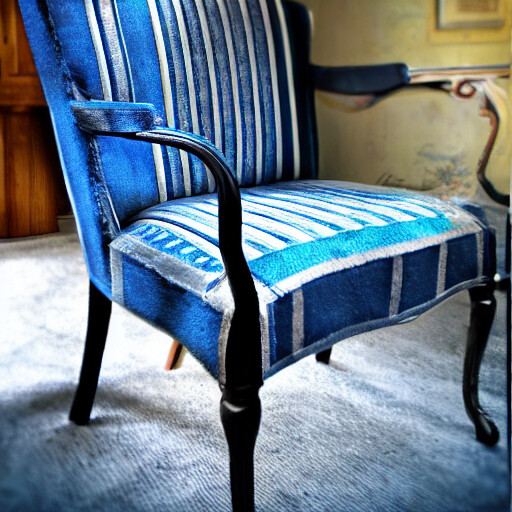}       &
        \includegraphics[width=14mm]{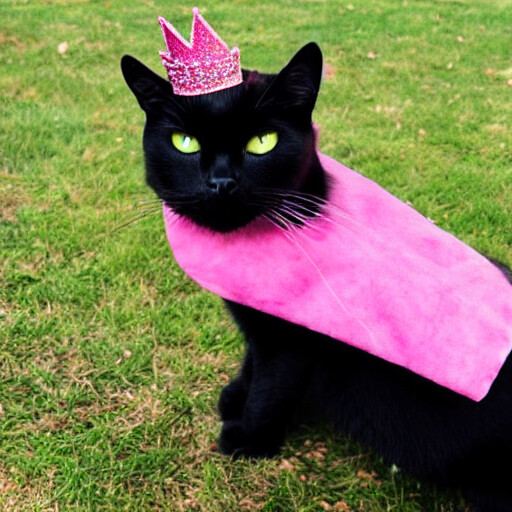}       &
        \includegraphics[width=14mm]{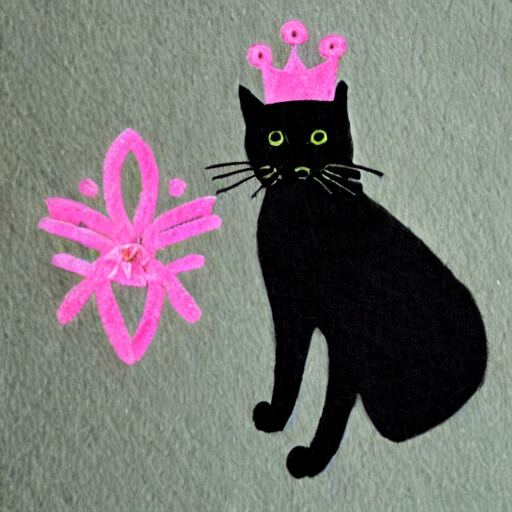}       &
        \includegraphics[width=14mm]{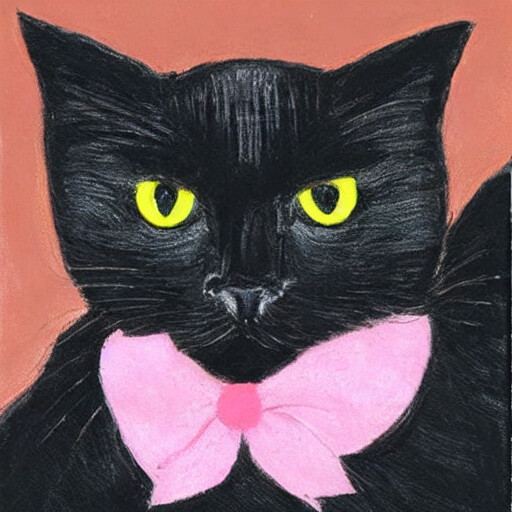}       &
        \includegraphics[width=14mm]{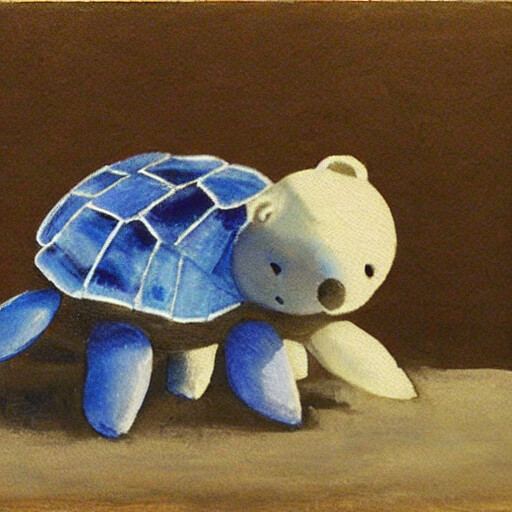}     &
        \includegraphics[width=14mm]{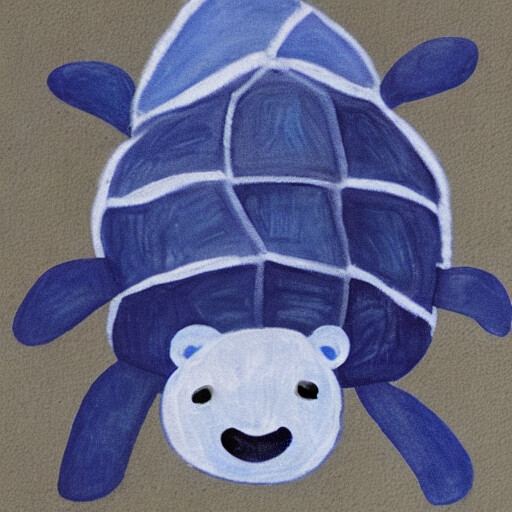}     &
        \includegraphics[width=14mm]{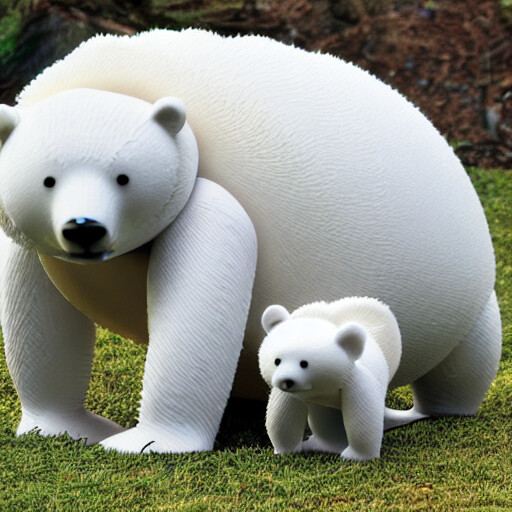}
        \\
\end{tabular}\\[-2mm]
    \caption{Additional results of Experiment \ref{ex:2} for one-to-one correspondence.
        See also Fig.~\ref{fig:experiment2}.}
    \label{fig:experiment2_additional}
\end{figure*}

\begin{figure*}[t]
    \tabcolsep=0.2mm
    \scriptsize
    \centering
    \begin{tabular}{c@{\hspace*{1mm}}cc@{\hspace*{2mm}}cc@{\hspace*{2mm}}cc@{\hspace*{2mm}}cc}
                                                                                                   &
        \multicolumn{2}{c}{\emph{a rabbit having a phone}}                                         &
        \multicolumn{2}{c}{\emph{a monkey having a bag}}                                           &
        \multicolumn{2}{c}{\emph{a woman holding a dog}}                                           &
        \multicolumn{2}{c}{\emph{a boy grasping a soccerball}}
        \\
        \rotatebox[origin=l]{90}{\parbox{13mm}{\centering Stable Diffusion}}                       &
        \includegraphics[width=14mm]{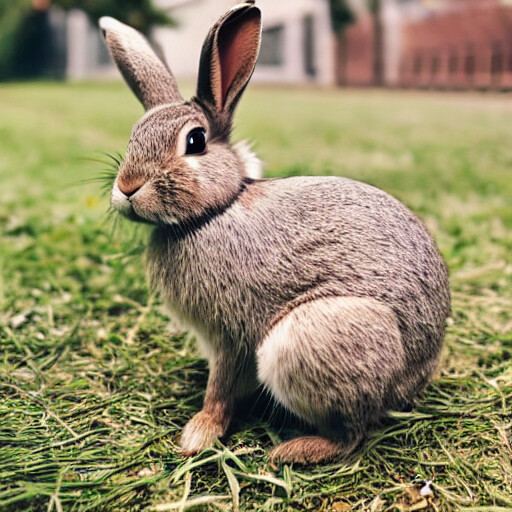}        &
        \includegraphics[width=14mm]{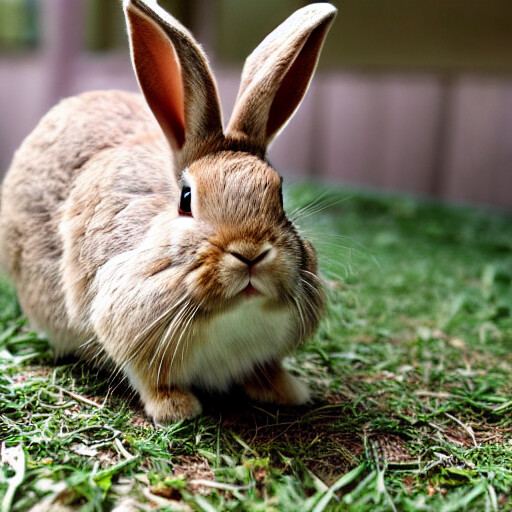}       &
        \includegraphics[width=14mm]{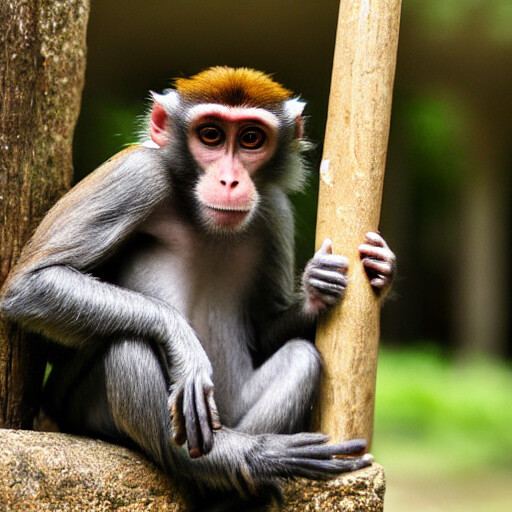}          &
        \includegraphics[width=14mm]{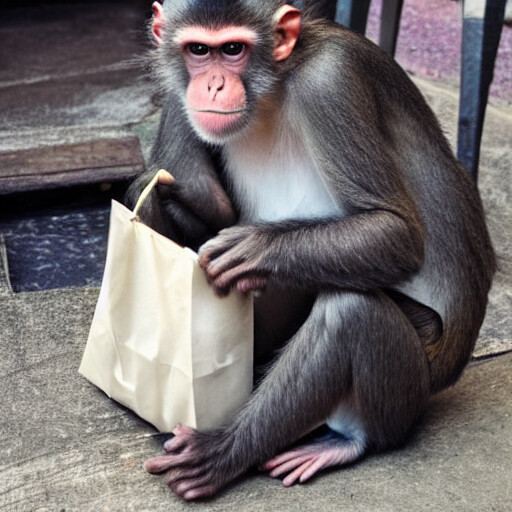}         &
        \includegraphics[width=14mm]{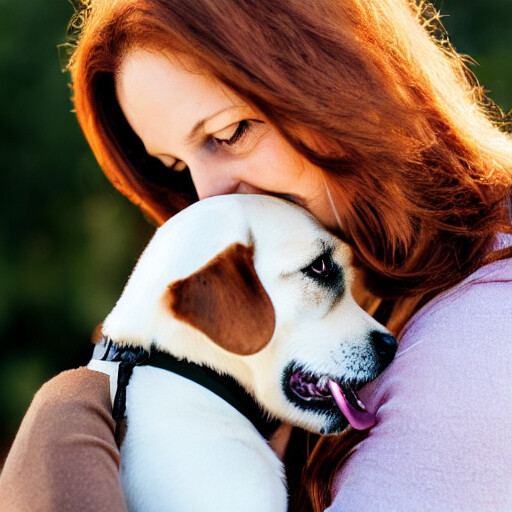}          &
        \includegraphics[width=14mm]{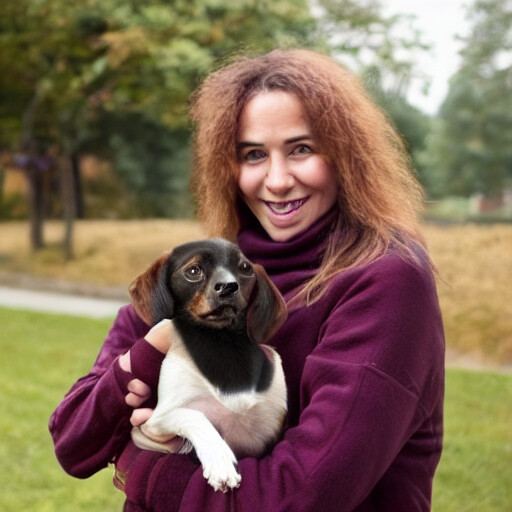}         &
        \includegraphics[width=14mm]{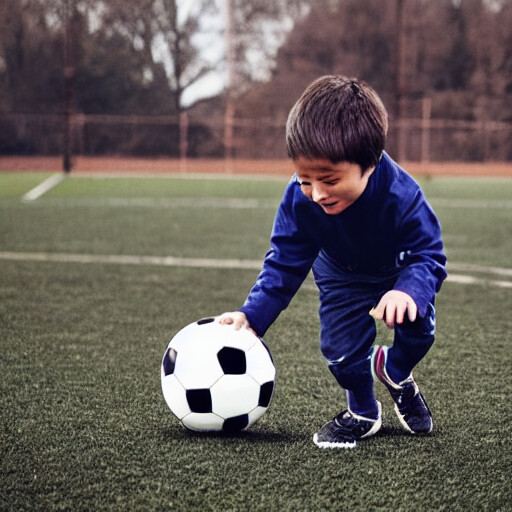}   &
        \includegraphics[width=14mm]{./fig/possession/A_boy_grasping_a_soccer_ball_11_Stable.jpg}
        \\
\end{tabular}\\[-2mm]
    \caption{Additional results of Experiment \ref{ex:3} for possession.
        See also Fig.~\ref{fig:experiment3} in the main body.}
    \label{fig:experiment3_additional}
\end{figure*}

\begin{figure*}[t]
    \tabcolsep=0.2mm
    \scriptsize
    \centering
    \begin{tabular}{c@{\hspace*{1mm}}ccc@{\hspace*{2mm}}ccc@{\hspace*{2mm}}ccc}
                                                                                                                                                                                                                                                                                                                                                                                                                                                                                                                   &
        \multicolumn{3}{c}{\parbox{42mm}{\centering\emph{Woman wearing a black coat holding~up~a~red~cellphone}}}                                                                                                                                                                                                                                                                                                                                                                                                  &
        \multicolumn{3}{c}{\parbox{42mm}{\centering\emph{A green and grey bird in tree with~white~leaves}}}                                                                                                                                                                                                                                                                                                                                                                                                        &
        \multicolumn{3}{c}{\parbox{42mm}{\centering\emph{a yellow vase with a blue and white bird}}}
        \\
\end{tabular}\\[-2mm]
    \caption{Example results of Experiment \ref{ex:4} using prompts in ABC-6K.
        See also Figs.~\ref{fig:experiment4} and \ref{fig:experiment4_additional2}.
    }
    \label{fig:experiment4_additional1}
    \vspace*{5mm}
    \tabcolsep=0.2mm
    \scriptsize
    \centering
    \begin{tabular}{c@{\hspace*{1mm}}ccc@{\hspace*{2mm}}ccc@{\hspace*{2mm}}cccc}

                                                                                                                                                                                                                                                                                                                                                                                                                                                                                                                                                                                                                                                                                                                                                                                                                                                                                                 &
        \multicolumn{3}{c}{\parbox{42mm}{\centering\emph{A purple bowl and a blue car and~a~green~sofa}}}                                                                                                                                                                                                                                                                                                                                                                                                                                                                                                                                                                                                                                                                                                                                                                                        &
        \multicolumn{3}{c}{\parbox{42mm}{\centering\emph{A brown dog and a boy holding~a~blue~suitcase}}}                                                                                                                                                                                                                                                                                                                                                                                                                                                                                                                                                                                                                                                                                                                                                                                        &
                                                                                                                                                                                                                                                                                                                                                                                                                                                                                                                                                                                                                                                                                                                                                                                                                                                                                                 &
        \multicolumn{3}{c}{\parbox{42mm}{\centering\emph{A polar bear (without snow)}}}
        \\
\end{tabular}\\[-2mm]
    \caption{Example results of Experiment \ref{ex:4} using our original prompts.
        See also Figs.~\ref{fig:experiment4} and \ref{fig:experiment4_additional1}.
    }
    \label{fig:experiment4_additional2}
\end{figure*}

\paragraph{Detailed Results of Experiment \ref{ex:3}}
For reference, we summarize the actual numbers out of 20 responses for each prompt in the right half of Table~\ref{tab:prompt3}.
The successful rate varies greatly depending on the prompt, but the proposed method, Predicated Diffusion, shows the best results in almost all combinations of metrics and prompts.
The sole exception is the prompt ``A man holding a rabbit,'' where Stable Diffusion already produced satisfactory results but Predicated Diffusion deteriorated the scores.
When the backbone, Stable Diffusion, can generate images faithful to the prompt, the additional guidance might disturb the generation process.

\paragraph{Additional Visualizations}
We summarize the additional visual examples of Experiments \ref{ex:1}--\ref{ex:4} in Figs.~\ref{fig:experiment1_additional}--\ref{fig:experiment4_additional2}.
While a quantitative comparison is difficult, Predicated Diffusion often retains the original layout when Stable Diffusion produces satisfactory results.
This is because, as long as logical operations like implications are satisfied, Predicated Diffusion does not trigger any further changes.

\subsection{Extensions to Other Logical Statements}\label{appendix:extensions}
\paragraph{Multi-color by Logical Disjunction}
Consider cases where multiple colors are specified for a single object.
For instance, the prompt ``a green and grey bird'' implies that every part of the bird is either green or grey, not both.
This statement can be represented using the disjunction by $\forall x.\,\prop{Bird}(x)\rightarrow\prop{Green}(x)\lor\prop{Grey}(x)$.
The corresponding loss function is:
\begin{equation}\label{eq:loss_multiadjective}
    \begin{aligned}
        \textstyle \mathcal{L}[\forall x.\,\prop{Bird}(x)\rightarrow\prop{Green}(x)\lor\prop{Grey}(x)] \hspace*{-25mm} \\
         & \hspace*{-25mm}\textstyle = -\sum_i\log (1-\att{\prop{Bird}}[i] \times (1-\att{\prop{Green}}[i])            \\
         & \hspace*{12mm} \textstyle \times (1-\att{\prop{Grey}}[i])).
    \end{aligned}
\end{equation}
When another object is introduced, one can replace the implication with a biimplication, as is the case with one-to-one correspondence.

The right six columns of Fig.~\ref{fig:experiment4_additional1} show example results.
SynGen produced birds with a mixed hue because it was designed to equalize the intensity distributions (that is, the regions) of both specified colors and that of the bird.
Conversely, Predicated Diffusion, based on predicate logic, can generate birds in the given combination of colors.

Note that, when multiple adjectives modify the same noun independently, they can be represented using the logical conjunction rather than the logical disjunction.
For instance, the prompt ``long, black hair'' can be decomposed into two statements that can hold simultaneously: ``The hair is long,'' and ``The hair is black.''
Then, the prompt is represented by the conjunction of two propositions that represent these statements.

\paragraph{Negation by Logical Negation}
In the right three columns of Fig.~\ref{fig:experiment4_additional2}, we explored the negation of a concept.
Sometimes, we might wish for certain concepts to be absent or negated in the generated images.
If given the prompt ``a polar bear,'' the output will typically be an image of a polar bear depicted with snowy landscapes because of their high co-occurrence rate in the dataset.
One can give the prompt ``a polar bear without snow,'' but Stable Diffusion often struggles to remove the snow, as depicted in the top row.
Alternatively, we could provide a negative prompt ``snow'' as proposed as part of Composable Diffusion~\citep{Liu2022b}.
We also examined Perp-Neg, which ensures a negative prompt not to interfere with a regular prompt by projecting the former's update to be orthogonal to the latter's update~\citep{Armandpour2023}; it often failed to remove the snow.
We consider an alternative way using predicate logic.
The absence of snow is represented by the proposition $\neg(\exists x.\,\prop{Snow}(x))=\forall x.\,\neg\prop{Snow}(x)$, leading the loss function:
\begin{equation}
    \textstyle \mathcal{L}[\neg(\exists x.\,\prop{Snow}(x))]=-\sum_i \log (1-\bar{A}_{\prop{Snow}}[i]),\label{eq:loss_negation}
\end{equation}
where $\bar{A}_w$ represents the attention map corresponding to a word $w$ in an auxiliary prompt like a negative prompt.
This approach did not show any clear advantages compared to the negative prompt but at least demonstrated the generality of Predicated Diffusion.

\begin{figure*}[t]
    \tabcolsep=0.2mm
    \scriptsize
    \centering
    \begin{tabular}{c@{\hspace*{1mm}}ccc@{\hspace*{2mm}}cccc}
                                                                                                      &
        \multicolumn{3}{c}{\parbox{42mm}{\centering\emph{A blue elephant}}}                           &
        \multicolumn{3}{c}{\parbox{42mm}{\centering\emph{A green dog}}}
        \\[1mm]
        \rotatebox[origin=l]{90}{\parbox{13mm}{\centering Noun$\rightarrow$ Adjective}}               &
        \includegraphics[width=14mm,height=14mm]{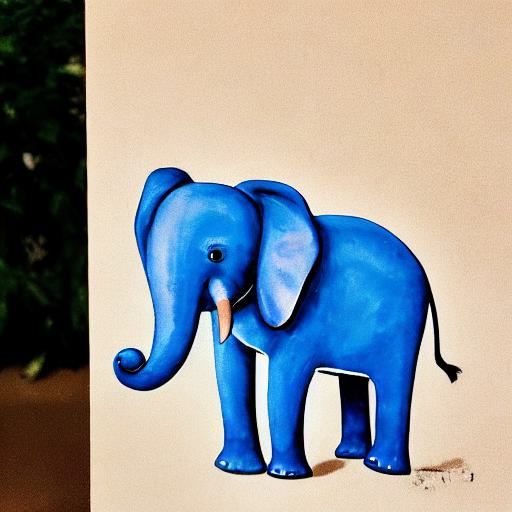}                &
        \includegraphics[width=14mm,height=14mm]{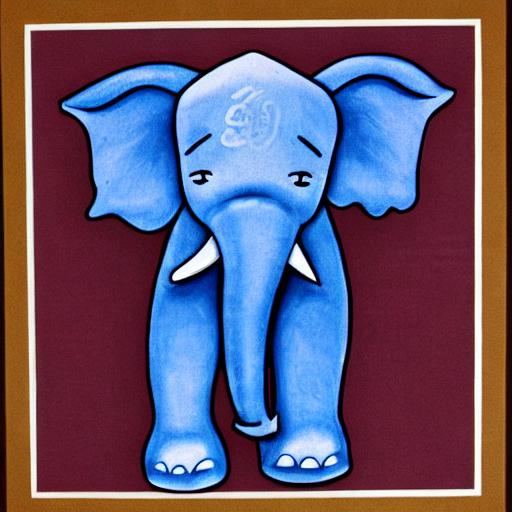}                &
        \includegraphics[width=14mm,height=14mm]{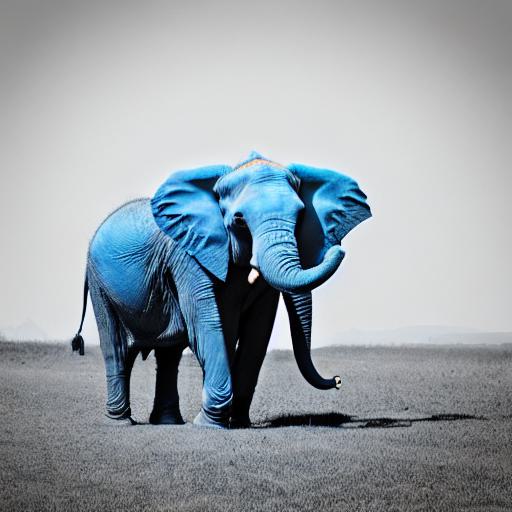}                &
        \includegraphics[width=14mm,height=14mm]{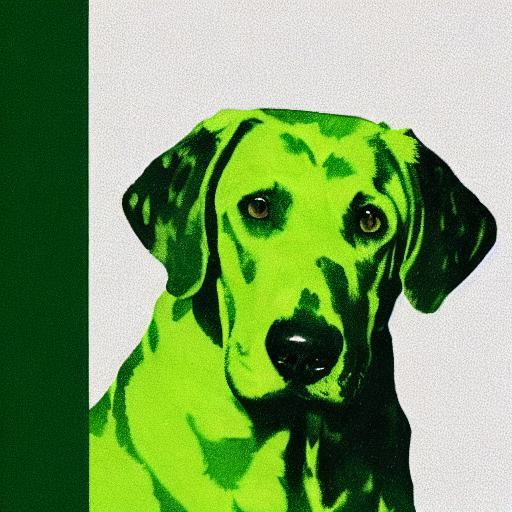}                    &
        \includegraphics[width=14mm,height=14mm]{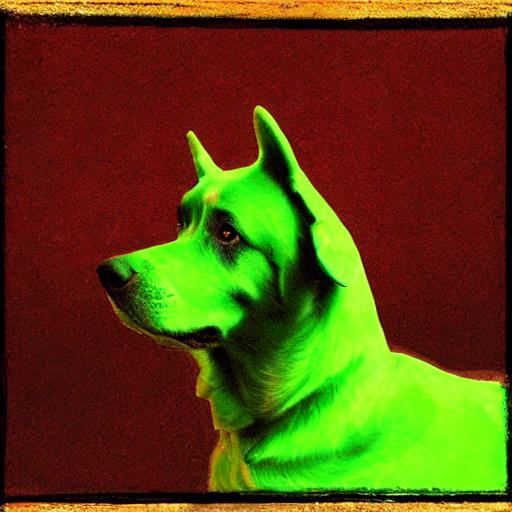}                    &
        \includegraphics[width=14mm,height=14mm]{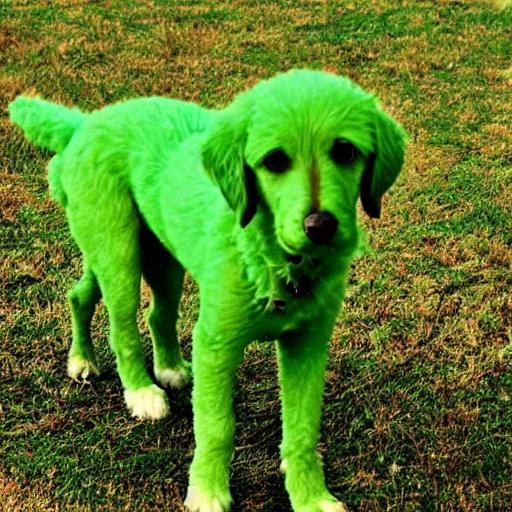}
        \\[-0.3mm]
        \rotatebox[origin=l]{90}{\parbox{13mm}{\centering Noun$\leftarrow$ Adjective}}                &
        \includegraphics[width=14mm,height=14mm]{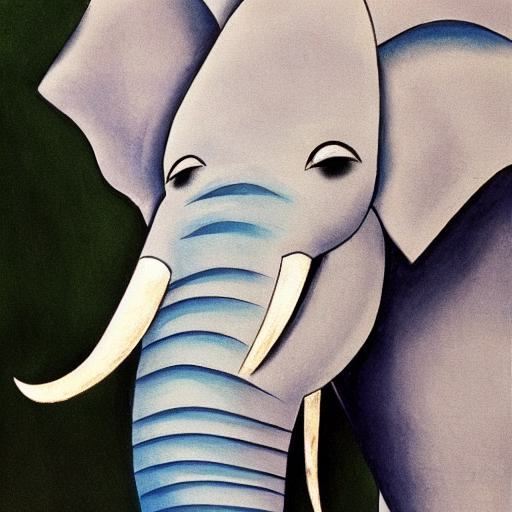}            &
        \includegraphics[width=14mm,height=14mm]{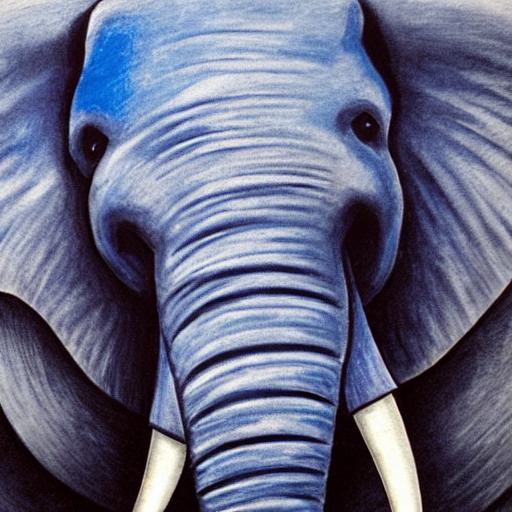}            &
        \includegraphics[width=14mm,height=14mm]{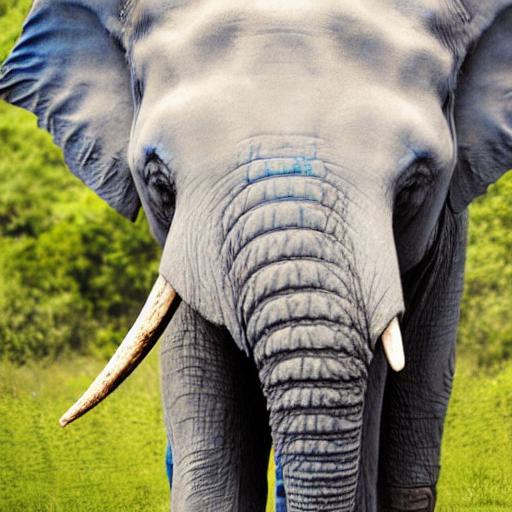}            &
        \includegraphics[width=14mm,height=14mm]{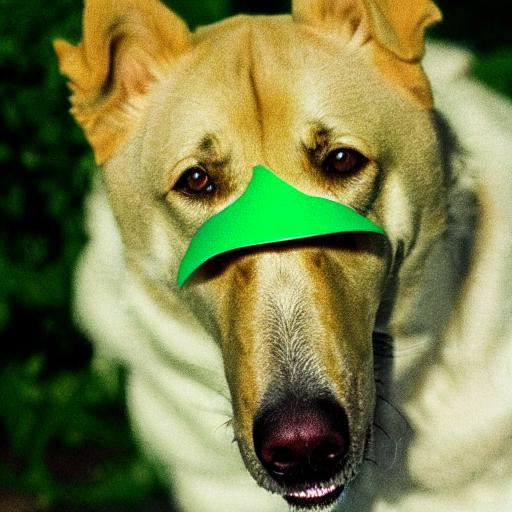}                &
        \includegraphics[width=14mm,height=14mm]{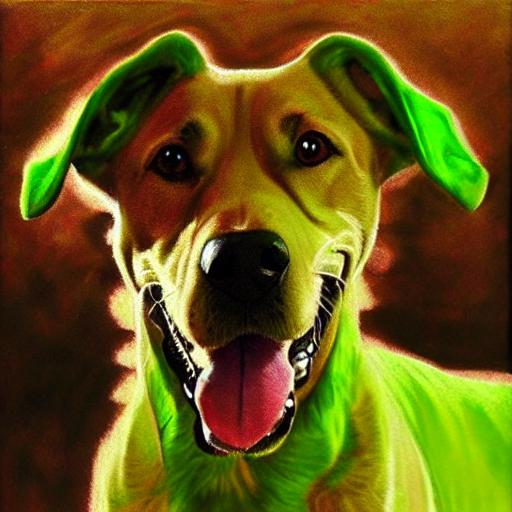}                &
        \includegraphics[width=14mm,height=14mm]{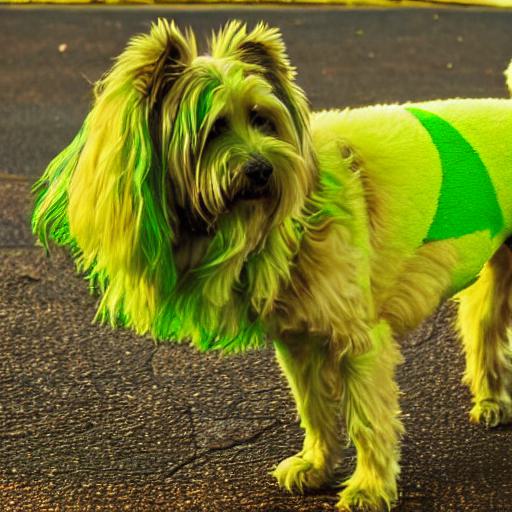}
        \\[-0.3mm]
        \rotatebox[origin=l]{90}{\parbox{13mm}{\centering Noun$\leftrightarrow$ Adjective}}           &
        \includegraphics[width=14mm,height=14mm]{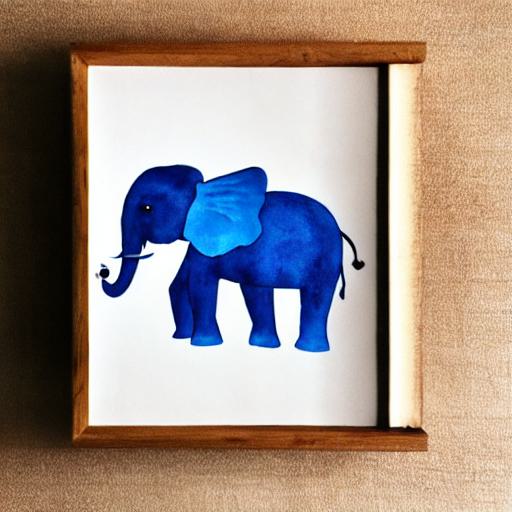} &
        \includegraphics[width=14mm,height=14mm]{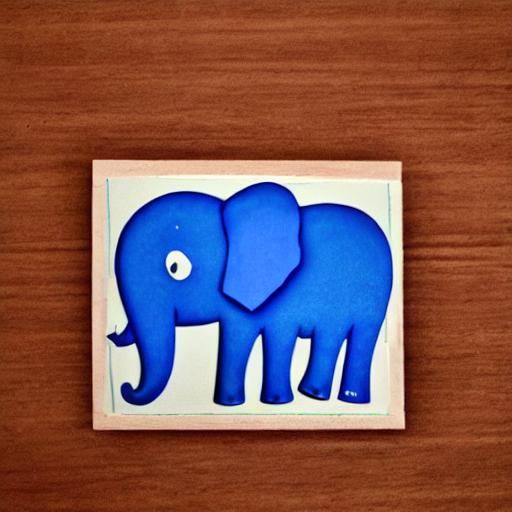} &
        \includegraphics[width=14mm,height=14mm]{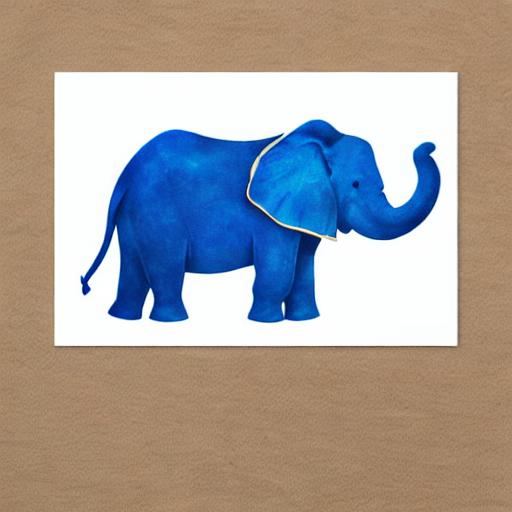} &
        \includegraphics[width=14mm,height=14mm]{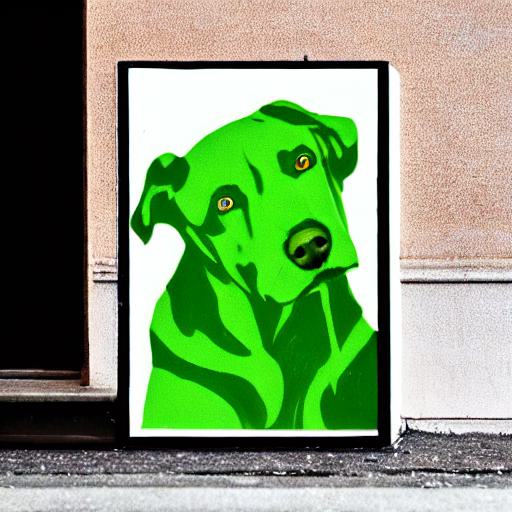}      &
        \includegraphics[width=14mm,height=14mm]{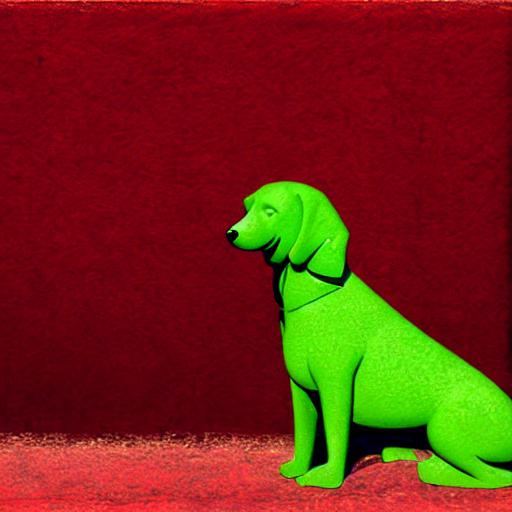}      &
        \includegraphics[width=14mm,height=14mm]{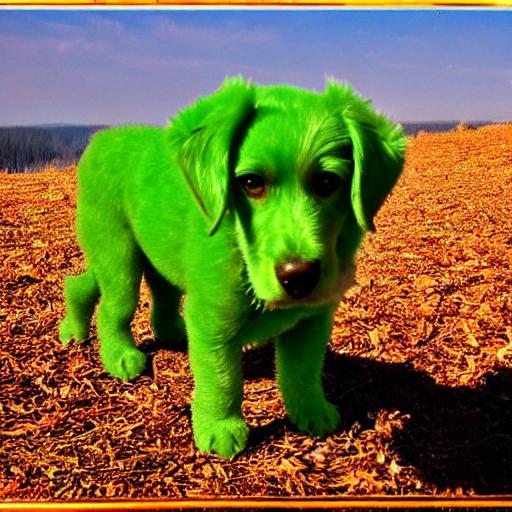}
        \\
    \end{tabular}\\[-2mm]
    \caption{Ablation study for the modification by the implication.
    }
    \label{fig:ablation1}
\end{figure*}

\begin{figure*}[t]
    \tabcolsep=0.2mm
    \scriptsize
    \centering
    \begin{tabular}{c@{\hspace*{1mm}}ccc@{\hspace*{2mm}}cccc}
                                                                                                                &
        \multicolumn{3}{c}{\parbox{42mm}{\centering\emph{A monkey having a bag}}}                               &
        \multicolumn{3}{c}{\parbox{42mm}{\centering\emph{A panda having a suitcase}}}
        \\[1mm]
        \rotatebox[origin=l]{90}{\parbox{13mm}{\centering Object$\rightarrow$ Subject}}                         &
        \includegraphics[width=14mm,height=14mm]{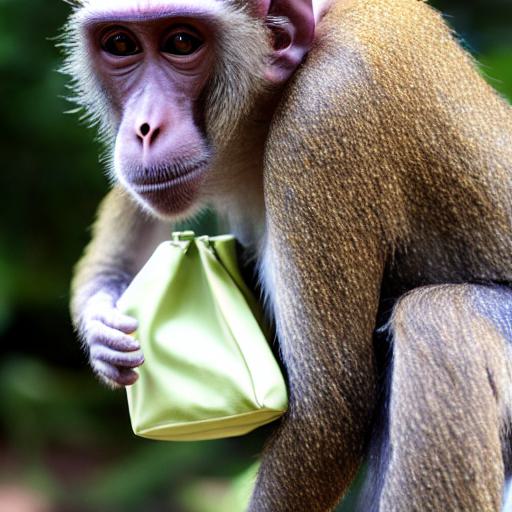}                    &
        \includegraphics[width=14mm,height=14mm]{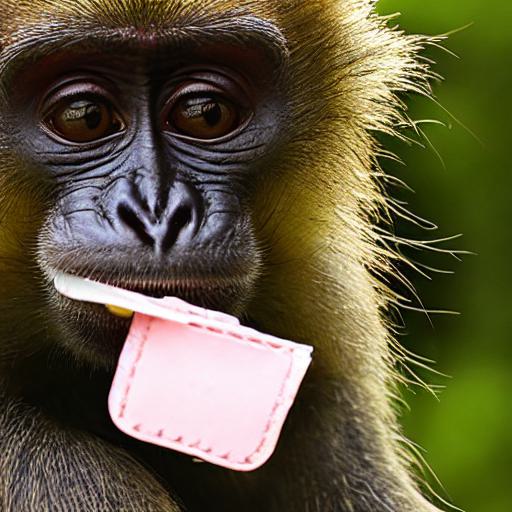}                    &
        \includegraphics[width=14mm,height=14mm]{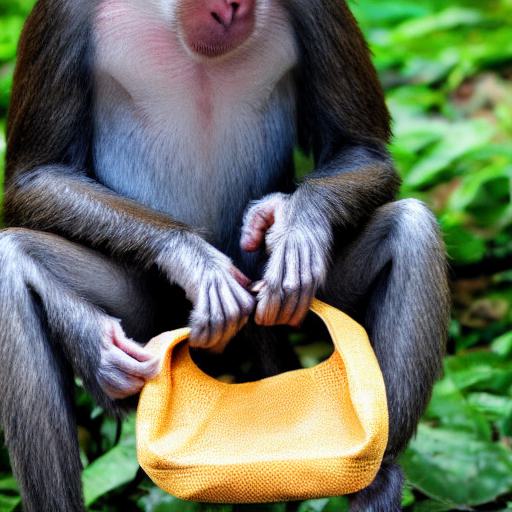}                    &
        \includegraphics[width=14mm,height=14mm]{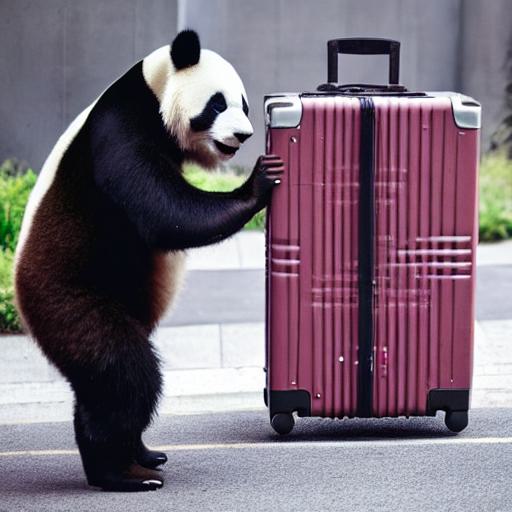}                &
        \includegraphics[width=14mm,height=14mm]{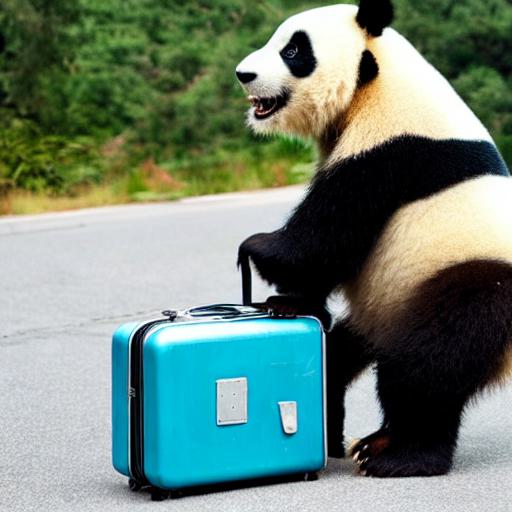}                &
        \includegraphics[width=14mm,height=14mm]{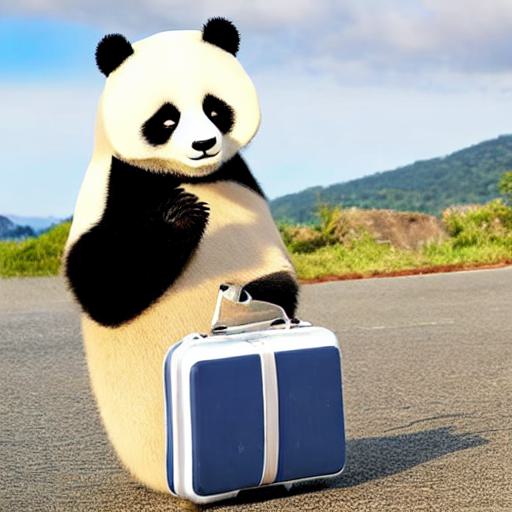}
        \\[-0.3mm]
        \rotatebox[origin=l]{90}{\parbox{13mm}{\centering Object$\leftarrow$ Subject}}                          &
        \includegraphics[width=14mm,height=14mm]{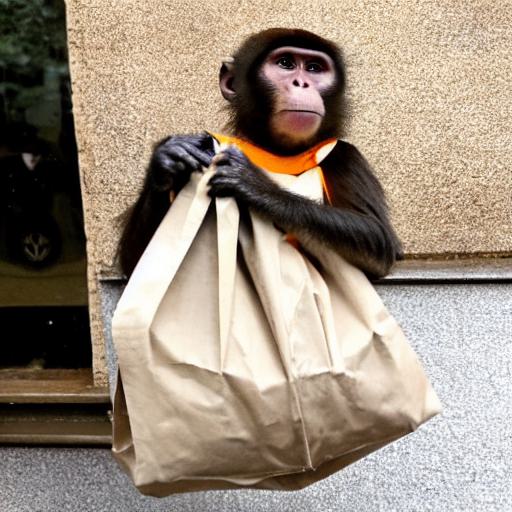}                &
        \includegraphics[width=14mm,height=14mm]{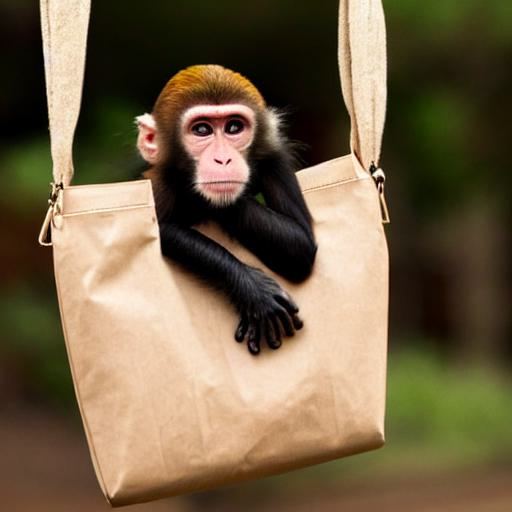}                &
        \includegraphics[width=14mm,height=14mm]{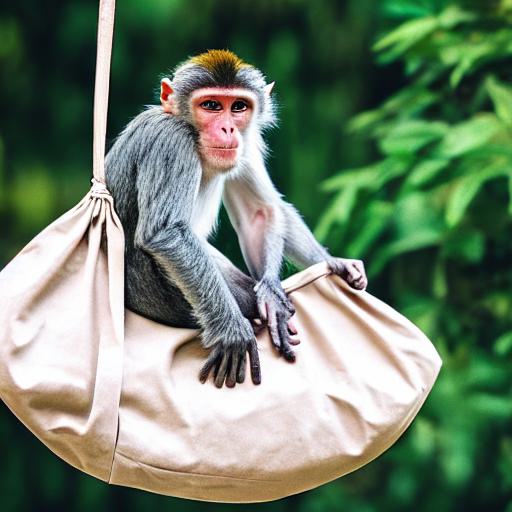}                &
        \includegraphics[width=14mm,height=14mm]{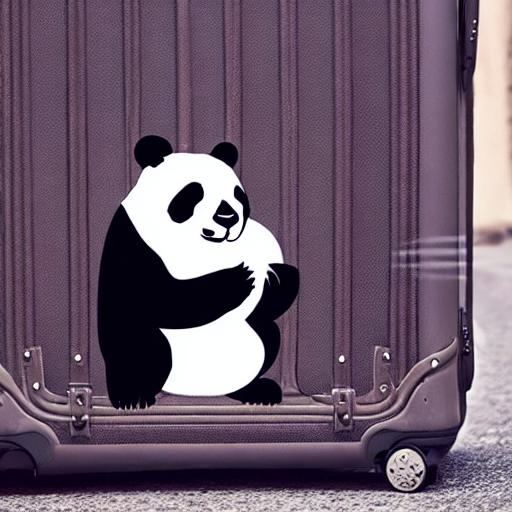}            &
        \includegraphics[width=14mm,height=14mm]{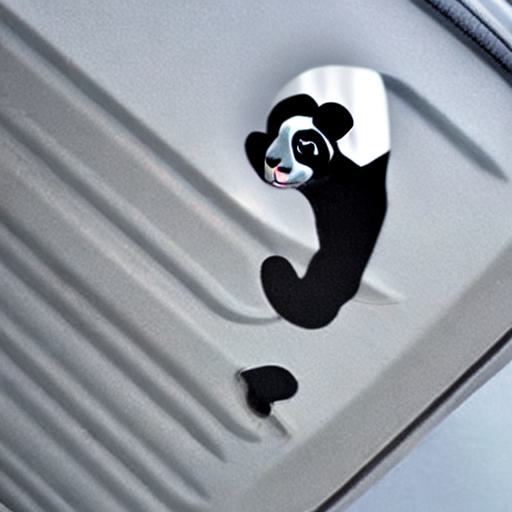}            &
        \includegraphics[width=14mm,height=14mm]{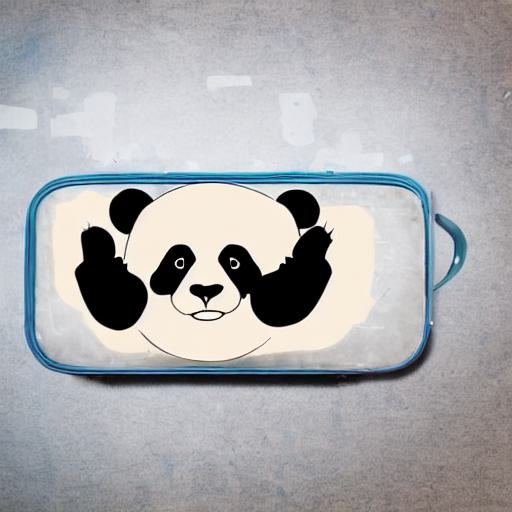}
        \\[-0.3mm]
        \rotatebox[origin=l]{90}{\parbox{13mm}{\centering Object$\leftrightarrow$ Subject}}                     &
        \includegraphics[width=14mm,height=14mm]{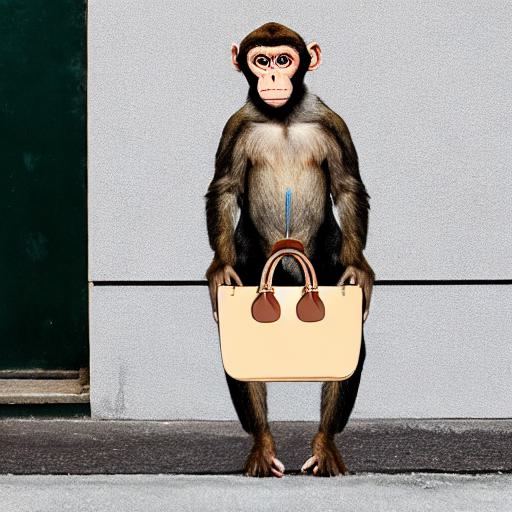}     &
        \includegraphics[width=14mm,height=14mm]{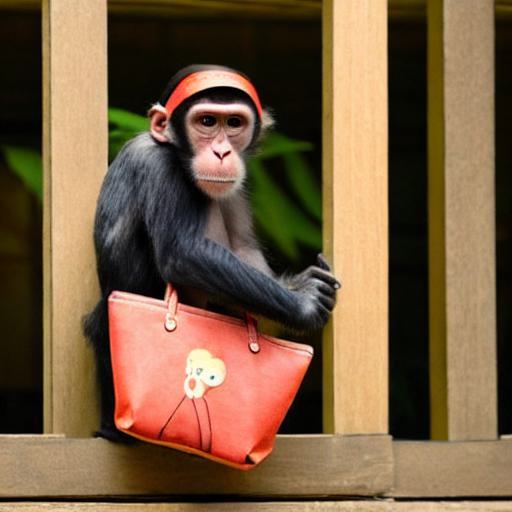}     &
        \includegraphics[width=14mm,height=14mm]{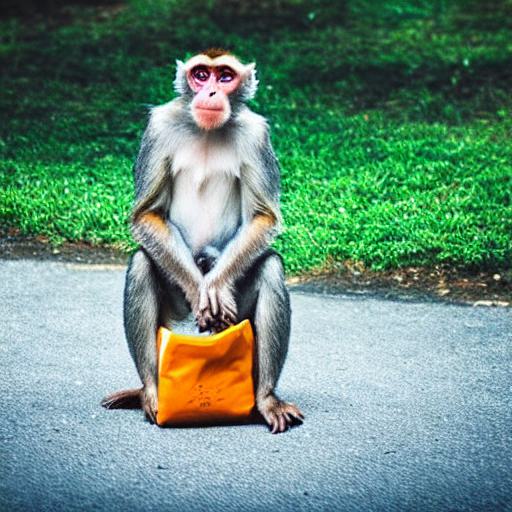}     &
        \includegraphics[width=14mm,height=14mm]{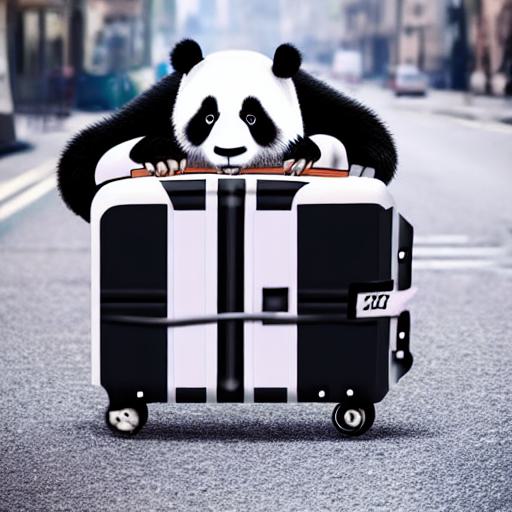} &
        \includegraphics[width=14mm,height=14mm]{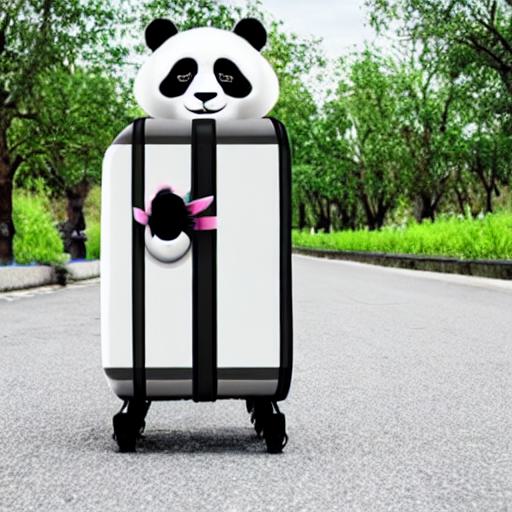} &
        \includegraphics[width=14mm,height=14mm]{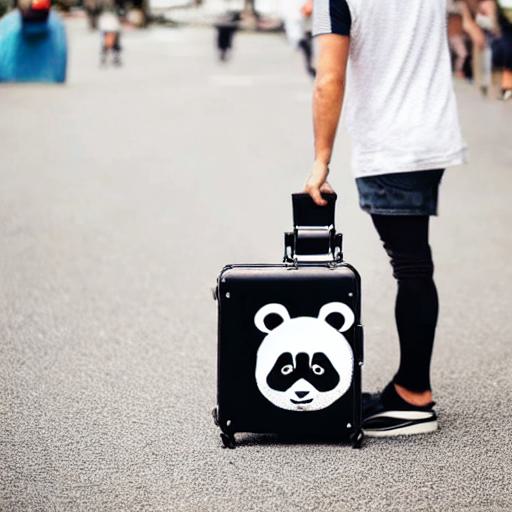}
        \\
    \end{tabular}\\[-2mm]
    \caption{Ablation study for the possession by the implication.
    }
    \label{fig:ablation2}
\end{figure*}

\subsection{Automatic Extraction of Propositions}\label{appendix:automatic_extraction}
In Experiment \ref{ex:4}, we manually extracted propositions from prompts.
However, in most cases, a syntactic dependency parser can be employed to automate this process.
To validate this approach, we used spaCy v3.0 to identify the following:
\begin{itemize}
    \item Nouns for propositions of concurrent existence can be identified as words tagged with \textsc{NOUN} (common noun) and \textsc{PROPN} (proper noun).
    \item Modifier-noun pairs for propositions of one-to-one correspondence can be identified as word pairs linked by grammatical dependencies, including \textsc{amod} (adjectival modifier), \textsc{nmod} (nominal modifier), \textsc{compound} (compound nouns)
          and \textsc{acomp} (adjectival complement).
    \item Possessor-possession pairs for propositions of possession can be identified as subject-object pairs where the verbs indicate possession, namely, `have,' `wear,' `grasp,' and `hold.'
\end{itemize}
For example, from the prompt "Woman wearing a black coat holding up a red cellphone," we successfully extracted:
\begin{itemize}
    \item Nouns: Woman, coat, cellphone
    \item Modifier-noun pairs: [black, coat], [red, cellphone]
    \item Possessor-possession pairs: [Woman, coat], [Woman, cellphone]
\end{itemize}
These results enabled us to automatically create propositions using a simple script.

From six text prompts in Figs.~\ref{fig:experiment4} and \ref{fig:experiment4_additional1}, we successfully extracted all propositions (shown below the images) with two exceptions: the possession relationships in prompts ``A black bird with a red beak'' and ``A white teddy bear with a green shirt and a smiling girl.''
The preposition ``with'' expresses possession in these cases but potentially expresses existence in other cases.
Given that ``with'' can have multiple meanings, more advanced syntactic analysis might be necessary.
Nonetheless, our findings generally support the sufficiency of simple syntactic analysis.

Furthermore, we believe it is crucial for users to explicitly use predicate logic to clarify their intentions that cannot be fully expressed in text.
The meaning of ``with,'' whether indicating a possession relationship or merely concurrent existence, can sometimes be ambiguous even for human readers.
By intentionally employing our proposed method, users can clearly express their intentions and eliminate such ambiguities.

\begin{table}[t]
        \centering
        \footnotesize
        \begin{tabular}{lccc}
            \toprule
            \textbf{Model}                        & \textbf{Similarity}$^\ddagger$             & \textbf{CLIP-IQA} \\
            \midrule
            Stable Diffusion & 0.353 / 0.699                   & 0.765                                  \\
            SynGen                                & 0.371 / 0.736                   & 0.753                                  \\
            Proposed                              & \textbf{0.387} / \textbf{0.745} & \textbf{0.777}                         \\
            \bottomrule
        \multicolumn{4}{l}{$^\ddagger$Text-image similarity and text-text similarity.}
        \end{tabular}
        \caption{Results of three objects and three attributes.}
        \label{tab:three}
\end{table}
\begin{figure}[t]
        \centering
        \footnotesize
        \tabcolsep=.3mm
        \caption{Style-Specifying Keywords.}
        \label{fig:ssk}
        \vspace*{-2mm}
        \begin{tabular}{C{12mm}@{\hspace*{1.5mm}}ccc}
            \raisebox{5.5mm}{\parbox{12mm}{\centering\emph{an apple                              \\and a lion}}}               &
            \includegraphics[width=12mm]{./fig/coexist/a_apple_and_a_lion_73_Predicated.jpg} &
            \includegraphics[width=12mm]{./fig/coexist/a_apple_and_a_lion_40_Predicated.jpg} &
            \includegraphics[width=12mm]{./fig/coexist/a_apple_and_a_lion_62_Predicated.jpg}   \\[-0.5mm]
            \raisebox{5.5mm}{\parbox{12mm}{\centering\emph{a photo of an apple and a lion}}}   &
            \includegraphics[width=12mm]{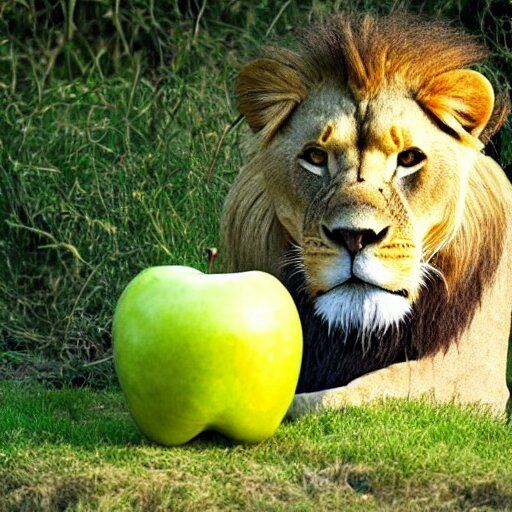}  &
            \includegraphics[width=12mm]{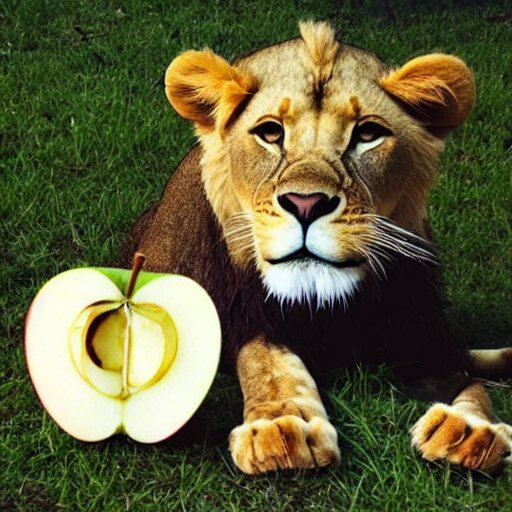}  &
            \includegraphics[width=12mm]{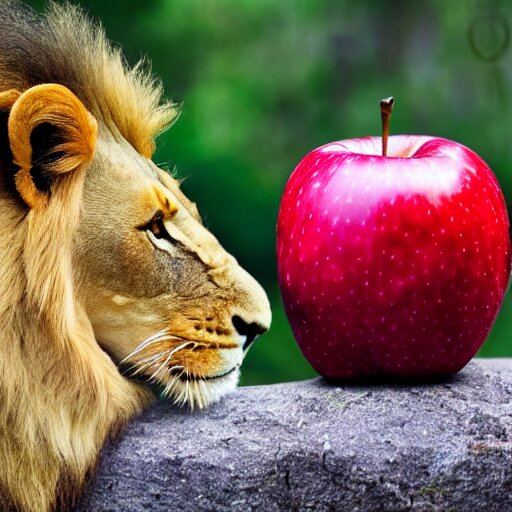}    \\
        \end{tabular}
\end{figure}
\begin{table}[t]
        \centering
        \footnotesize
        \begin{tabular}{lcccc}
            \toprule
                           & \multicolumn{2}{c}{\textbf{Experiment (i)}} & \multicolumn{2}{c}{\textbf{Experiment (ii)}}                                                                \\
            \cmidrule(lr){2-3}\cmidrule(lr){4-5}
            \textbf{Model} & \textbf{Similarity}$^\ddagger$                         & \textbf{CLIP-IQA}       & \textbf{Similarity}$^\ddagger$ & \textbf{CLIP-IQA} \\
            \midrule
            SynGen         & --                                          & --                                           & 72.2/70.1           & 48.9                                   \\
            Proposed       & 67.4/73.0                                   & 55.2                                         & 81.2/73.5           & 54.9                                   \\
            \bottomrule
        \multicolumn{5}{l}{$^\ddagger$Text-image similarity and text-text similarity.}
        \end{tabular}
        \caption{Percentages of Improvement.}
        \label{tab:percentage}
\end{table}

\subsection{Additional Analyses}\label{appendix:analyses}
Our proposed Predicated Diffusion incorporates a variety f loss functions, which might raise concerns about achieving suboptimal solutions and thus potentially diminishing the fidelity and quality of generated images.
However, the results in Figs.~\ref{fig:experiment4}, \ref{fig:experiment4_additional1}, and \ref{fig:experiment4_additional2} visually suggest these concerns are unwarranted.
For a quantitative evaluation, we generated images of three objects and three attributes, using 15 different loss functions under the same experimental settings as those used in Experiment \ref{ex:2}.
The propositions are exemplified in the leftmost column of Fig.~\ref{fig:experiment4_additional2}
The results in Table~\ref{tab:three} confirm that our optimization process is robust and successfully handles multiple loss functions without failure.
Despite the complexity of the scenarios, objective evaluations through CLIP-IQA rated our method as producing the highest quality images, surpassing those generated by SynGen.
Our primary goal is to guide the image generation process rather than strictly fulfill logical constraints, which is why we used fuzzy logic.
As such, we view suboptimal results as acceptable within the context of our framework.

Our Predicated Diffusion yielded cartoonish results in the lion/apple case shown in Fig.~\ref{fig:experiment1}.
However, this does not suggest that Predicated Diffusion degrades the naturalness of the generated images.
Since the dataset comprises both photos and paintings, what style appears is random.
The vanilla Stable Diffusion also generated images of painting style in the cases of bird/cat in Fig.~\ref{fig:experiment1}, and cat/crown and turtle/bear in Fig.~\ref{fig:experiment2_additional}; there is no consistent trend across methods.
Notably, our proposed method supports the use of style-specifying keywords (like ``a photo of X''), which allows users to intentionally choose the image style, as demonstrated in Fig.~\ref{fig:ssk}.

In Table~\ref{tab:percentage}, we present the percentage at which each method improves the fidelity and quality of images compared to vanilla Stable Diffusion.
Our Predicated Diffusion exhibited statistically significant improvements in all metrics, with $p\ll 0.0001$.
Especially, it increased the text-image similarity in 81.2\% of 10,000 images generated in Experiment \ref{ex:2}, indicating its consistent efficacy across a broad range of prompts, rather than being limited to particular ones.
Moreover, it improved the quality of 55\% of images, in contrast to SynGen, which reduced quality in more than 50\% of the cases.

\subsection{Ablation Study}\label{appendix:ablation}
In Section~\ref{sec:loss}, we used implications to represent the modification by adjectives and the possession of objects as $\forall x.\,\prop{Noun}(x)\rightarrow\prop{Adjective}(x)$ and $\forall x.\,\prop{Object}(x)\rightarrow\prop{Subject}(x)$, respectively.
In this section, we explored the effects of reversing the direction of these implications as an ablation study.

Figure~\ref{fig:ablation1} summarizes the results for modification.
In the first row, a noun implicates an adjective, indicating that the object specified by the noun is uniformly colored by the hue specified by the adjective.

This implication allows other objects, such as the background, to share the same color, making it suitable for representing modifications in general.
As another example, consider the prompt ``a sunbathed car,'' which indicates that the car should be depicted as sunbathed and allows other objects to also appear sunbathed.
Conversely, when an adjective implicates a noun, the area colored by the adjective becomes a subset of that of the object, suggesting partial coloring of the object, as shown in the second row.
With the biimplication in the last row, the color and object regions perfectly overlap.
Therefore, biimplication is preferable to avoid attribute leakage.
A clear correspondence can be found between the semantics of propositions and the generated results.

Figure~\ref{fig:ablation2} summarizes the results for possession.
In the first row, a grammatical object implicates a grammatical subject, and the results show the subjects in possession of the objects.
Conversely, the second row shows that when the subject implicates the object, the subject becomes part of the object.
For example, instead of ``A monkey having a bag,'' the situation resembles ``A bag envelops a monkey.''
Similarly, instead of ``A panda having a suitcase,'' the scene is more like ``A panda serves as a pattern on the suitcase.''
In cases of biimplication, the subject and object are often mixed together to form one object.
Thus, implicating the subject by the object most accurately represents the subject in possession of the object.

\begin{figure*}[p]
    \tabcolsep=0.2mm
    \scriptsize
    \centering
    \begin{tabular}{c@{\hspace{1mm}}C{14mm}C{14mm}C{14mm}@{\hspace{2mm}}C{14mm}C{14mm}C{14mm}@{\hspace{2mm}}C{14mm}C{14mm}C{14mm}}
                                                                                                                                           &
        \multicolumn{3}{c}{\parbox{42mm}{\centering\emph{A black bird with red beak}}}                                                     &
        \multicolumn{3}{c}{\parbox{42mm}{\centering\emph{A white teddy bear with a green shirt and~a~smiling girl}}}                       &
        \multicolumn{3}{c}{\parbox{42mm}{\centering\emph{A baby with green hair laying in~a~black~blanket next to a teddy bear}}}

        \\[1mm]
        \rotatebox[origin=c]{90}{\parbox{28mm}{\centering Stable Diffusion XL v1.0}}                                                       &
        \includegraphics[width=14mm,height=14mm]{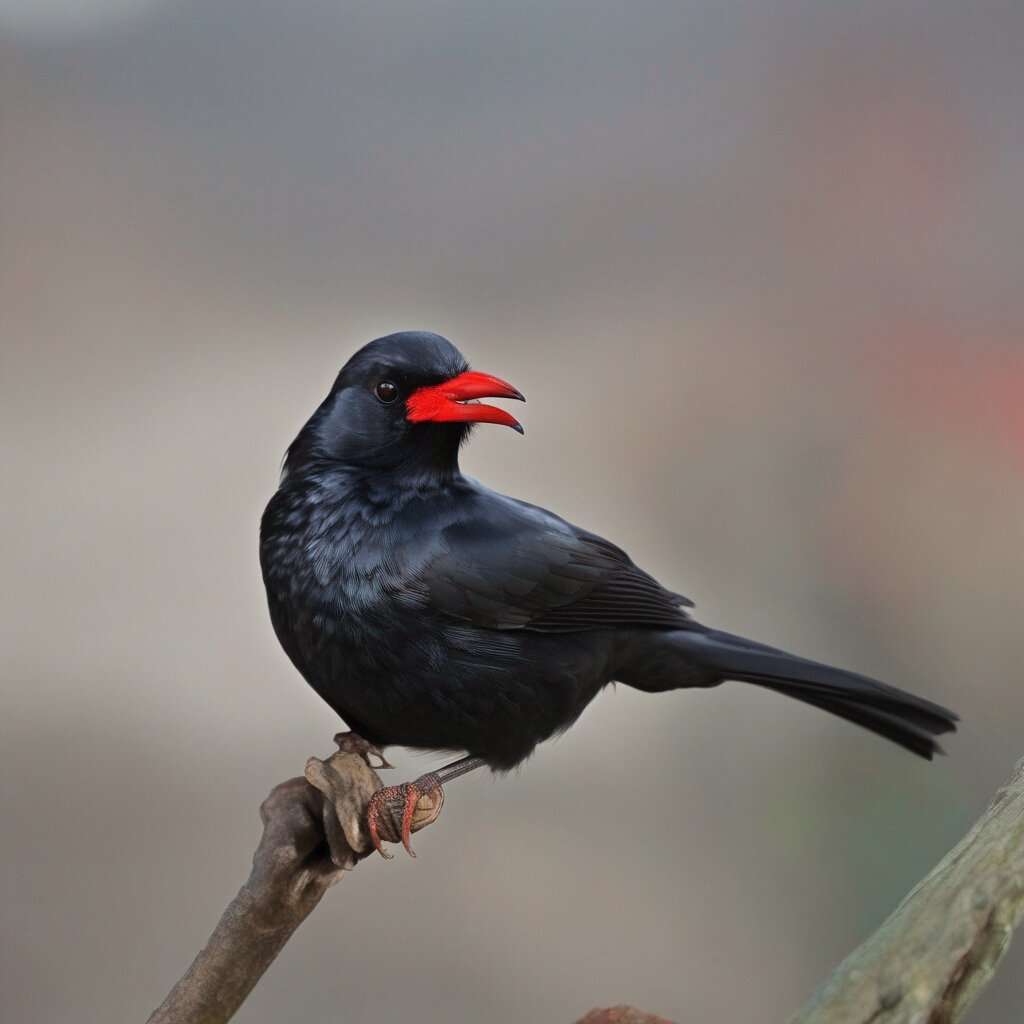}
        \includegraphics[width=14mm,height=14mm]{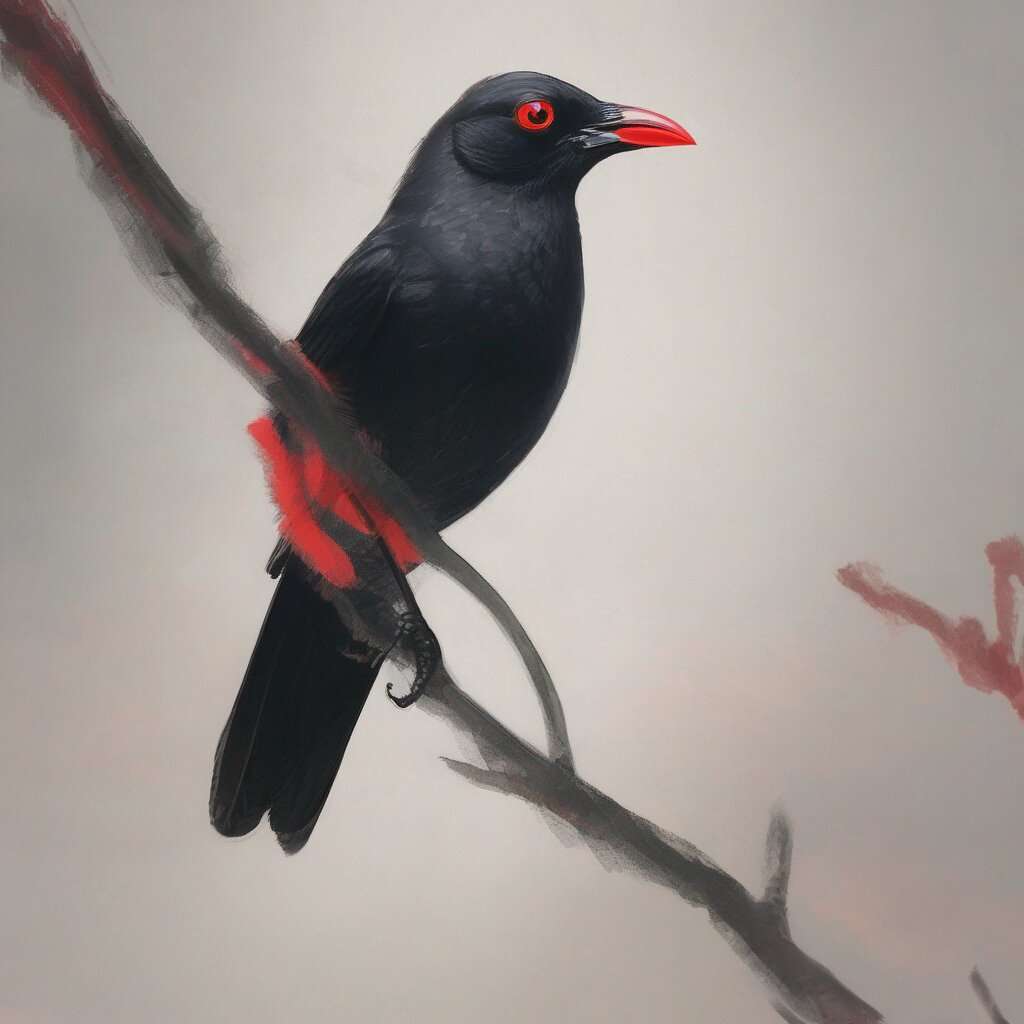}                                           &
        \includegraphics[width=14mm,height=14mm]{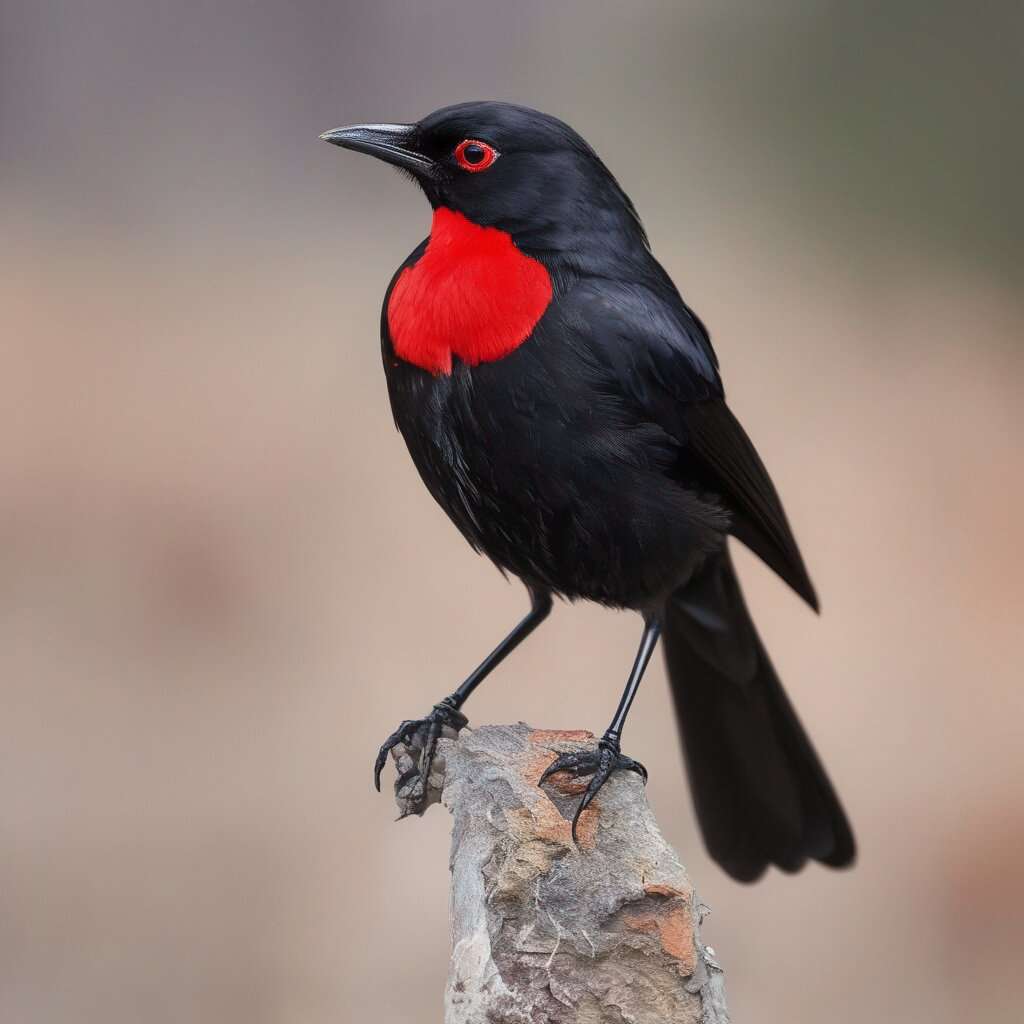}
        \includegraphics[width=14mm,height=14mm]{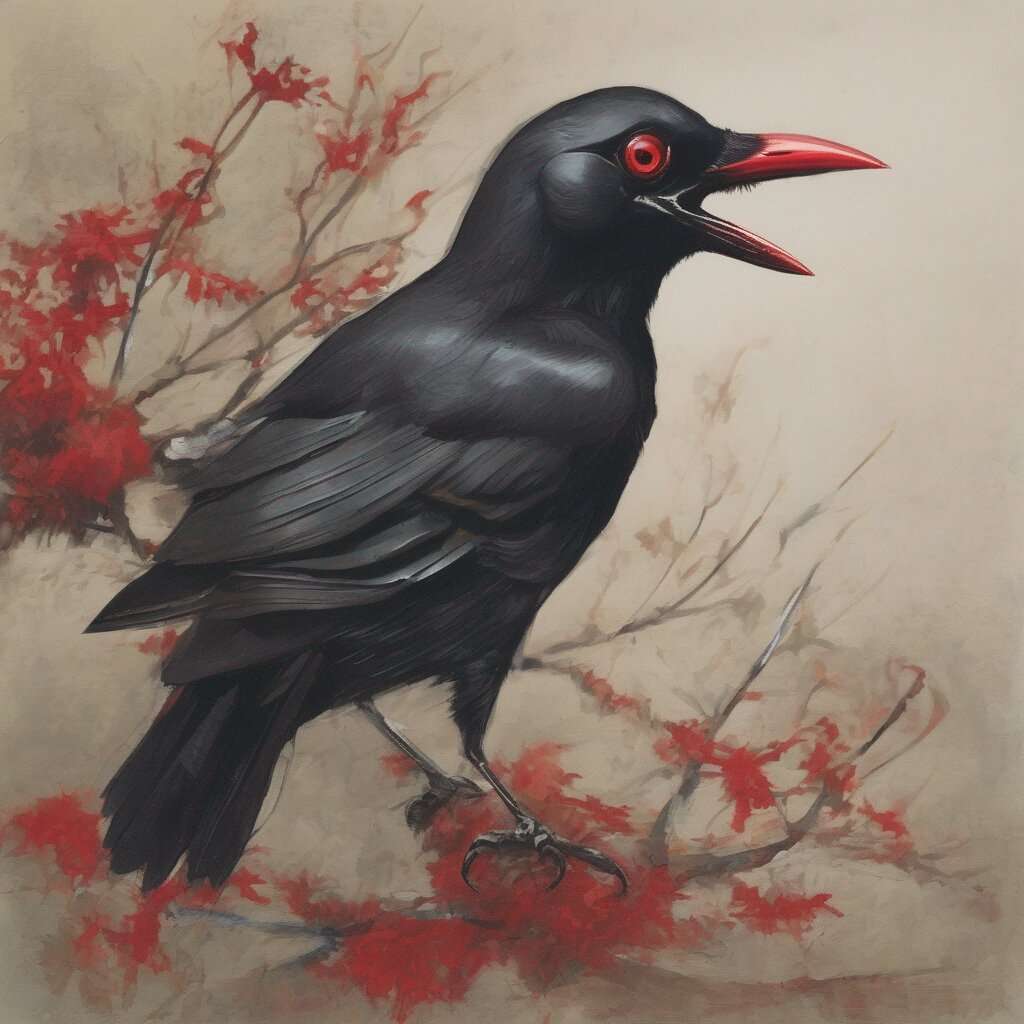}                                           &
        \includegraphics[width=14mm,height=14mm]{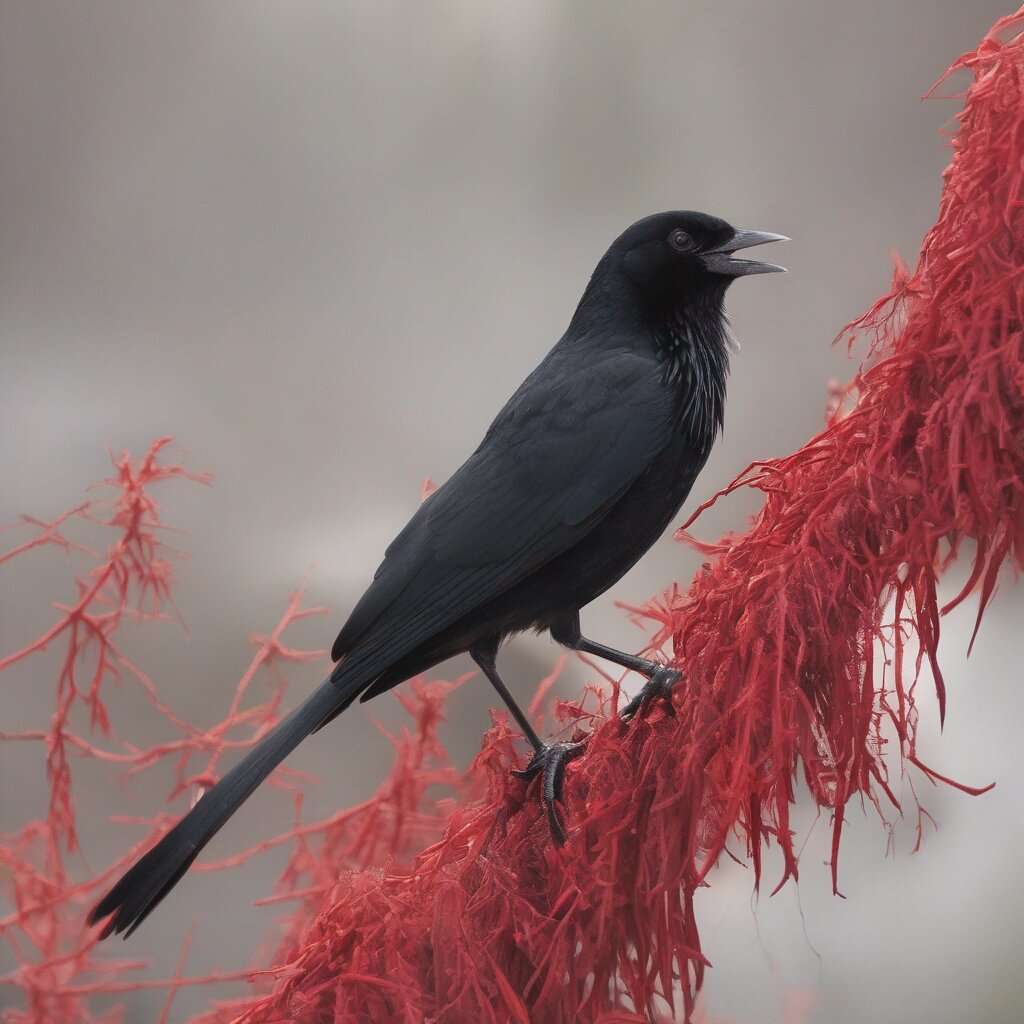}
        \includegraphics[width=14mm,height=14mm]{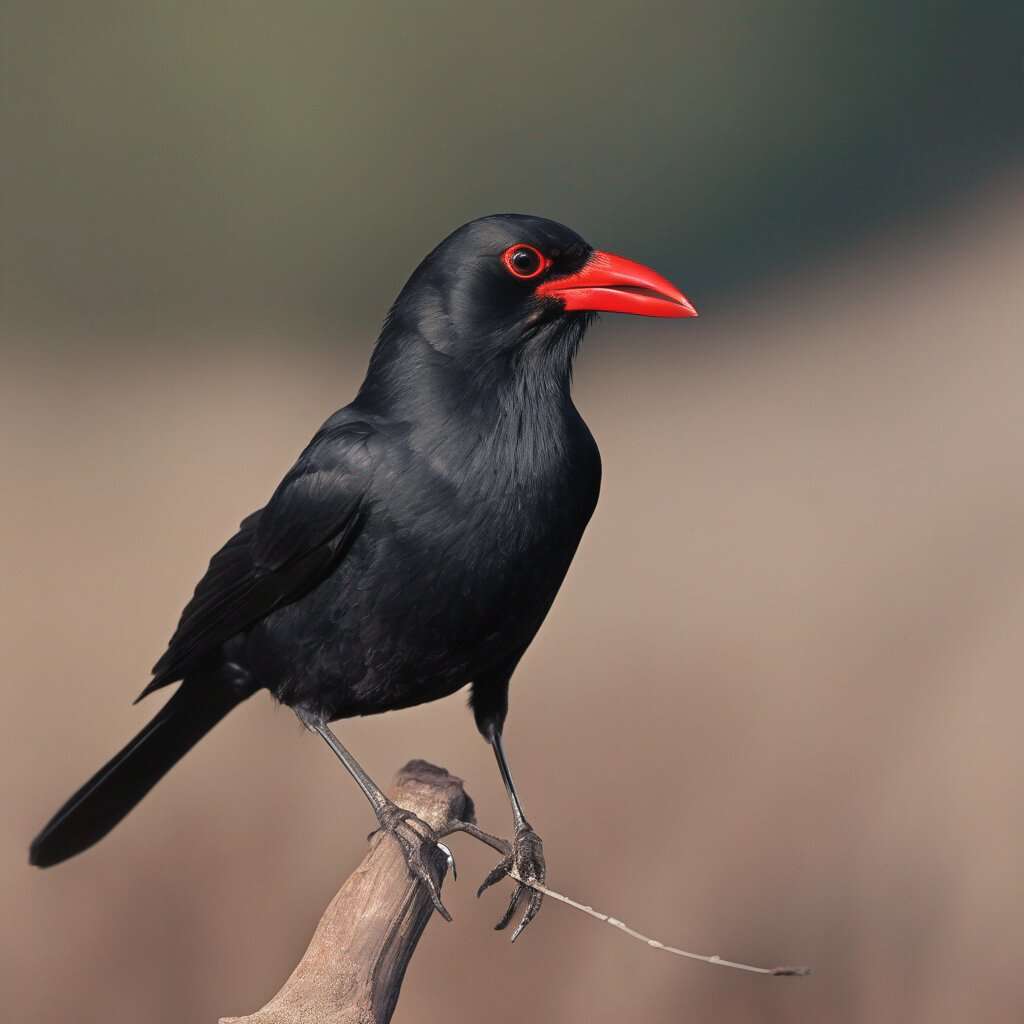}                                           &
        \includegraphics[width=14mm,height=14mm]{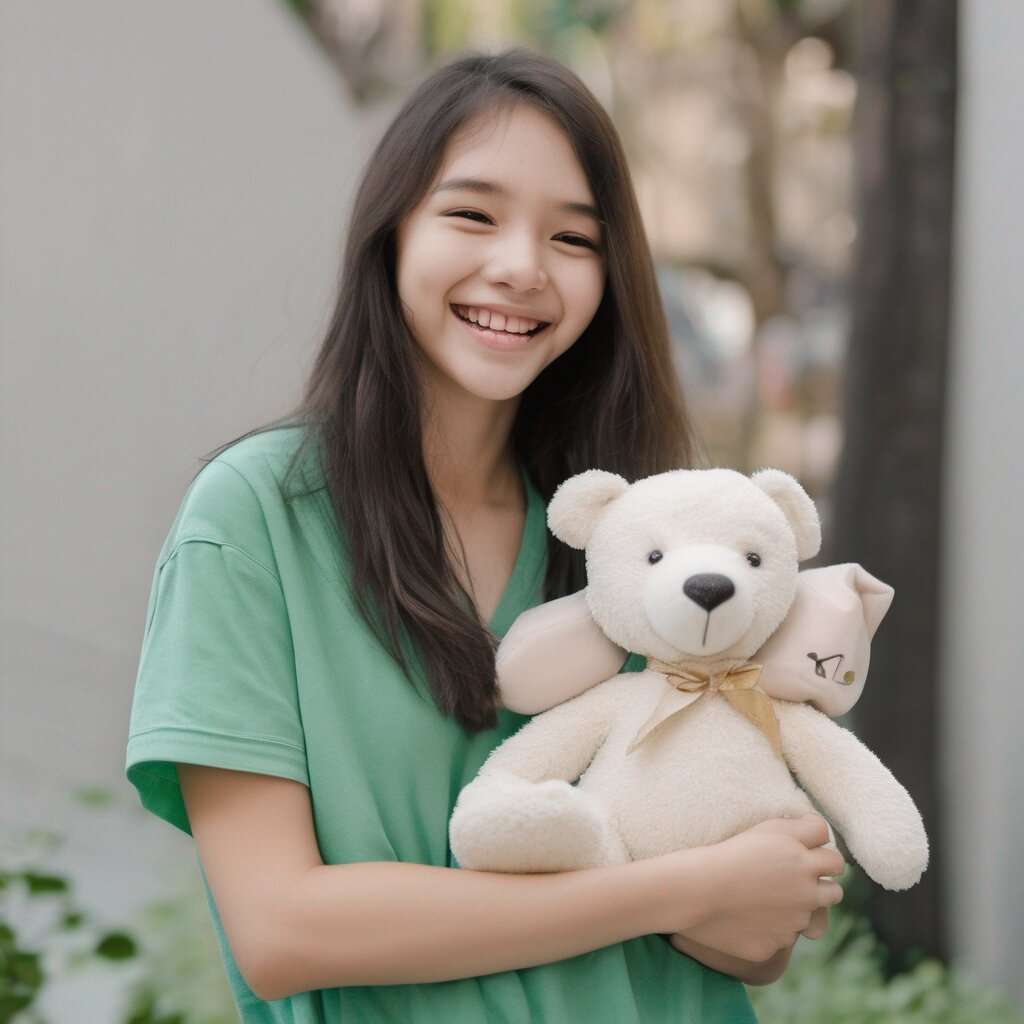}
        \includegraphics[width=14mm,height=14mm]{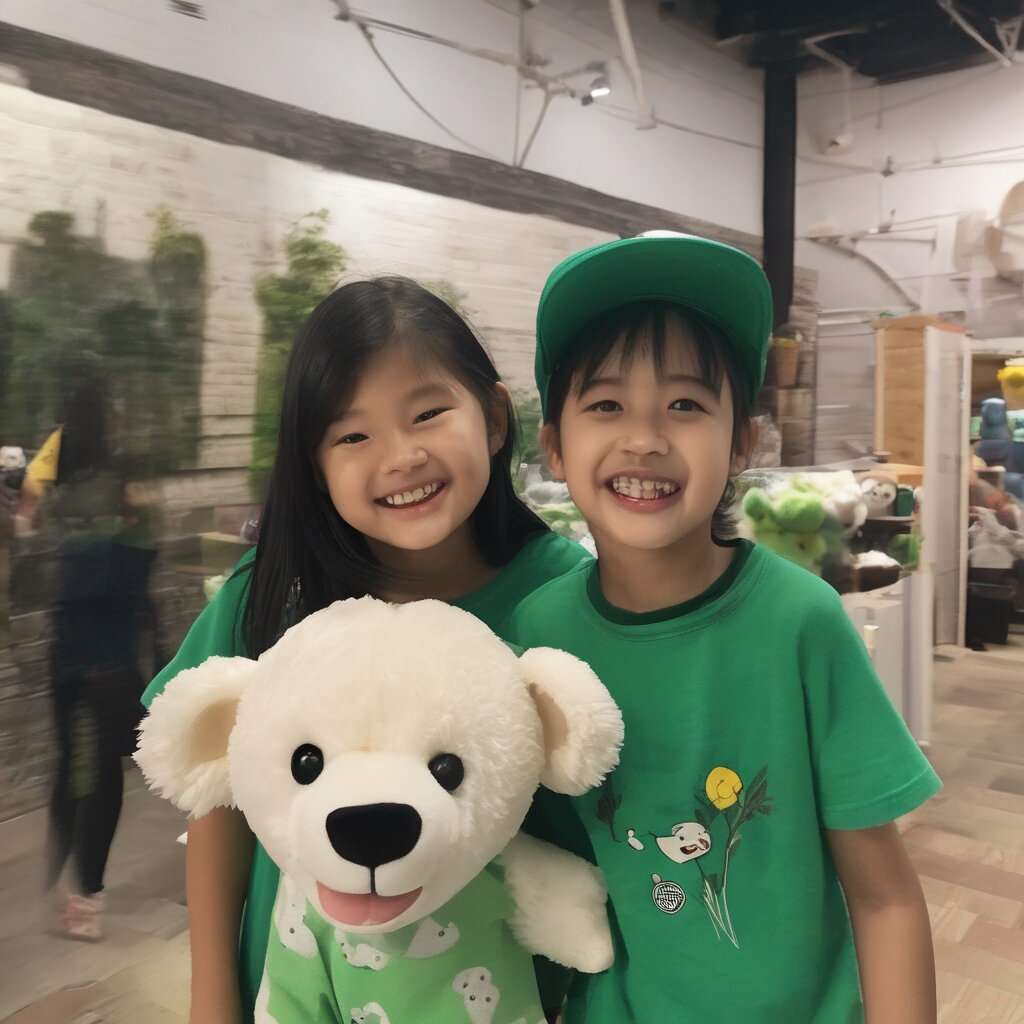}              &
        \includegraphics[width=14mm,height=14mm]{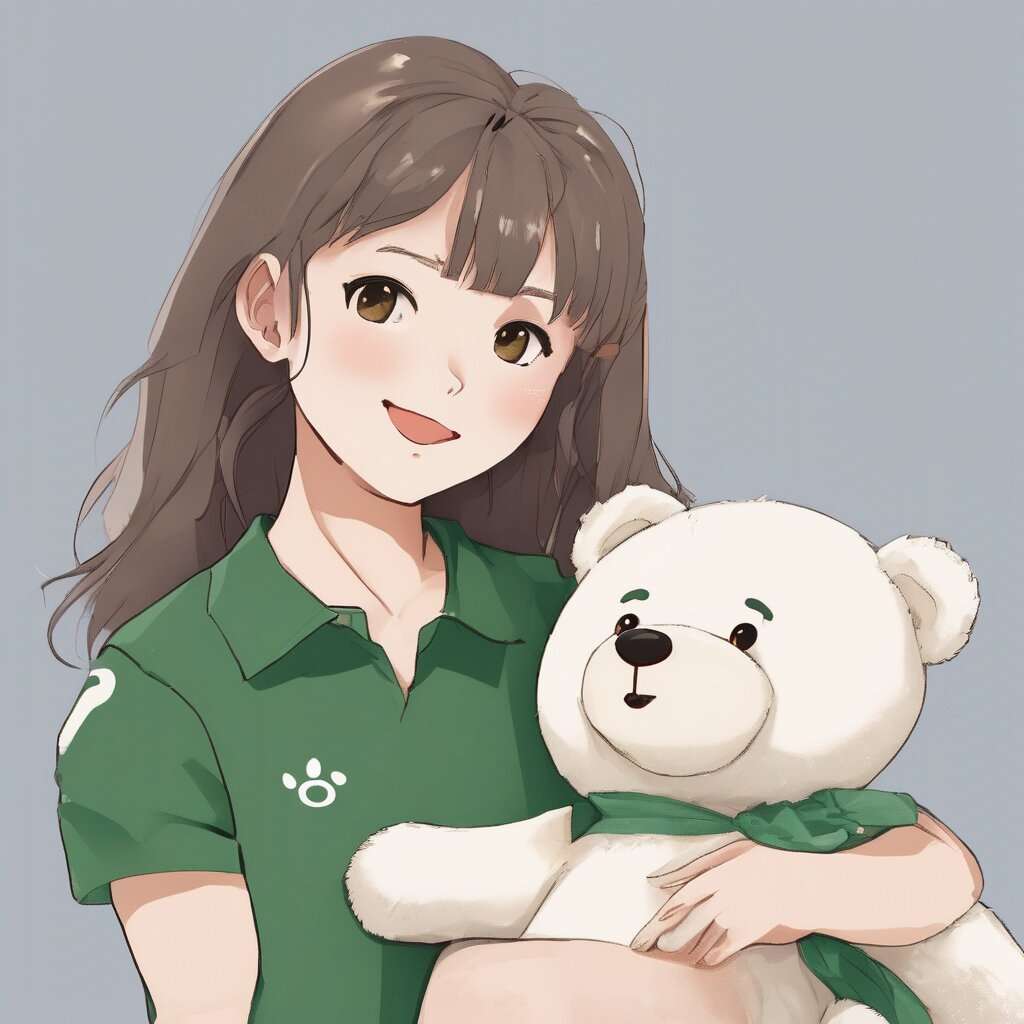}
        \includegraphics[width=14mm,height=14mm]{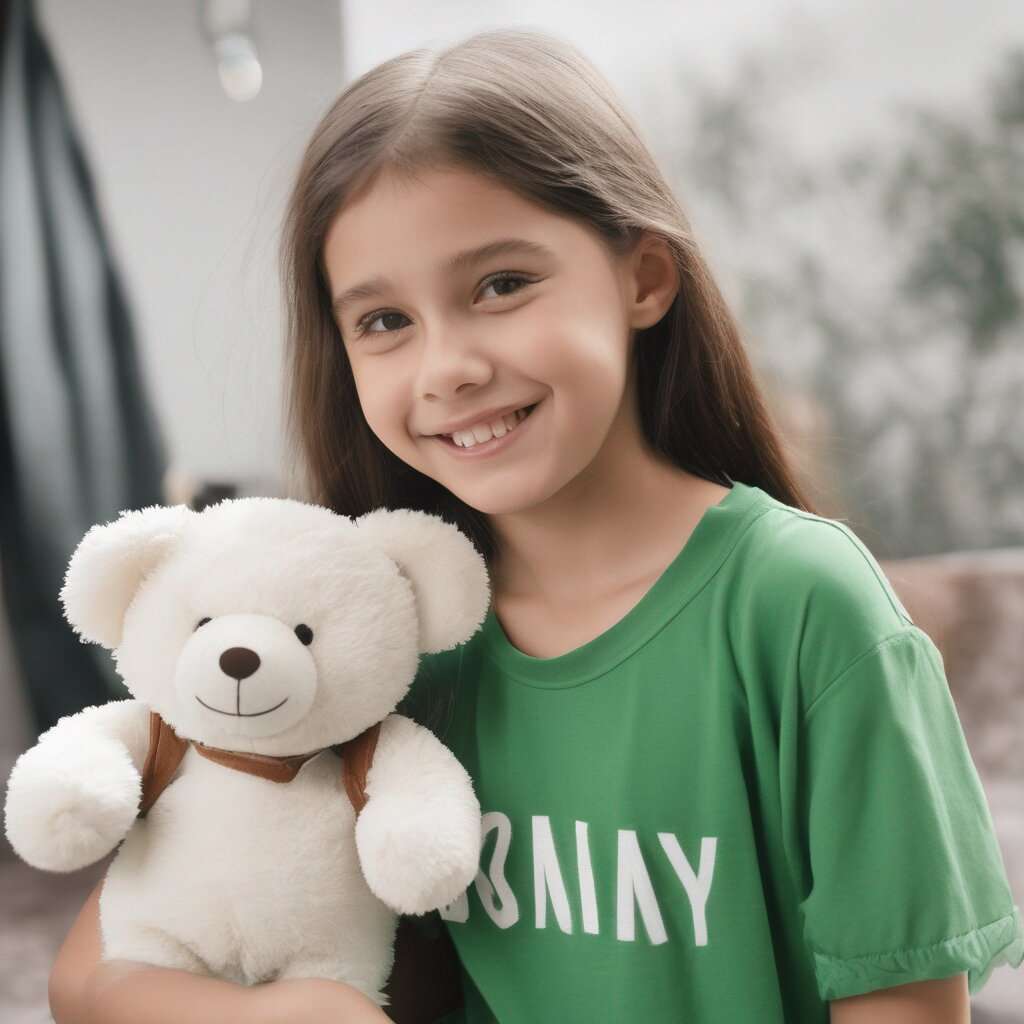}              &
        \includegraphics[width=14mm,height=14mm]{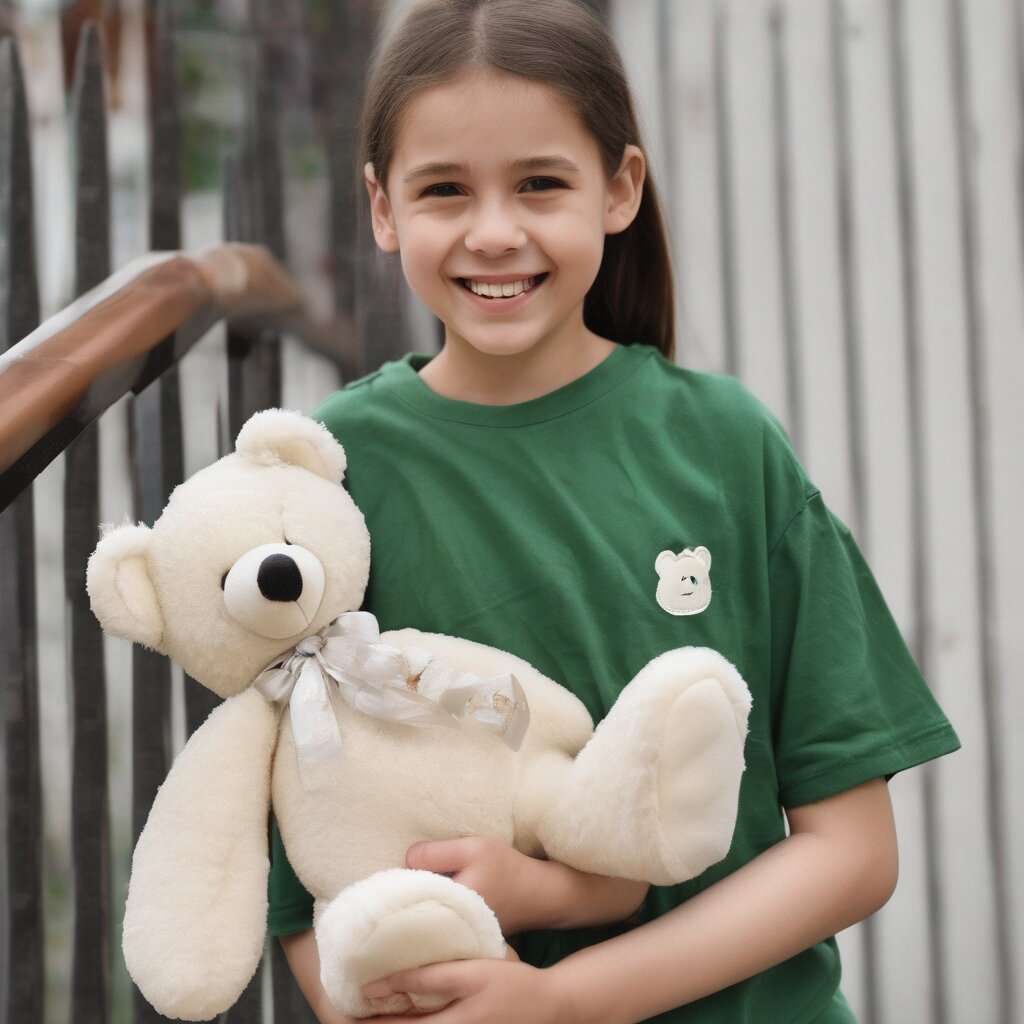}
        \includegraphics[width=14mm,height=14mm]{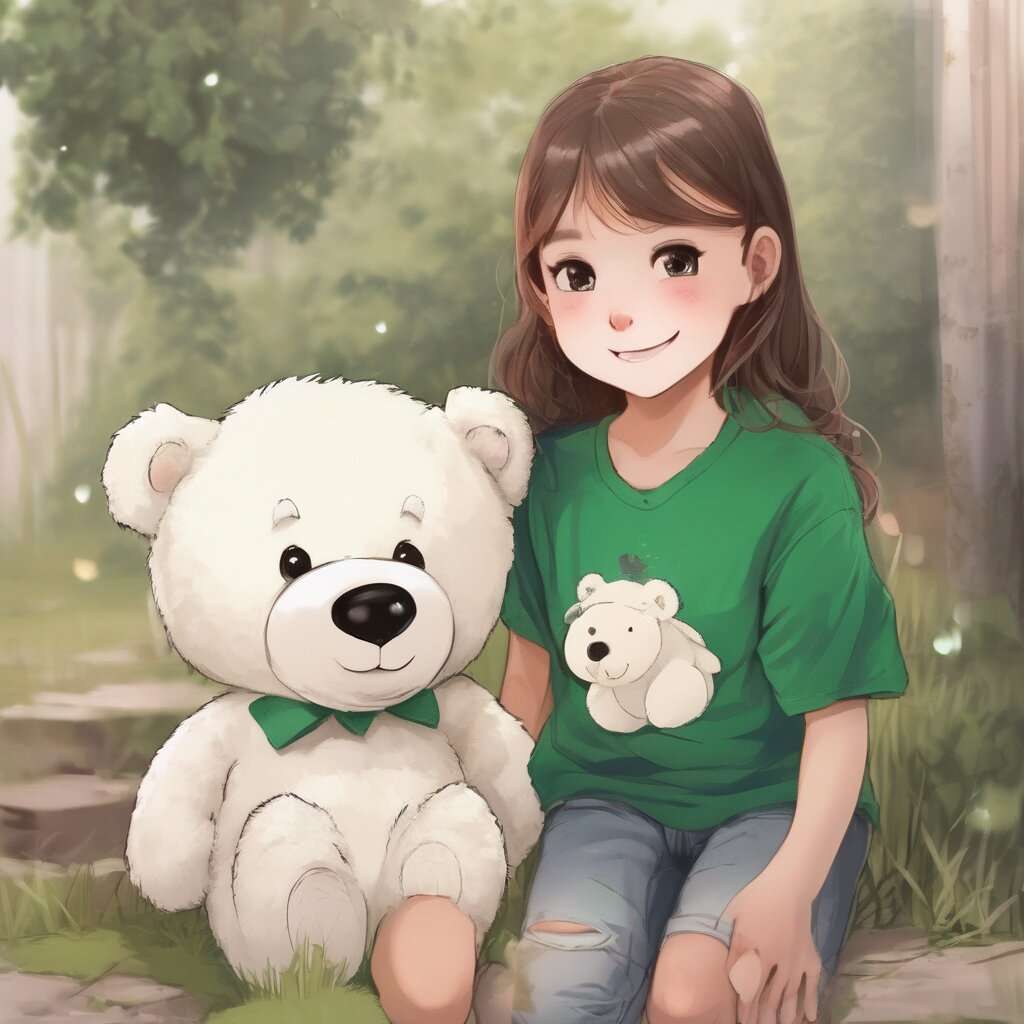}              &
        \includegraphics[width=14mm,height=14mm]{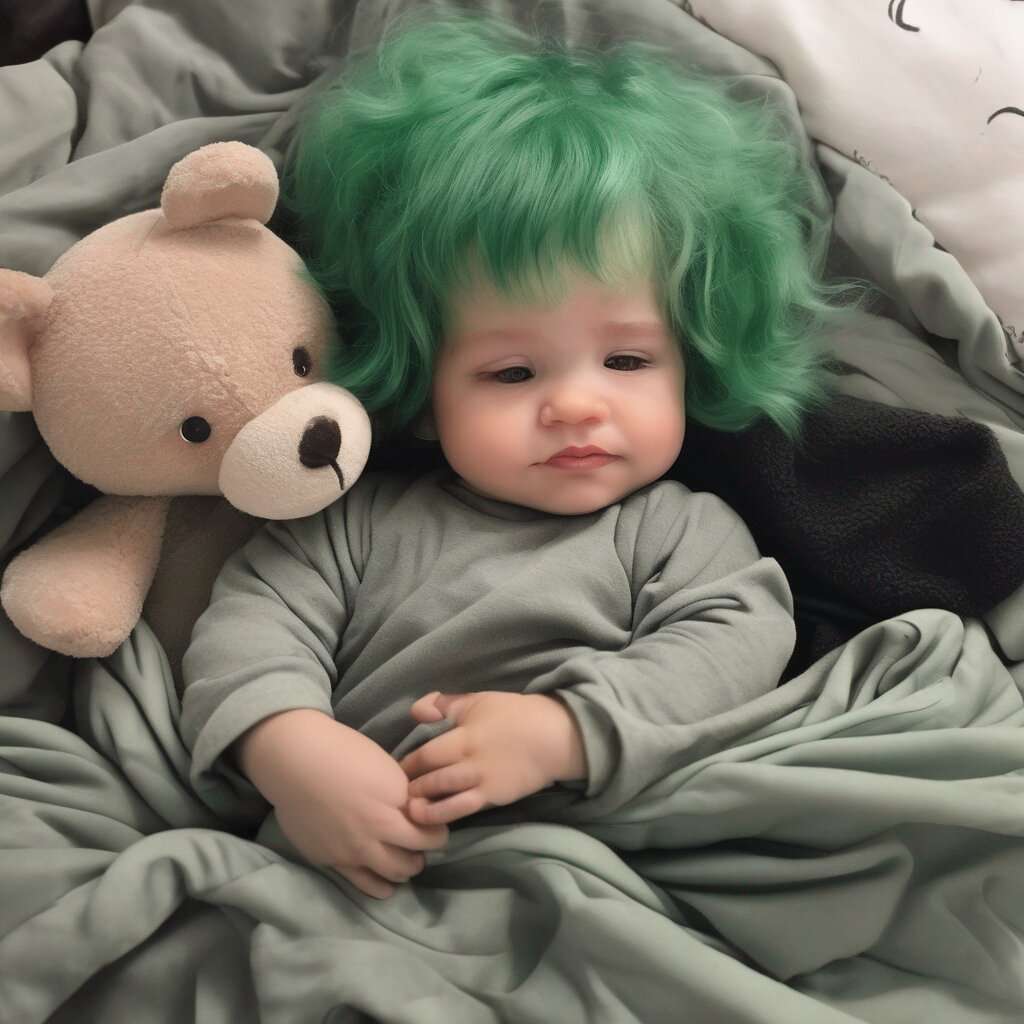}
        \includegraphics[width=14mm,height=14mm]{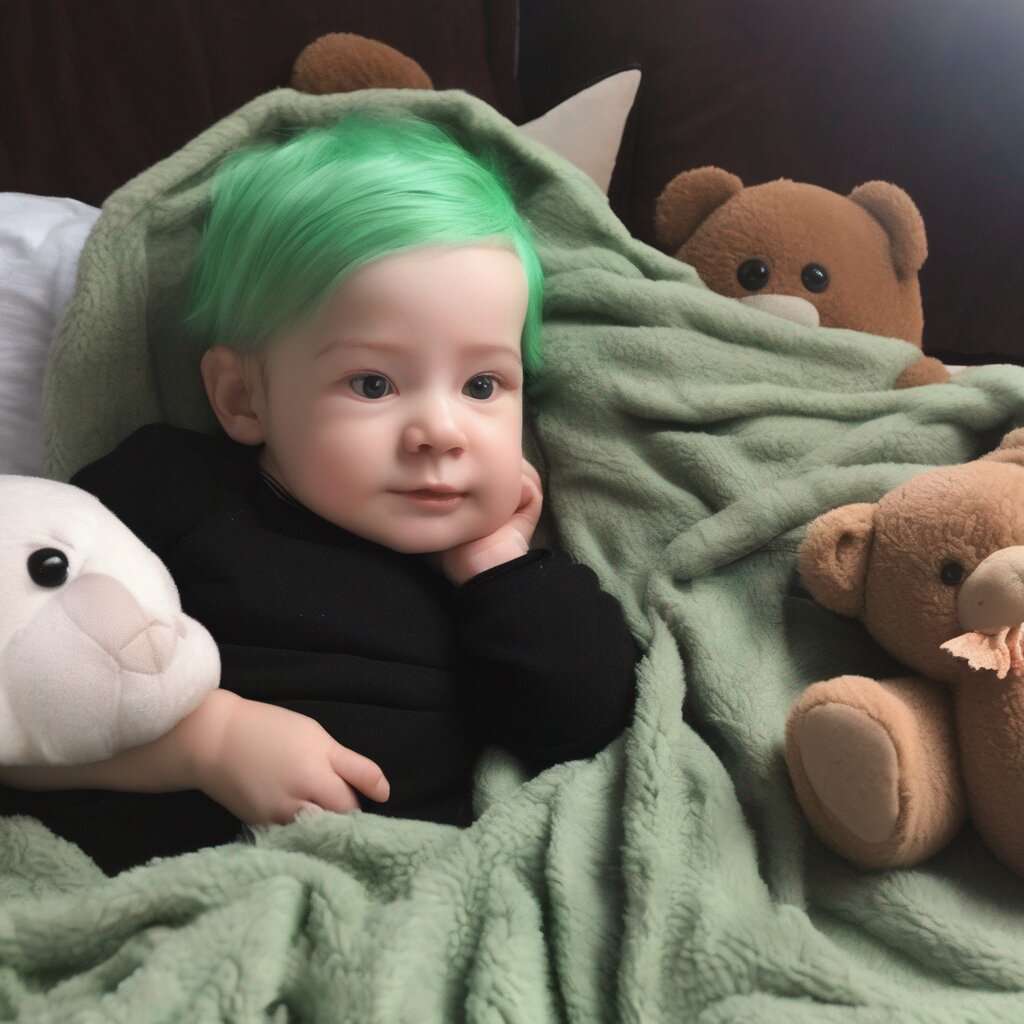} &
        \includegraphics[width=14mm,height=14mm]{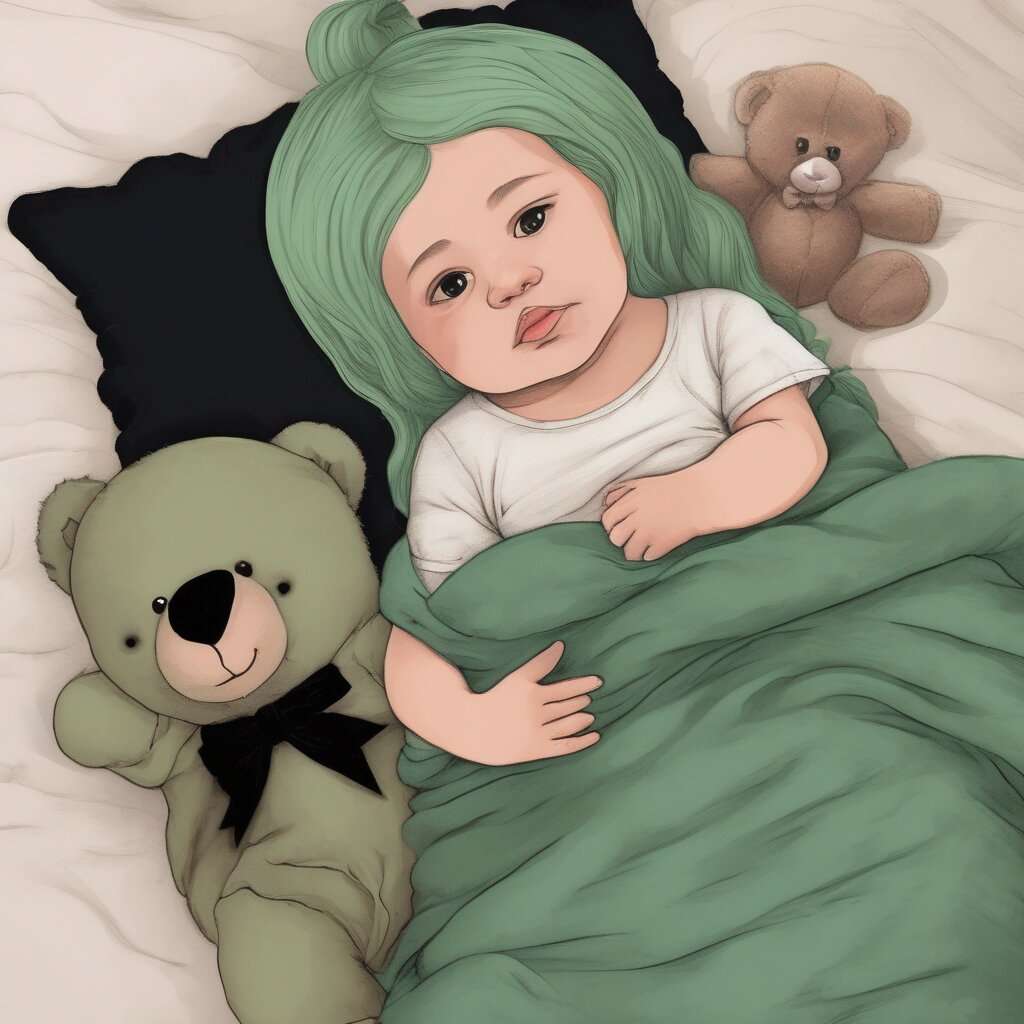}
        \includegraphics[width=14mm,height=14mm]{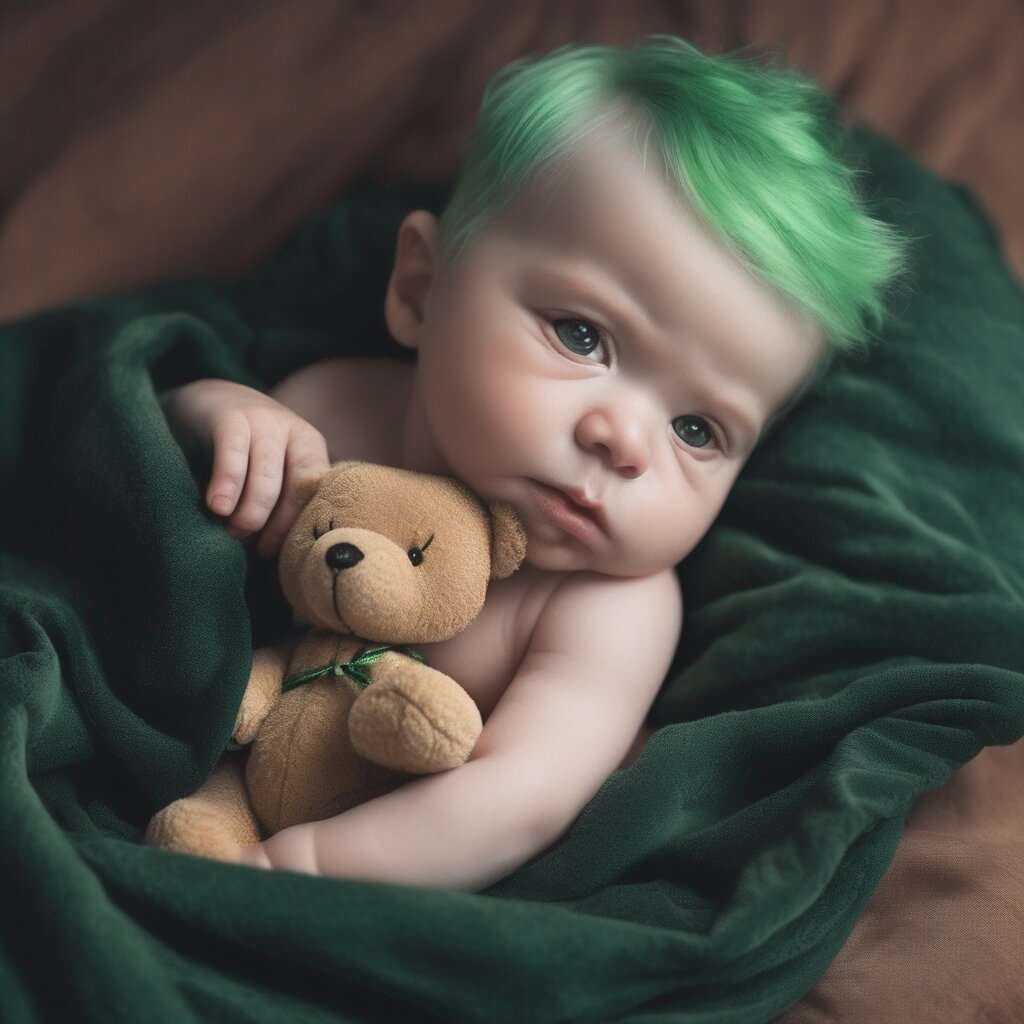} &
        \includegraphics[width=14mm,height=14mm]{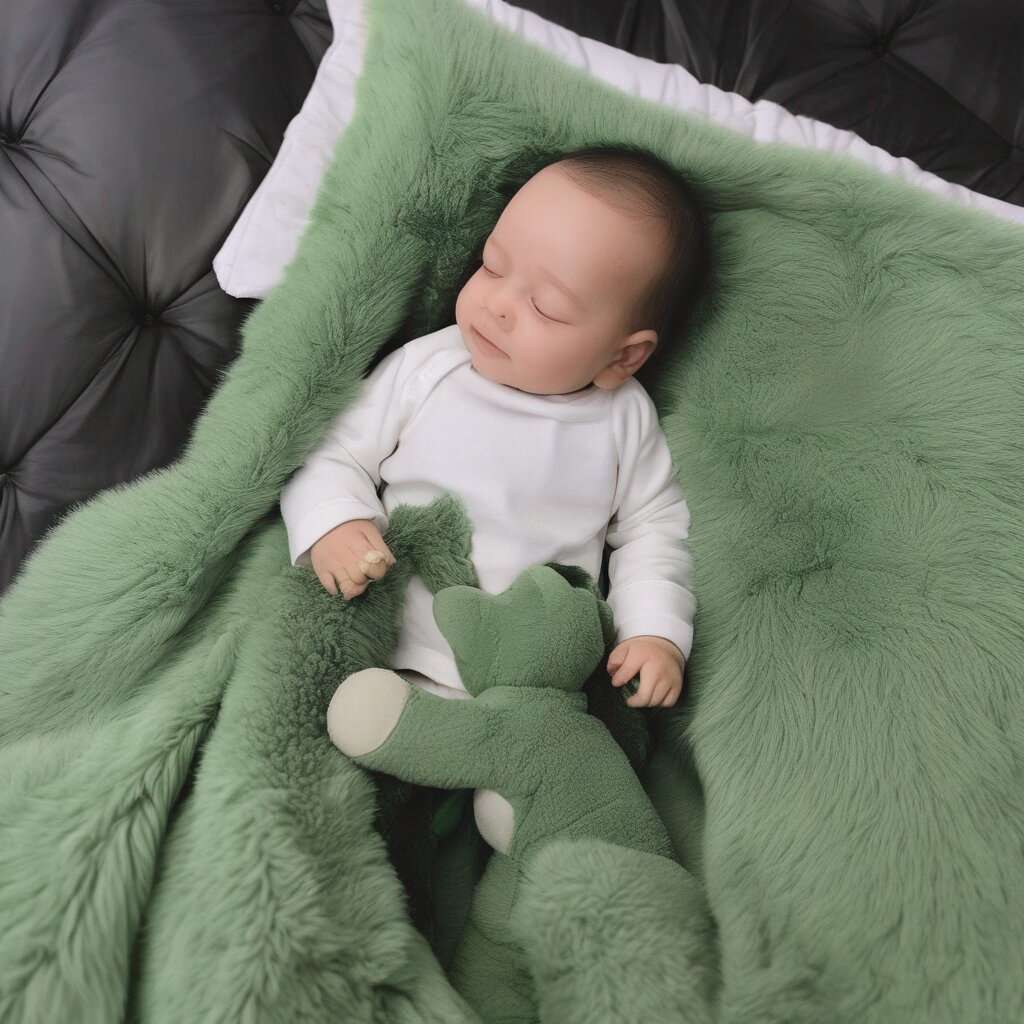}
        \includegraphics[width=14mm,height=14mm]{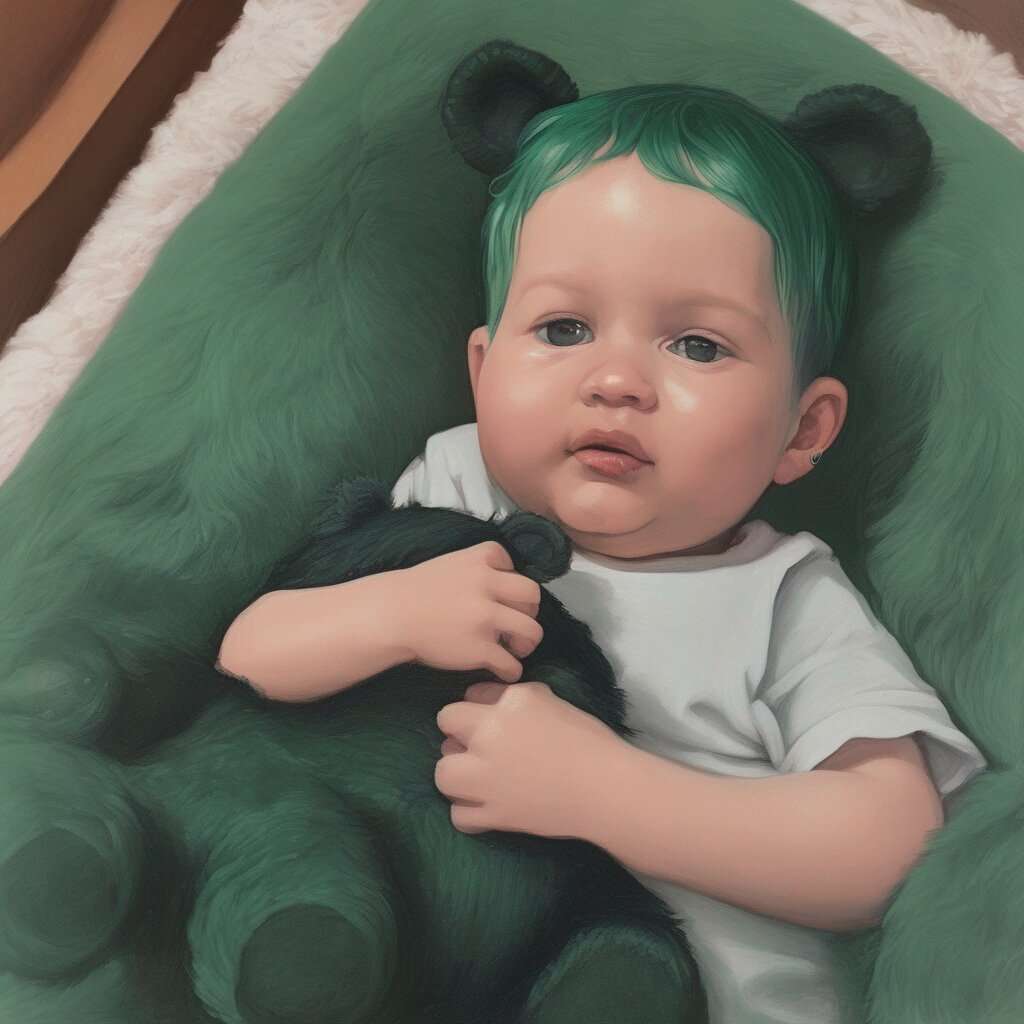}
        \\[-0.3mm]
        \rotatebox[origin=c]{90}{\parbox{28mm}{\centering Midjourney v5.2}}                                                                &
        \includegraphics[width=14mm,height=14mm]{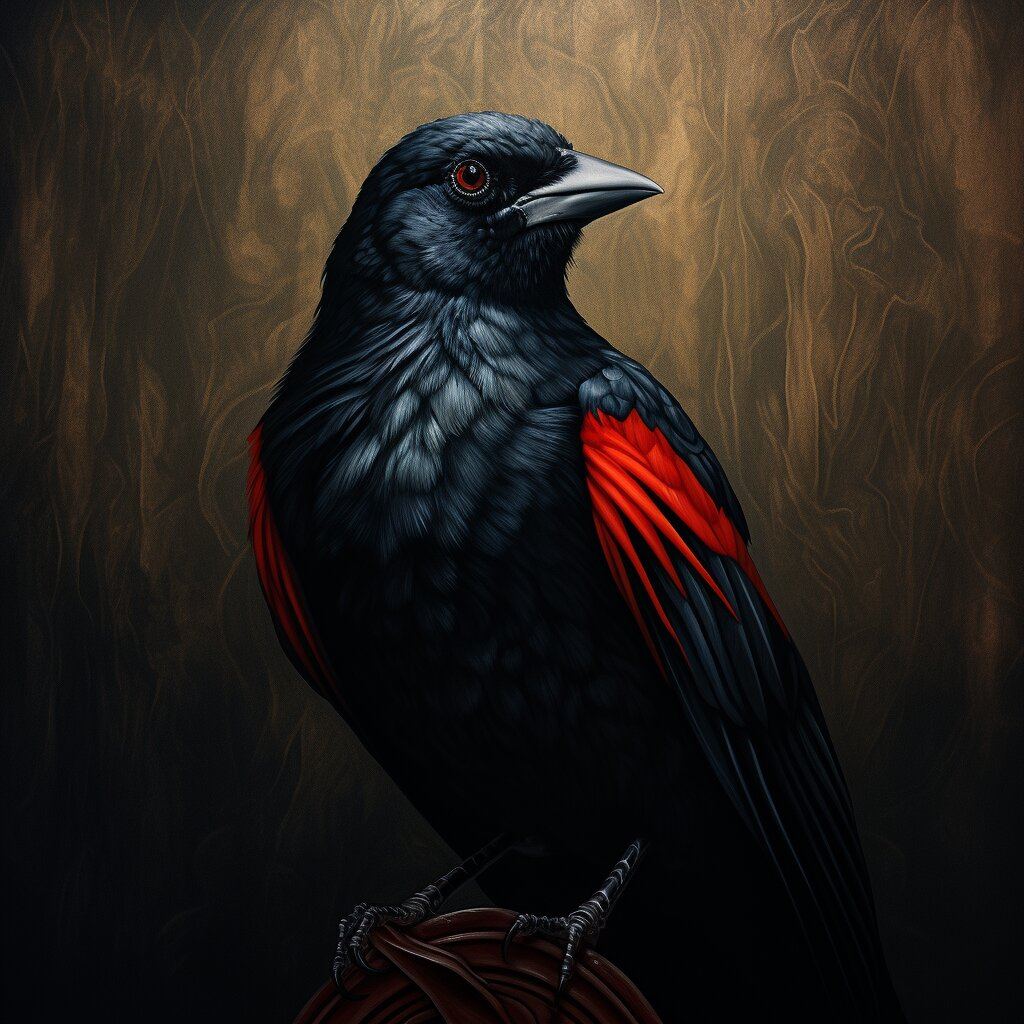}
        \includegraphics[width=14mm,height=14mm]{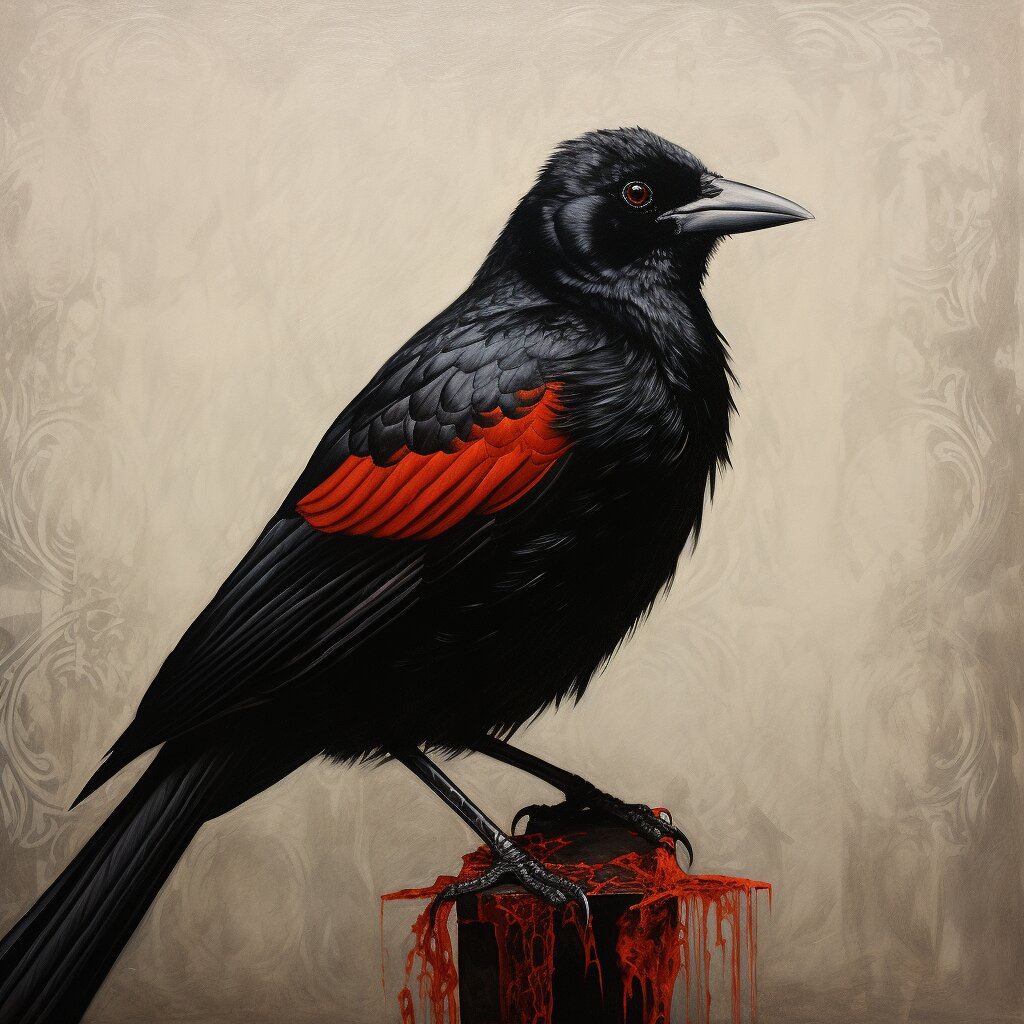}                                     &
        \includegraphics[width=14mm,height=14mm]{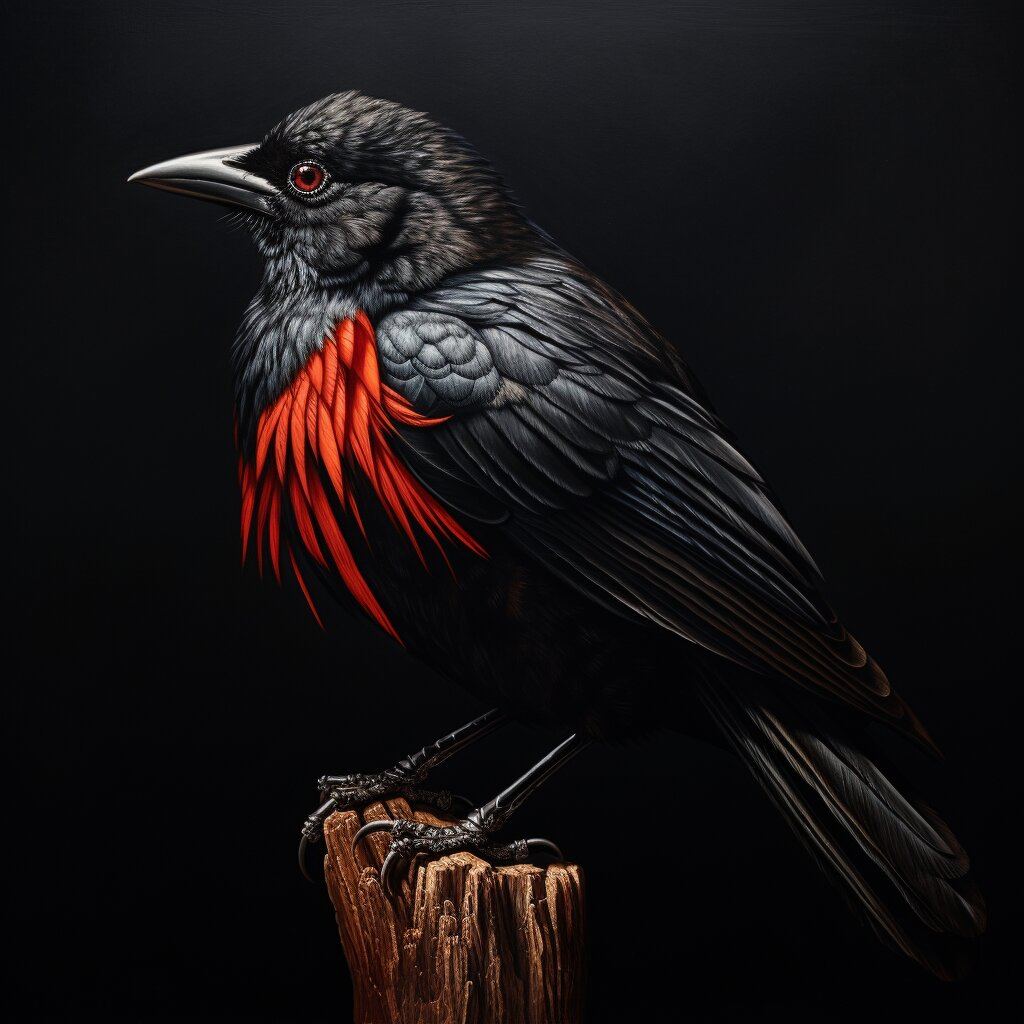}
        \includegraphics[width=14mm,height=14mm]{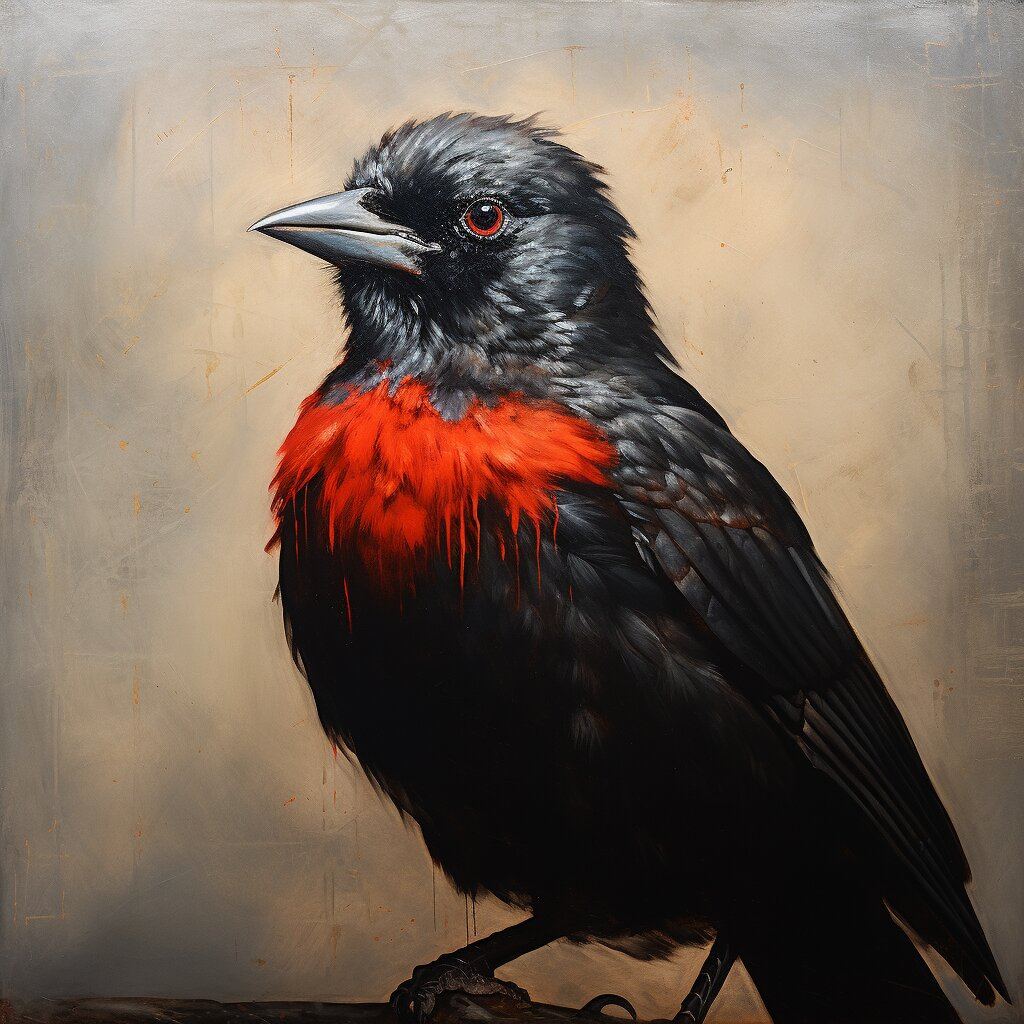}                                     &
        \includegraphics[width=14mm,height=14mm]{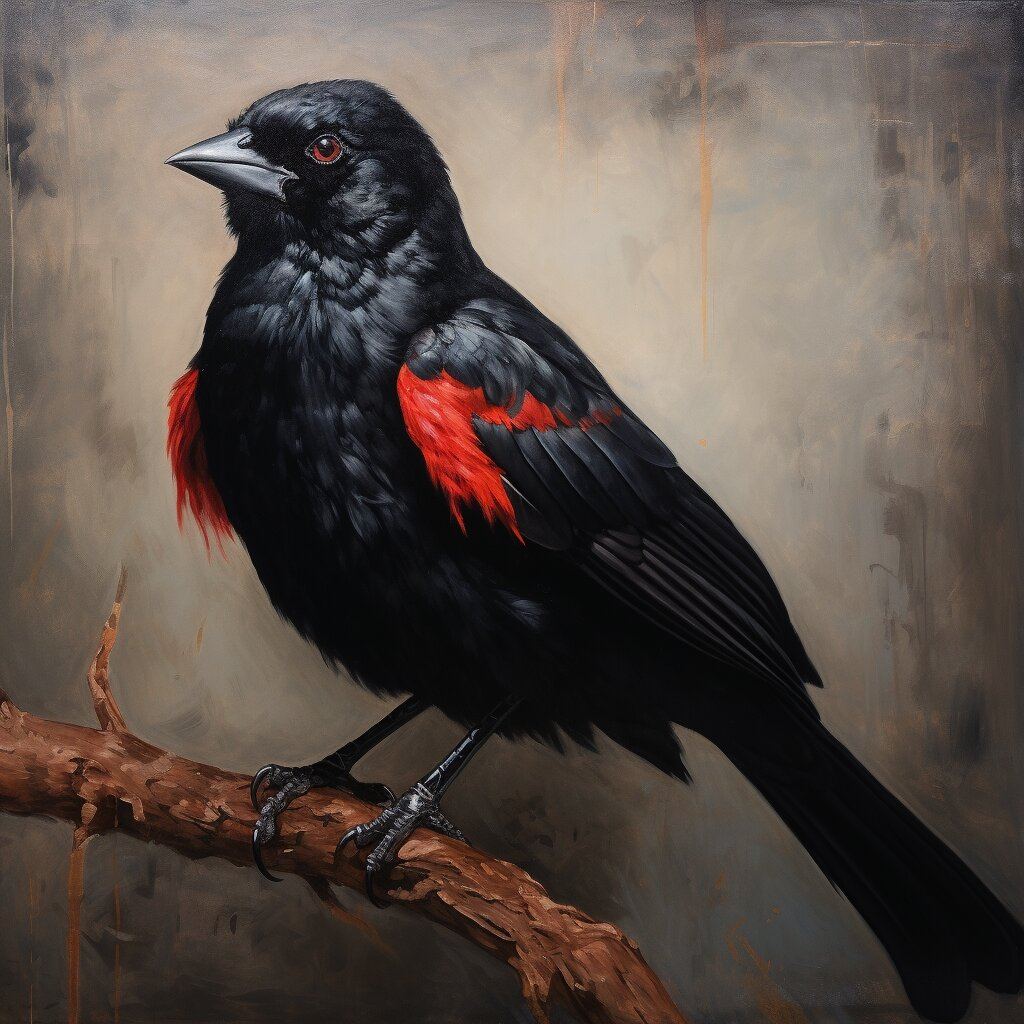}
        \includegraphics[width=14mm,height=14mm]{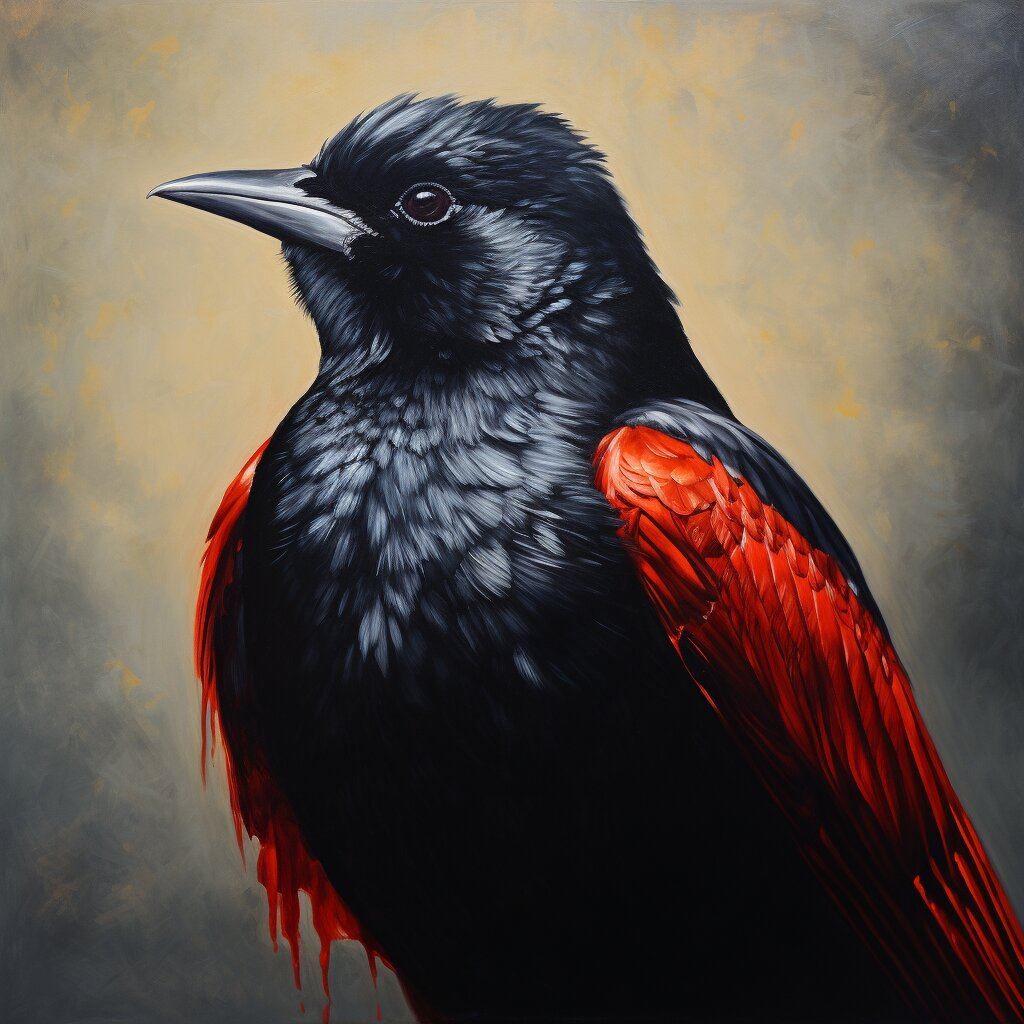}                                     &
        \includegraphics[width=14mm,height=14mm]{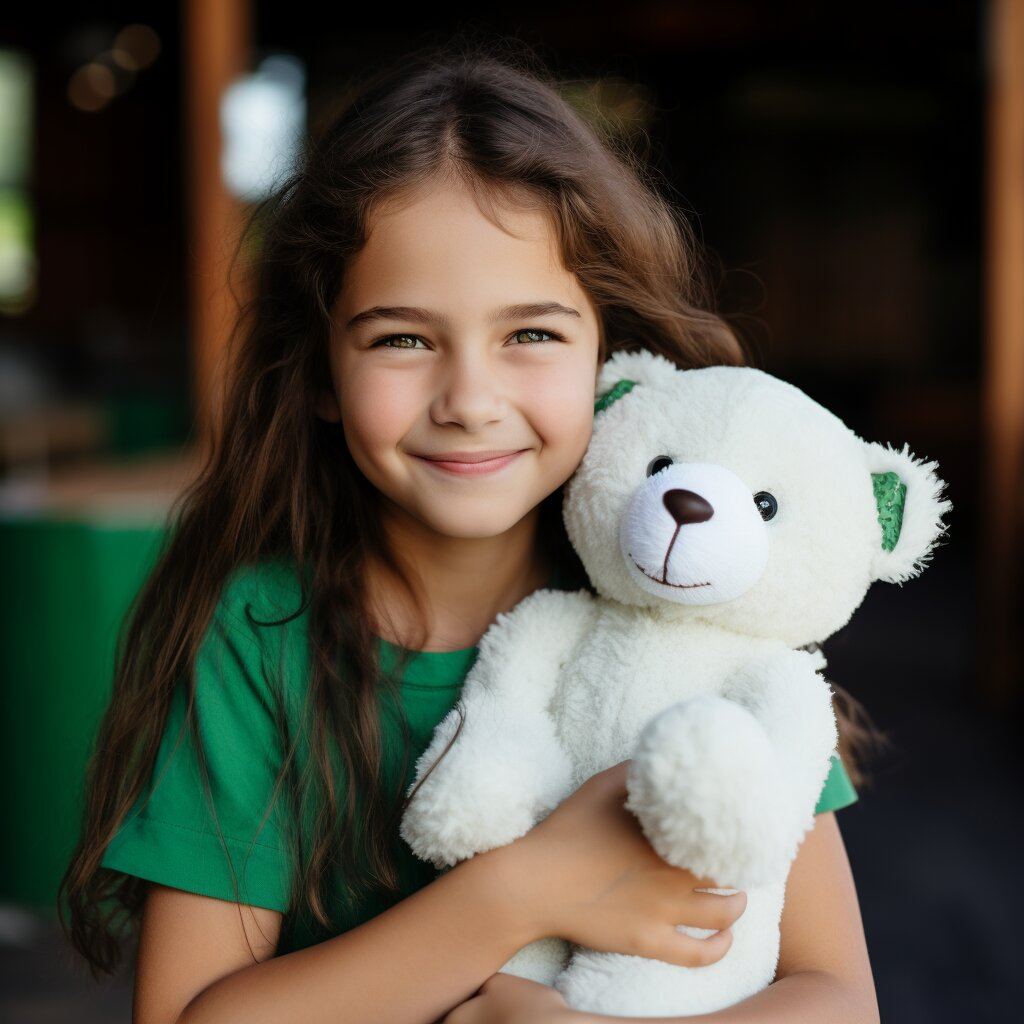}
        \includegraphics[width=14mm,height=14mm]{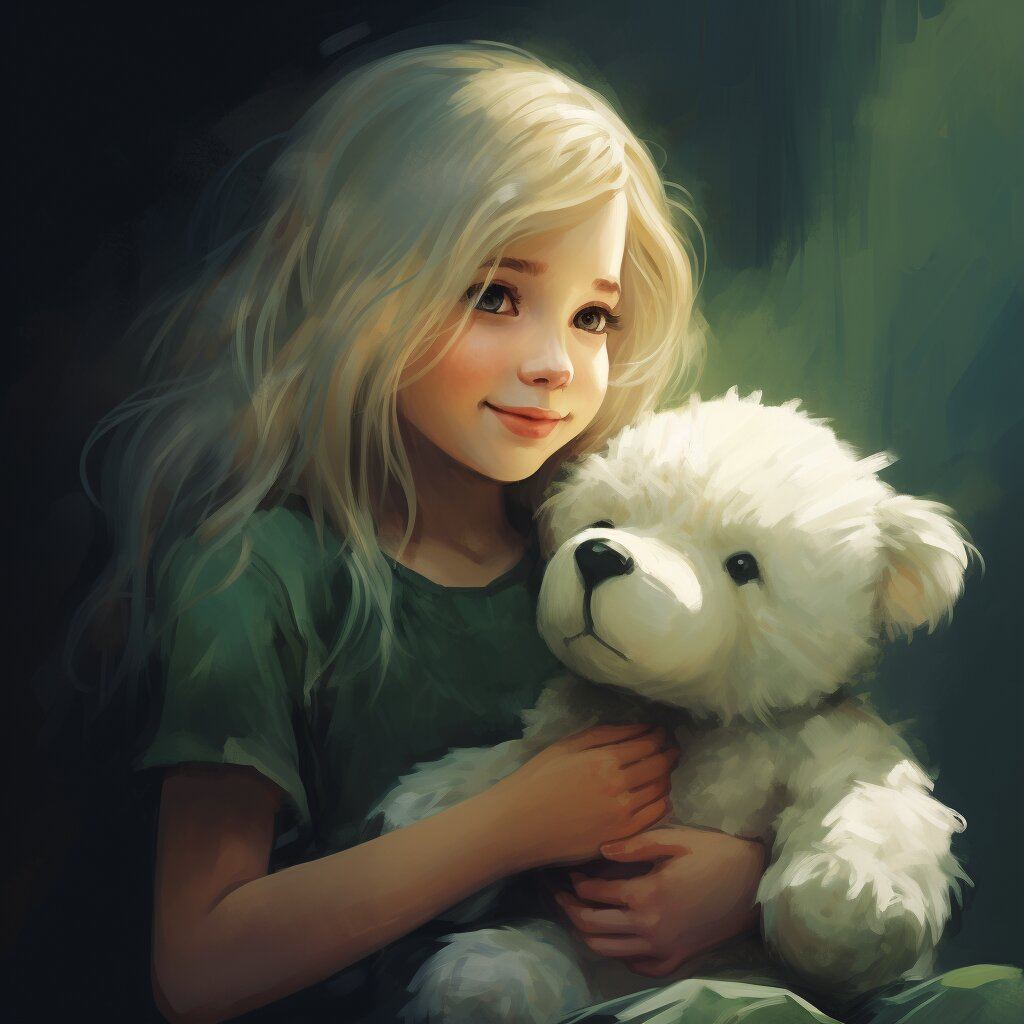}                 &
        \includegraphics[width=14mm,height=14mm]{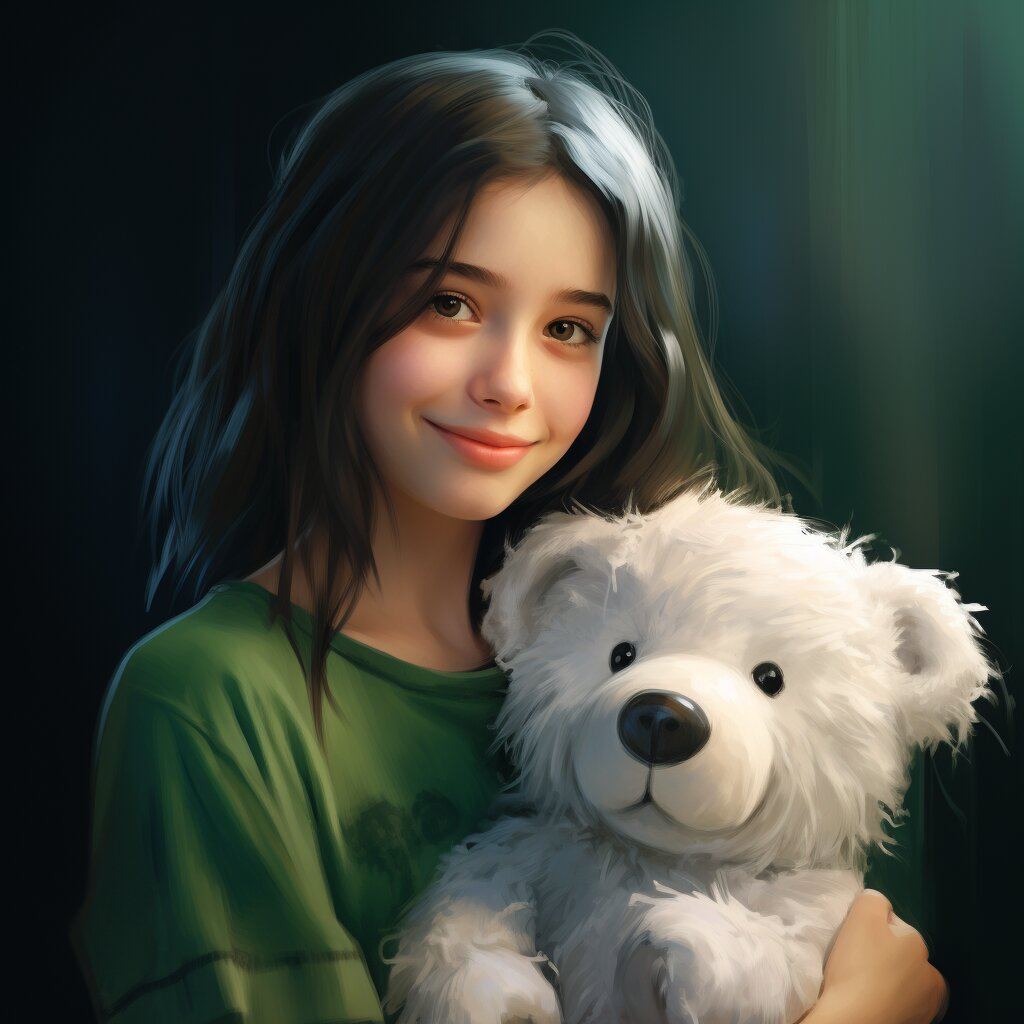}
        \includegraphics[width=14mm,height=14mm]{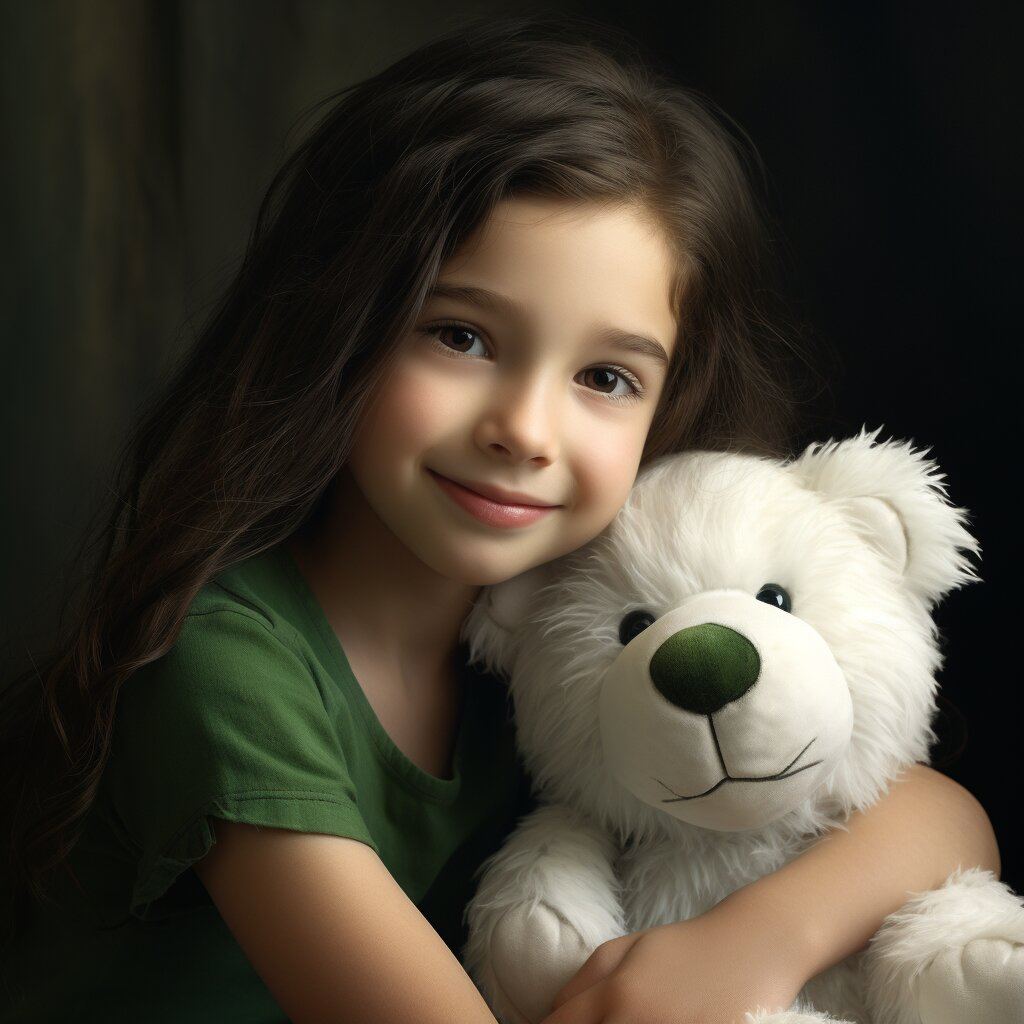}                 &
        \includegraphics[width=14mm,height=14mm]{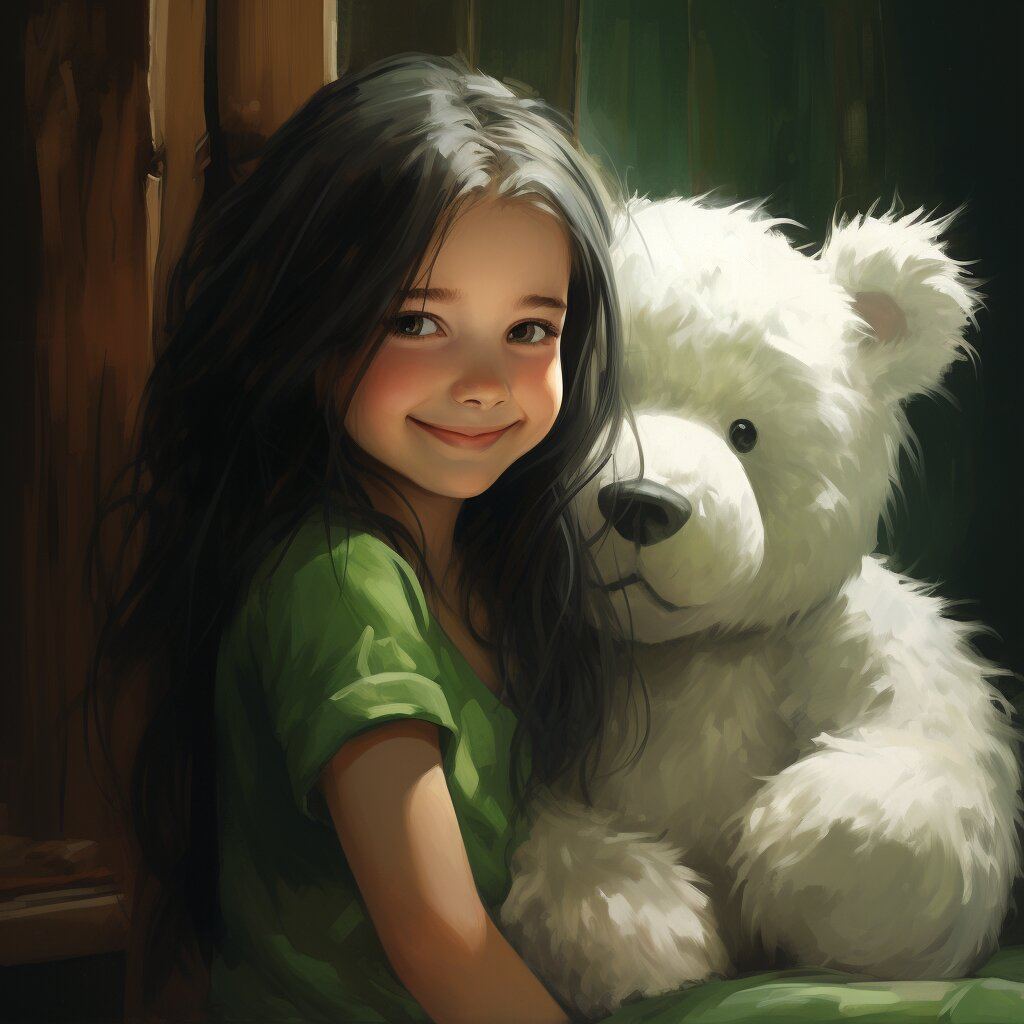}
        \includegraphics[width=14mm,height=14mm]{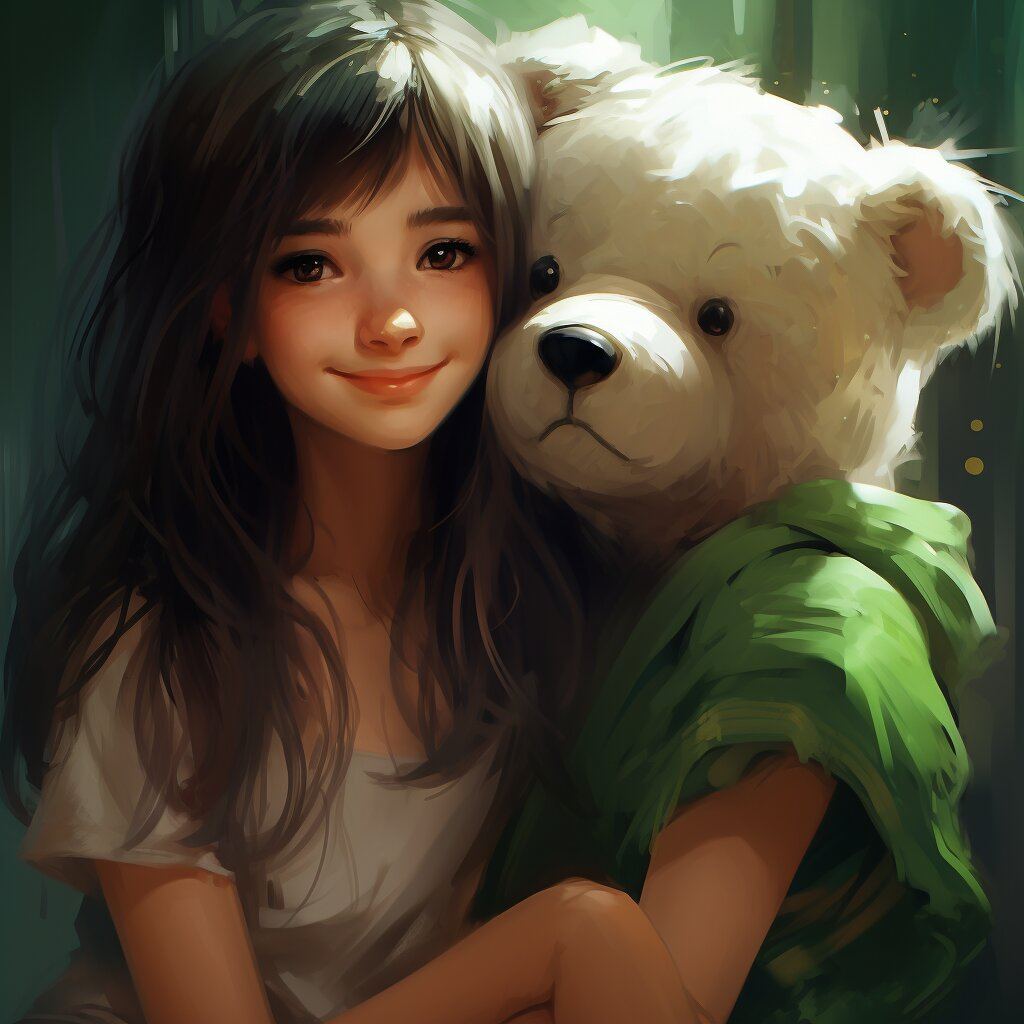}                 &
        \includegraphics[width=14mm,height=14mm]{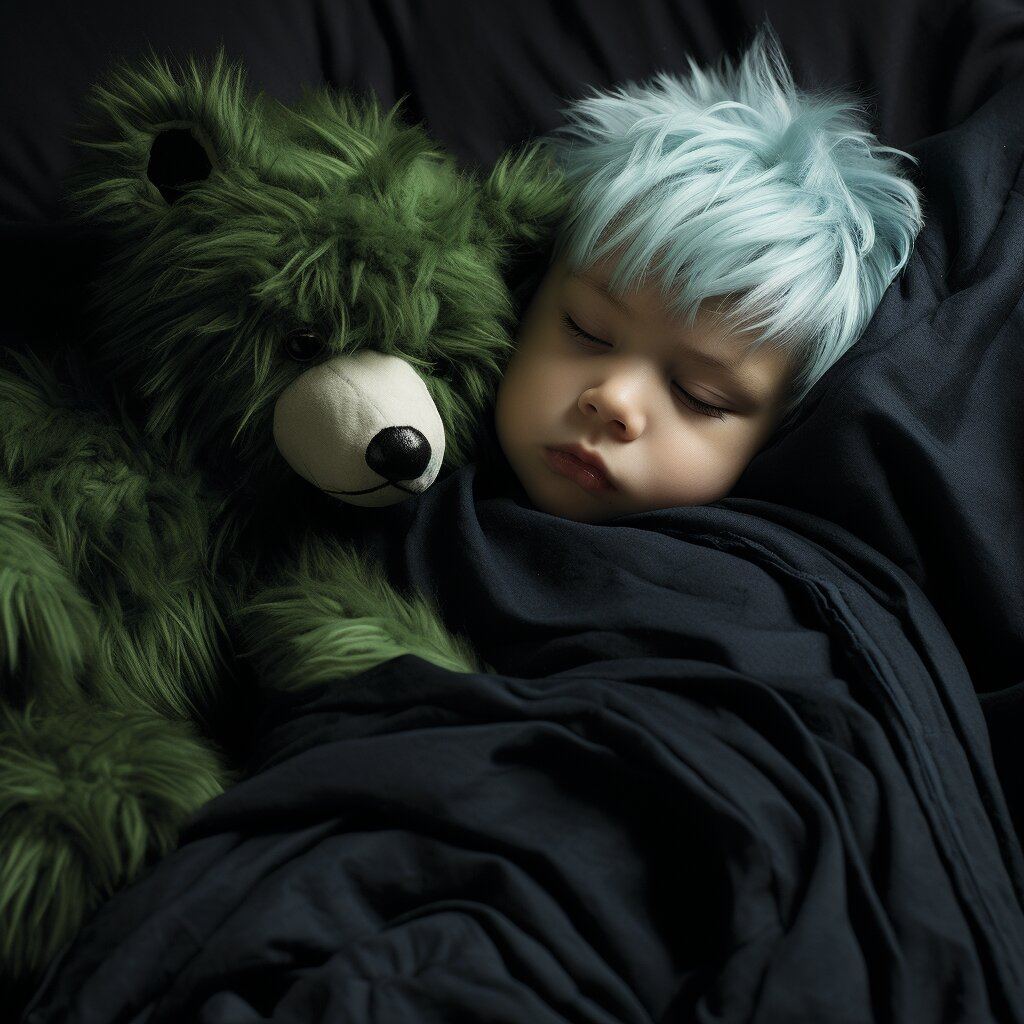}
        \includegraphics[width=14mm,height=14mm]{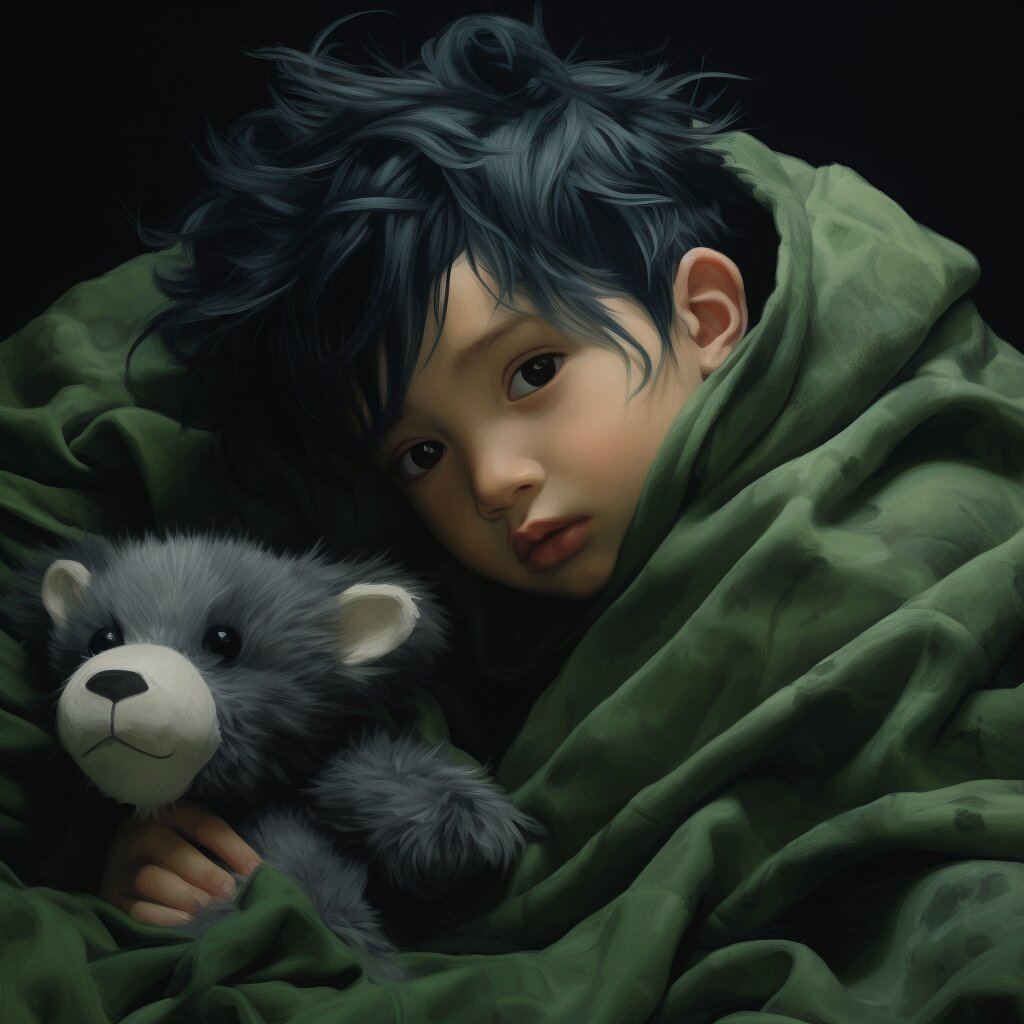}                 &
        \includegraphics[width=14mm,height=14mm]{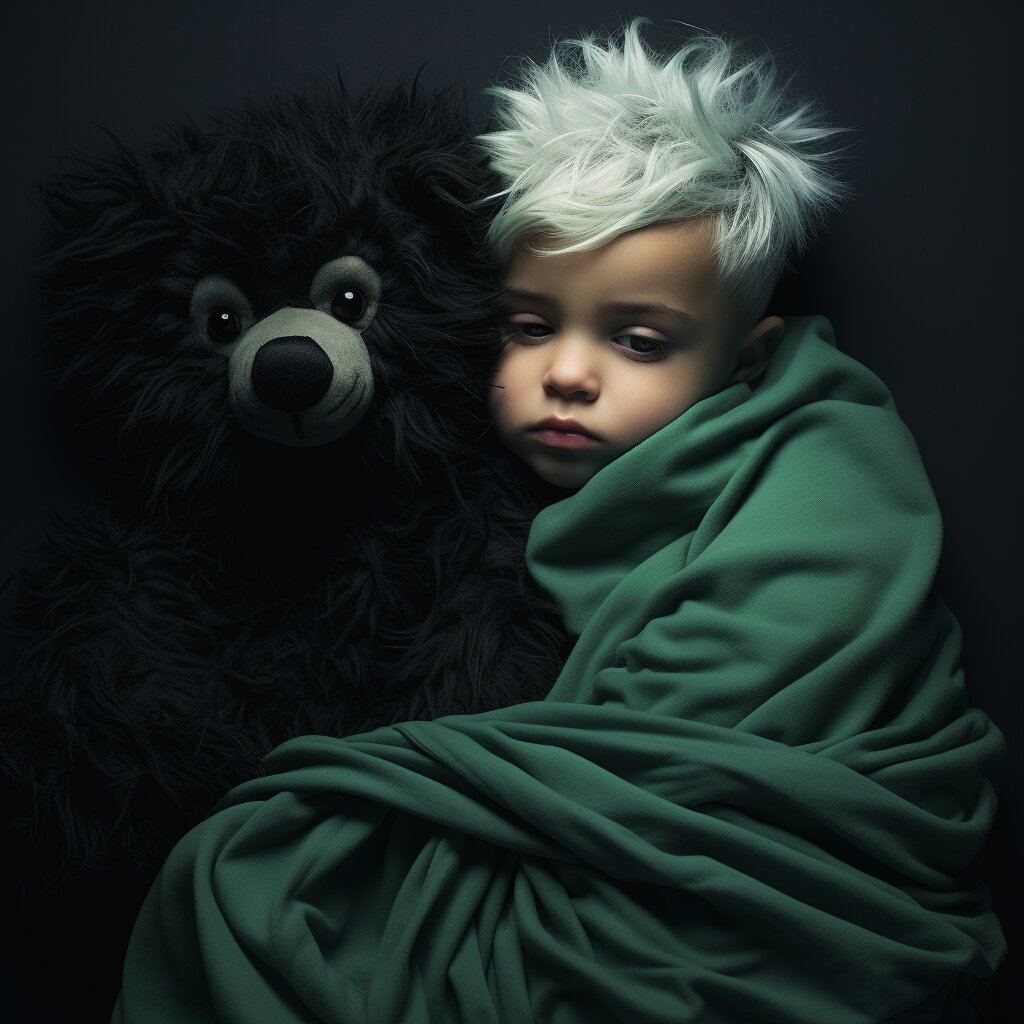}
        \includegraphics[width=14mm,height=14mm]{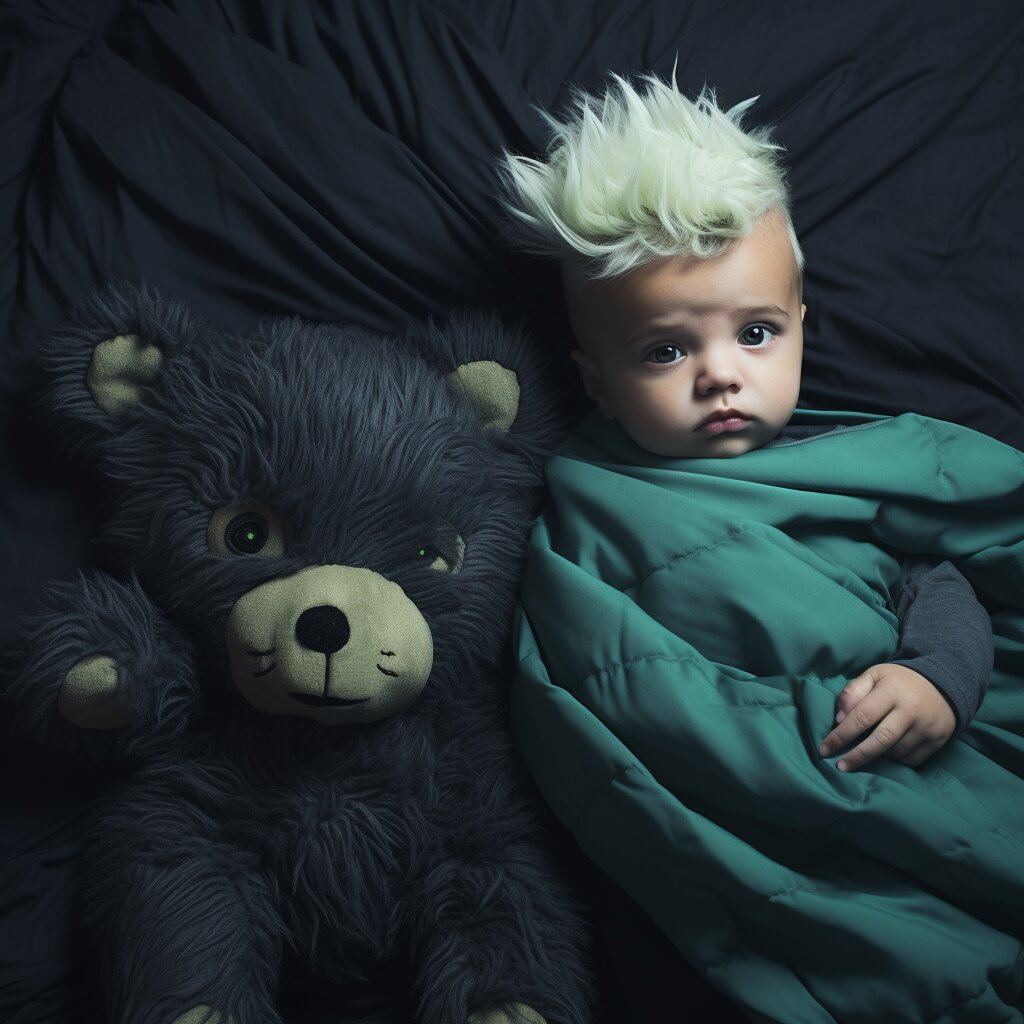}                 &
        \includegraphics[width=14mm,height=14mm]{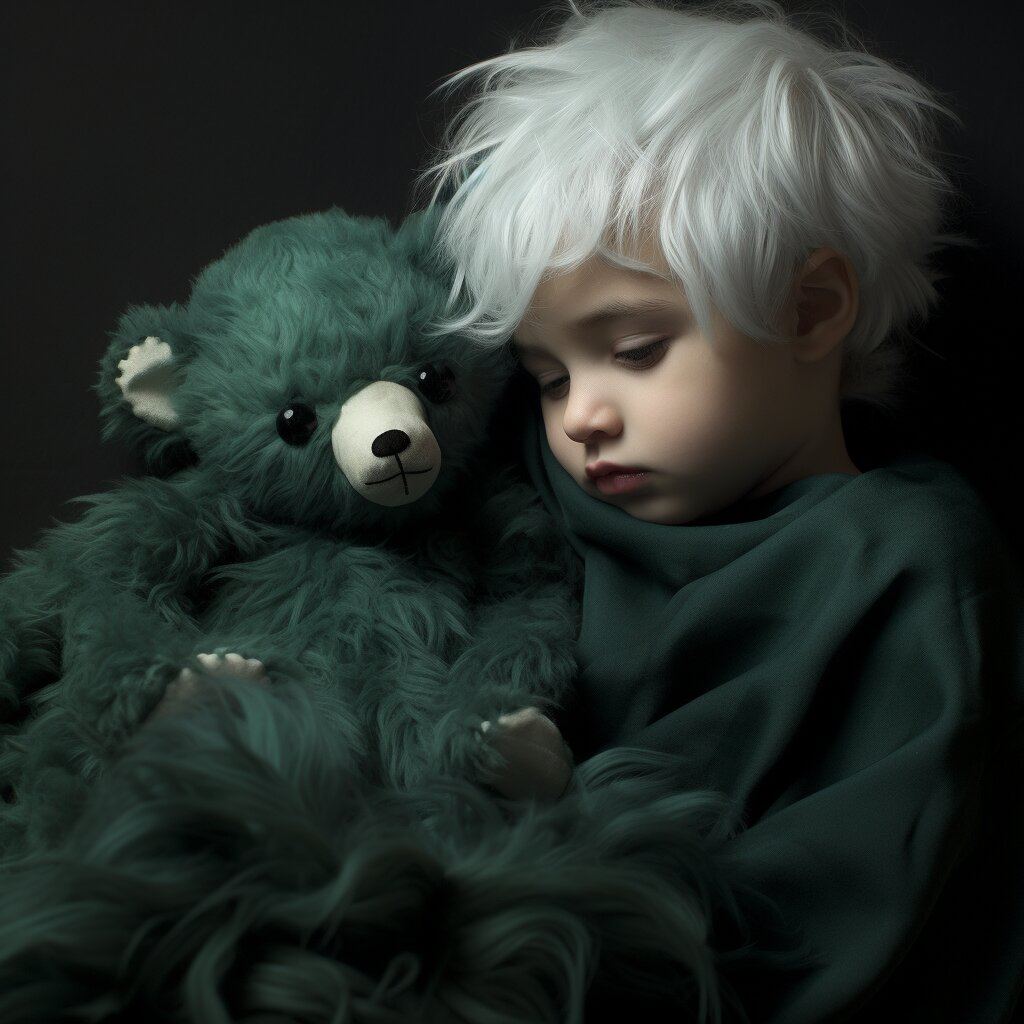}
        \includegraphics[width=14mm,height=14mm]{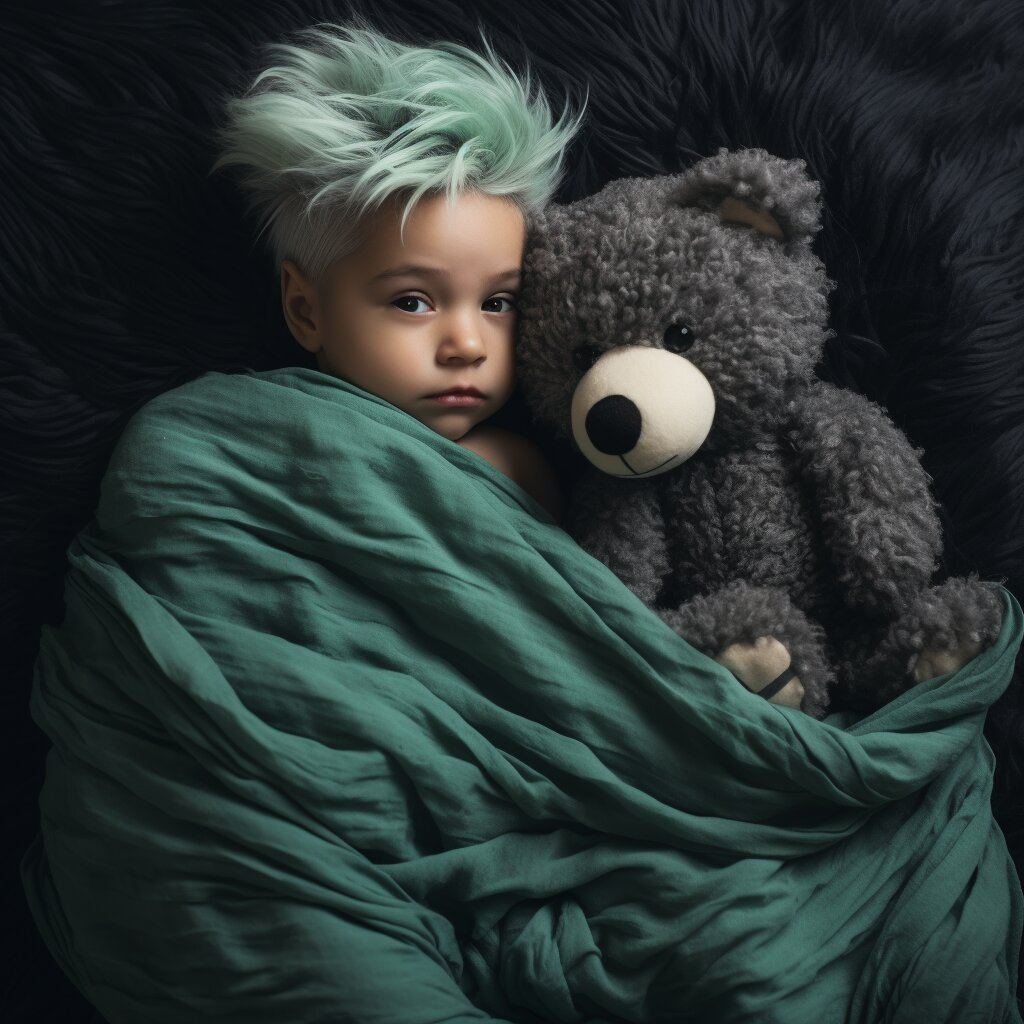}
    \end{tabular}\\[3mm]
    \begin{tabular}{c@{\hspace{1mm}}C{14mm}C{14mm}C{14mm}@{\hspace{2mm}}C{14mm}C{14mm}C{14mm}@{\hspace{2mm}}C{14mm}C{14mm}C{14mm}}
                                                                                                                           &
        \multicolumn{3}{c}{\parbox{42mm}{\centering\emph{Woman wearing a black coat holding~up~a~red~cellphone}}}          &
        \multicolumn{3}{c}{\parbox{42mm}{\centering\emph{A green and grey bird in tree with~white~leaves}}}                &
        \multicolumn{3}{c}{\parbox{42mm}{\centering\emph{a yellow vase with a blue and white bird}}}
        \\[1mm]
        \rotatebox[origin=c]{90}{\parbox{28mm}{\centering Stable Diffusion XL v1.0}}                                       &
        \includegraphics[width=14mm,height=14mm]{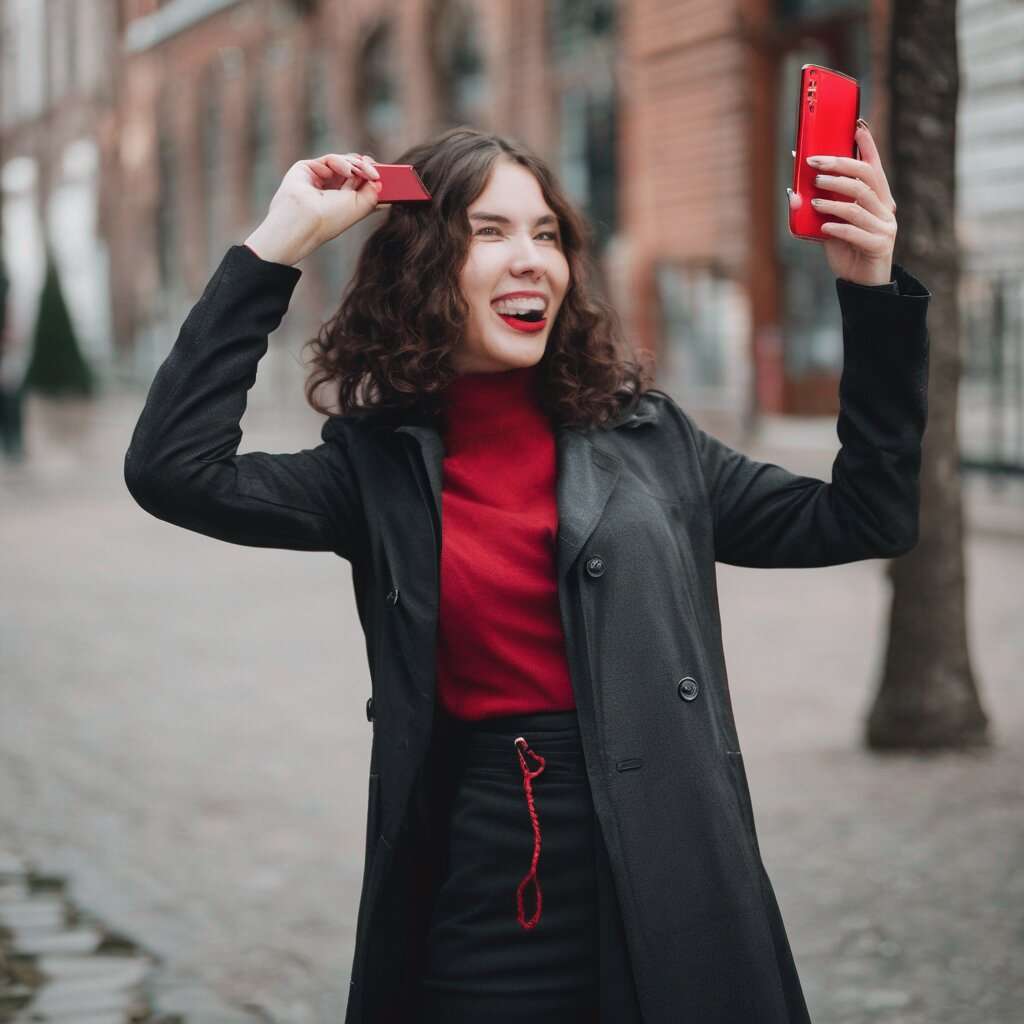}
        \includegraphics[width=14mm,height=14mm]{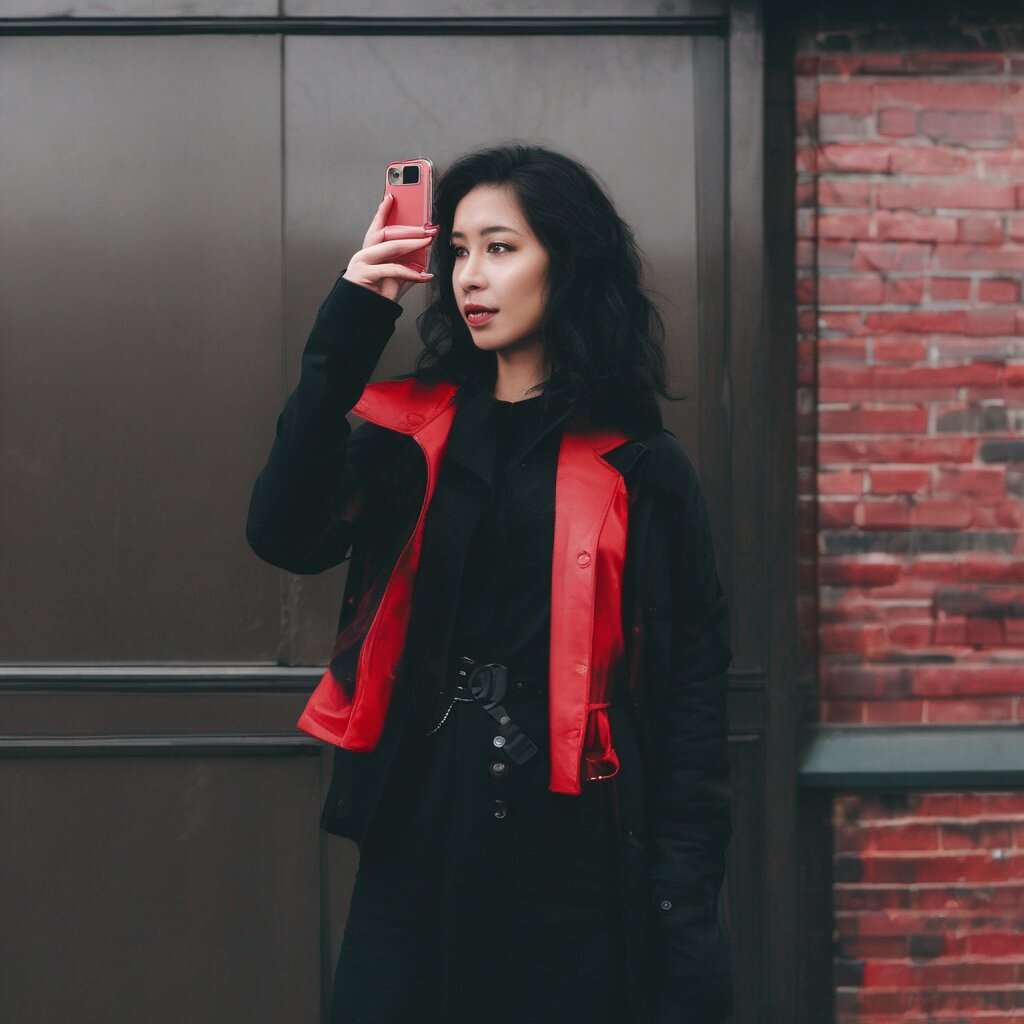} &
        \includegraphics[width=14mm,height=14mm]{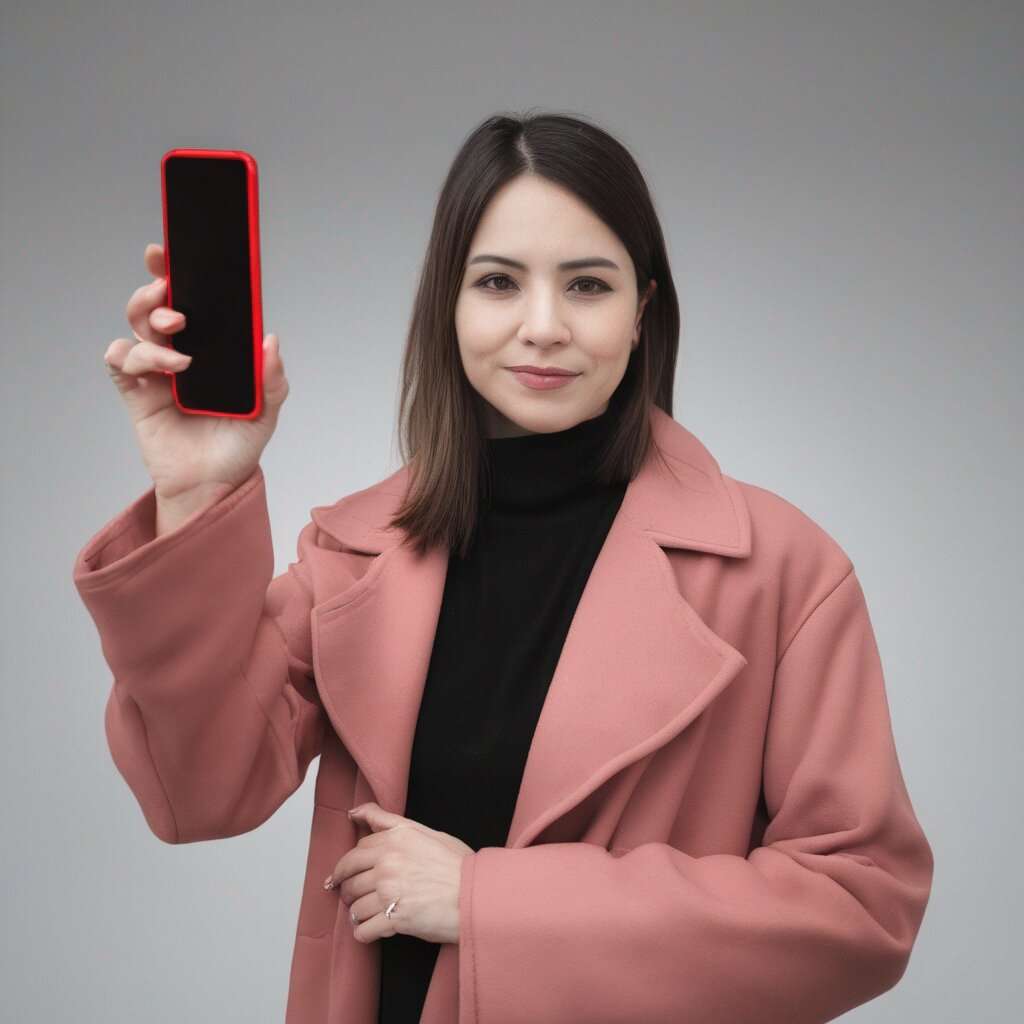}
        \includegraphics[width=14mm,height=14mm]{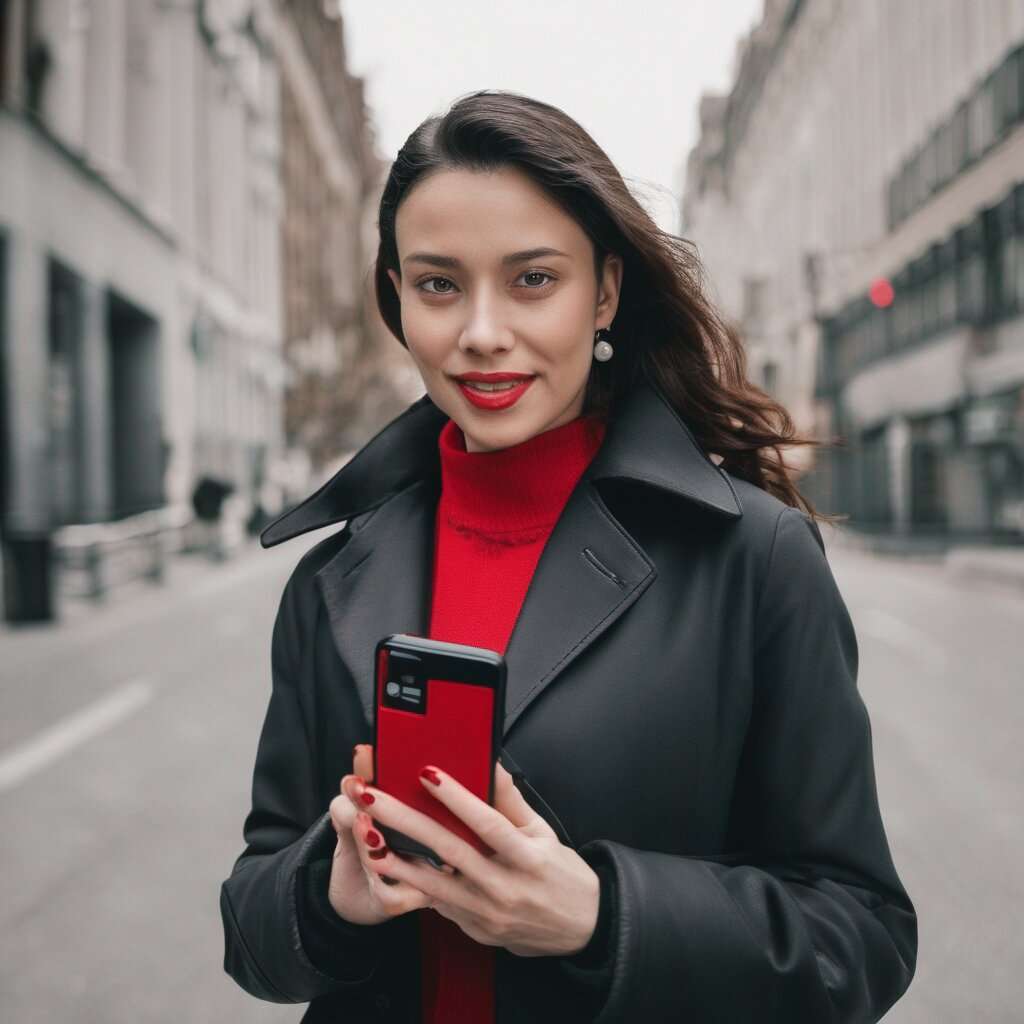} &
        \includegraphics[width=14mm,height=14mm]{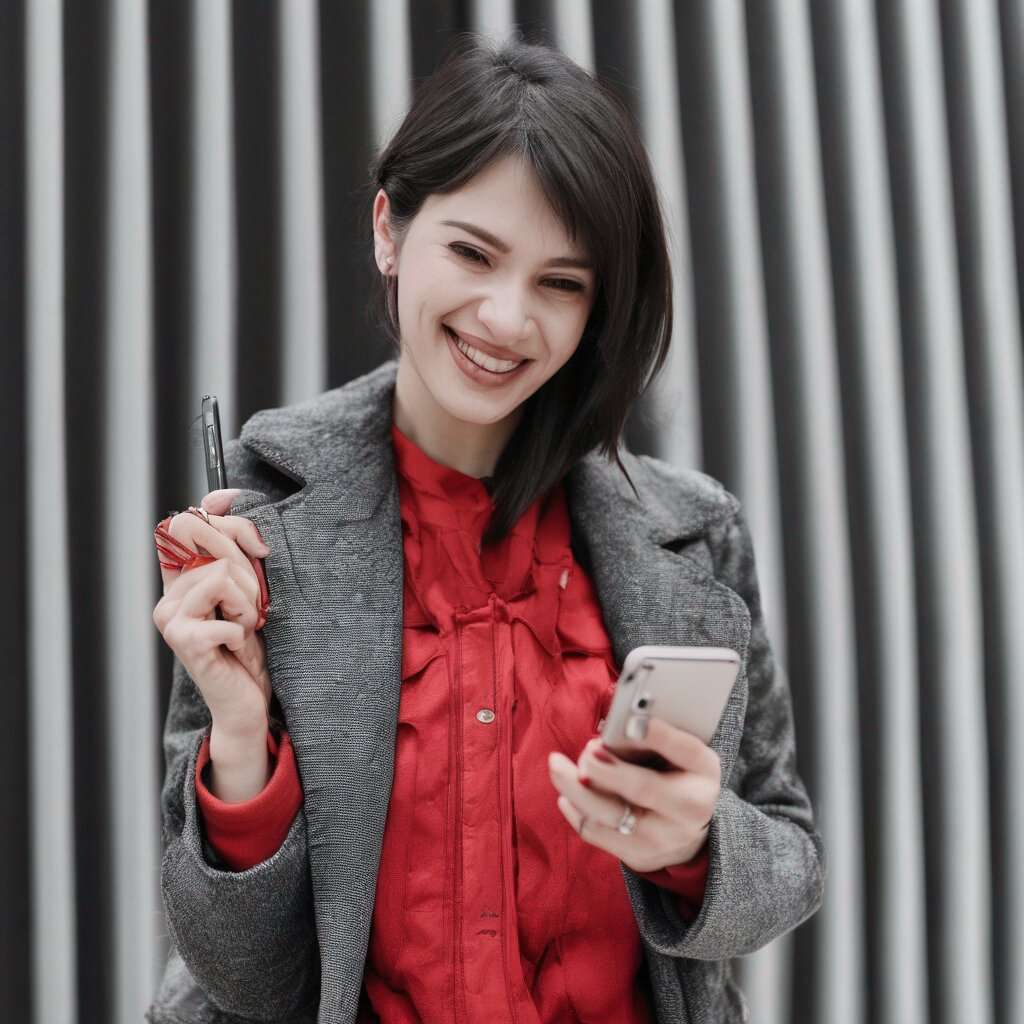}
        \includegraphics[width=14mm,height=14mm]{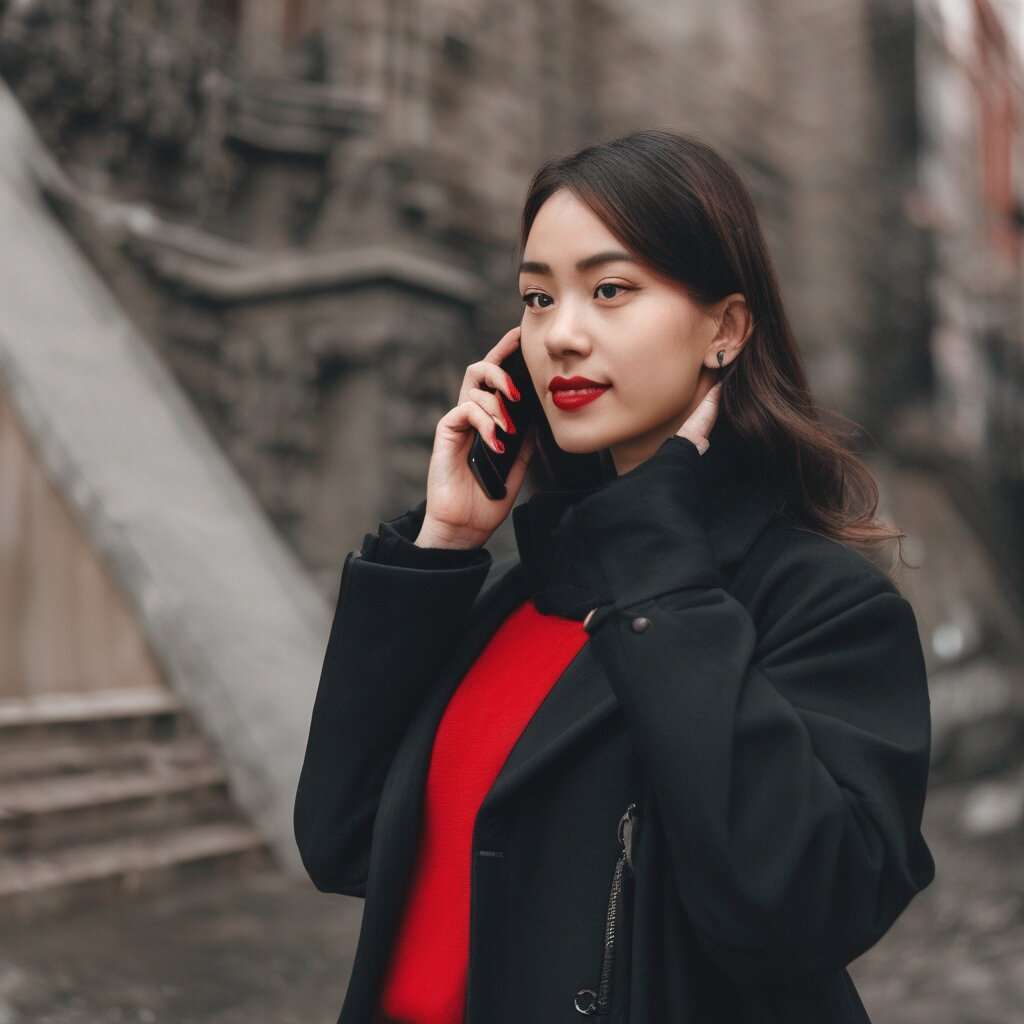} &
        \includegraphics[width=14mm,height=14mm]{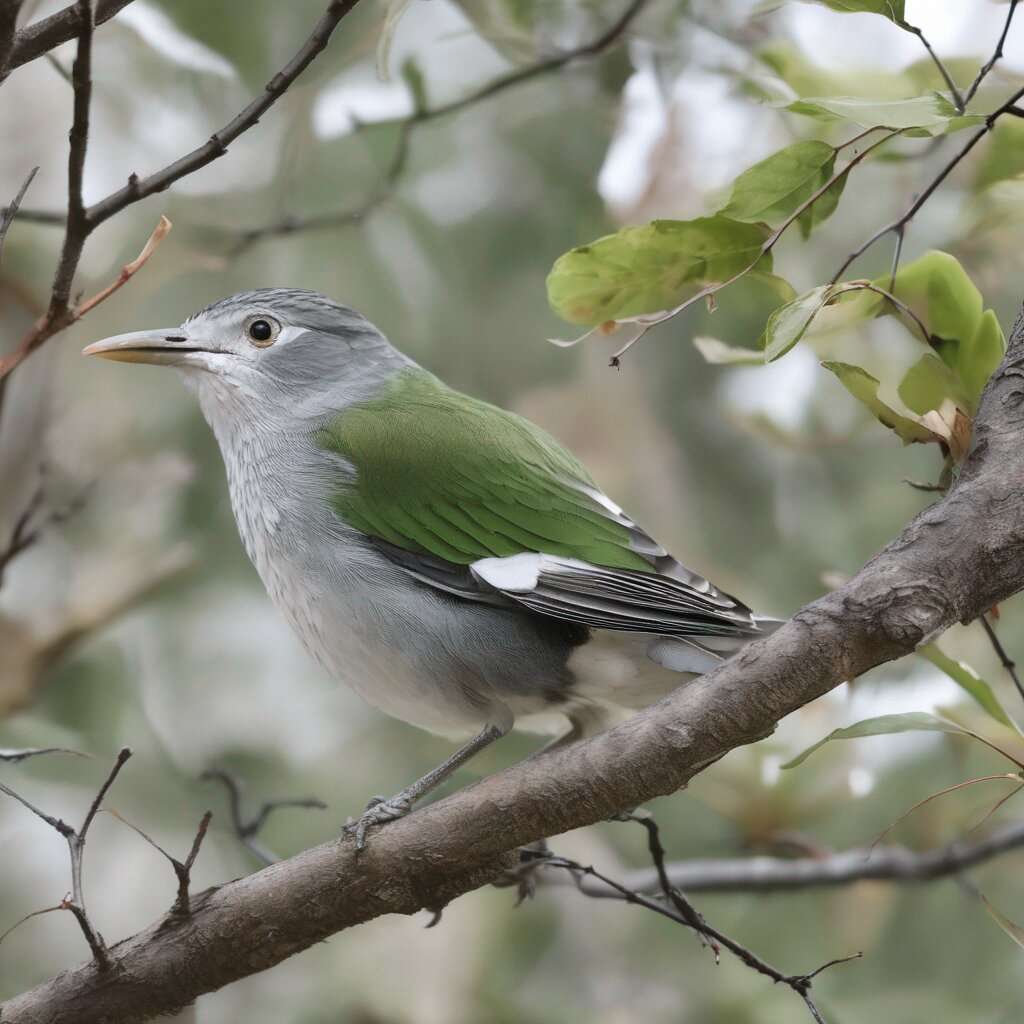}
        \includegraphics[width=14mm,height=14mm]{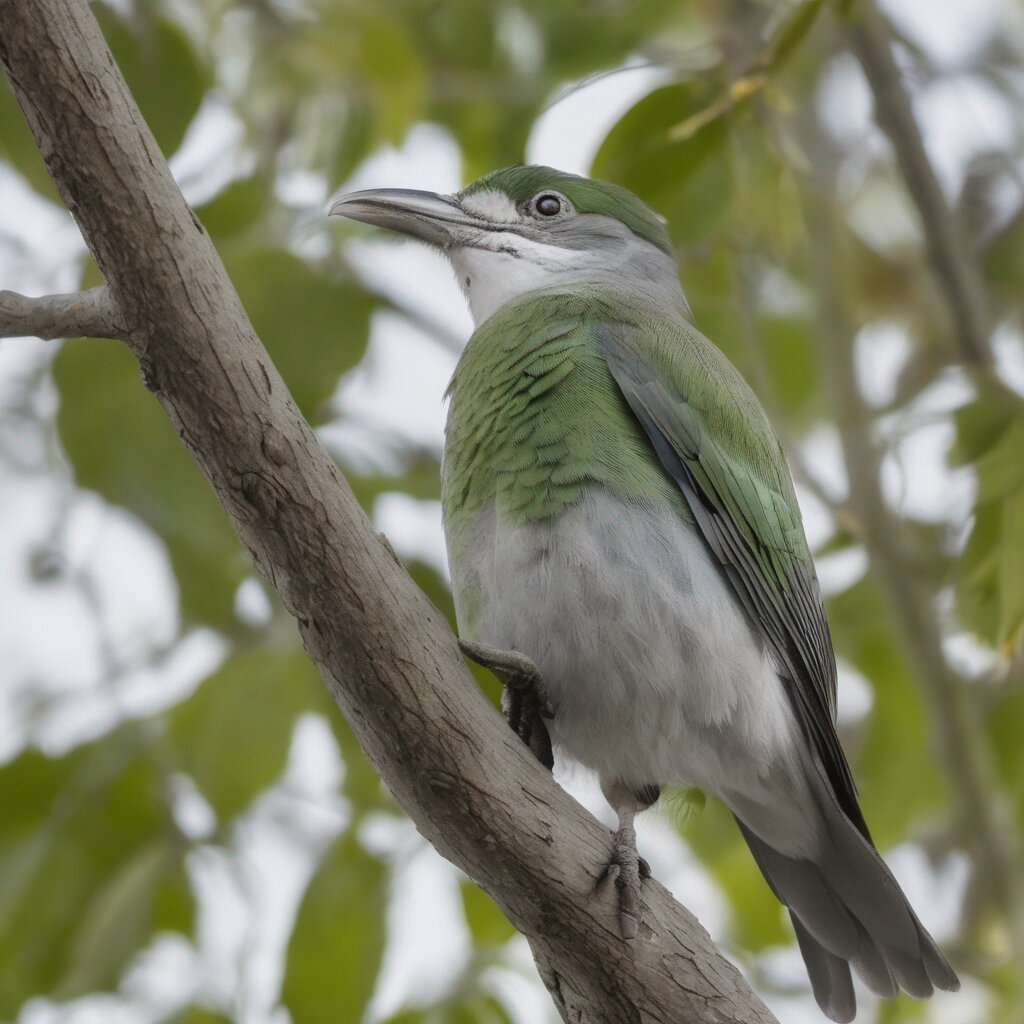}       &
        \includegraphics[width=14mm,height=14mm]{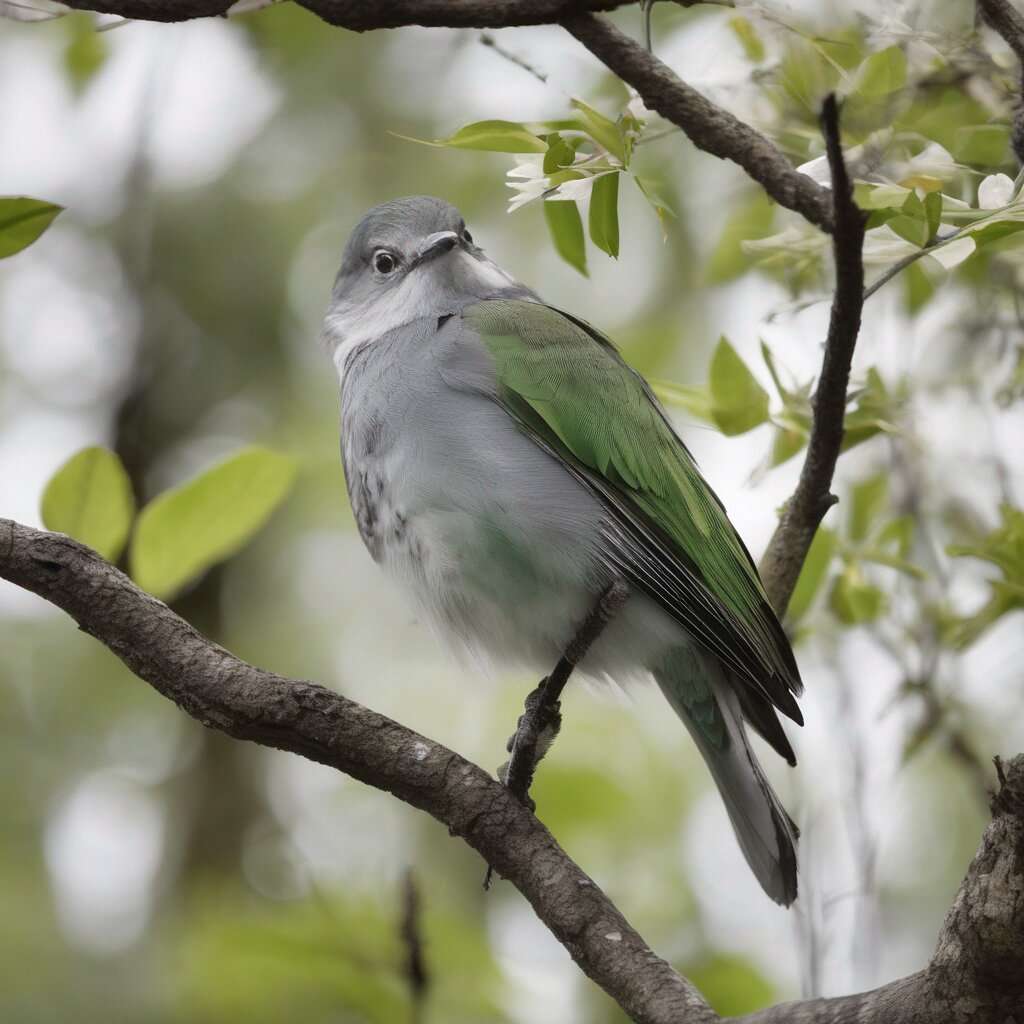}
        \includegraphics[width=14mm,height=14mm]{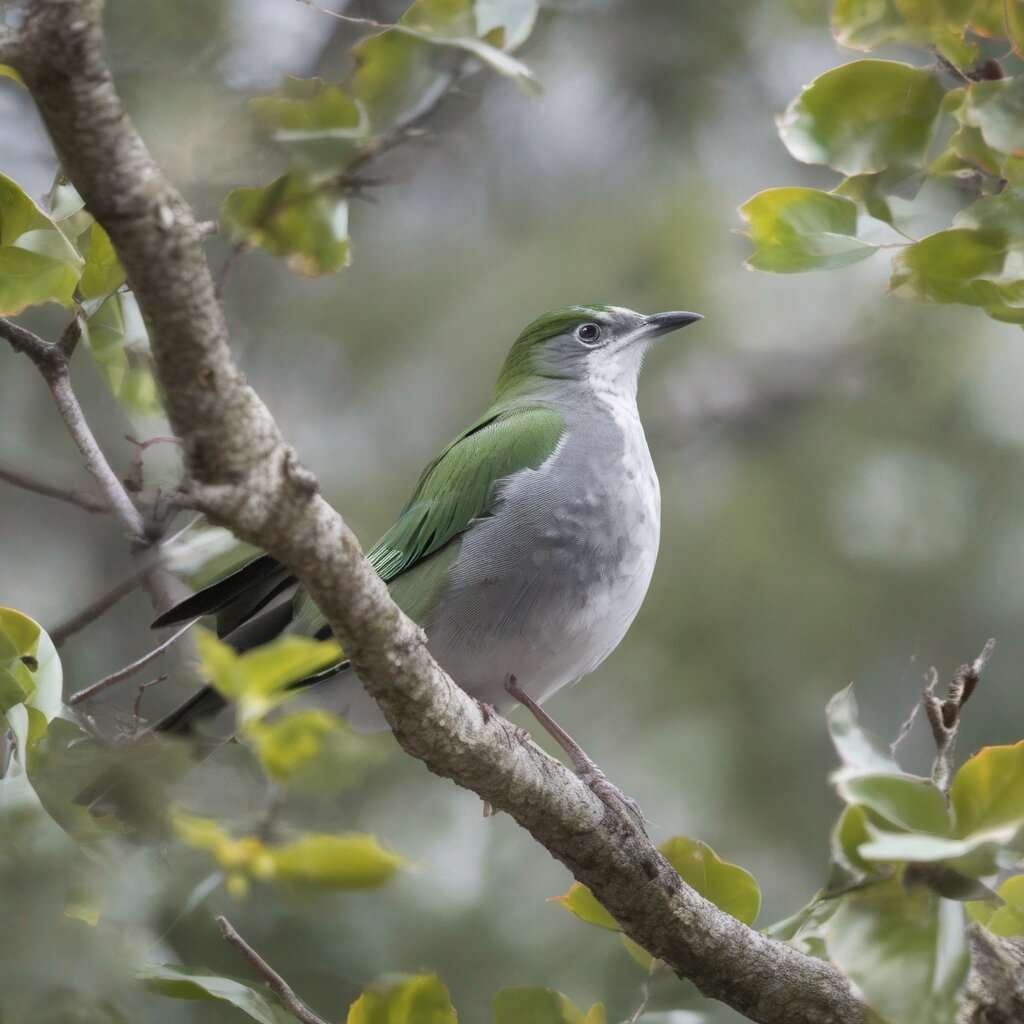}       &
        \includegraphics[width=14mm,height=14mm]{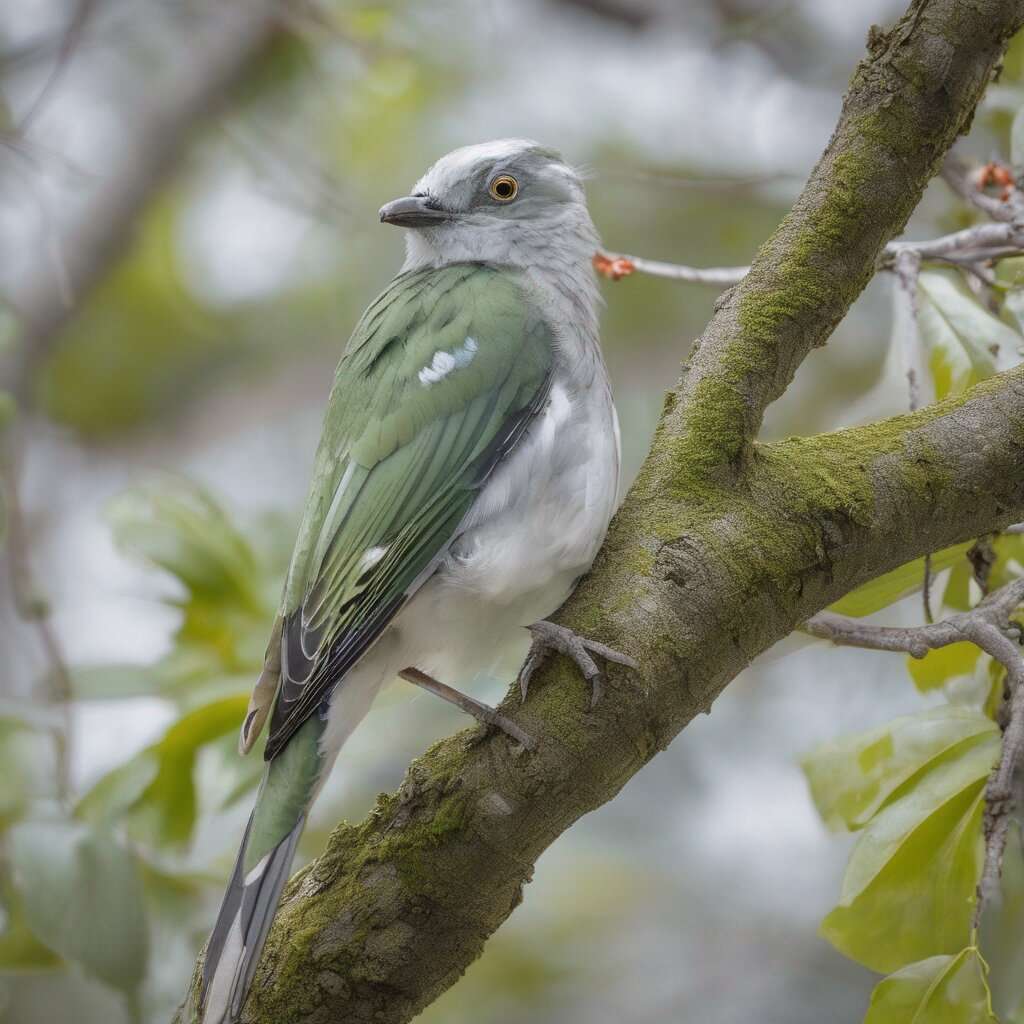}
        \includegraphics[width=14mm,height=14mm]{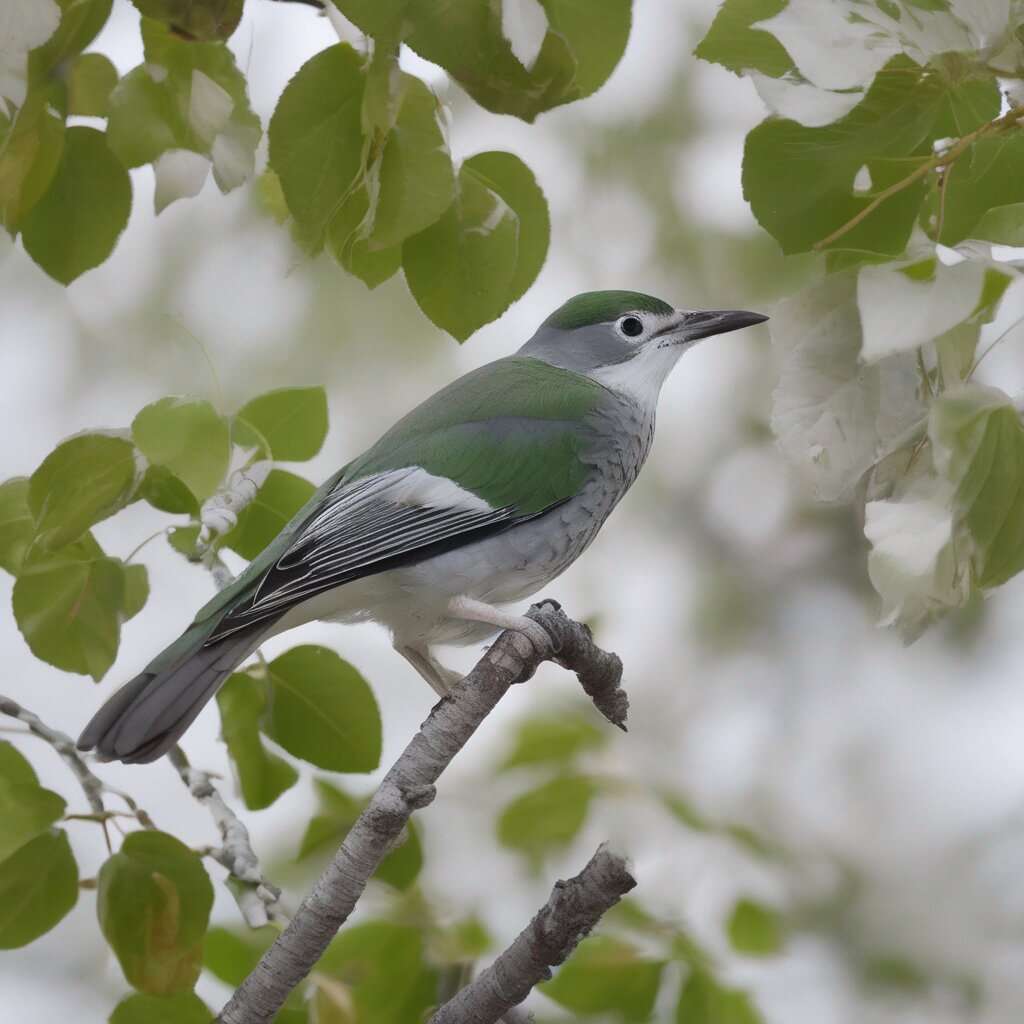}      &
        \includegraphics[width=14mm,height=14mm]{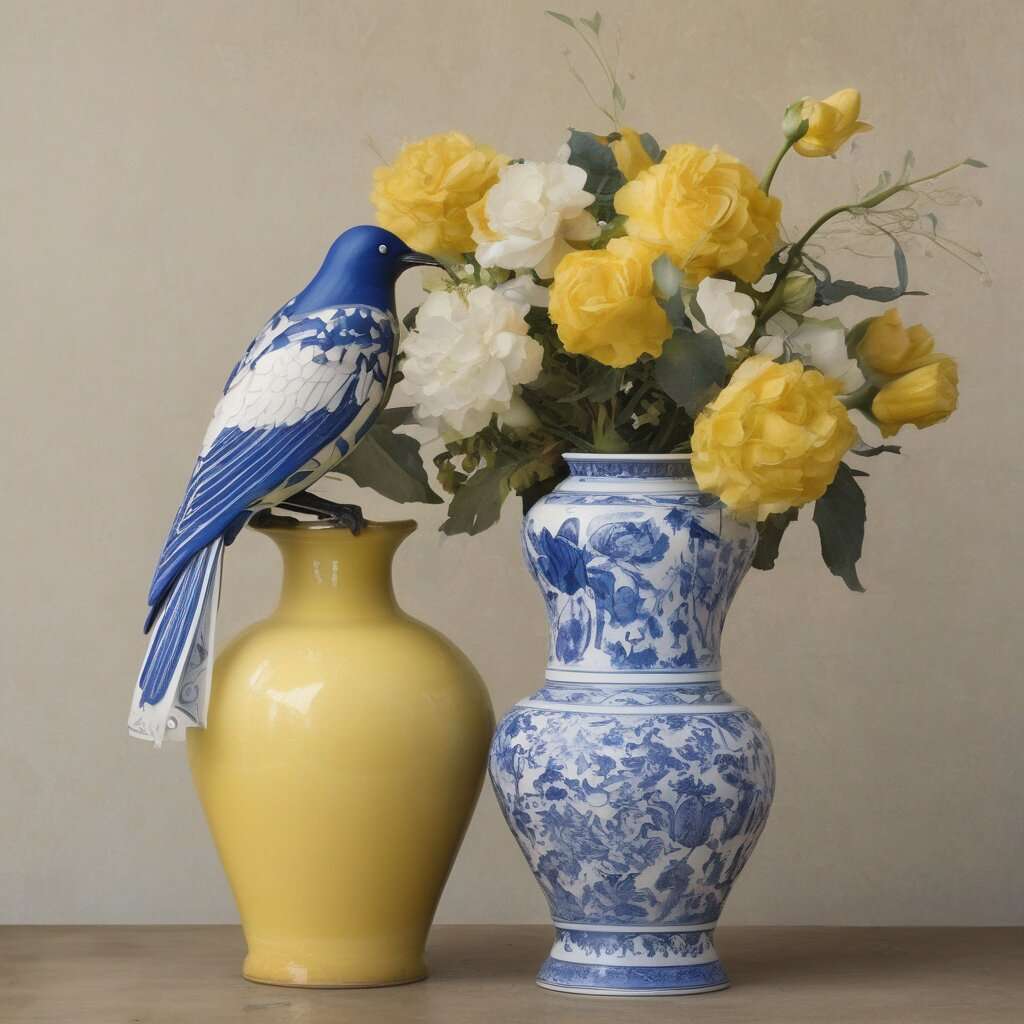}
        \includegraphics[width=14mm,height=14mm]{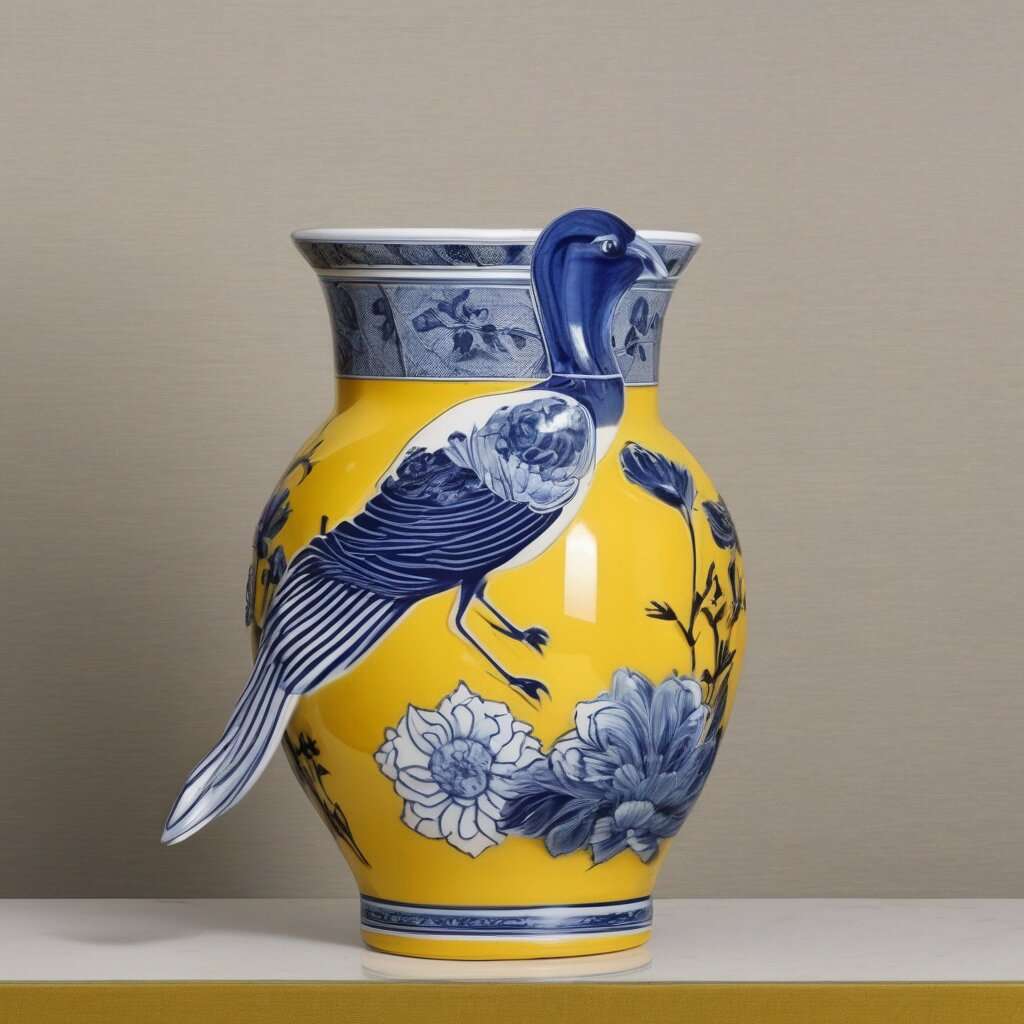}              &
        \includegraphics[width=14mm,height=14mm]{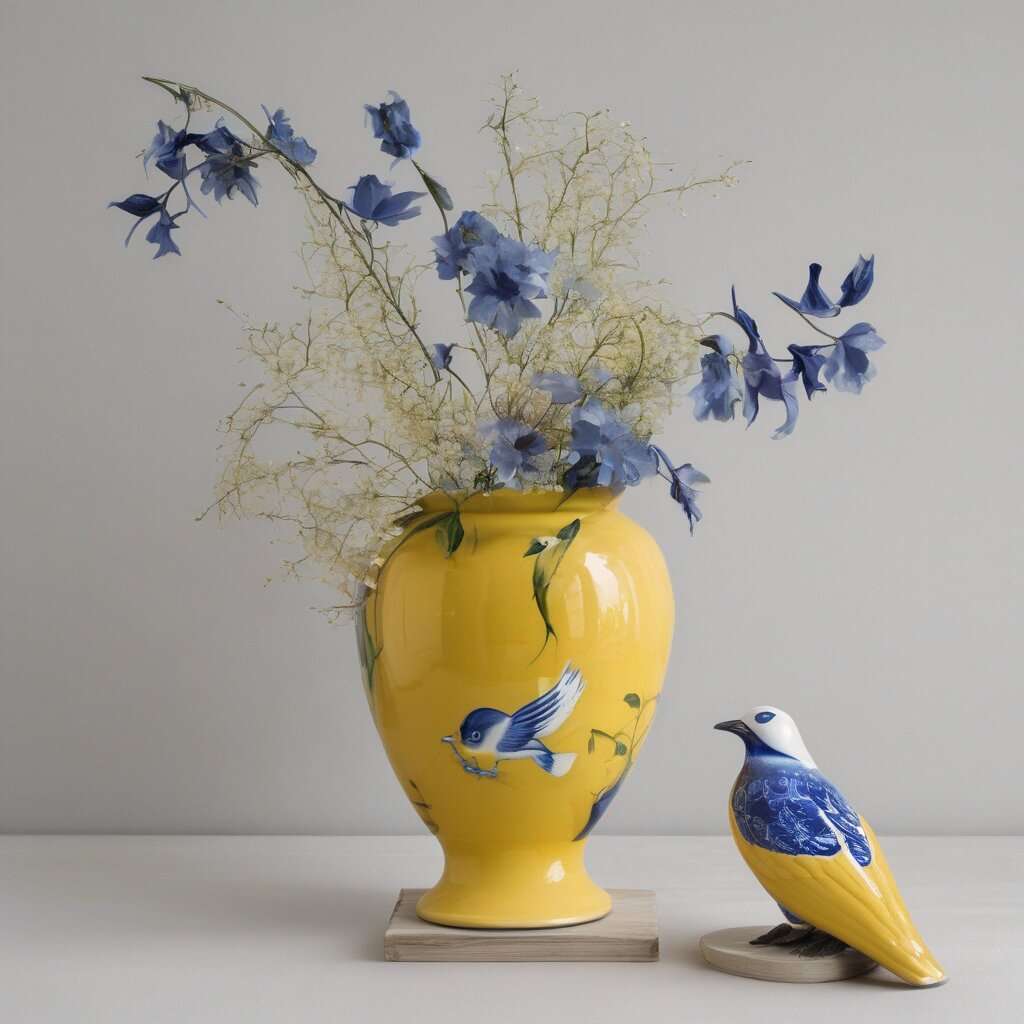}
        \includegraphics[width=14mm,height=14mm]{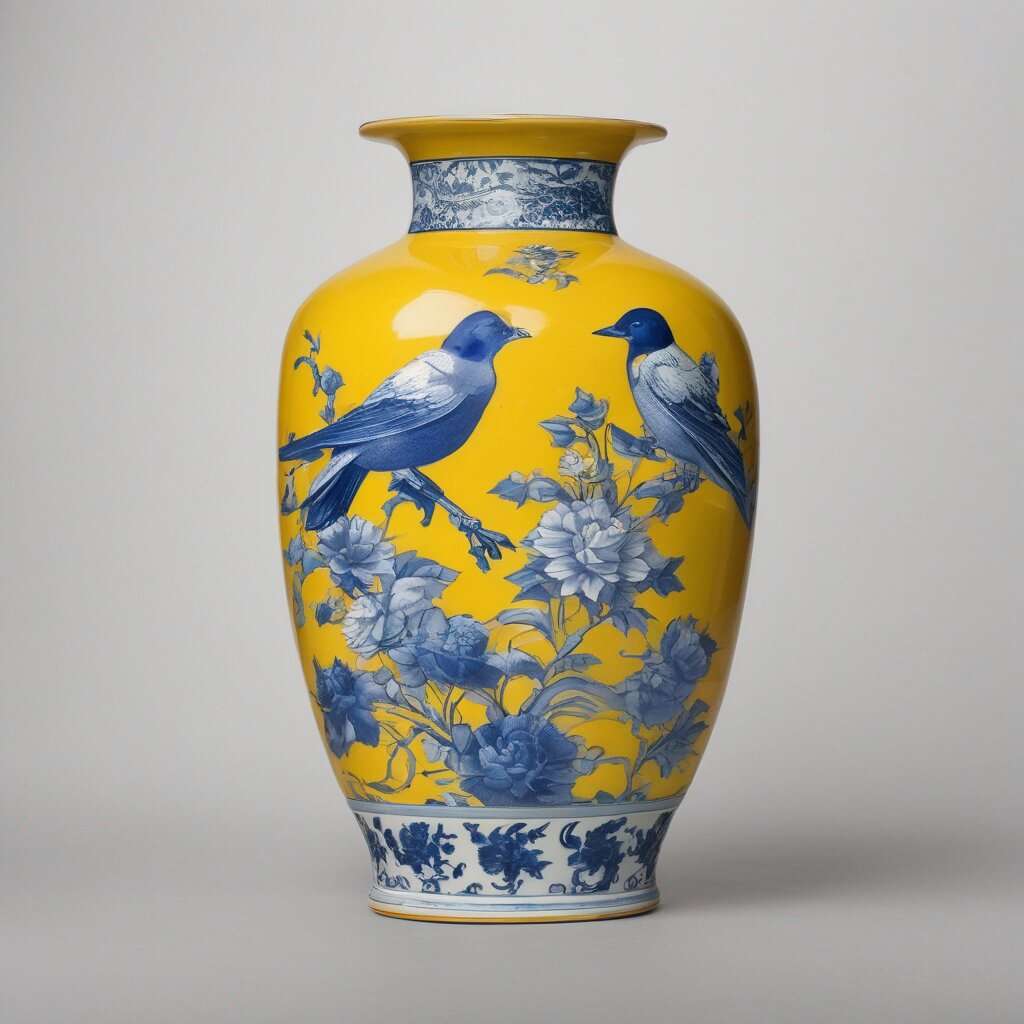}              &
        \includegraphics[width=14mm,height=14mm]{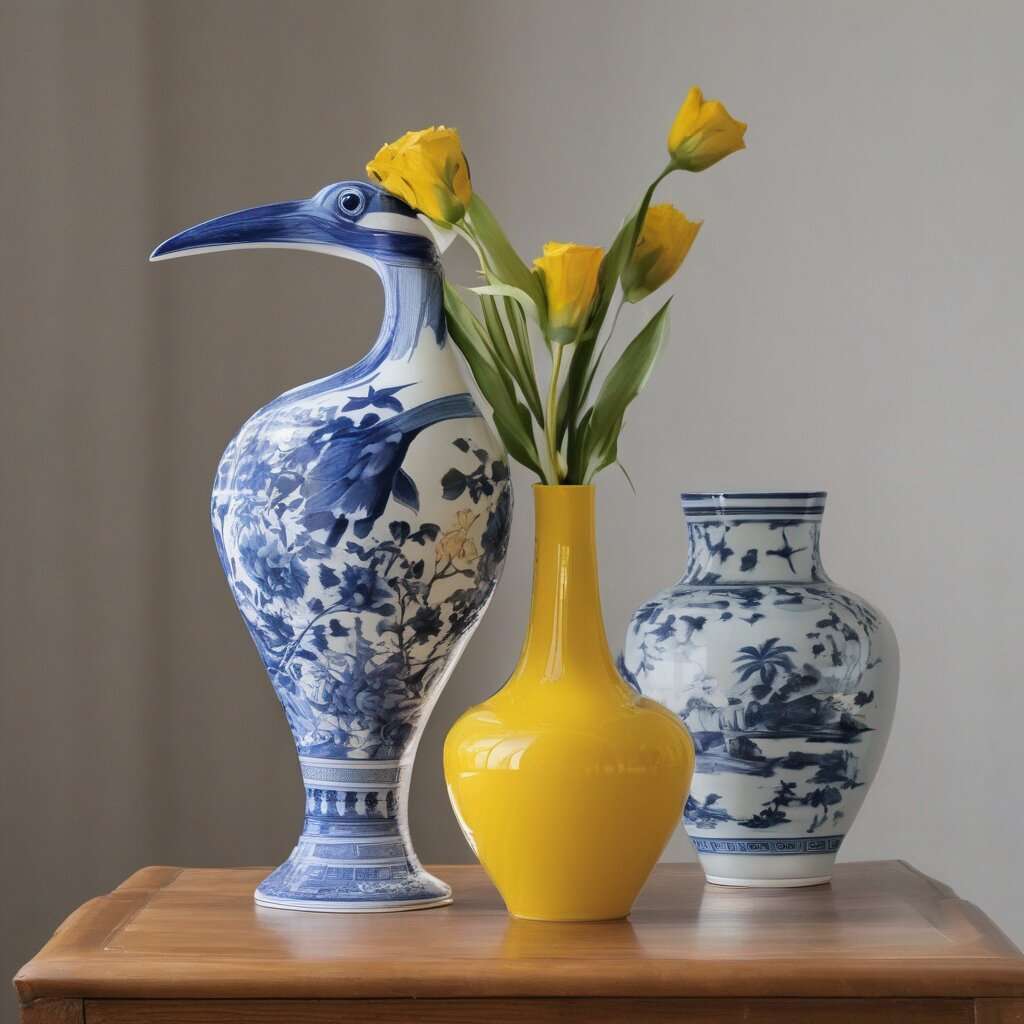}
        \includegraphics[width=14mm,height=14mm]{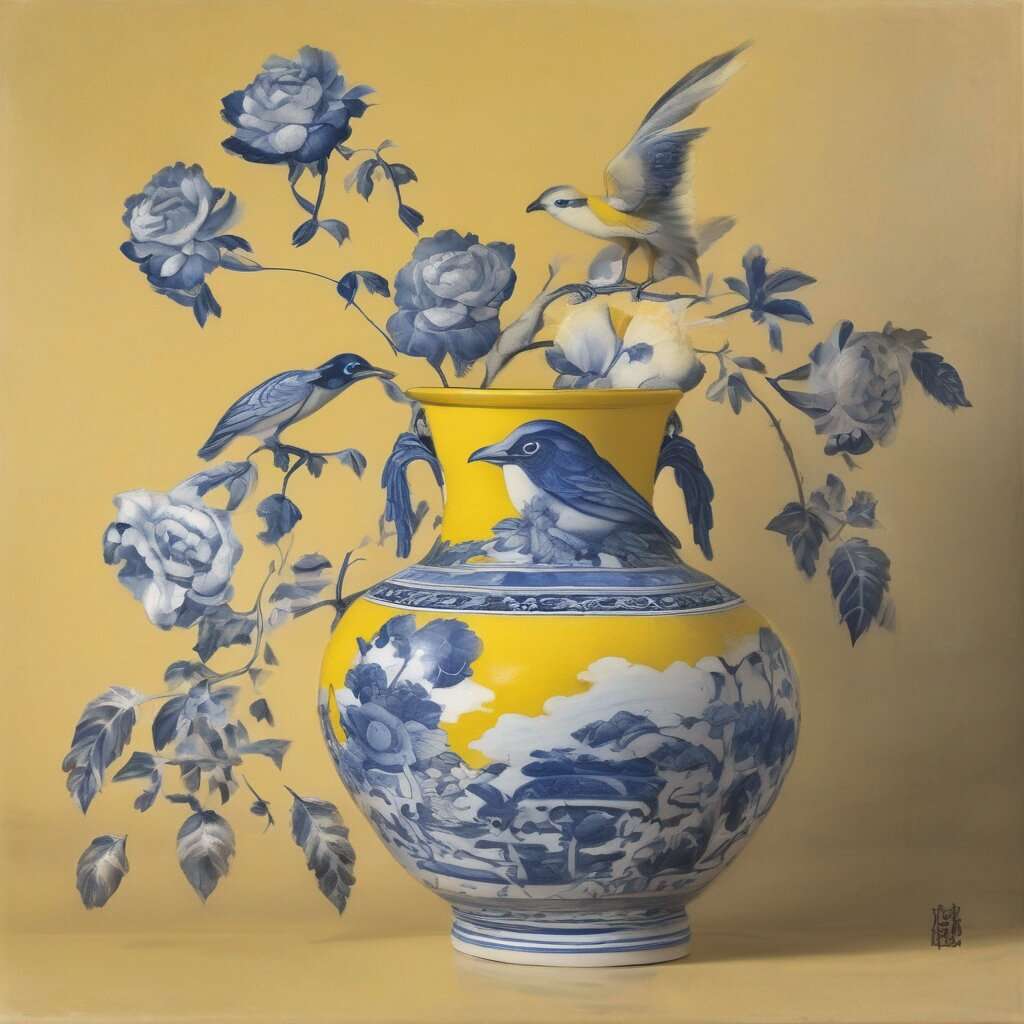}
        \\[-0.3mm]
        \rotatebox[origin=c]{90}{\parbox{28mm}{\centering Midjourney v5.2}}                                                &
        \includegraphics[width=14mm,height=14mm]{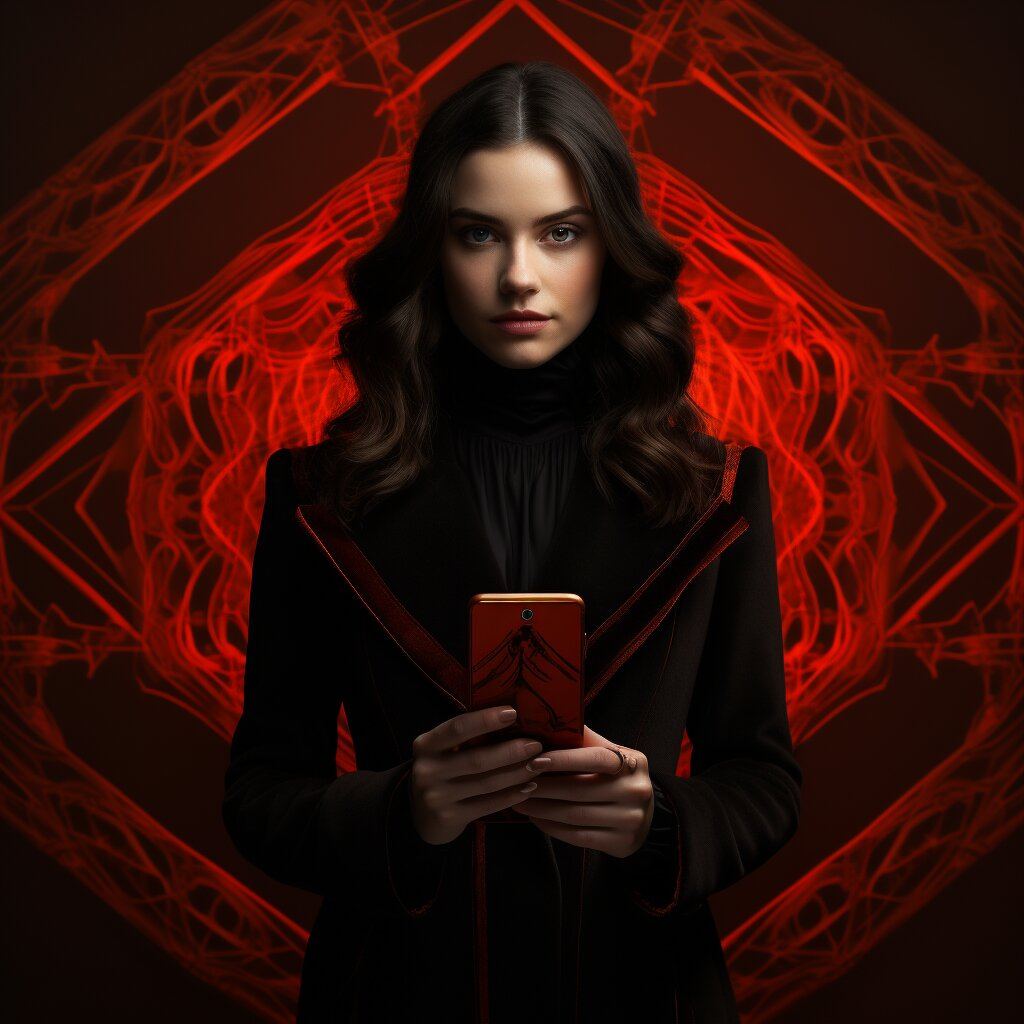}
        \includegraphics[width=14mm,height=14mm]{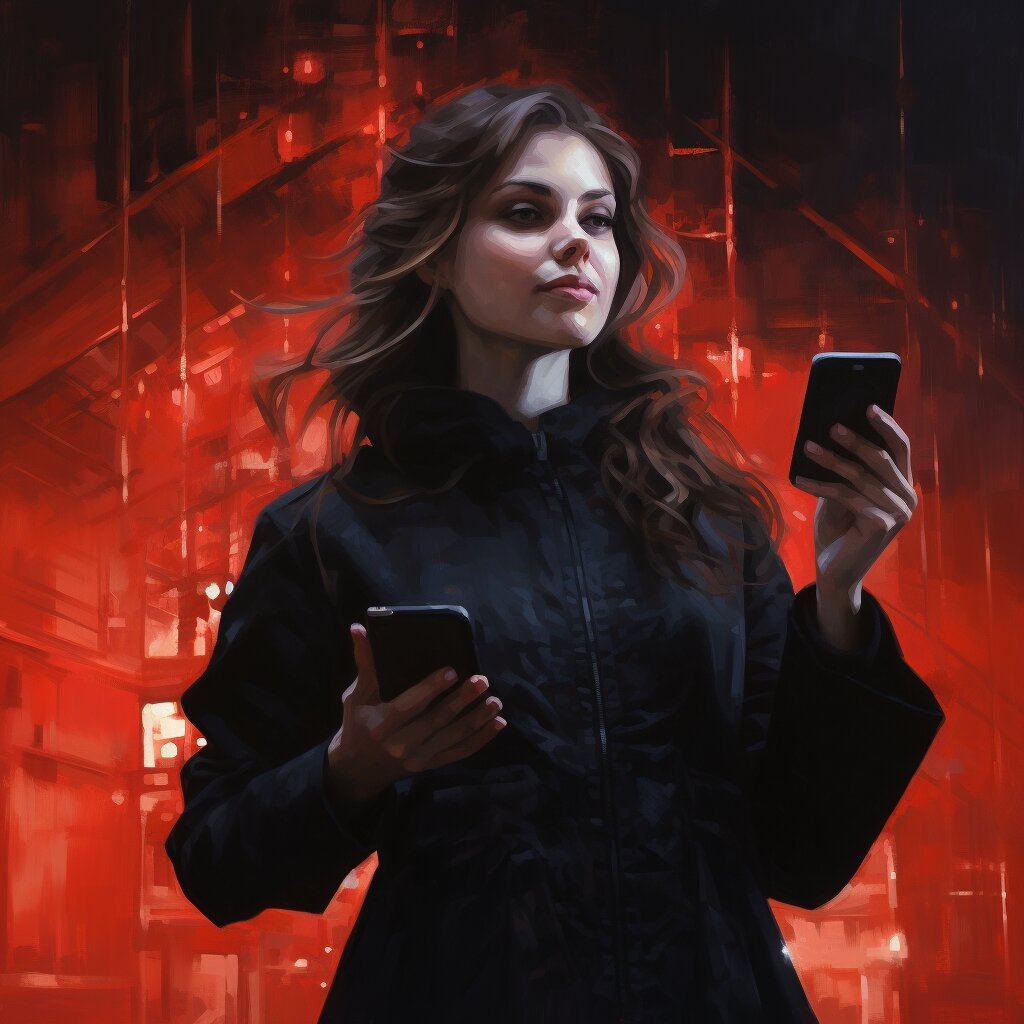} &
        \includegraphics[width=14mm,height=14mm]{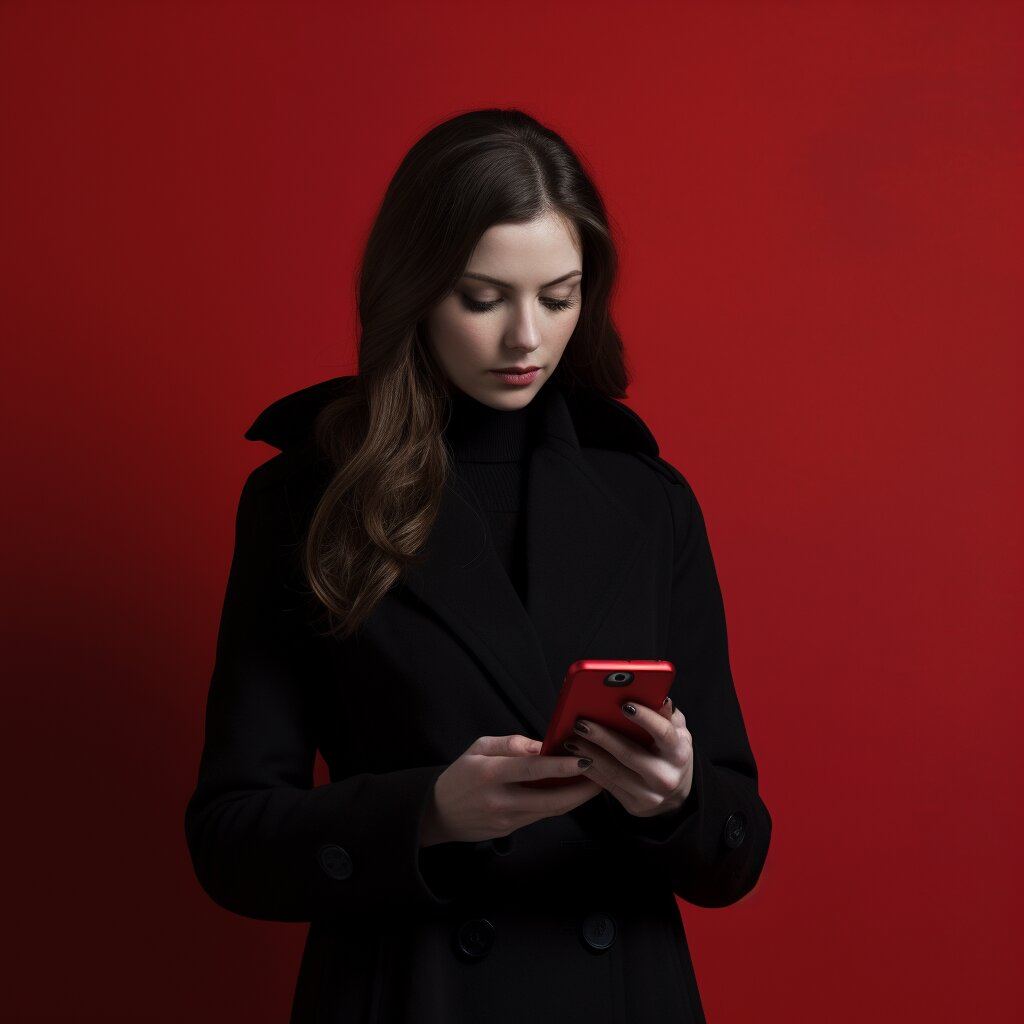}
        \includegraphics[width=14mm,height=14mm]{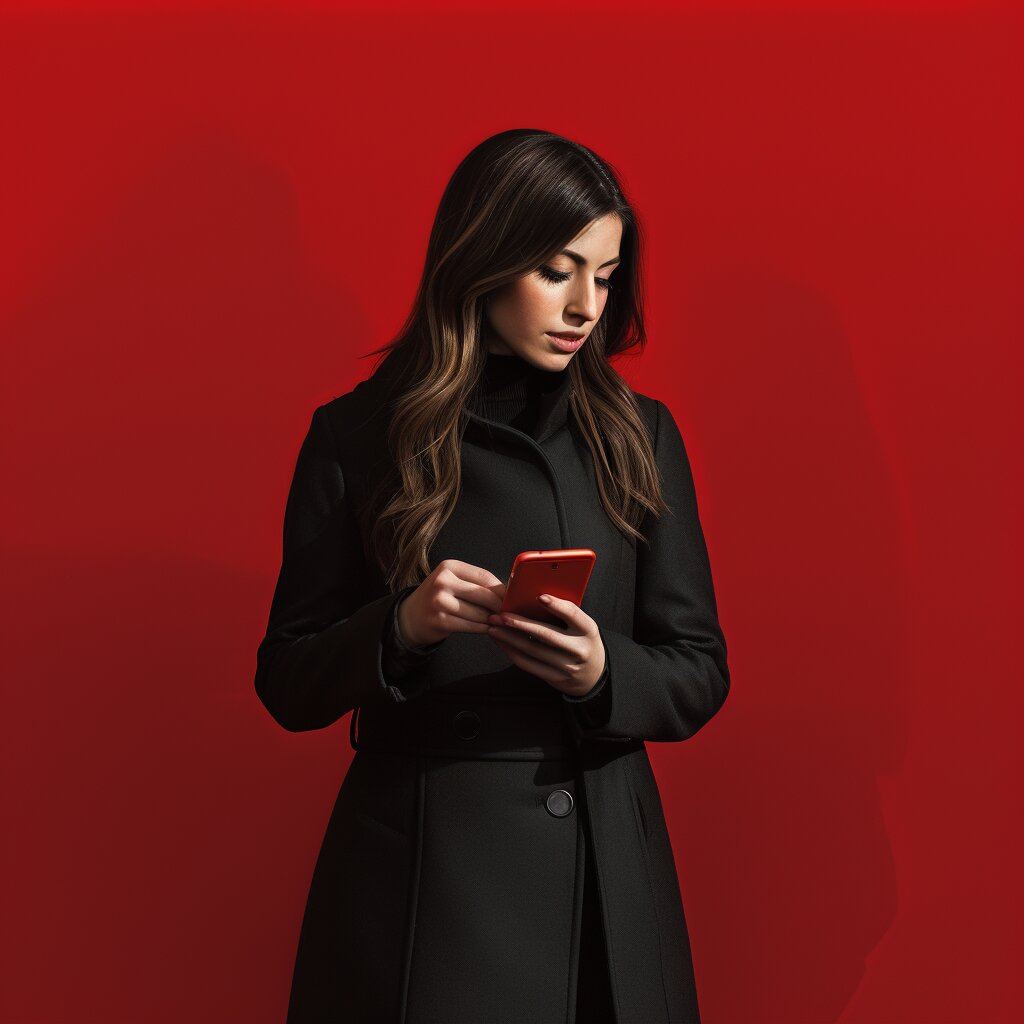} &
        \includegraphics[width=14mm,height=14mm]{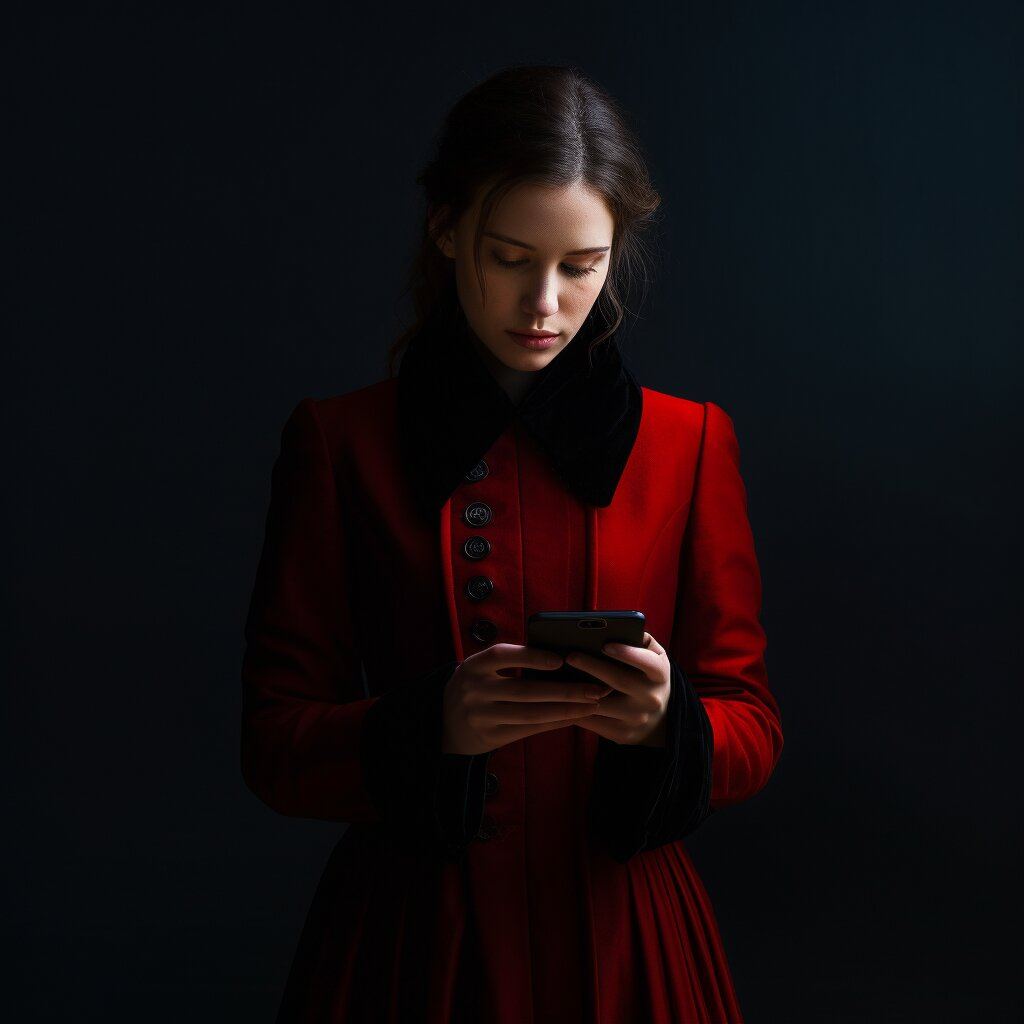}
        \includegraphics[width=14mm,height=14mm]{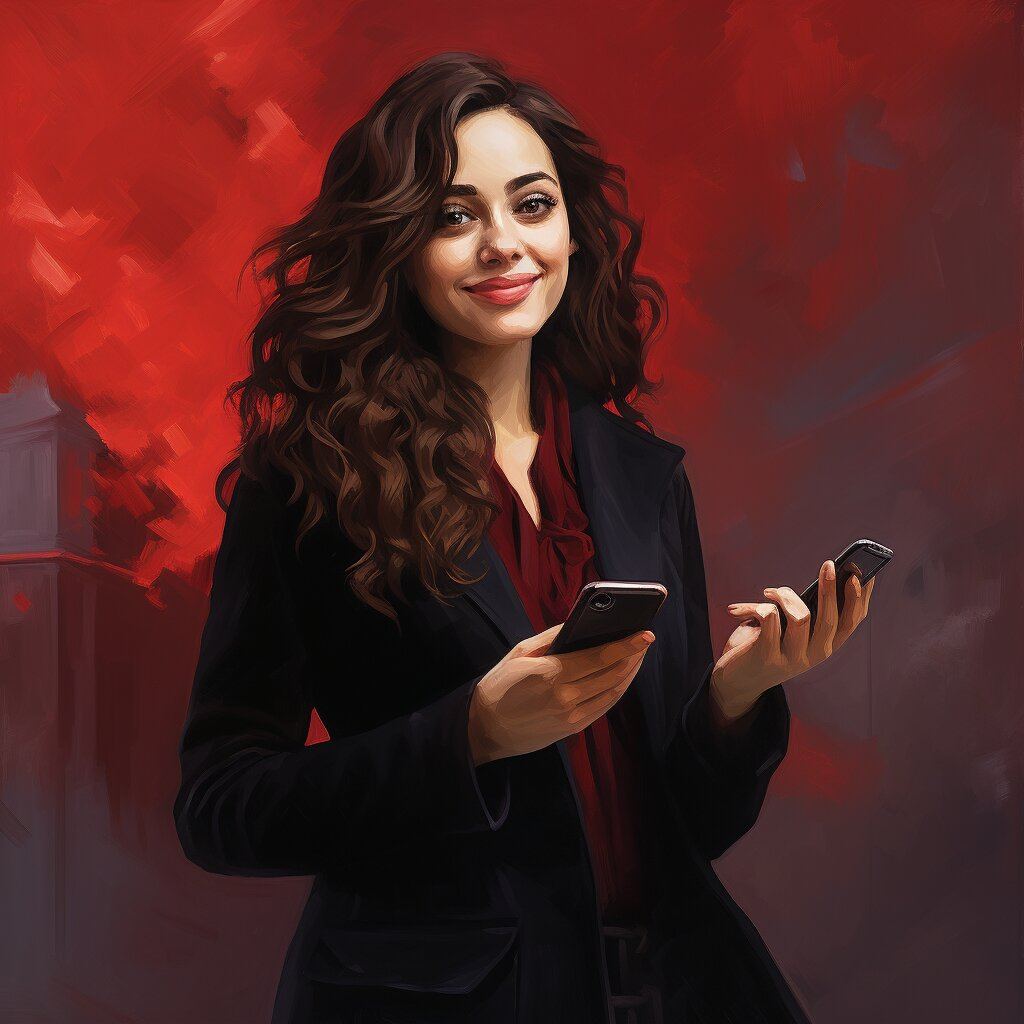} &
        \includegraphics[width=14mm,height=14mm]{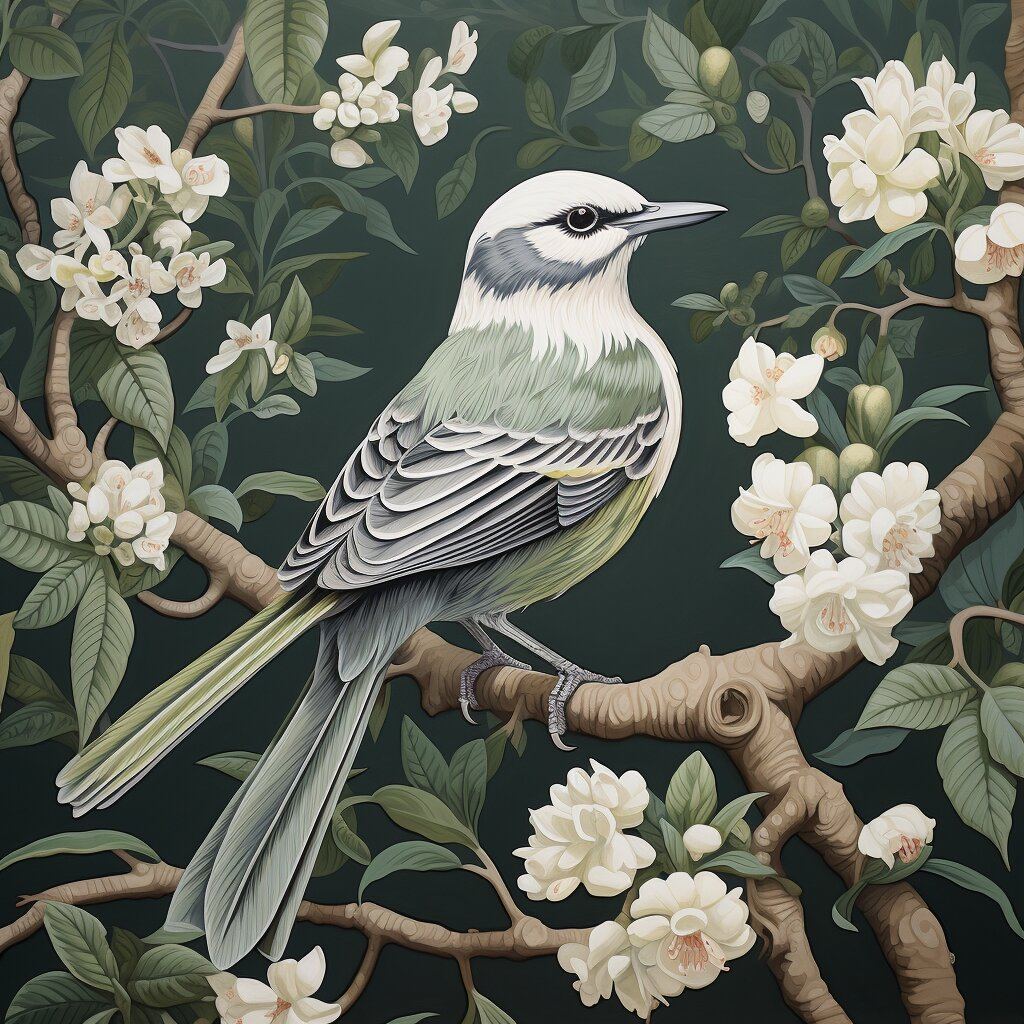}
        \includegraphics[width=14mm,height=14mm]{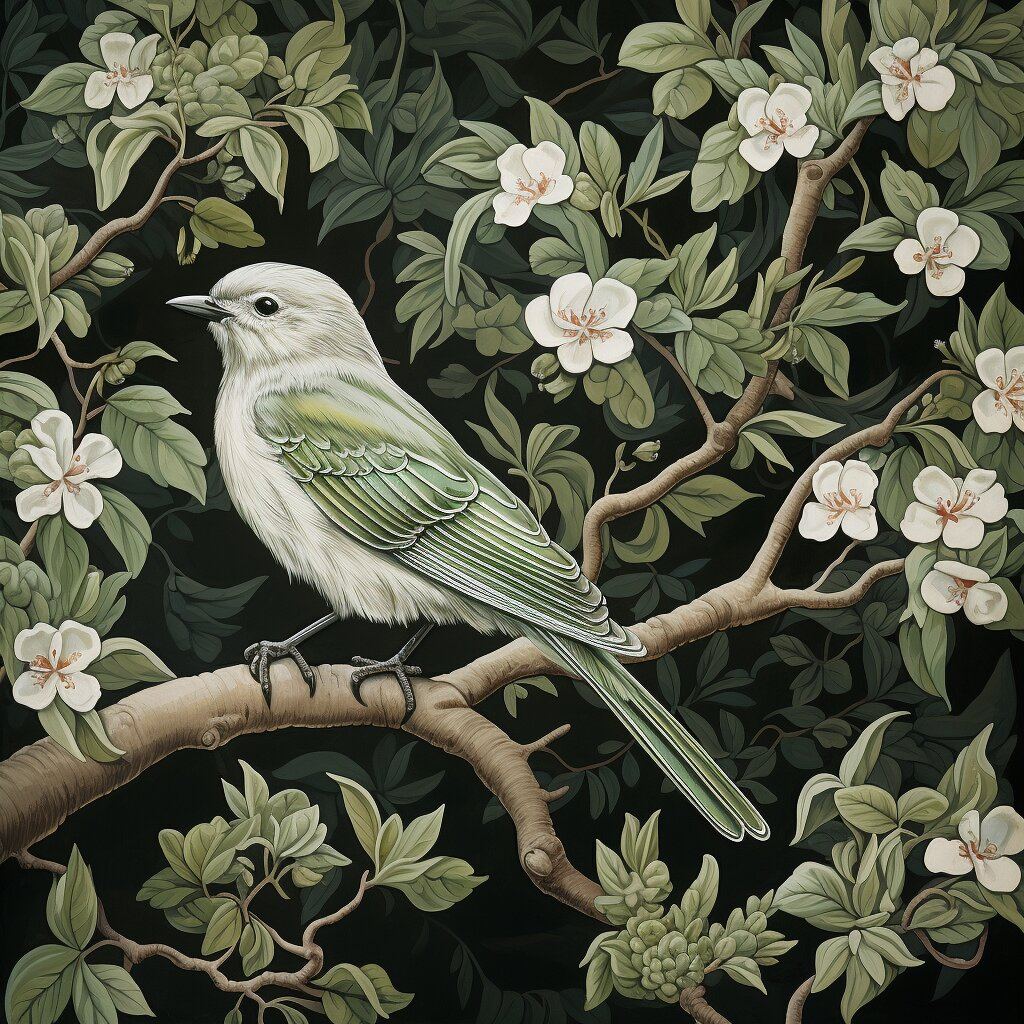} &
        \includegraphics[width=14mm,height=14mm]{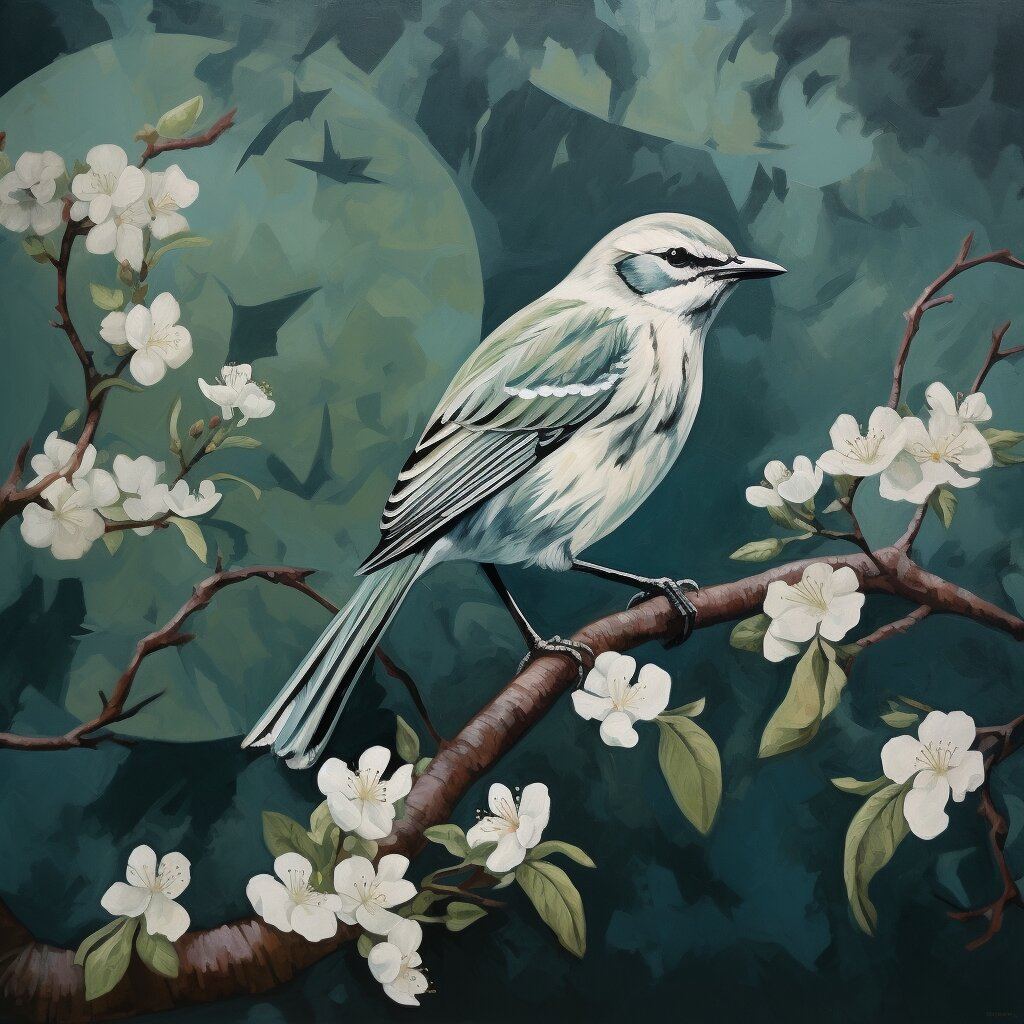}
        \includegraphics[width=14mm,height=14mm]{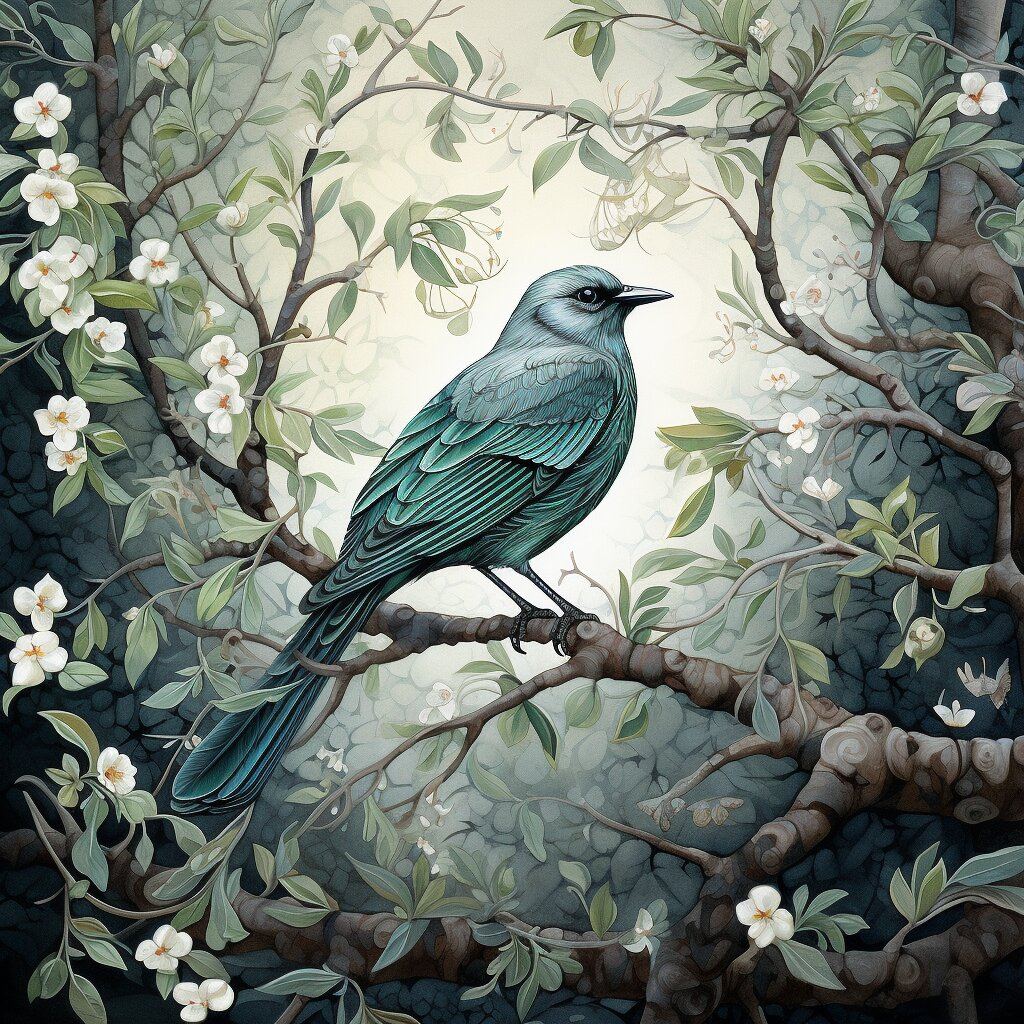} &
        \includegraphics[width=14mm,height=14mm]{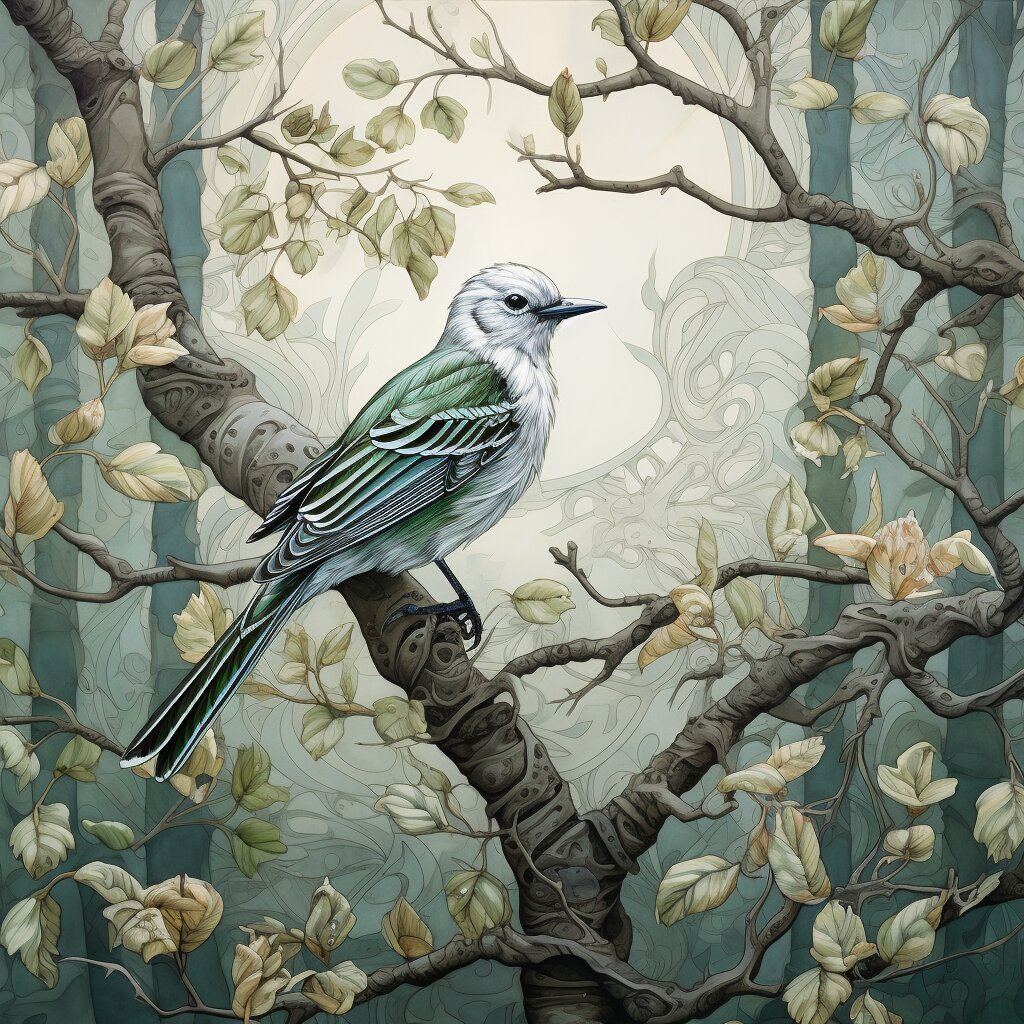}
        \includegraphics[width=14mm,height=14mm]{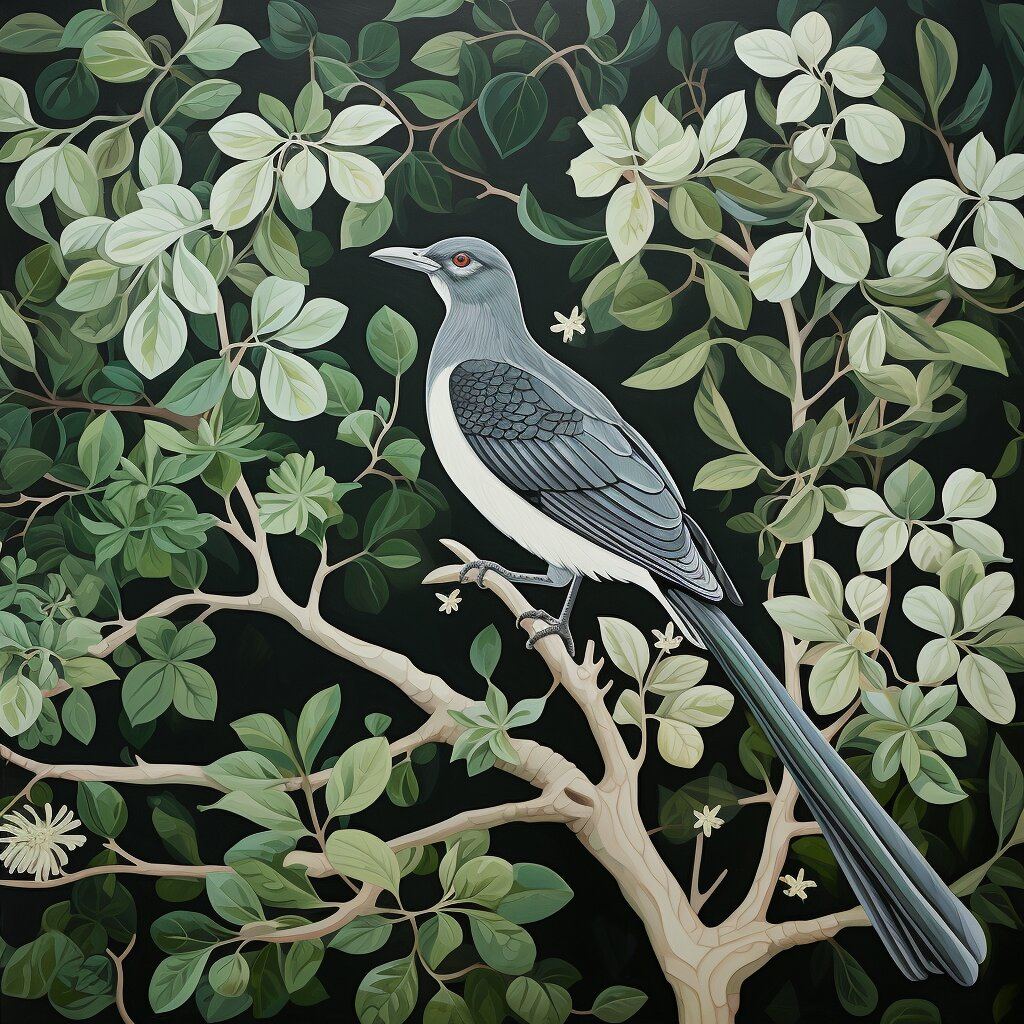} &
        \includegraphics[width=14mm,height=14mm]{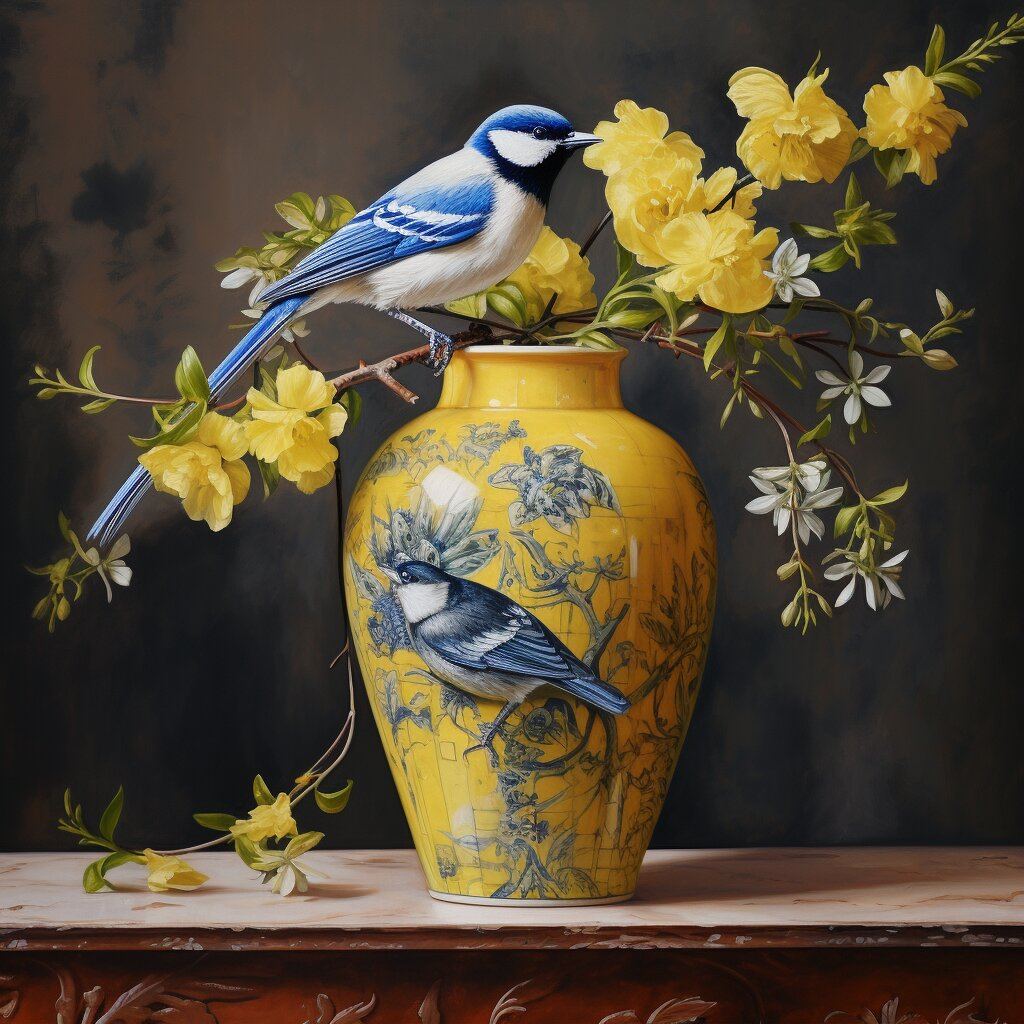}
        \includegraphics[width=14mm,height=14mm]{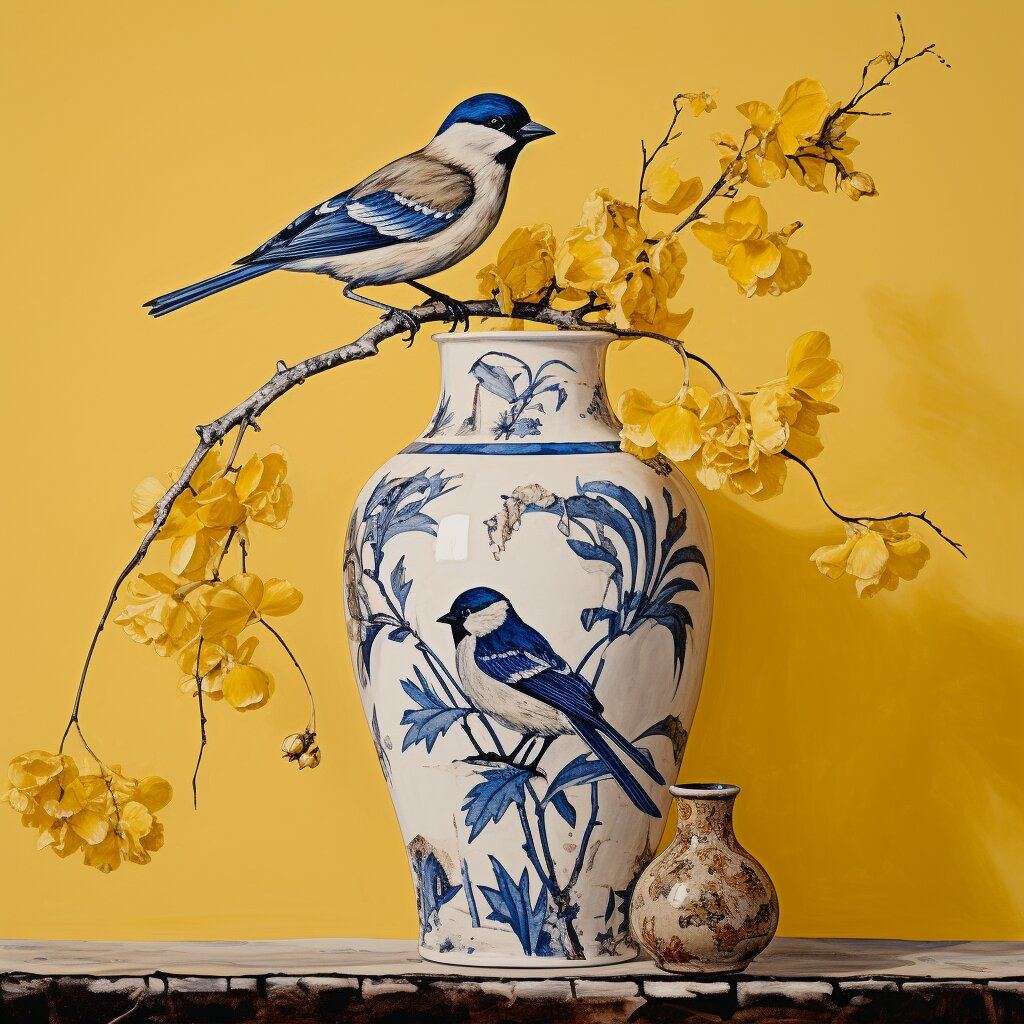}       &
        \includegraphics[width=14mm,height=14mm]{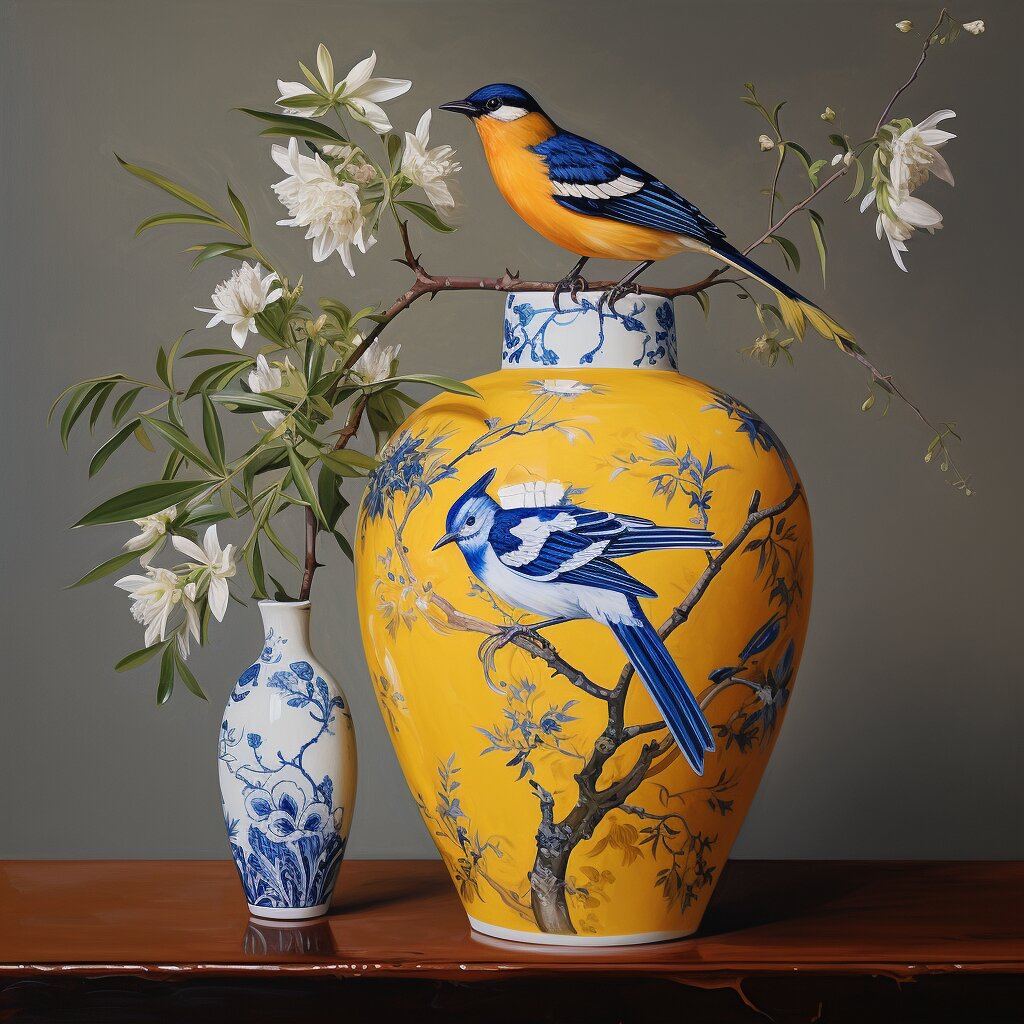}
        \includegraphics[width=14mm,height=14mm]{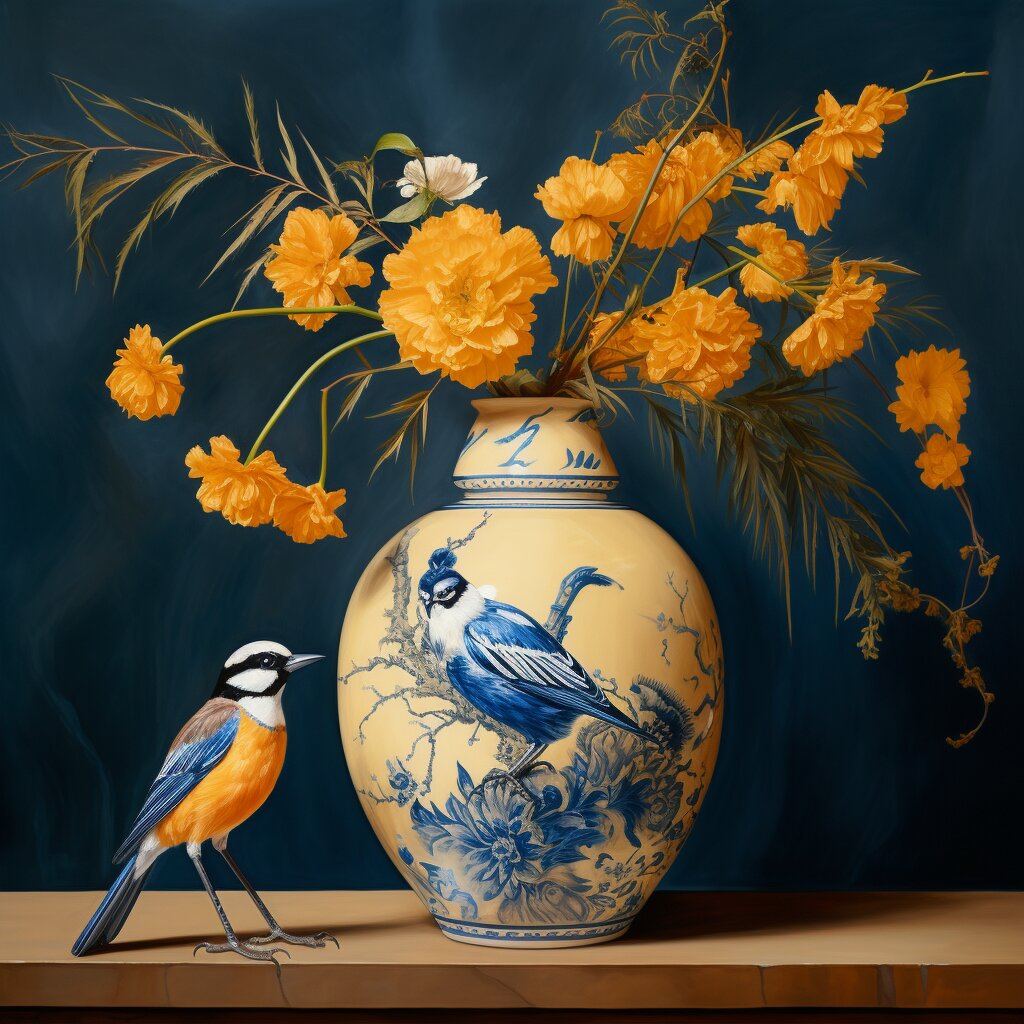}       &
        \includegraphics[width=14mm,height=14mm]{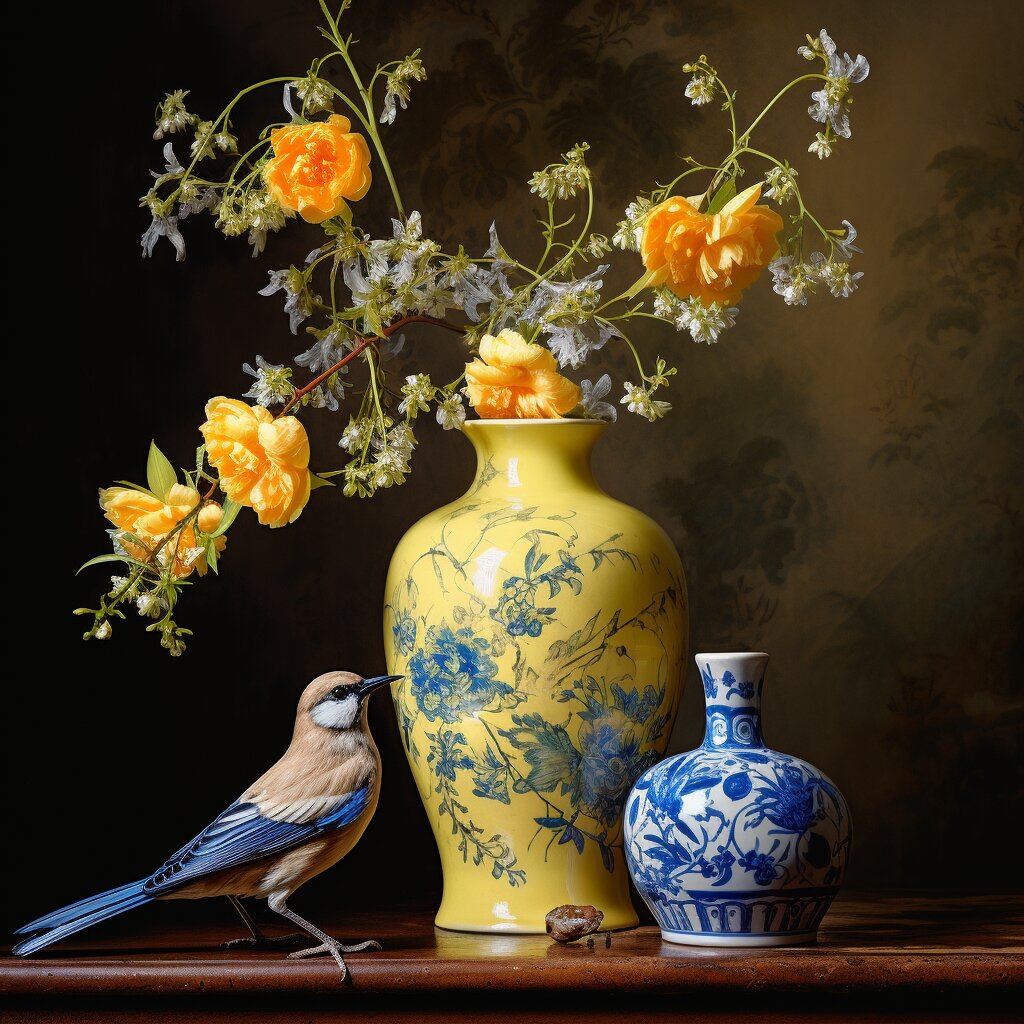}
        \includegraphics[width=14mm,height=14mm]{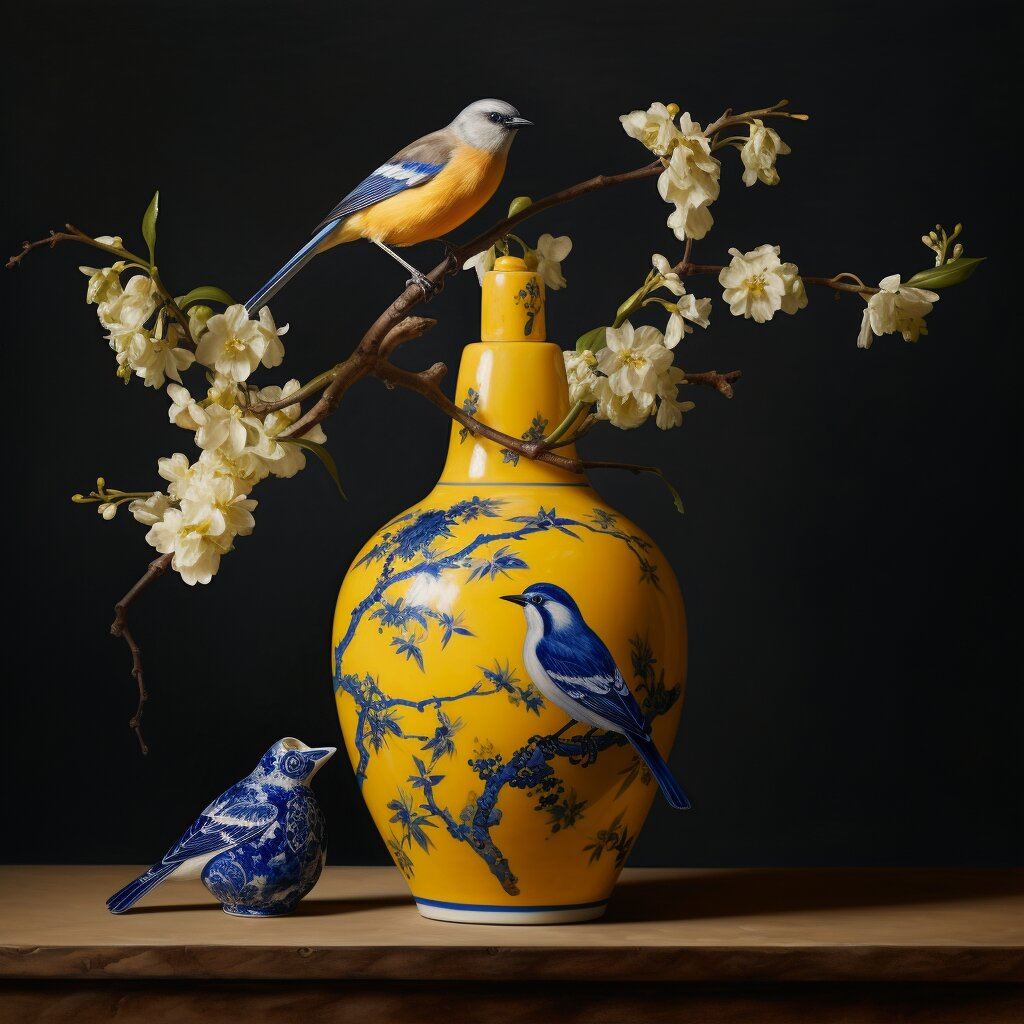}
    \end{tabular}\\[3mm]
    \begin{tabular}{c@{\hspace{1mm}}C{14mm}C{14mm}C{14mm}@{\hspace{2mm}}C{14mm}C{14mm}C{14mm}}
                                                                                                                          &
        \multicolumn{3}{c}{\parbox{42mm}{\centering\emph{A purple bowl and a blue car and~a~green~sofa}}}                 &
        \multicolumn{3}{c}{\parbox{42mm}{\centering\emph{A brown dog and a boy holding~a~blue~suitcase}}}
        \\
        \rotatebox[origin=c]{90}{\parbox{28mm}{\centering Stable Diffusion XL v1.0}}                                      &
        \includegraphics[width=14mm,height=14mm]{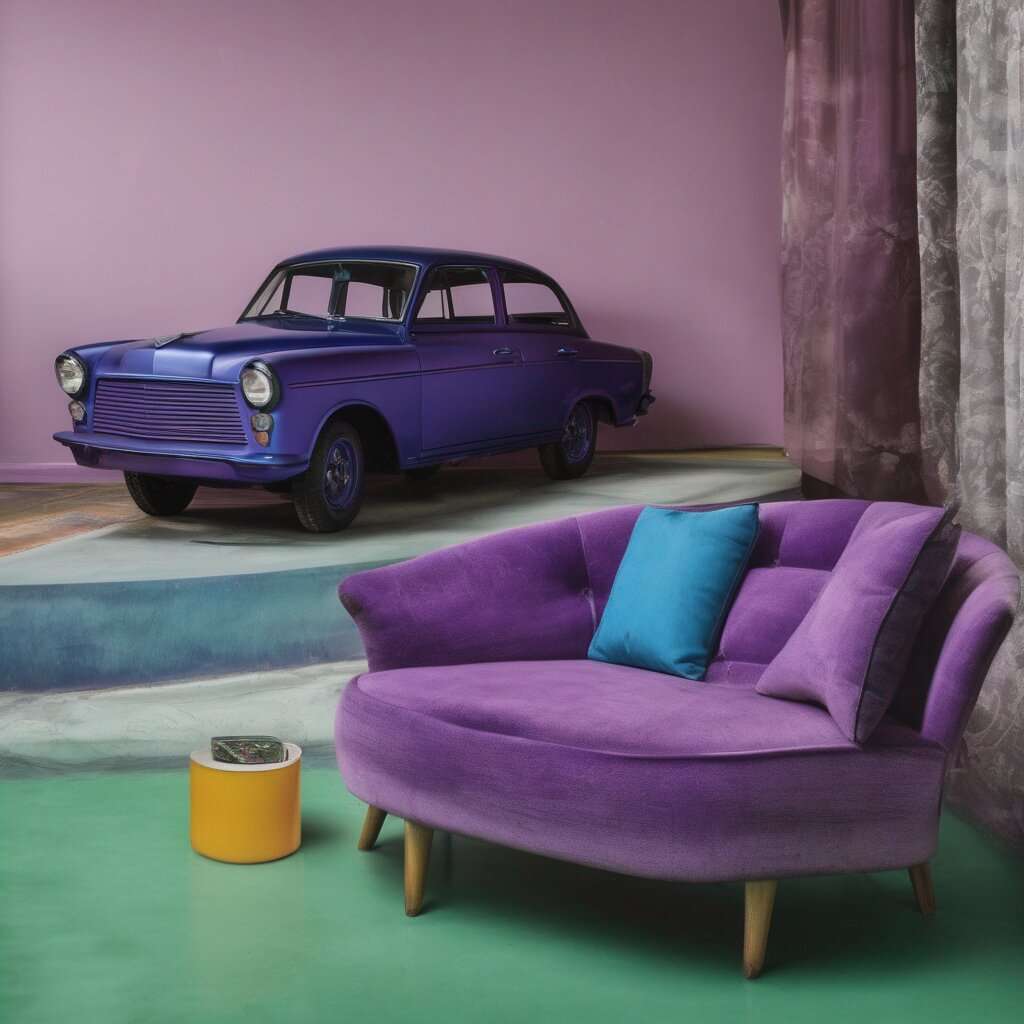}
        \includegraphics[width=14mm,height=14mm]{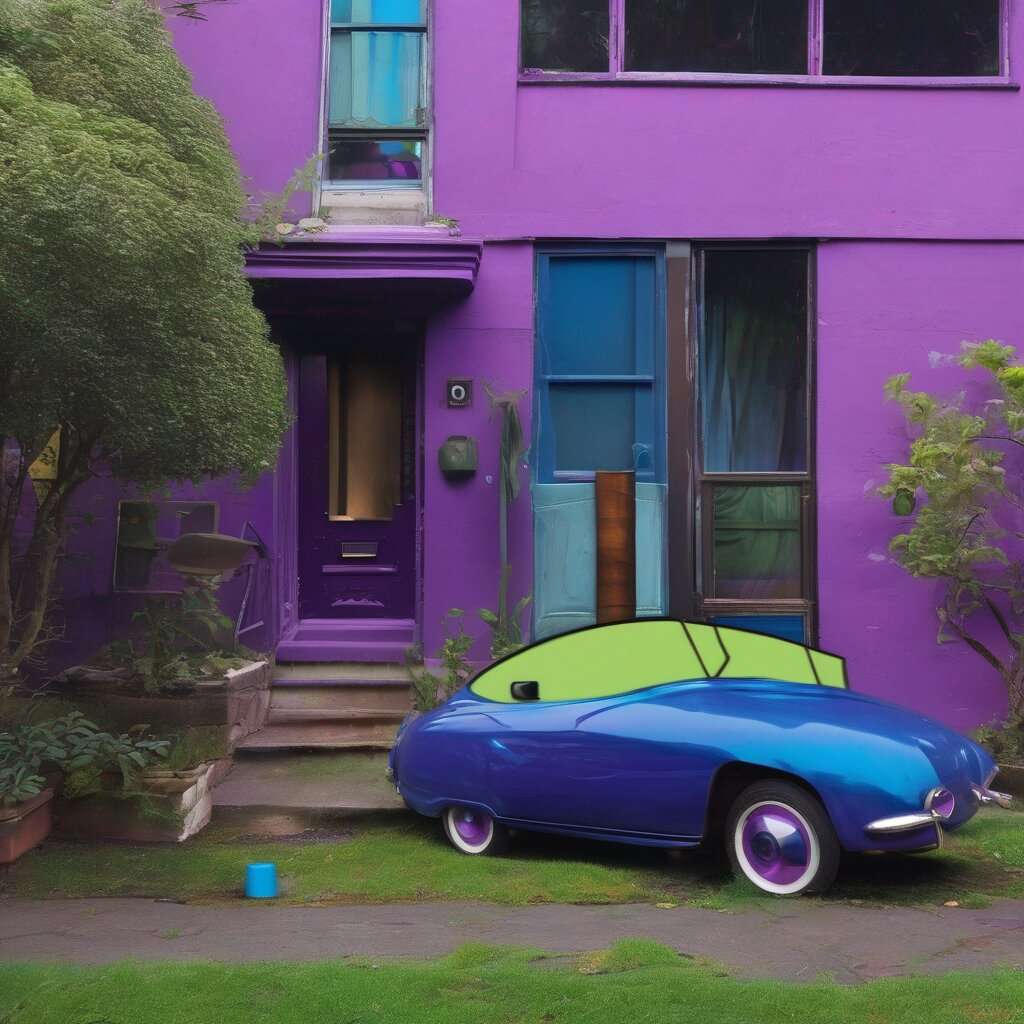}        &
        \includegraphics[width=14mm,height=14mm]{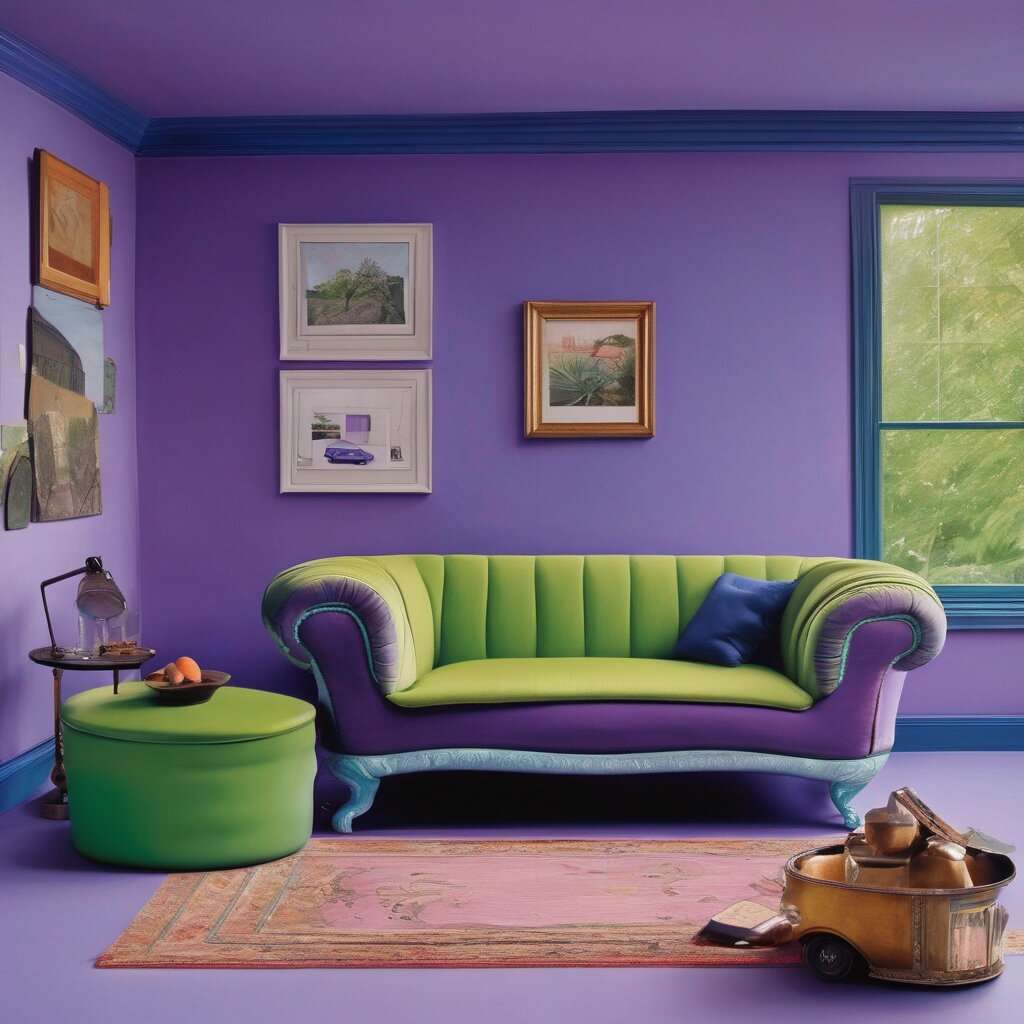}
        \includegraphics[width=14mm,height=14mm]{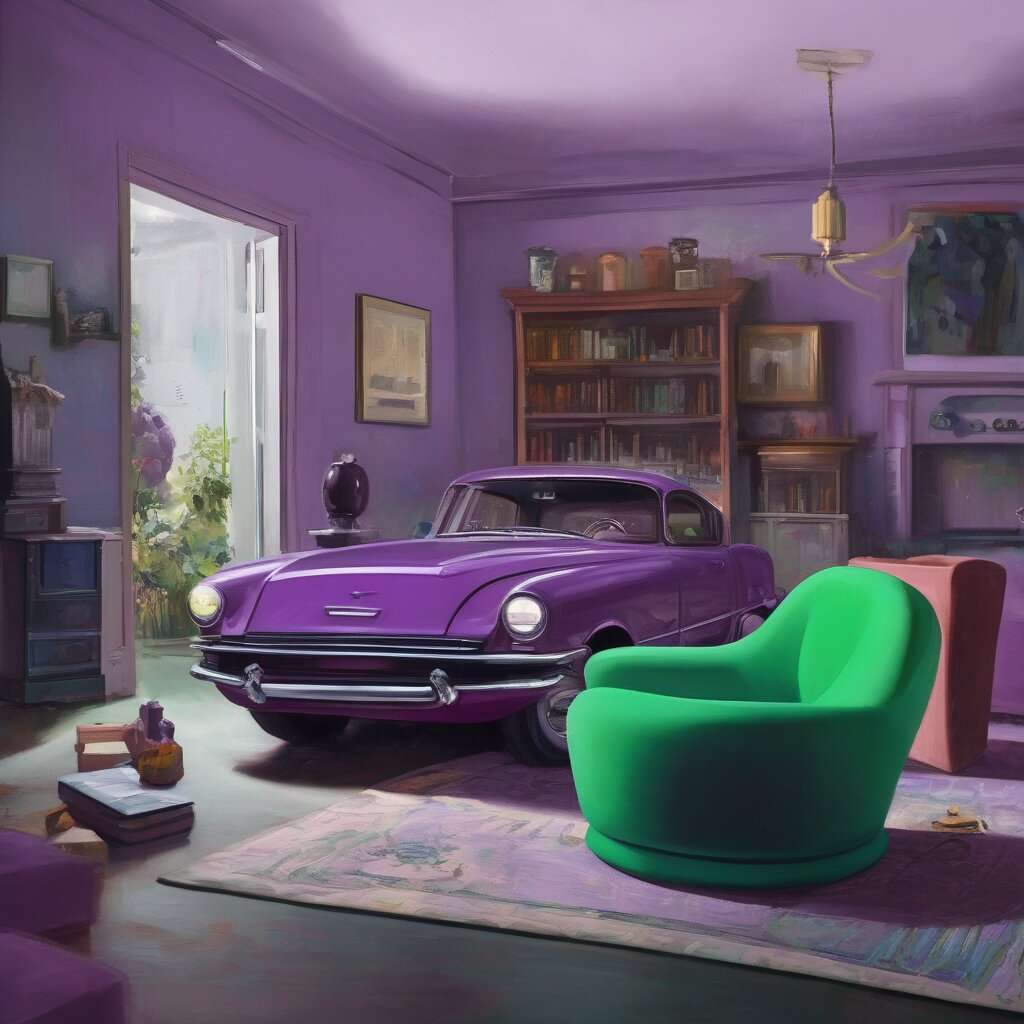}        &
        \includegraphics[width=14mm,height=14mm]{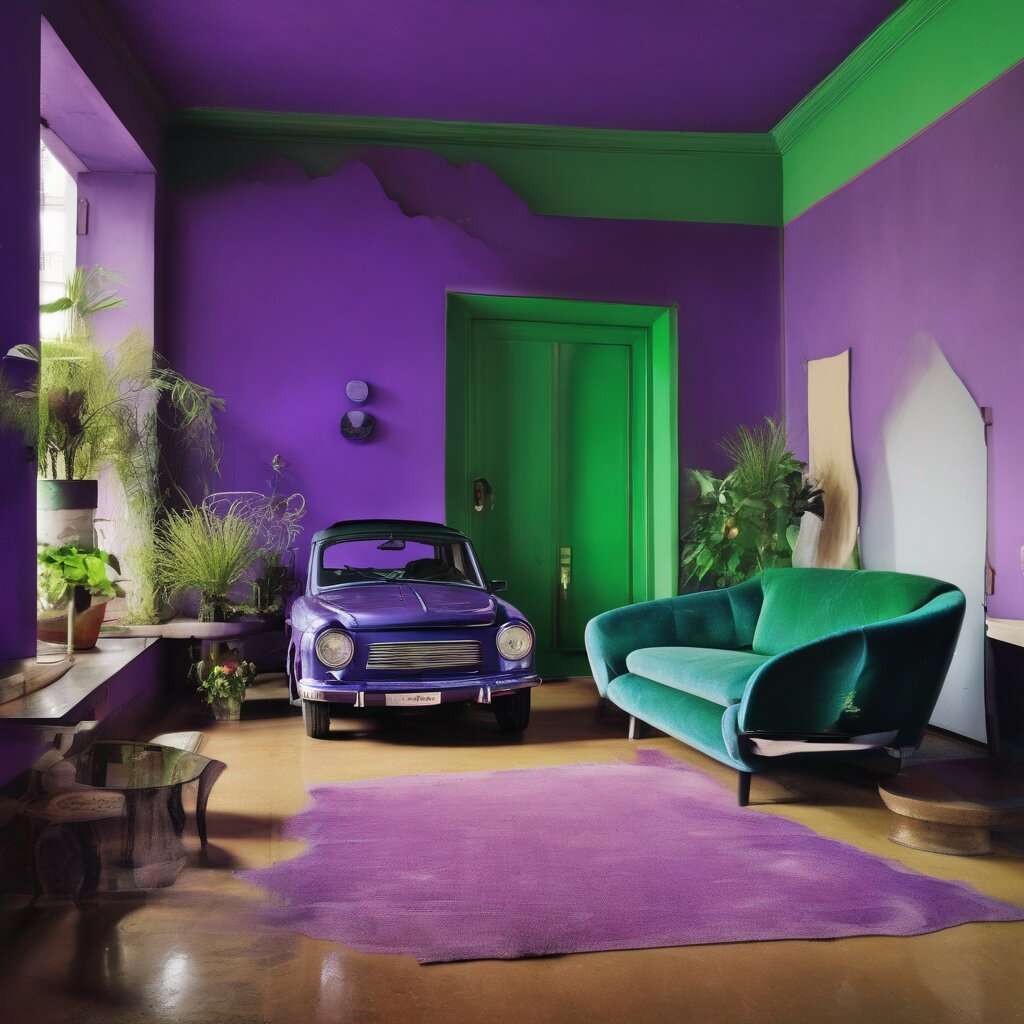}
        \includegraphics[width=14mm,height=14mm]{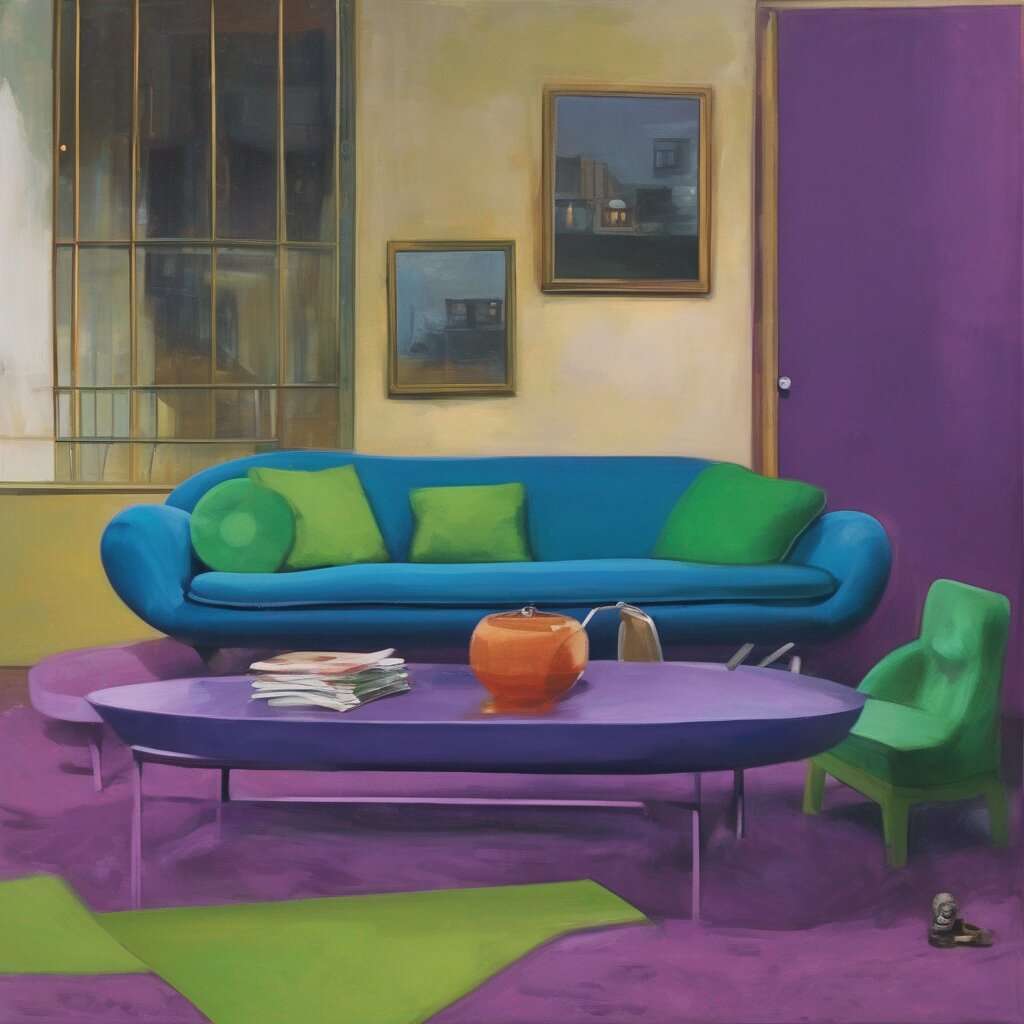}        &
        \includegraphics[width=14mm,height=14mm]{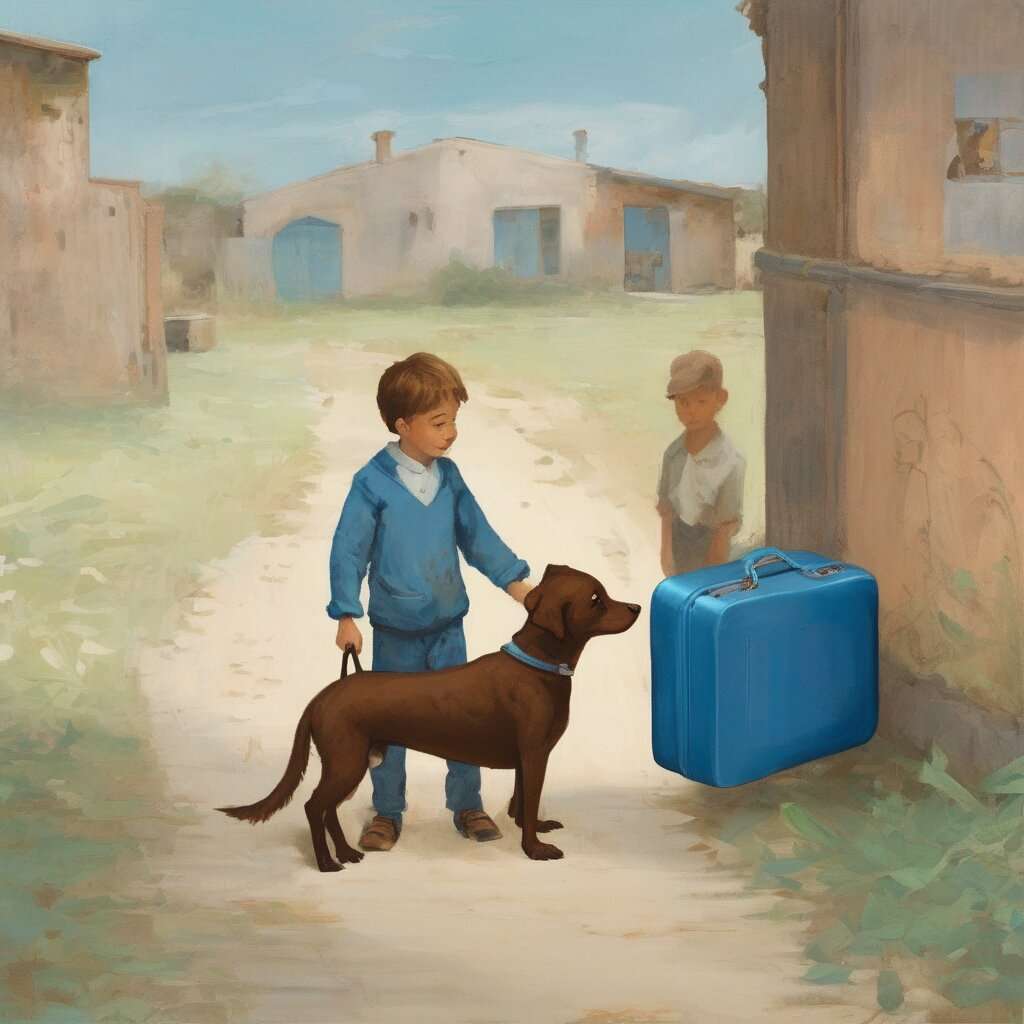}
        \includegraphics[width=14mm,height=14mm]{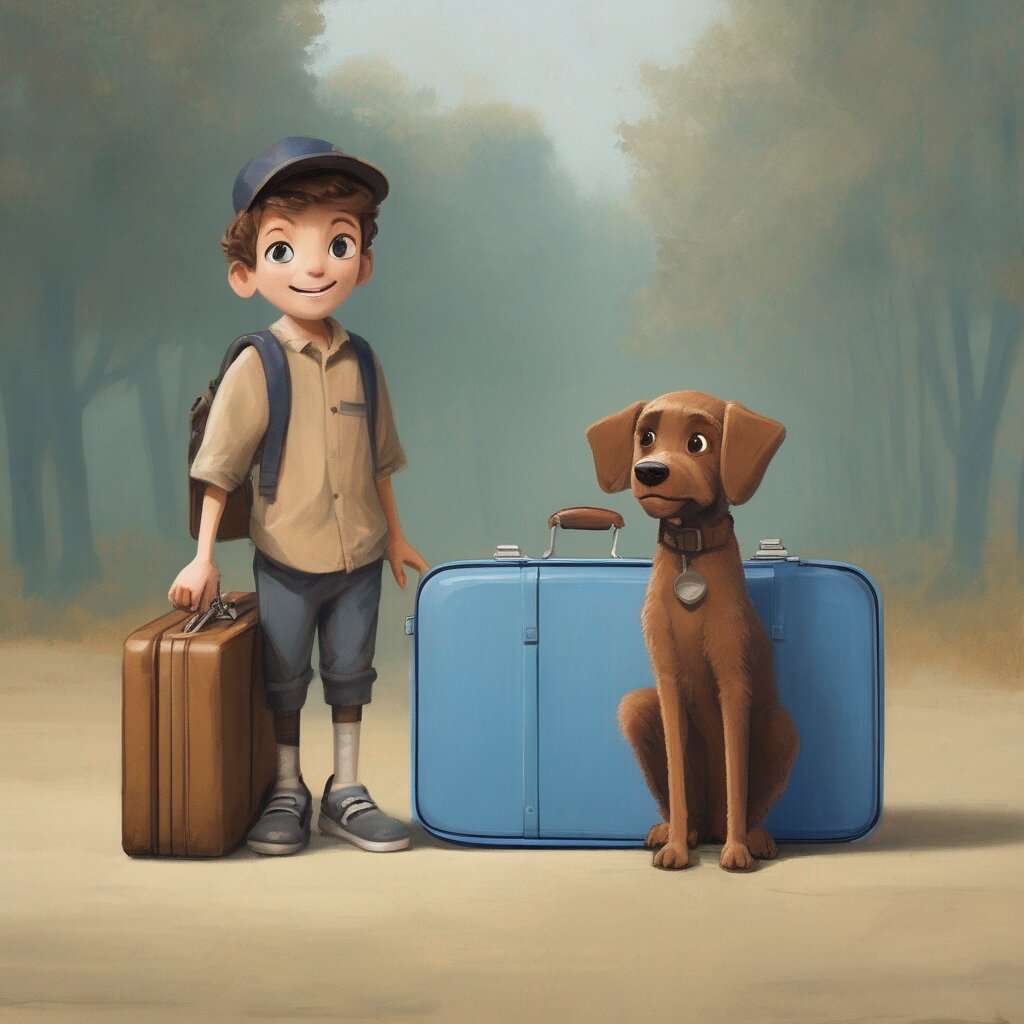}       &
        \includegraphics[width=14mm,height=14mm]{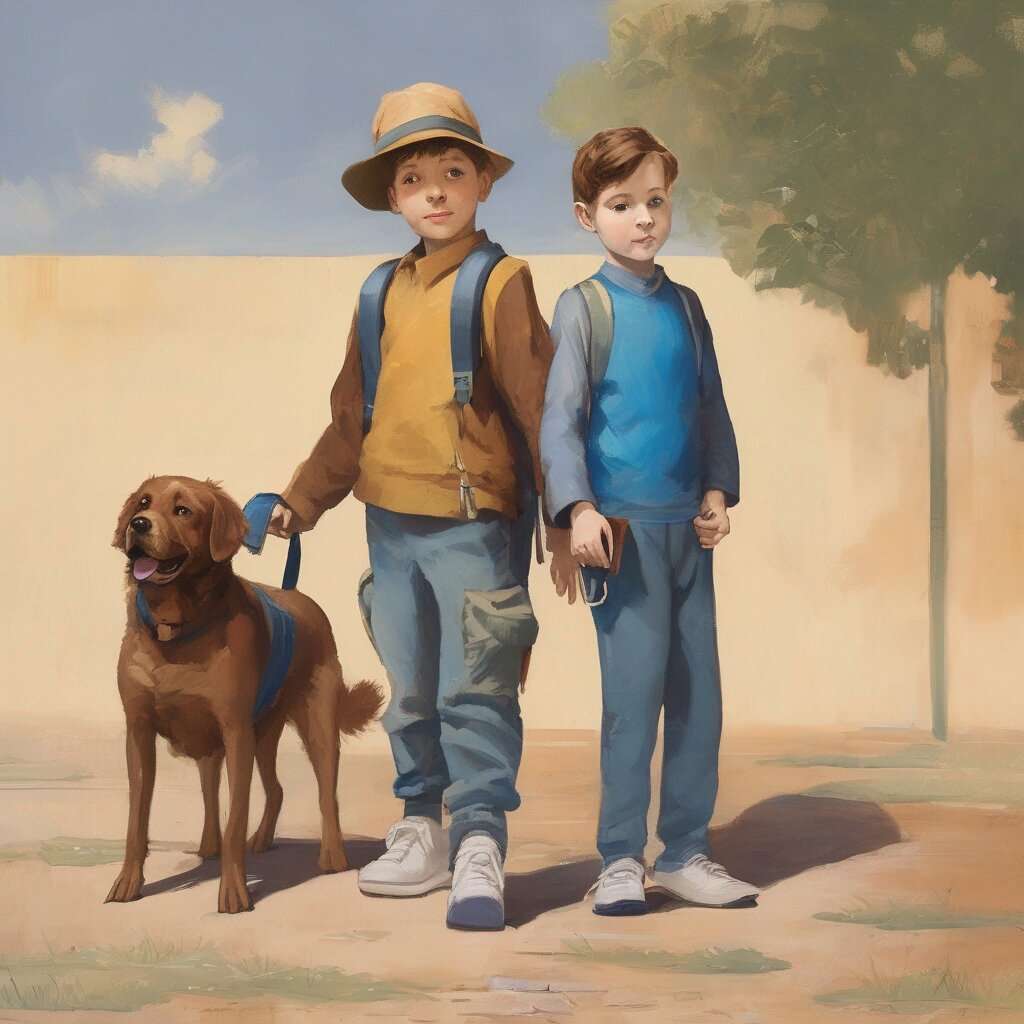}
        \includegraphics[width=14mm,height=14mm]{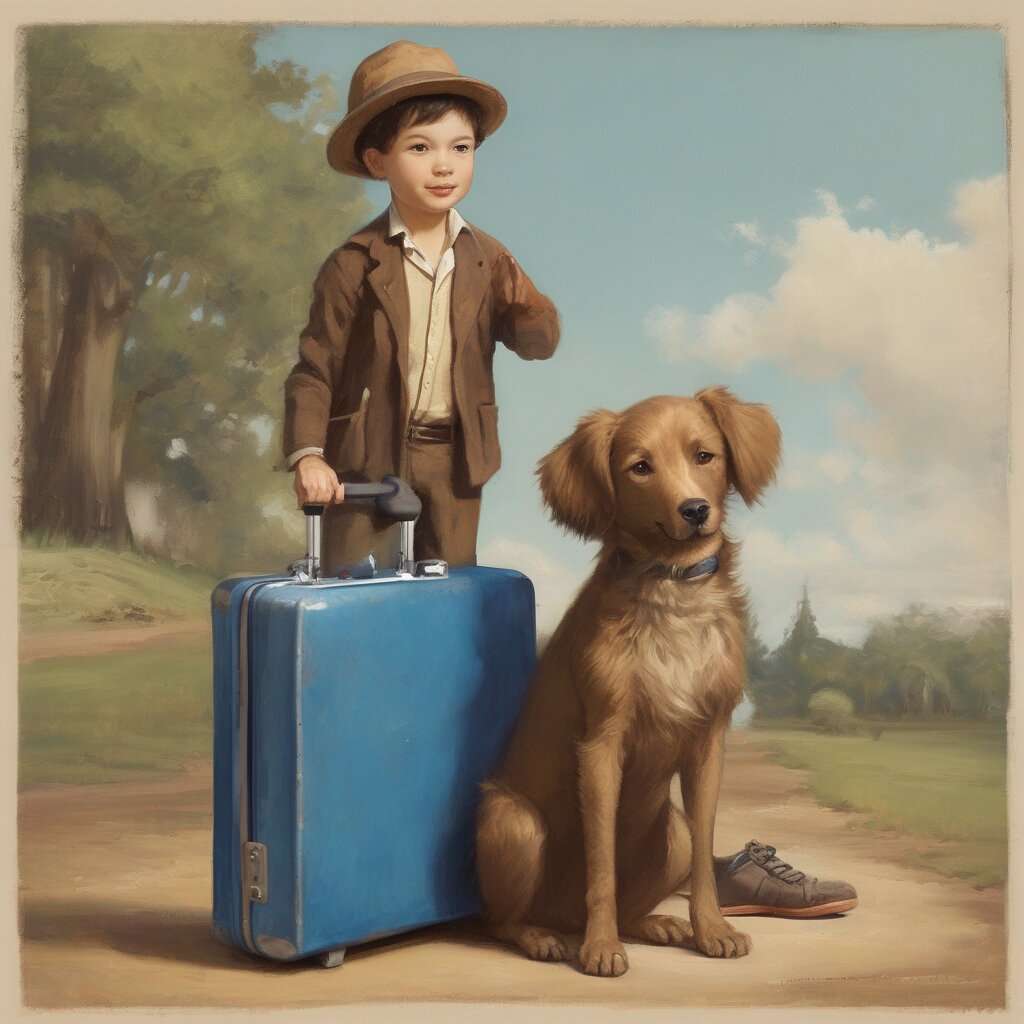}       &
        \includegraphics[width=14mm,height=14mm]{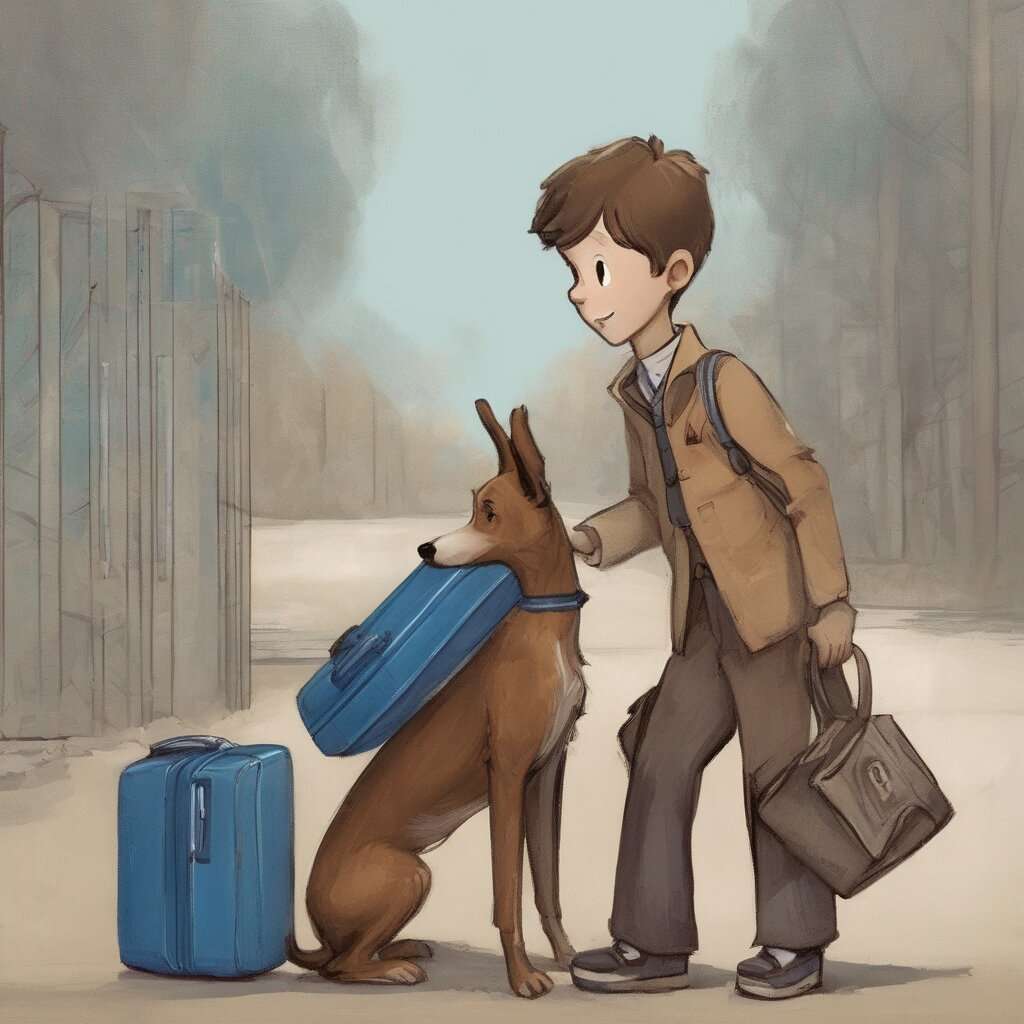}
        \includegraphics[width=14mm,height=14mm]{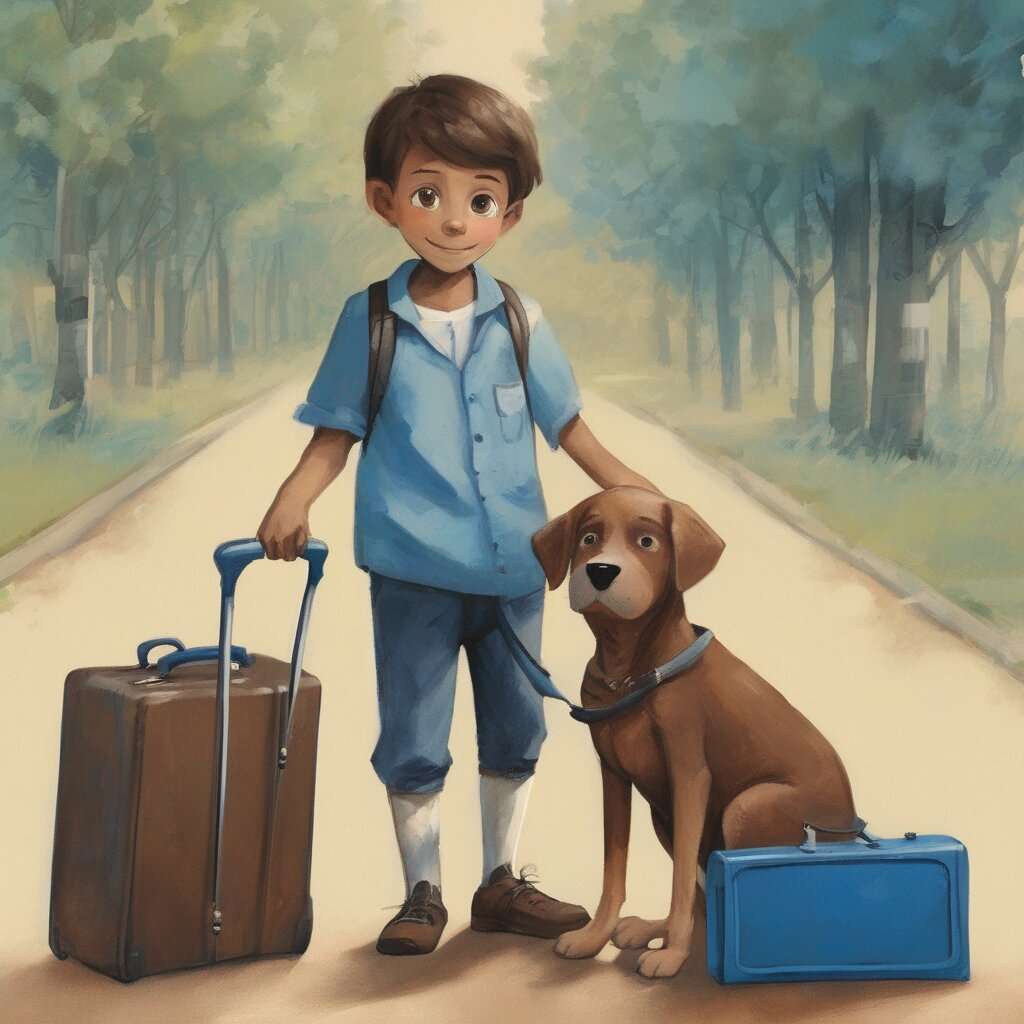}
        \\
        \rotatebox[origin=c]{90}{\parbox{28mm}{\centering Midjourney v5.2}}                                               &
        \includegraphics[width=14mm,height=14mm]{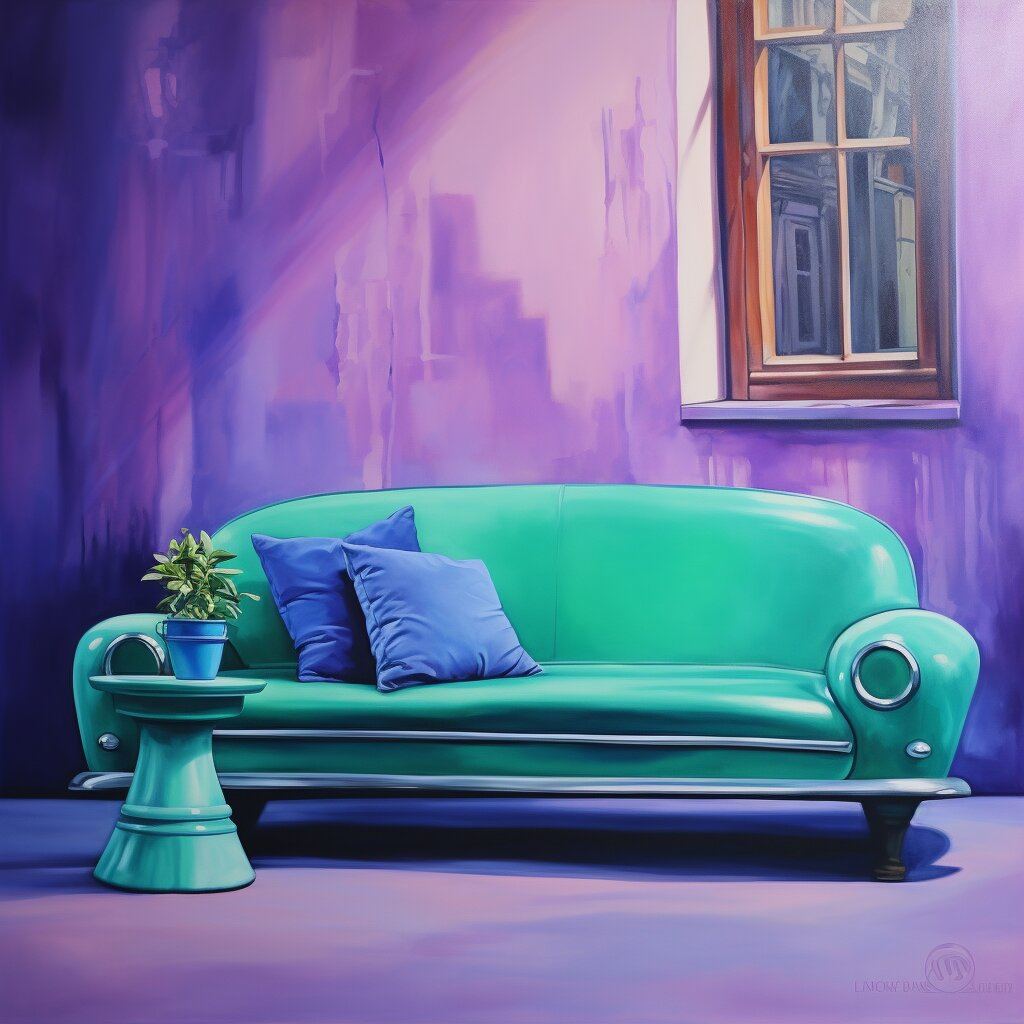}
        \includegraphics[width=14mm,height=14mm]{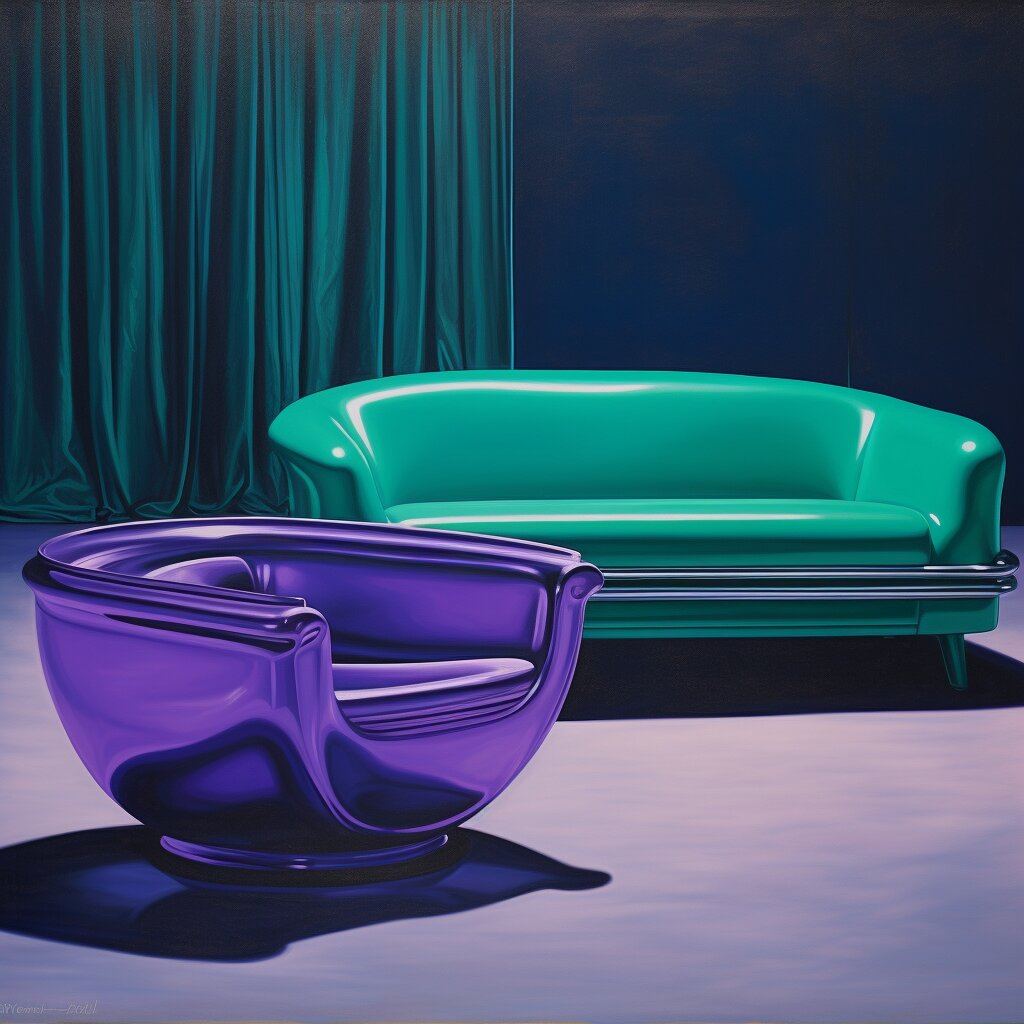} &
        \includegraphics[width=14mm,height=14mm]{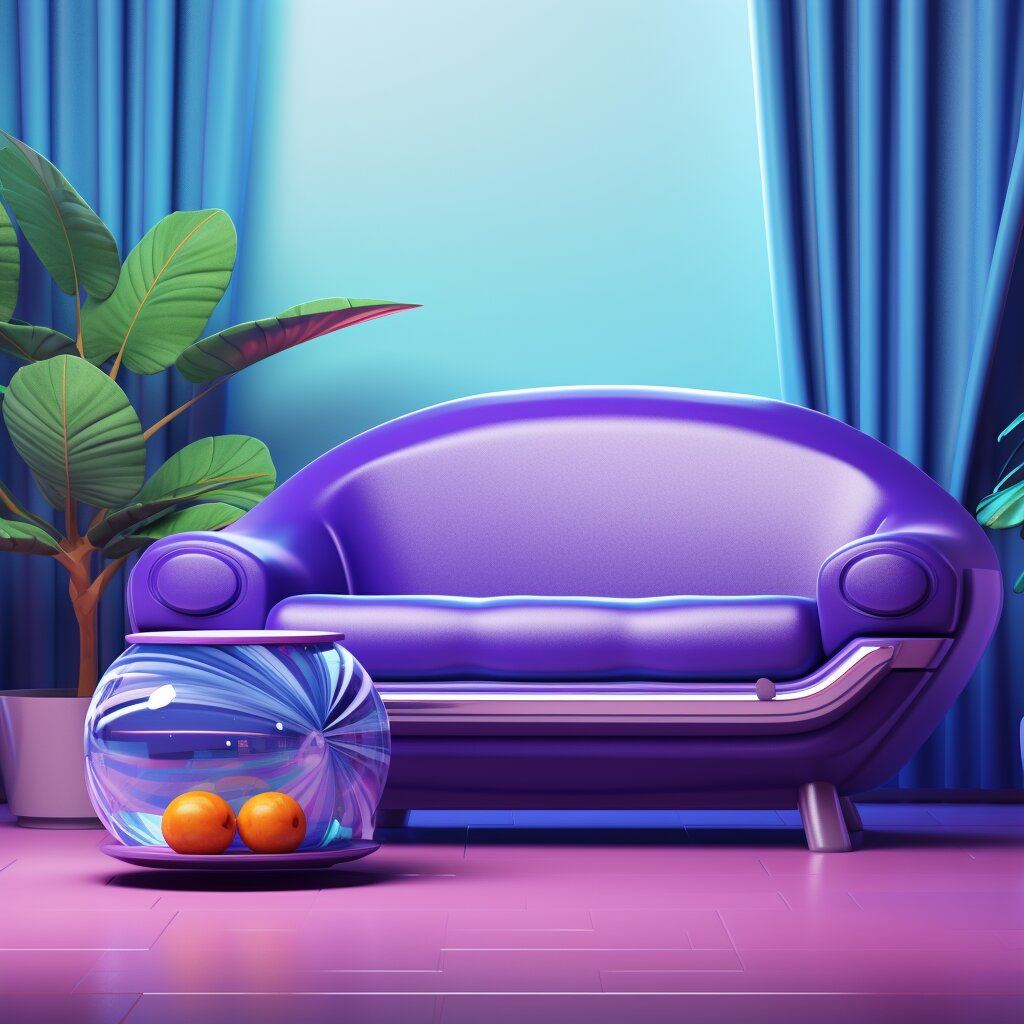}
        \includegraphics[width=14mm,height=14mm]{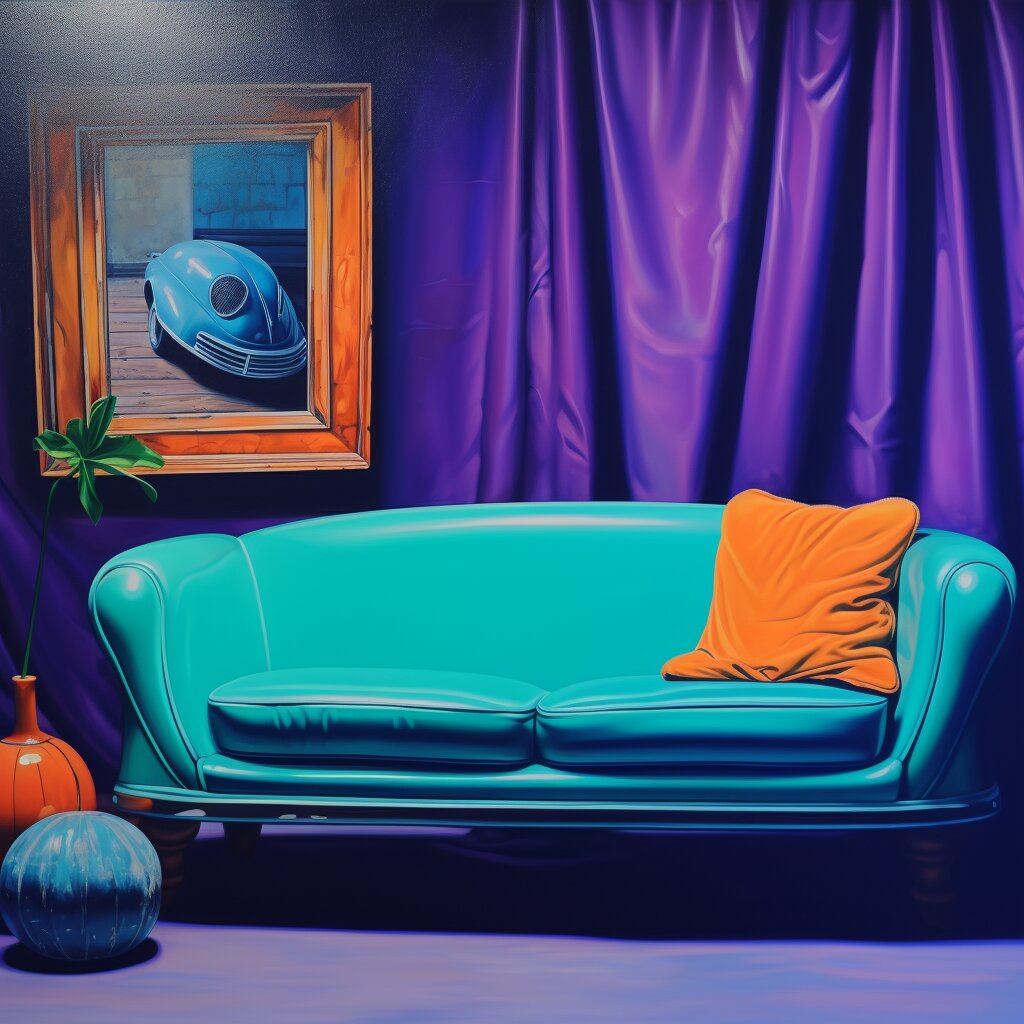} &
        \includegraphics[width=14mm,height=14mm]{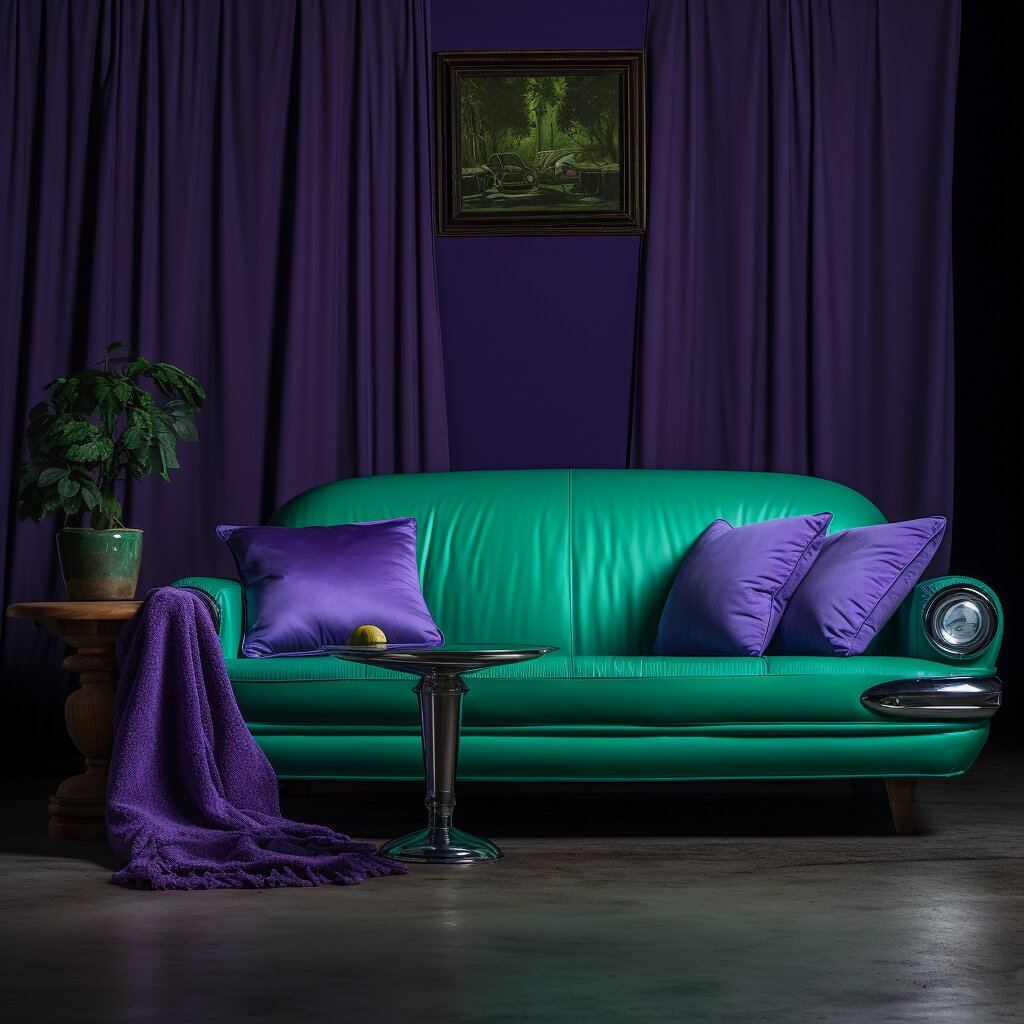}
        \includegraphics[width=14mm,height=14mm]{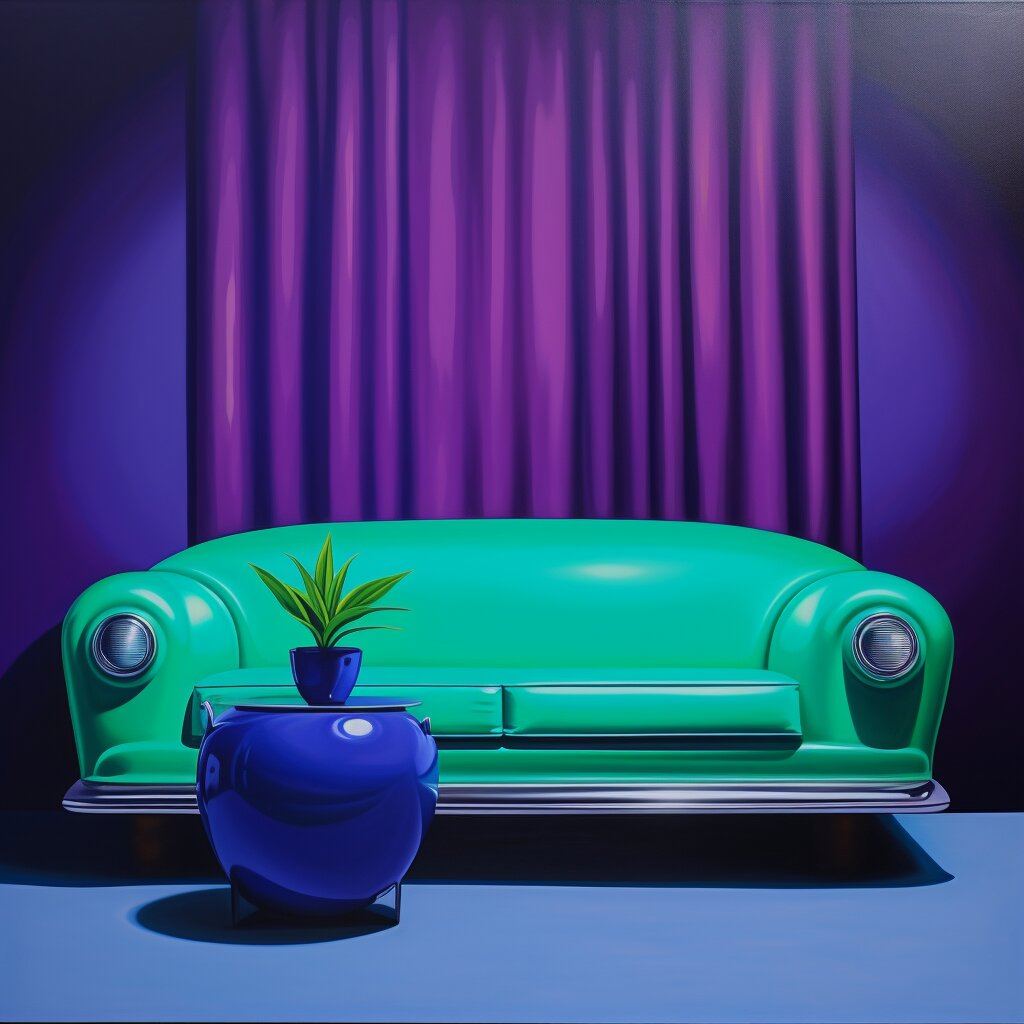} &
        \includegraphics[width=14mm,height=14mm]{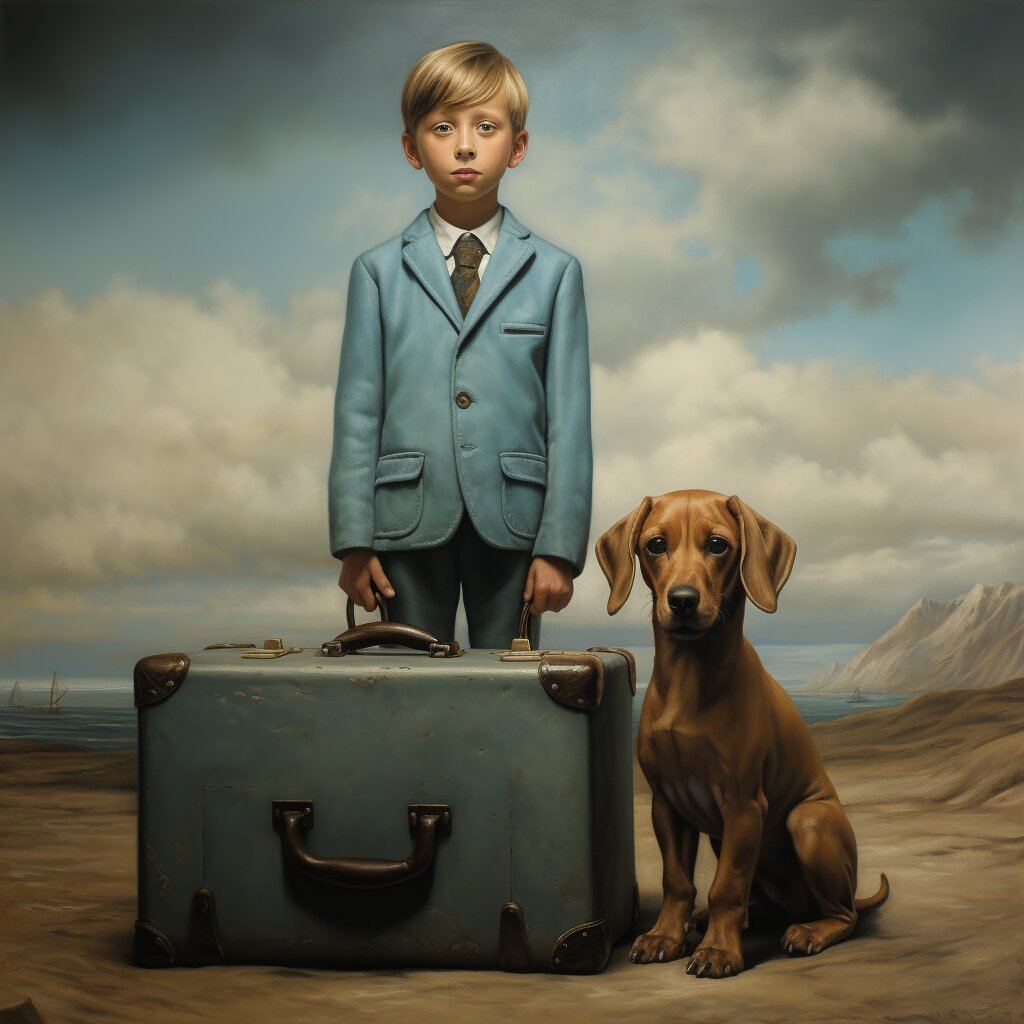}
        \includegraphics[width=14mm,height=14mm]{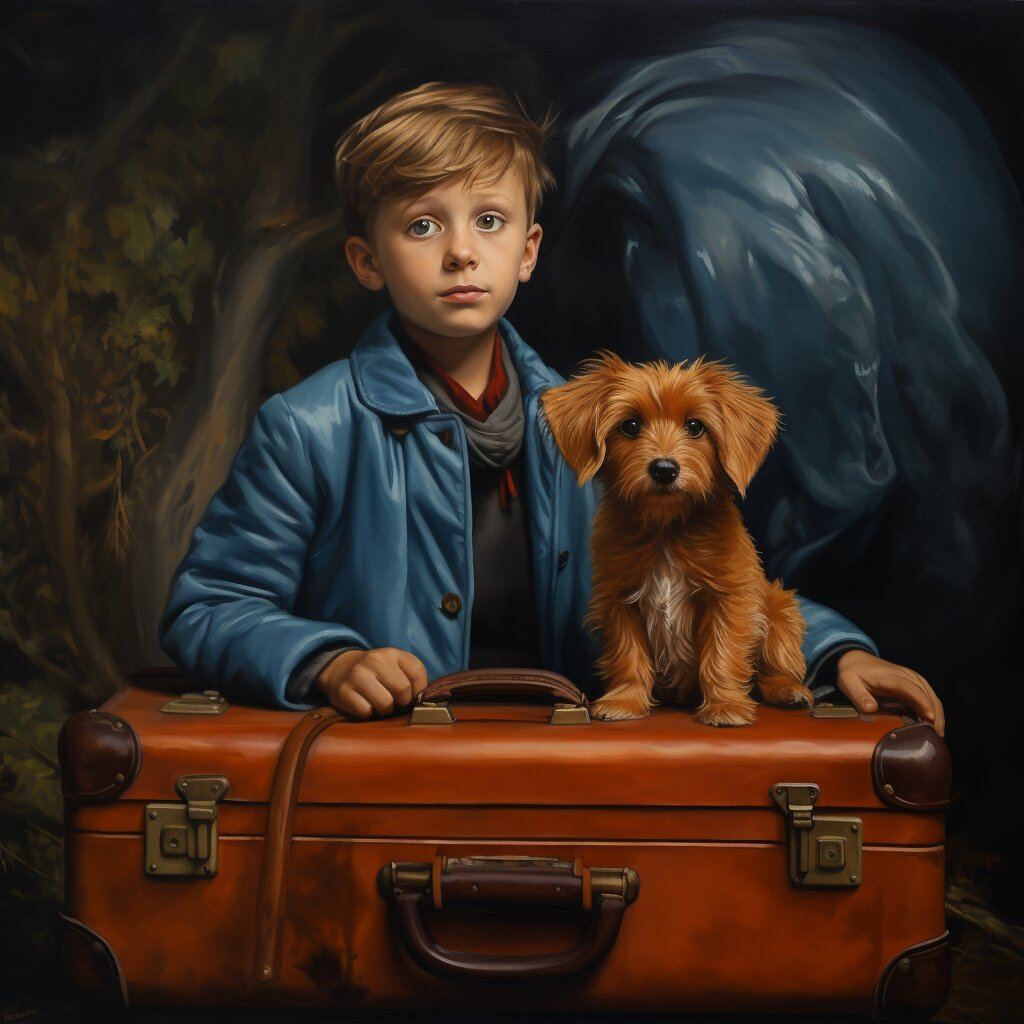} &
        \includegraphics[width=14mm,height=14mm]{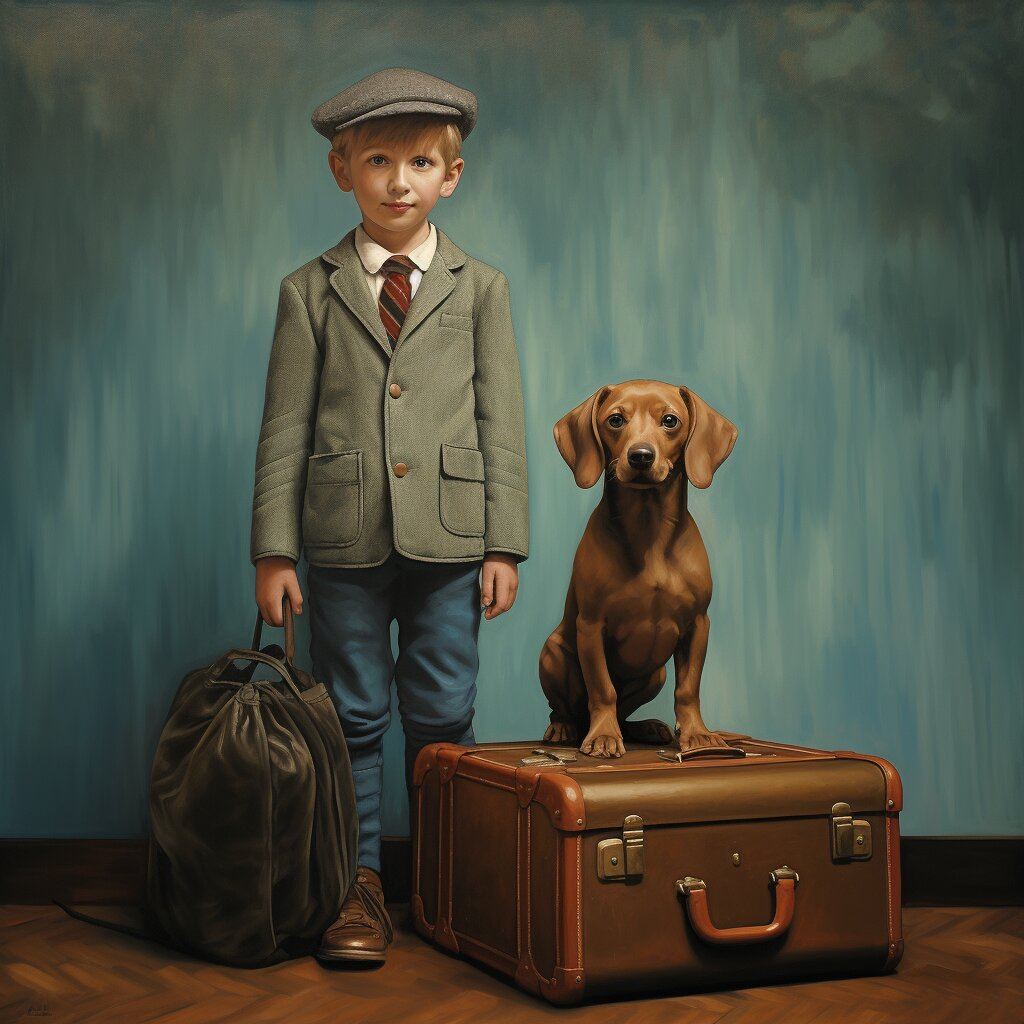}
        \includegraphics[width=14mm,height=14mm]{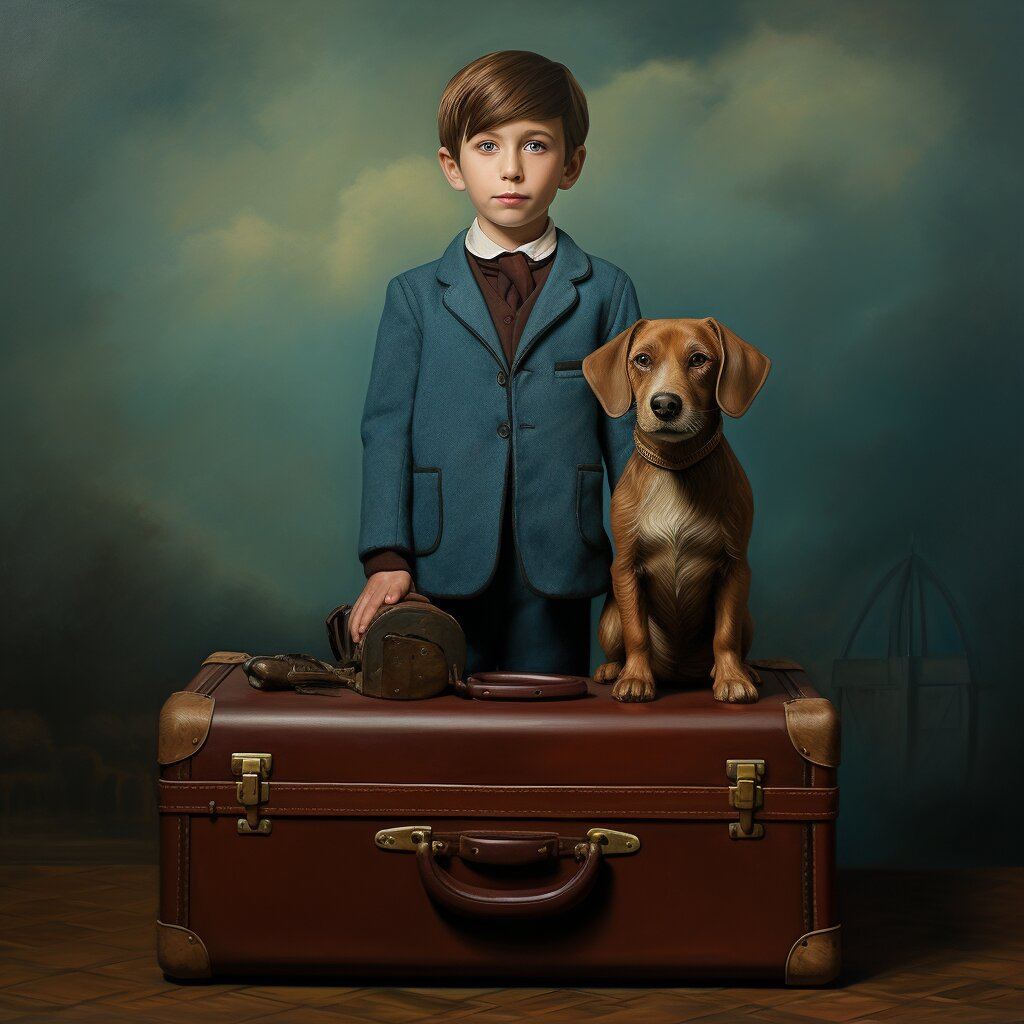} &
        \includegraphics[width=14mm,height=14mm]{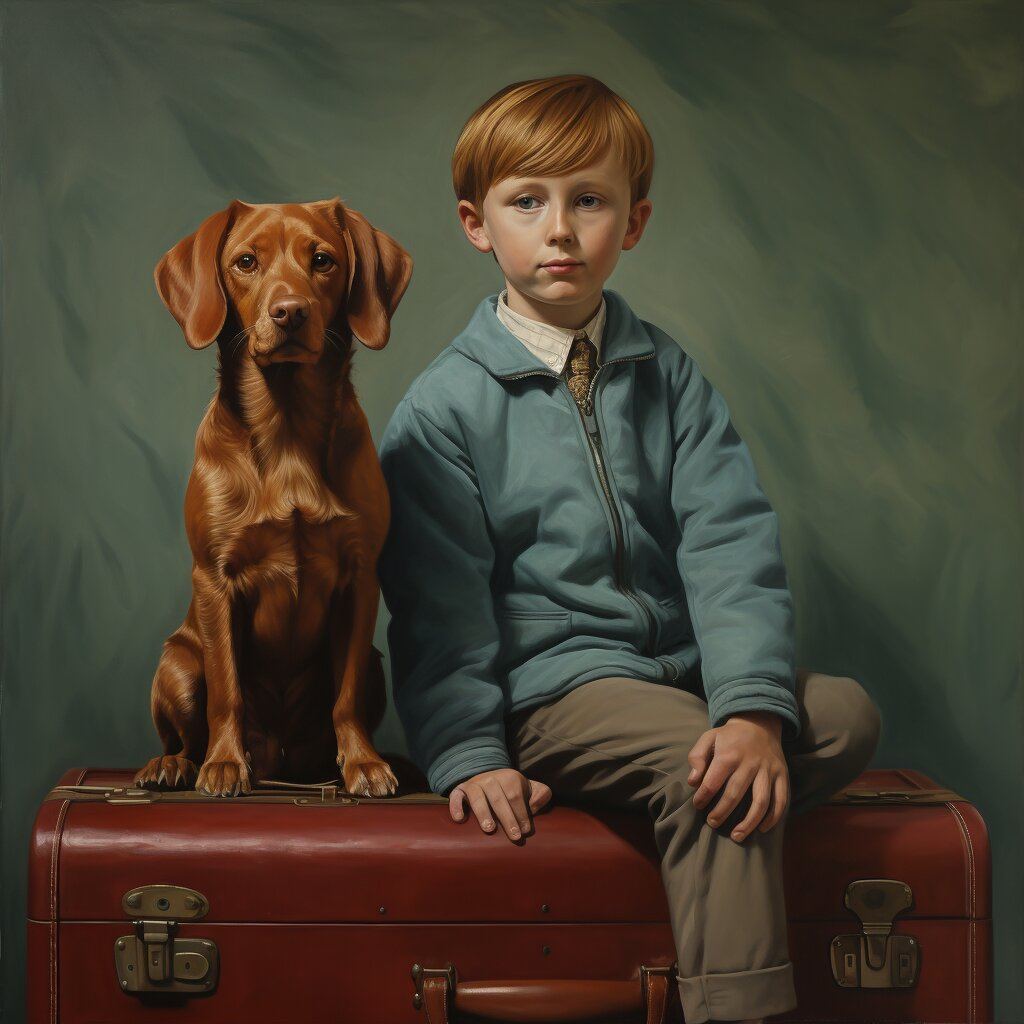}
        \includegraphics[width=14mm,height=14mm]{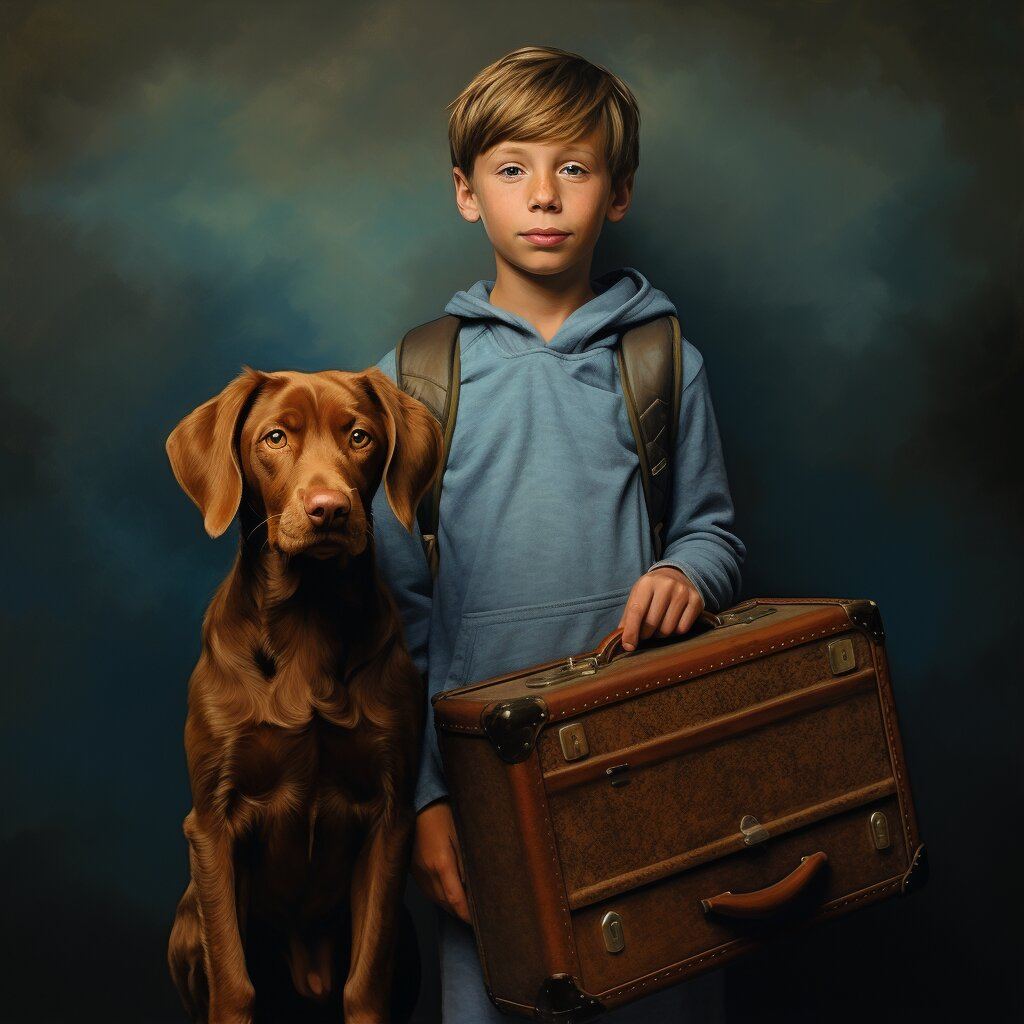}
    \end{tabular}\\[-2mm]
    \caption{Examination of more recently proposed text-to-image diffusion models.
    }
    \label{fig:sota_models}
\end{figure*}

\begin{table*}[t]
    \caption{Comparison with Stable Diffusion XL.}
    \label{tab:sdxl}
    \footnotesize
    \centering
    \tabcolsep=1.5mm
    \begin{tabular}{lcccccc}
        \toprule
                                                          & \multicolumn{2}{c}{Experiment \ref{ex:1}}    & \multicolumn{2}{c}{Experiment \ref{ex:2}} & \multicolumn{2}{c}{Experiment \ref{ex:3}}                                                                           \\
                                                          & \multicolumn{2}{c}{for Concurrent Existence} &
        \multicolumn{2}{c}{for One-to-One Correspondence} &
        \multicolumn{2}{c}{for Possession}                                                                                                                                                                                                                                 \\
        \cmidrule(lr){2-3}\cmidrule(lr){4-5}\cmidrule(lr){6-7}
        \textbf{Methods}                                   & \textbf{Similarity}$^\ddagger$               & \textbf{CLIP-IQA}                         & \textbf{Similarity}$^\ddagger$            & \textbf{CLIP-IQA} & \textbf{Similarity}$^\ddagger$  & \textbf{CLIP-IQA} \\
        \midrule
        Stable Diffusion                                  & 0.326 / 0.767                                & 0.761                                     & 0.345 / 0.744                             & 0.756             & 0.320 / 0.811                   & 0.762             \\
        Stable Diffusion + Predicated Diffusion                              & \textbf{0.348} / \textbf{0.825}              & 0.775                                     & \textbf{0.379} / \textbf{0.811}           & 0.769             & 0.345 / 0.855 & 0.765             \\
        Stable Diffusion XL                              & 0.340 / 0.820                                & \textbf{0.777}                            & 0.369 / 0.793                             & \textbf{0.773}    & \textbf{0.353} / \textbf{0.862}                   & \textbf{0.780}    \\
        \bottomrule
        \multicolumn{7}{l}{$^\dagger$Using the lenient and strict criterions. $^\ddagger$Text-image similarity and text-text similarity.}
    \end{tabular}
\end{table*}

\subsection{Generality of Challenges}\label{appendix:sdxl}
In the main body, we focused on Stable Diffusion v1.4.
This is because all comparison methods employed it as their backbone, allowing for fair and accurate comparisons.
However, one might raise concerns that the challenges identified are specific to Stable Diffusion alone.
To address this and demonstrate the generality of the challenges, we conducted Experiment \ref{ex:4} on more recently proposed text-to-image diffusion models, Stable Diffusion XL (SDXL) v1.0\footnote{\url{https://github.com/Stability-AI/generative-models} (MIT license)} and Midjourney v5.2\footnote{\url{https://www.midjourney.com/} (Terms of Service: \url{https://docs.midjourney.com/docs/terms-of-service})}, that is, we generated images using the prompts as in Figs.~\ref{fig:experiment4}, \ref{fig:experiment4_additional1}, and \ref{fig:experiment4_additional2}.
The results are summarized in Fig.~\ref{fig:sota_models}.

Midjourney and SDXL are clearly superior to Stable Diffusion v1.4 in terms of the quality of the generated images, but they are still prone to challenges discussed in this paper; missing objects, object mixture, attribute leakage, and possession failure.
In our first prompt ``A black bird with red beak,'' Midjourney incorrectly colored the bird's wings red instead of its beak, exemplifying a case of attribute leakage.
SDXL sometimes succeeded but also made the same mistake.
For the second prompt ``A white teddy bear with a green shirt and a smiling girl,'' both SDXL and Midjourney often misidentified the owner of the green shirt as the girl instead of the teddy bear, a typical instance of possession failure.
Interestingly, all images generated by Midjourney are remarkably similar in layout, suggesting the limitation of diversity.
In the third prompt ``A baby with green hair laying in a black blanket next to a teddy bear,'' Midjourney altered the color of the teddy bear or blanket to green, rather than the baby's hair.
SDXL often successfully gave the baby green hair, but still dyed the blanket or teddy bear green as well.
With the fourth prompt ``Woman wearing a black coat holding up a red cellphone,'' both SDXL and Midjourney often incorrectly switched the colors of the coat and cellphone.
The fifth prompt ``A green and grey bird in a tree with white leaves,'' resulted in both SDXL and Midjourney rarely depicting white leaves.
Midjourney often substituted them with white flowers, and the bird was not distinctly grey.
For the sixth prompt ``A yellow vase with a blue and white bird,'' both SDXL and Midjourney consistently mixed object colors.
In the seventh prompt ``A purple bowl and a blue car and a green sofa,'' the bowl was often missing.
In Midjourney, the car was often missing as well, with only something resembling headlights attached to the sofa, suggesting the object mixture.
The colors were also incorrect in most cases.
Finally, in the eighth prompt ``A brown dog and a boy holding a blue suitcase,''  the boy was typically depicted placing the suitcase on the ground, another case of possession failure.
In Midjourney, the suitcase was rarely blue, often replaced by the boy's clothes being blue.
Additionally, the boy and dog appeared almost identical in each trial, highlighting a lack of diversity.


In addition, we conducted Experiments \ref{ex:1}--\ref{ex:3} on SDXL as well, evaluating similarities and CLIP-IQA, with the results summarized in Table \ref{tab:sdxl}.
Note that, even when using the same random seeds, the images generated by Stable Diffusion and SDXL are totally different due to the differences in image resolution and network structure.
In terms of image quality (CLIP-IQA), SDXL consistently outperforms both Stable Diffusion and Predicated Diffusion, as also observed in Fig.~\ref{fig:sota_models}.
However, in terms of fidelity (similarity), Predicated Diffusion with Stable Diffusion as its backbone surpasses SDXL in Experiments \ref{ex:1} and \ref{ex:2}.
SDXL has approximately three times the number of parameters as Stable Diffusion and incorporates various improvements, such as an additional text encoder and pooled text embeddings.
Moreover, it has been trained on a more extensive dataset.
Despite these enhancements, Predicated Diffusion proves to be more effective in preventing missing objects and attribute leakage.
In Experiment \ref{ex:3}, while SDXL outperforms Predicated Diffusion, the margin is smaller than the improvement achieved by Predicated Diffusion over vanilla Stable Diffusion.

Our findings demonstrate that Predicated Diffusion effectively overcomes a variety of challenges.
While cutting-edge models excel in generating high-quality images, they still struggle with these challenges.
Furthermore, the fundamental concept of Predicated Diffusion has the potential to improve these models, providing a promising direction for future research.

\end{document}